\documentclass[a4paper, 12pt, oneside, BCOR1cm,toc=chapterentrywithdots]{scrbook}
\usepackage{graphicx}           
\usepackage{makeidx} 
\usepackage {hyperref}
\hypersetup{
	colorlinks=true,
	linkcolor=black,
}
\usepackage{mathtools}
\usepackage{tocbibind}
\usepackage{blindtext}
\usepackage{subfigure} 
\usepackage{acronym}
\usepackage{amssymb}
\usepackage{graphicx}
\usepackage[bottom]{footmisc}
\usepackage{siunitx}
\newcounter{examplecounter}
\newenvironment{example}{\begin{quote}%
    \refstepcounter{examplecounter}%
  \textbf{Example \arabic{examplecounter}}%
  \quad
}{%
\end{quote}%
}

\usepackage[font=scriptsize]{caption}

\usepackage{amsthm}

\newtheorem{problem}{Problem}
\newtheorem{theorem}{Theorem}
\usepackage{pgfgantt}
\usepackage{tikz}
\usetikzlibrary{arrows, shapes}
\newlength{\mytw}
\DeclareMathOperator*{\argmax}{argmax}
\DeclareMathOperator*{\argminA}{arg\,min} 

\usepackage{tikz}
\usetikzlibrary{arrows.meta}
\tikzset{%
	>={Latex[width=2mm,length=2mm]},
	base/.style = {rectangle, rounded corners, draw=black,
		minimum width=4cm, minimum height=1cm,
		text centered, font=\sffamily},
	activityStarts/.style = {base, fill=white!30},
	activityStarts_d/.style = {base,draw=white , fill=white!30},
	startstop/.style = {base, fill=white!30},
	activityRuns/.style = {base, fill=white!30},
	process/.style = {base, minimum width=2.5cm, fill=white!15,
		font=\ttfamily},
	nonblock/.style = {base,draw=white ,fill=white!30},
}

\renewcommand*{\tableofcontents}{%
  	\begingroup
  	\tocsection
  	\tocfile{\contentsname}{toc}
  	\endgroup
}
\renewcommand*{\listoffigures}{%
  	\begingroup
  	\tocsection
  	\tocfile{\listfigurename}{lof}
  	\endgroup
}

\usepackage{tikz}
\usetikzlibrary{shapes,arrows}
\usepackage{caption}

\usepackage[english]{babel}
\usepackage[utf8]{inputenc}
\usepackage{algorithm}
\usepackage{algorithmicx}
\usepackage[noend]{algpseudocode}
\usepackage{hyperref}
 
\usepackage{subfiles}

\providecommand{\keywords}[1]
{
	\small	
	\textbf{{Keywords --- }} #1 
}

\begin{document}

\begin{titlepage}

{
    \begin{center}
    \raisebox{-1ex}{\includegraphics[scale=1.5]{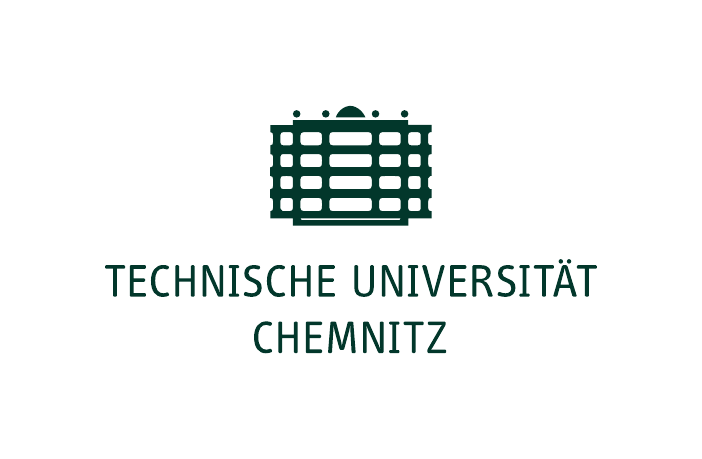}}\\
    \end{center}
    \vspace{0.5cm}
}

\begin{center}

\LARGE{\textbf{Master Thesis}}\\
\vspace{1cm}

\Large{\textbf{Evaluation of Robust Point Set Registration Applied to Automotive Doppler Radar}}\\ 
\vspace{0.3cm}
Faculty of Computer Science$ ^\mathtt{a}$.\\
\vspace{0.5cm}
Faculty of Electrical Engineering and Information Technology.\\
Chair of Automation Technology$ ^\mathtt{b}$.\\

\end{center}
\vspace{1cm}
{Submitted by: Karim Haggag}\\
Student ID: 462756 \\
Submission Date: 19.02.2020 \\
\vspace{0.1cm}\\
Supervising Tutor:\\ 
Prof. Dr.-Ing. Alejandro Masrur$ ^\mathtt{a}$\\
Prof. Dr.-Ing. Peter Protzel$ ^\mathtt{b}$\\
\vspace{0.1cm}\\
Further Supervisors \\
Dr.-Ing. Sven Lange$ ^\mathtt{b}$

\end{titlepage}

\addchap*{Abstract}

Point set registration is the process of finding the best alignment between two point sets, and it is a common task in different domains, especially in the automotive and mobile robotics domains. Lots of approaches are proposed in the literature, where the iterative closest point ICP is a well-known approach in this vein, which builds an explicit correspondence between both point sets to achieve the registration task. However, this work is interested in achieving the registration without building any explicit correspondence between both point sets, following a probabilistic framework.\\

The most critical task in point set registration is how to elaborate the cost function, which measures the distance between both point sets. The probabilistic framework includes two possible ways to build the cost function: The summing and the likelihood. The main focus of this work is to analyze and compare the behavior of both approaches. 
Therefore, a 1D synthetic scenario is used to build the cost function step by step, besides the estimation error. Finally, this work uses two data sets for evaluation: A 2D synthetic data set and a real data set. The evaluation process compares and analyzes the estimation error and estimated uncertainty. Thus, two different methods are used in the evaluation process: The normalized estimation error squared NEES and noncredibility index NCI. A 77 GHz automotive Doppler radar provides the real data set, and in the real evaluation, we evaluate the ego-motion estimation of a robot as an application for the registration.

\vspace*{2 cm}


\tableofcontents
\listoffigures

\chapter{Motivation}

Motion estimation is an essential task in mobile robotics. It aims to know how much the agent moves in its environment. Various sensors and methods aim to provide accurate information about the robot state, which can be categorized into two categories: Local and global. The global category estimates the motion based on absolute measurements, such as the global navigation satellite system (GNSS), active beacons, landmark navigation, or model matching. The local category tries to estimate the motion based on relative measurements, such as wheel odometry, an inertial navigation system (INS), or range sensors, like radar and LIDAR. Each method and sensor has limitations and drawbacks. Thus, sensor fusion is a useful way to optimize the motion state based on different sensors estimation. The motivation of this work is to estimate the motion state of the robot with its covariance based on an automotive Doppler radar to calculate a consistent input for subsequent sensor fusion algorithm.\\

Point set registration is a possible way to estimate the motion state from a range sensor's consecutive scans. Literature includes different algorithms for point set registration problems. The main difference between these algorithms is how to build the cost function. The cost function is the metric to measure the distance between the two-point sets. There are two approaches to address point set registration based on a probabilistic framework: the summing and the likelihood approach. Most researchers propose to use the summing approach; however, it is an approximated version from the likelihood approach. Therefore, this work analyzes the behavior of both approaches to point out why most of the researchers used the summing approach, and then it builds a robust cost function. Moreover, it provides a method to estimate the uncertainty for the estimated motion state.\\

Eventually, the work comes up with a robust cost function and stresses it with different test cases to evaluate its behavior. This work builds the cost function based on a probabilistic framework; thus, chapter 2 reviews some probability background. Then, chapter 3 explains the point set registration problem, and it introduces some typical approaches to solve it. Chapter 4 introduces the cost function and addresses each challenging aspect and builds the robust cost function step by step, with a 1-D scenario for elaboration. Chapter 5 addresses the estimation problem after building a robust cost function. The complete evaluation process for the robust cost function with a 2-D synthetic scenario with providing some credibility results is in chapter 6. Lastly, chapter 7 discusses the Ego-motion problem as an application for point set registration, with a real scenario.


\printindex

\chapter{Gaussian Probability Distribution}

Gaussian distribution is a common way to model a continuous random variable, where the variable can be a scalar or a vector, and the observations of that variable could be well distributed so one Gaussian distribution can model it, or poorly distributed. Hence, it needs a mixture of Gaussians to model it. Thus, this chapter explains some conventional Gaussian distributions models: The  univariate distribution,  the multivariate distribution, and the Gaussian Mixture Model (GMM)as well as illustrates different probability terms, such as the uncertainty or the covariance and the mean. 
\section{Univariate Normal Distribution}

The univariate normal distribution can be used to model a single random variable. Assume $x$ is a one-dimensional random variable and for N observations, as shown in Fig. \ref{fig:uni_1}. The normal distribution for $x$ written as $\mathcal{N}(\mu,\sigma^2)$, where $\mu$ and $\sigma$ are the model parameters, the mean and the standard deviation, respectively.\\

The Maximum Likelihood Estimation (MLE) is one possible way to estimate the parameters ($\mu , \sigma$) from observed data $\{x_i\}$, utilizing the equation below:

\begin{equation}
 \hat{\mu},\hat{\sigma} = \argmax_{\mu , \sigma}p(\{x_i\}|\mu,\sigma)
\end{equation}

The complete derivation of MLE can be found in ~\cite[ch. 2]{nasrabadi2007pattern}. The MLE solution to obtain the parameters is as follows:

\begin{equation}
\hat{\mu} = \frac{1}{N} \sum_{i=1}^{N} x_i
\end{equation}
\begin{equation}
\hat{\sigma}^2 =  \frac{1}{N} \sum_{i=1}^{N} (x_i - \hat{\mu} )^2
\end{equation}

Fig. \ref{fig:uni_1} includes the Gaussian distribution that fits this pseudo scenario, plotted on red. Thus, the probability density function for $\hat{x}$ given as follows:
\begin{figure}[htb!]
	\centering
	\includegraphics[width=0.6\textwidth]{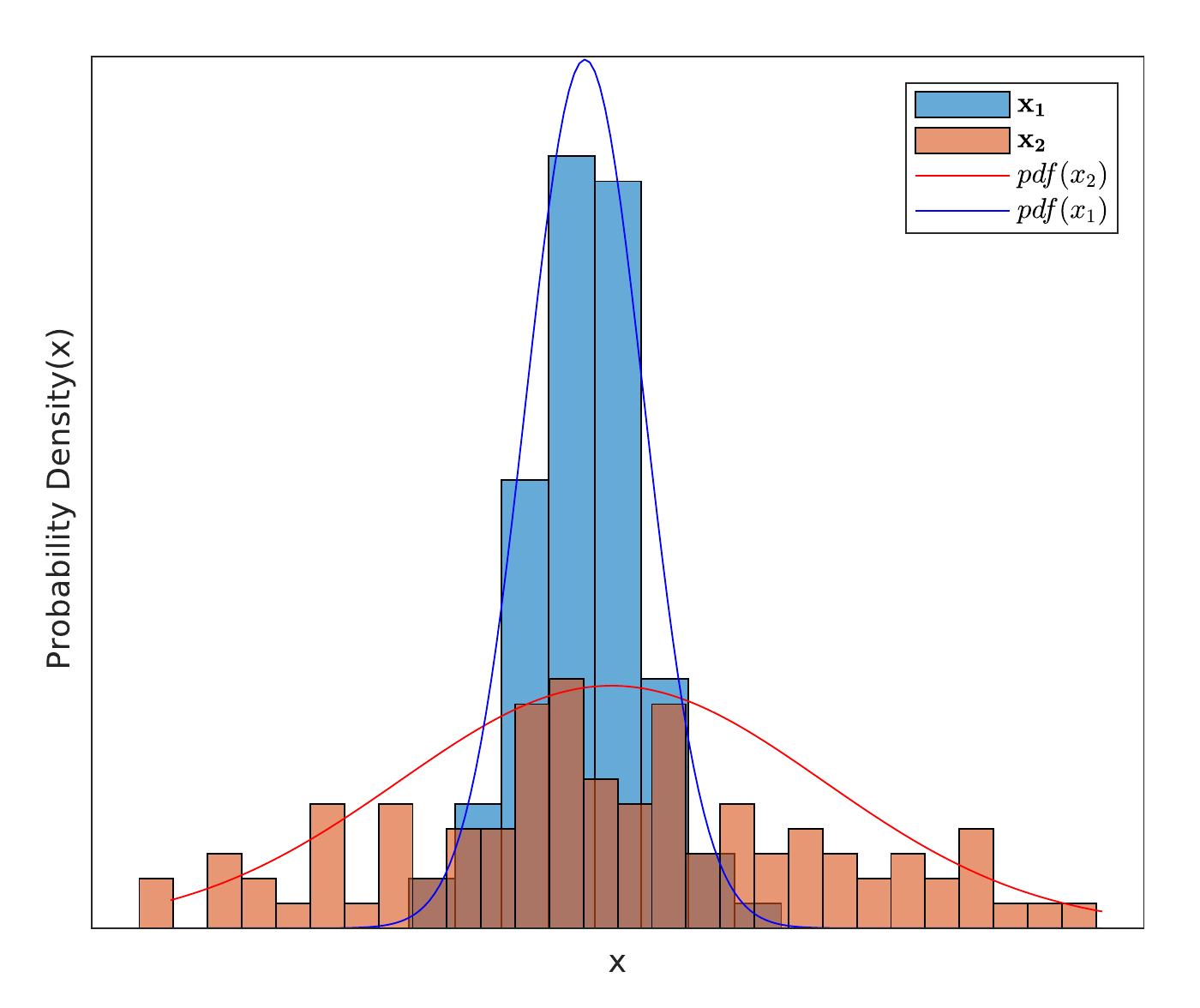}
	\caption{A pseudo univariate scenario, where it shows how is the shape of the distribution behaves with the random event ${\mathbf{x}}$.}
	\label{fig:uni_1}
\end{figure}

\begin{equation}
\mathcal{N}(\hat{x}|\mu,\sigma) = \frac{1}{\sqrt{(2\pi)(\sigma^2)}} exp {[-\frac{(\hat{x} - \mu)^2}{2\sigma^2} ]} .
\end{equation}

where is: $\mu$ the mean, $\sigma$: the standard deviation and $\sigma^2$: the variance. \\

\begin{figure}[htb!]
	\centering
	\includegraphics[width=0.6\textwidth]{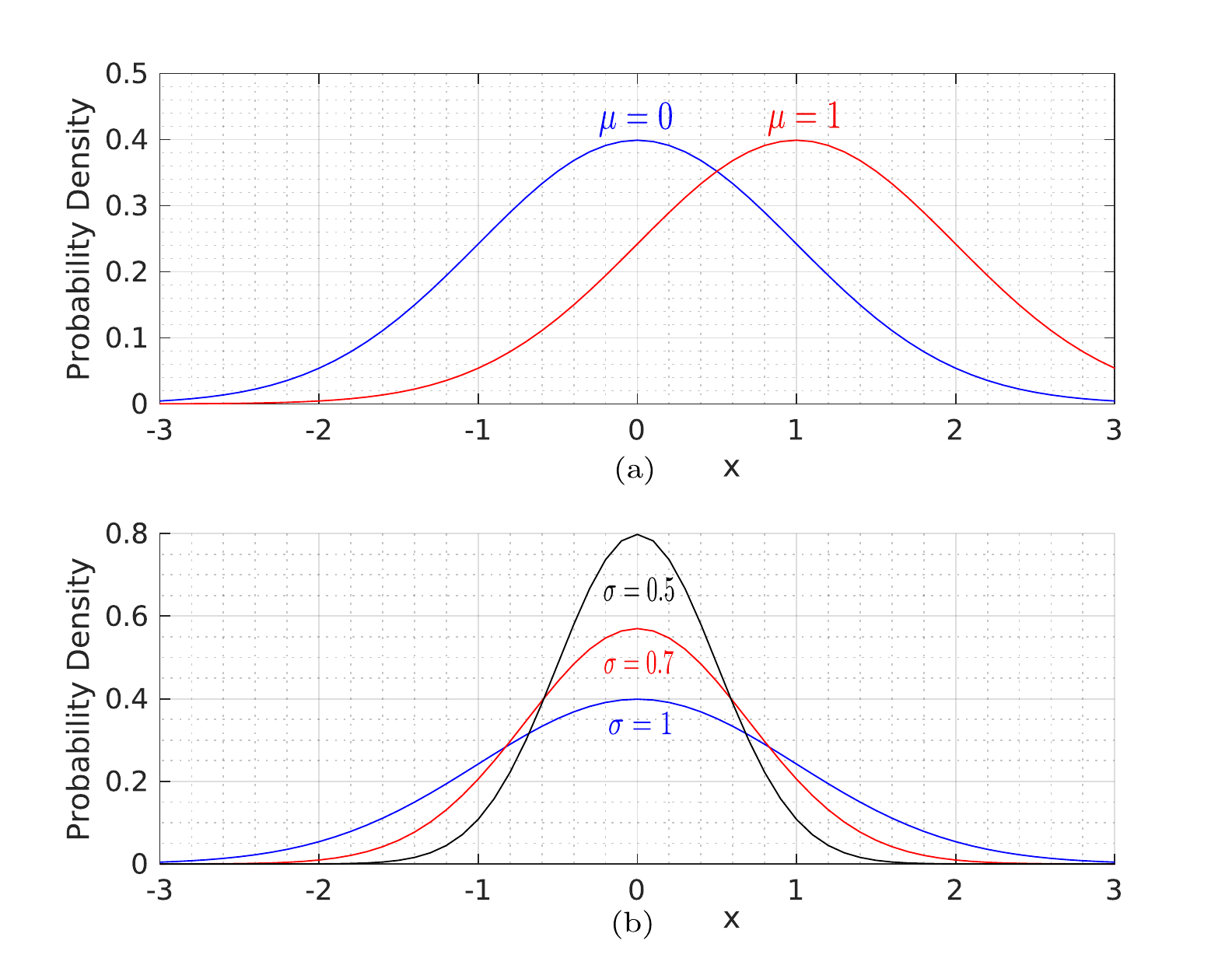}
	\caption{Univariate PDF plot for $x$ with different scenarios for $\mu$ and $\sigma$:(a) The center of the plot changes with the changing of the mean $\mu$ .(b)  The spread of the curve changes with the changing of std-deviation $\sigma$.
		 }\label{fig:uni_2} 
\end{figure}

\vspace*{1cm}

Fig. \ref{fig:uni_2} exhibits some properties of the distribution of $\mathcal{N}(\mu,\sigma^2)$. Where the mean $\mu$ represents the center of the distribution, and the change in the mean will, in turn, change the center of the distribution. The variance $\sigma^2$ represents the spread of the distribution. The spread of the curve is directly proportional to the variance. The variance indicates the uncertainty, which means, the higher the variance, the higher the uncertainty is.\\

\section{Multivariate Normal Distribution}

\begin{figure}[htb!]
\centering
\subfigure[\textbf{ 2-d scenario  \textbf{x}}]{\label{fig:a}\includegraphics[width=60mm]{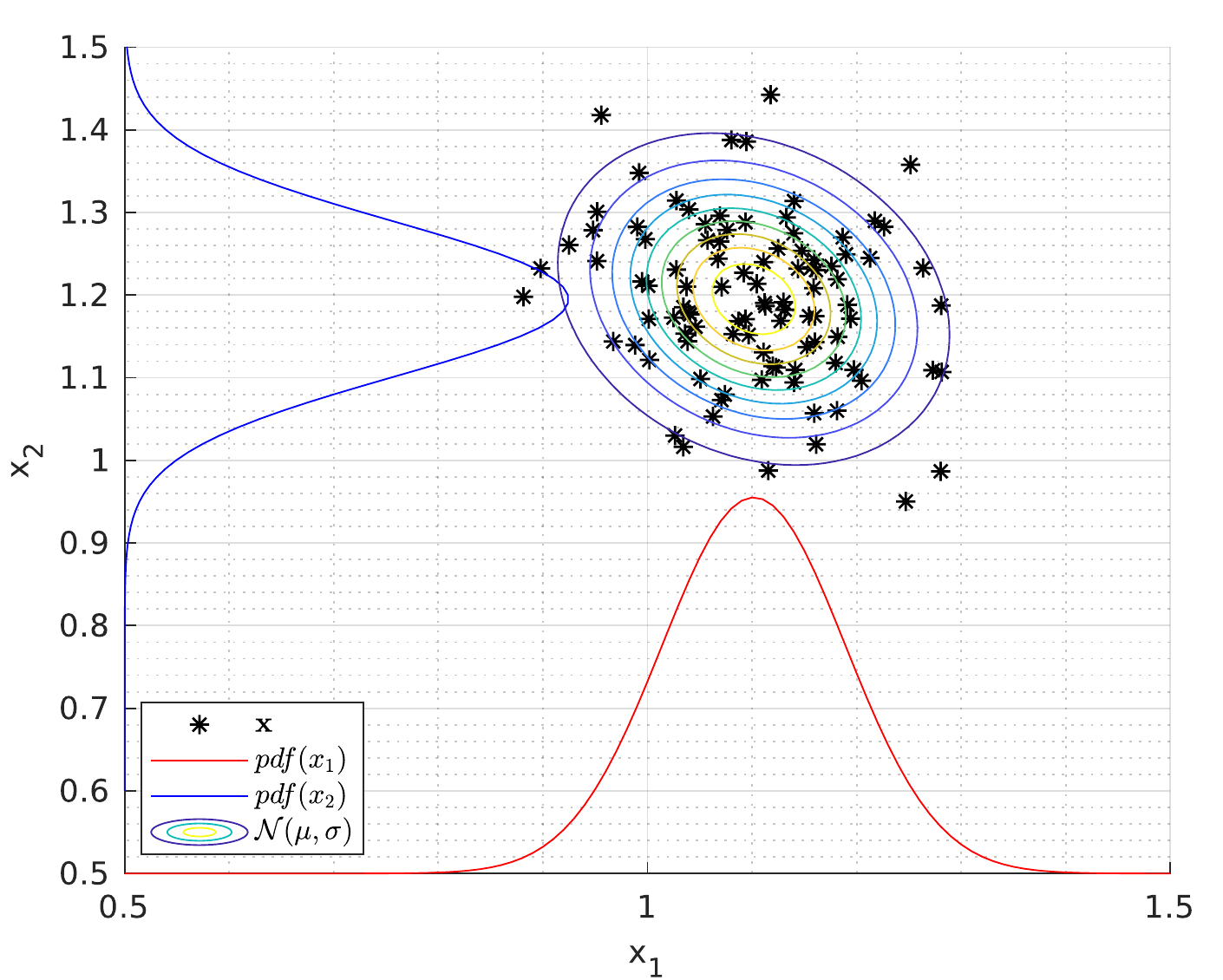}}
\subfigure[\textbf{Probability Density Function, top view }]{\label{fig:b}\includegraphics[width=60mm]{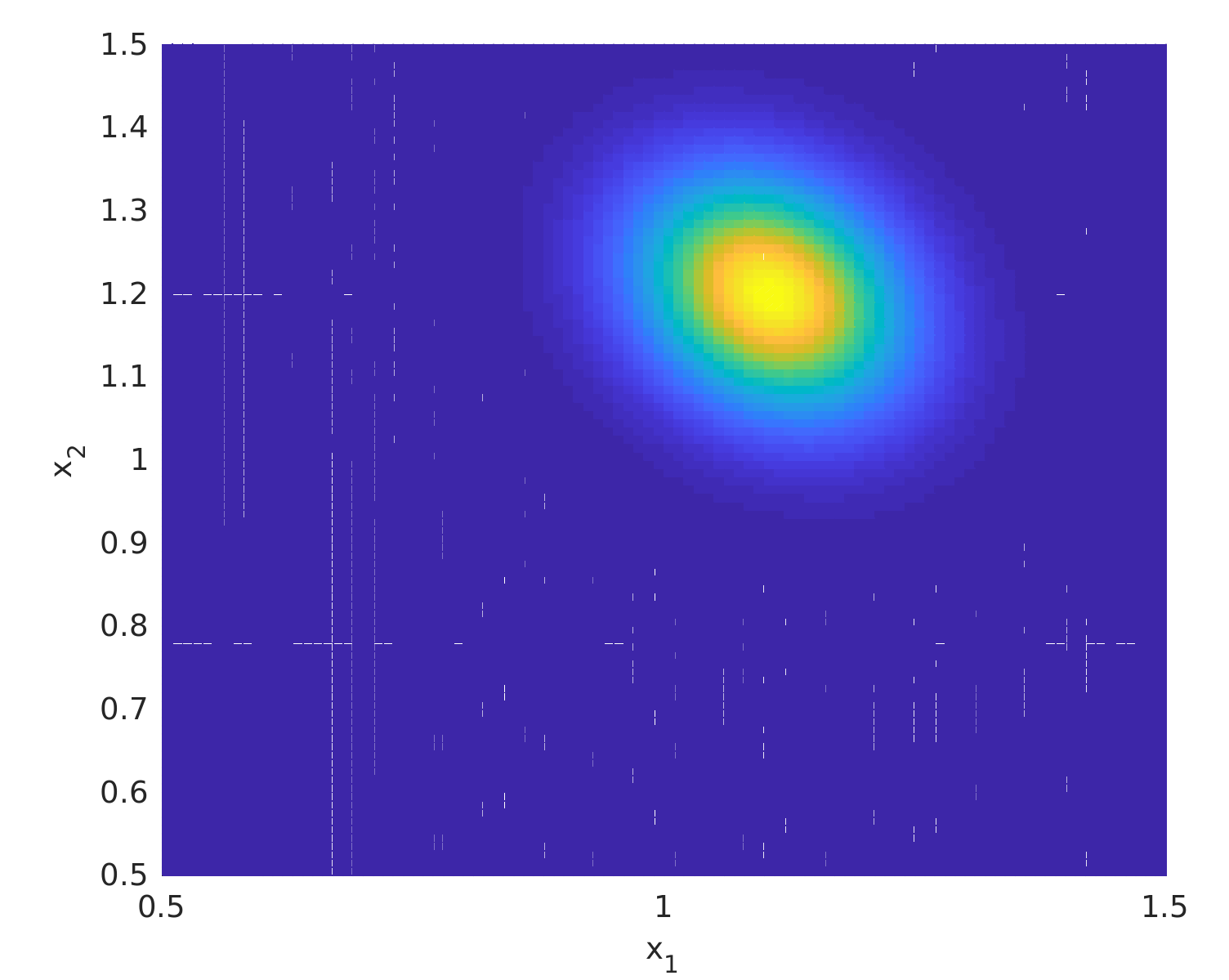}}
\subfigure[\textbf{Probability Density Function}]{\label{fig:b}\includegraphics[width=60mm]{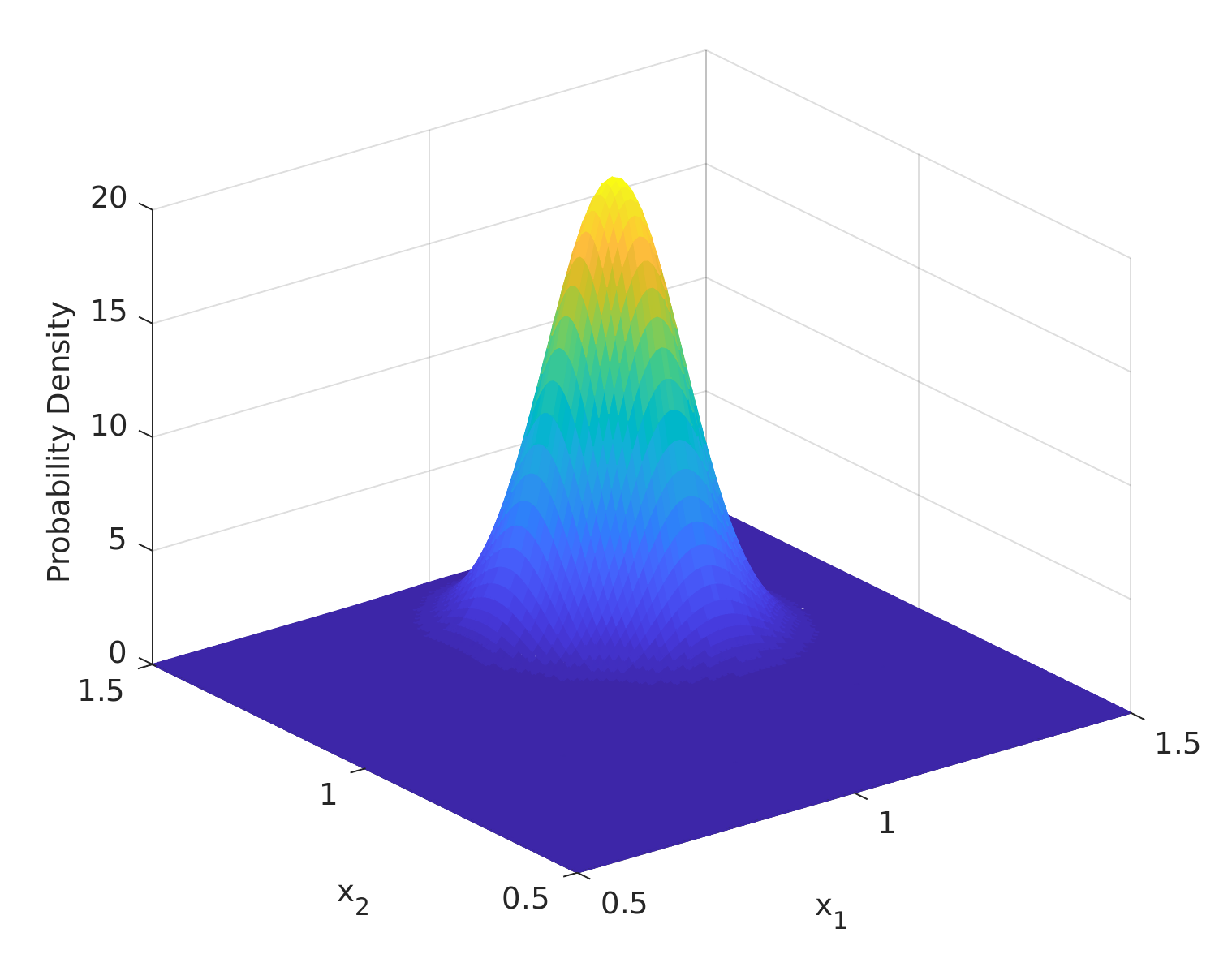}}
\caption{A 2-D example shows the pattern of the distribution in each dimension, and the contour plot represents the same information but in two dominations. Moreover, it shows how the shape of the probability density function in two dimensions.}
\label{fig:Mult_1}
\end{figure}

Assuming that, $\boldsymbol{\mathrm{x}}$ is in a higher dimension $\{n\}$ rather than a scalar, so the univariate normal distribution cannot be applied. Where each dimension has a univariates distribution shown in Fig. \ref{fig:Mult_1} (a). Therefore, the multivariate Gaussian distribution is able to represent the higher dimension variables $\boldsymbol{\mathrm{x}} $ as $\mathcal{N}(\boldsymbol{\mu},\boldsymbol{\Sigma})$, where $\boldsymbol{\mu}$ is vector with n-by-1 and the uncertainty is a covariance matrix with n-by-n. Fig. \ref{fig:Mult_1} shows an pseudo example, where $n \ = 2$, the model parameters written as:

\begin{center}
$ \boldsymbol{\mu} = [\mu_{x_1} \ \ \mu_{x_2} ] $
\end{center}
\begin{center}
$ \boldsymbol{\Sigma} =  \begin{bmatrix}
    \sigma_{x_1}^2& \sigma_{x_1 x_2}  \\
    \sigma_{x_2 x_1} & \sigma_{x_2}^2
  \end{bmatrix}
$
\end{center}.

\begin{figure}[h!]
	\centering
	\includegraphics[width=0.7\textwidth]{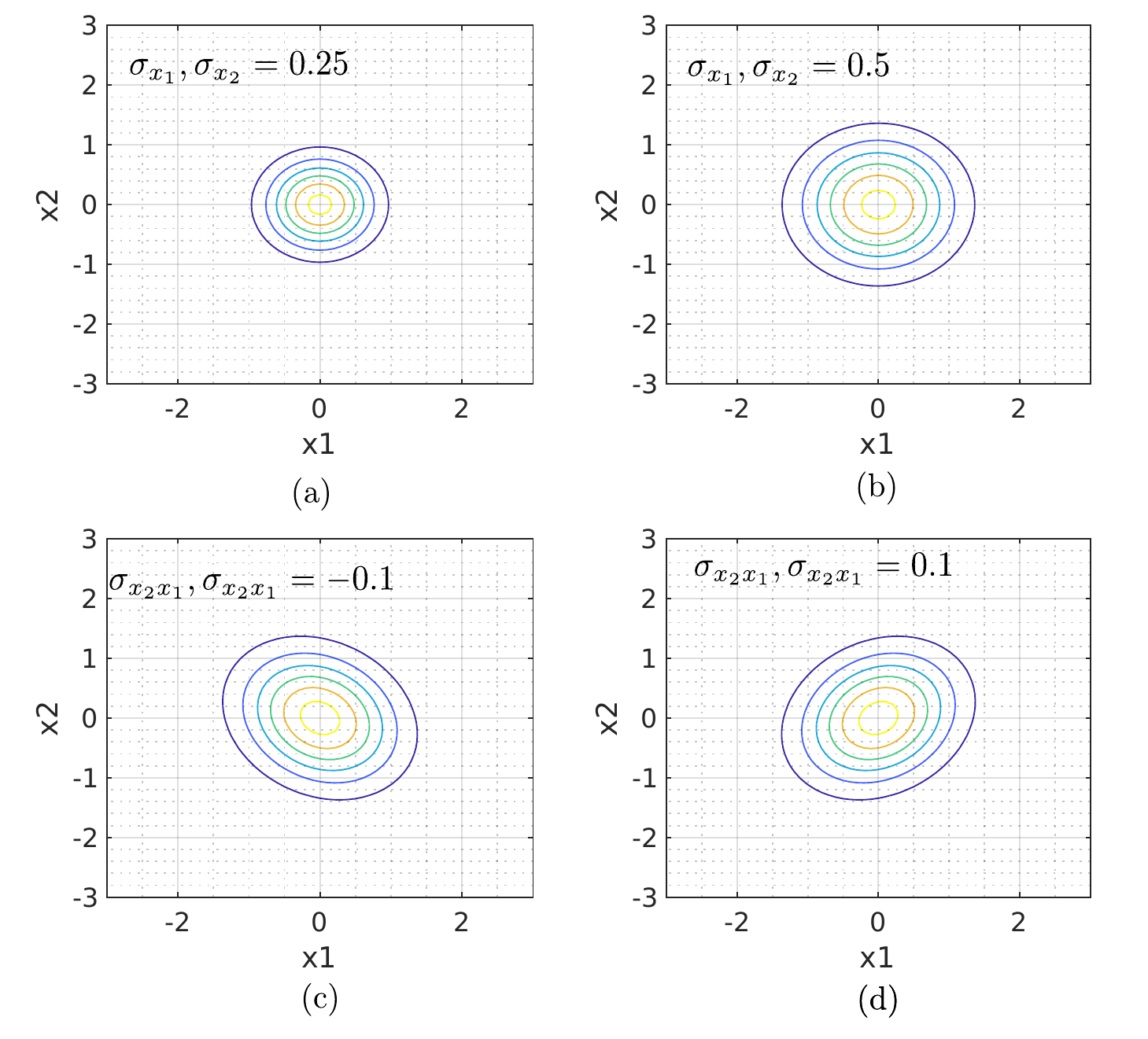}
	\caption{Contour plots of constant PDF for 2d data $\boldsymbol{\mathrm{x}}$, with different covariance matrix scenarios:(a) represent zero correlation and low variance  (b) represent zero correlation and high variance, (c) negative correlation (d) positive correlation.}
\label{Mult_2}
\end{figure}

\vspace{2cm}

Moreover, the distribution is represented using the contour plot, as depicted in Fig. \ref{Mult_2} (a) and (b). The covariance matrix composites of two components: the diagonal, which represents the variance in each dimension and the off-diagonal, which represents the correlation between the two dimensions. When the variances increase the uncertainties increase, as shown in subplot (a) and (b) in Fig. \ref{Mult_2}. Moreover, at zero correlation and equal variances, the contour turns to be a circle centered around the mean. Subplot(c) shows a negative correlation which means when $x_1$ increase $x_2$ decrease and vice-versa. However, subplot(d) shows a positive correlation.\\

Like the univariate distribution, the maximum likelihood estimation is one possibility to estimate the model parameters from the observed data $\boldsymbol{\mathrm\{\mathrm{x}_i\}} $, as follows :
\begin{equation}
\boldsymbol{\hat{\mu}} = \frac{1}{N} \sum_{i=1}^{N} \boldsymbol{\mathrm{x}_i}
\end{equation}

\begin{equation}
\boldsymbol{\hat{\Sigma}} =  \frac{1}{N} \sum_{i=1}^{N} (\boldsymbol{\mathrm{x}_i} - \boldsymbol{\hat{\mu}} )(\boldsymbol{\mathrm{x}_i} - \boldsymbol{\hat{\mu}} )^T
\end{equation}\\

Lastly, Fig. \ref{Mult_2} (c) represents the probability density function for a random variable $\boldsymbol{\mathrm{x}}$, which can be written as:

\begin{equation}
\mathcal{N}(\boldsymbol{\mathrm{\hat{x}}} |\boldsymbol{\mu},\boldsymbol{\Sigma}) = \frac{1}{\sqrt{(2\pi)^d  det(\boldsymbol{\Sigma})}}  exp {\{- \frac{1}{2} (\boldsymbol{\mathrm{\hat{x}}} - \boldsymbol{\mu} )^T {\boldsymbol{\Sigma}}^{-1} (\boldsymbol{\mathrm{\hat{x}}} - \boldsymbol{\mu})\}}
\end{equation}


\section{Gaussian Mixture Model}

\begin{figure}[htb!]
\centering
\subfigure[\textbf{ One distribution }]{\label{fig:a}\includegraphics[width=50mm]{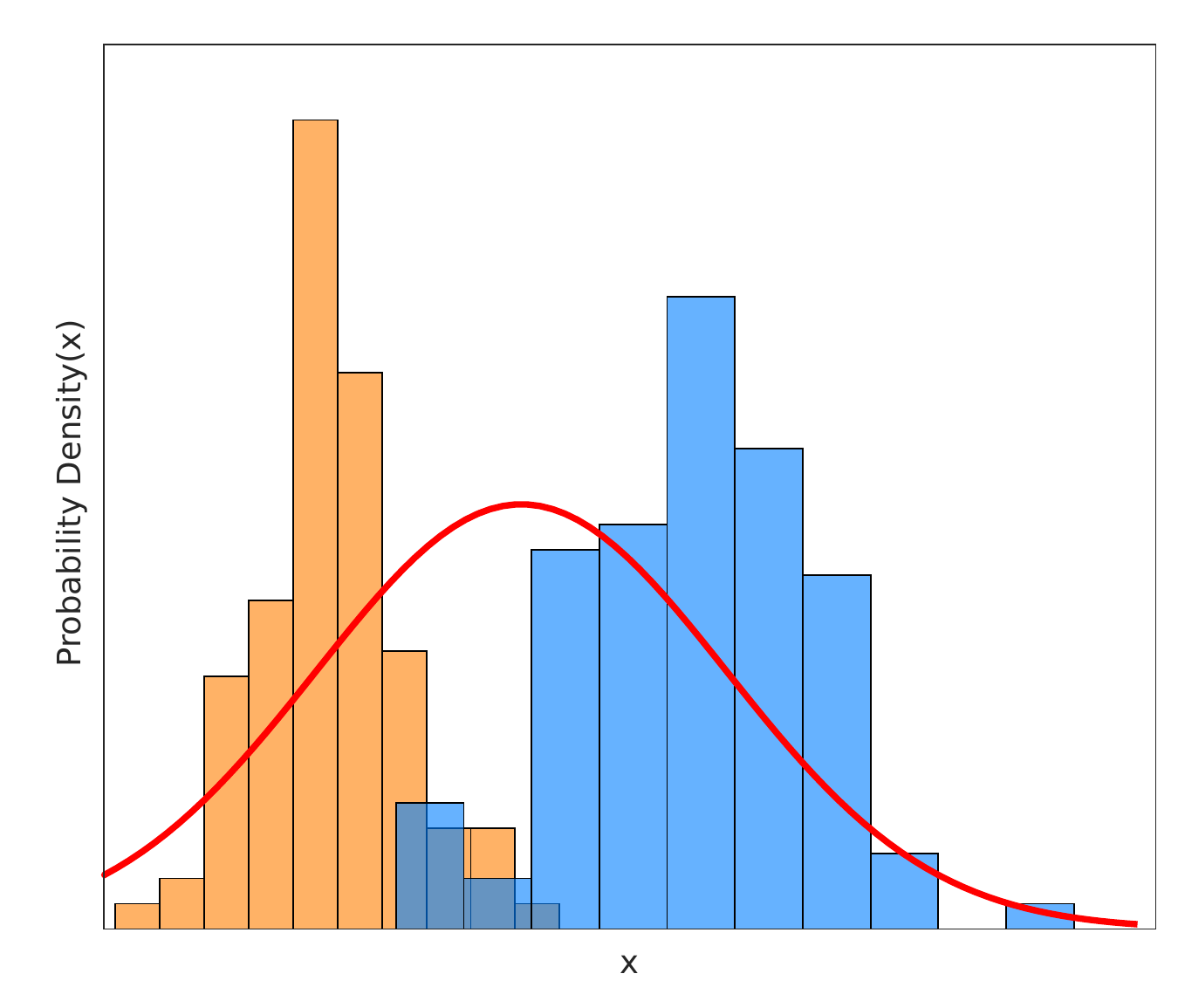}}
\subfigure[\textbf{Two distribution}]{\label{fig:b}\includegraphics[width=50mm]{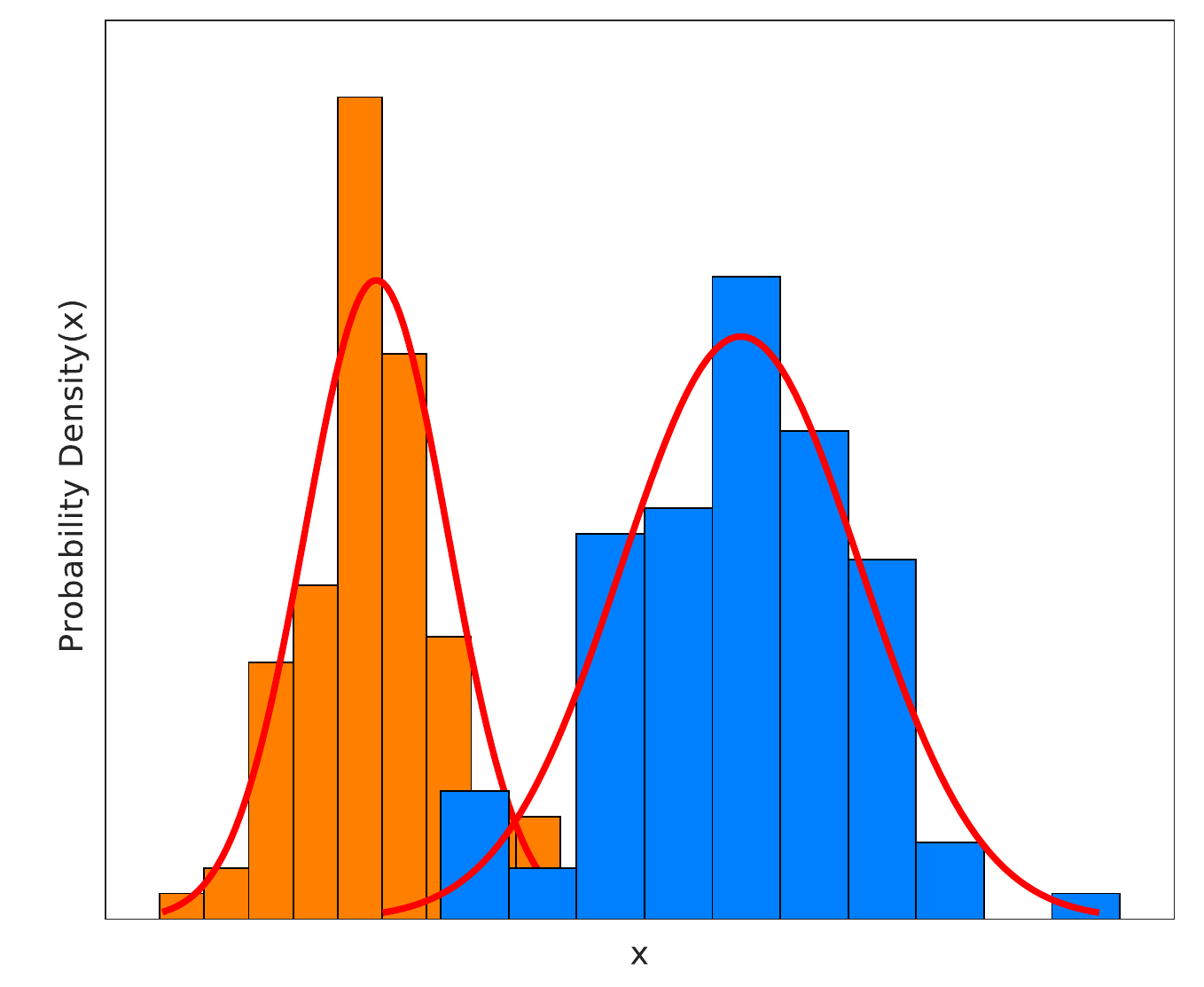}}
\subfigure[\textbf{GMM distribution}]{\label{fig:b}\includegraphics[width=60mm]{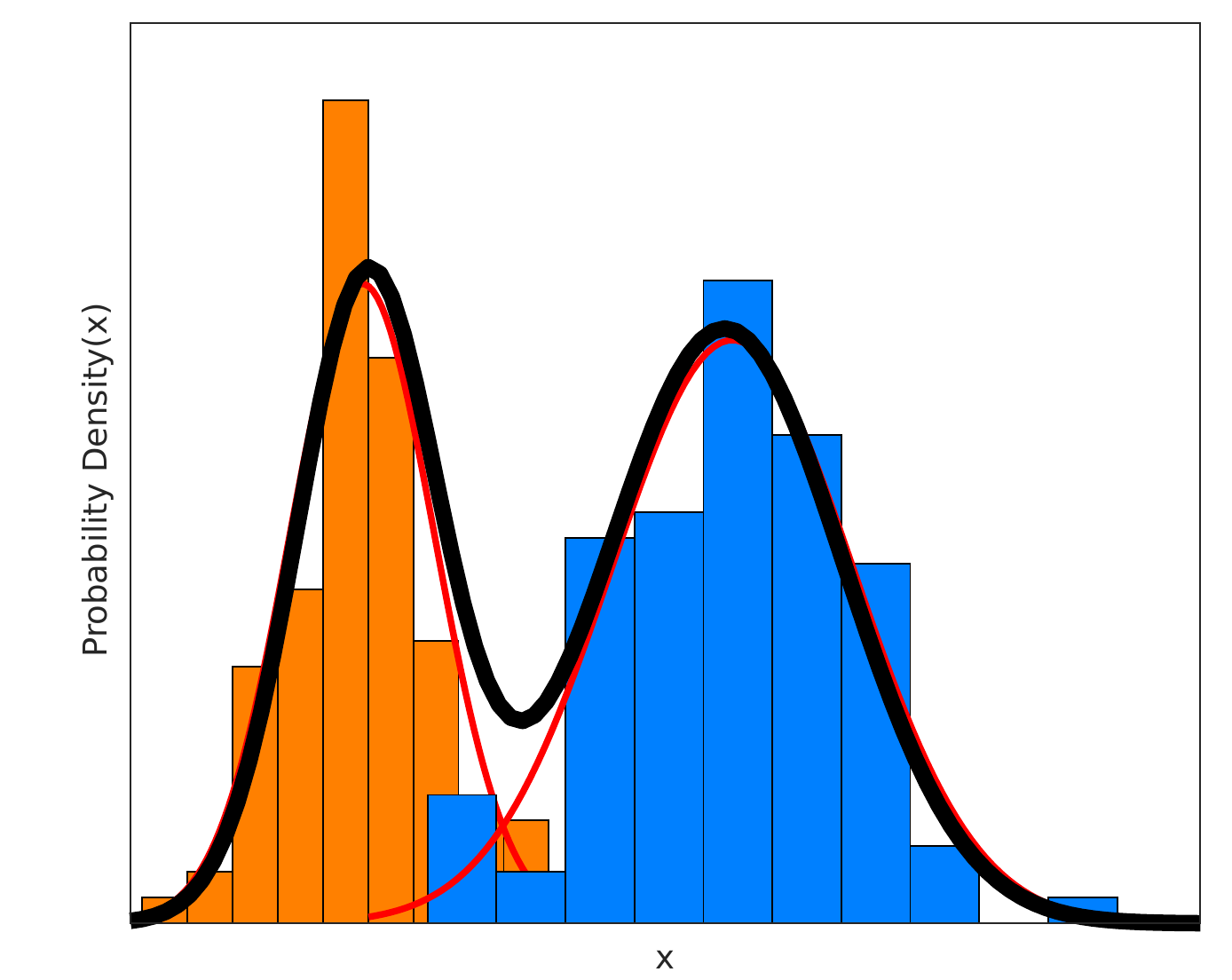}}
\caption{This scenario shows representing the data as a Gaussian Mixture Model (GMM) is better than using one component (a). However, using two Gaussians is better,  where each one called a component (b). Thus, GMM shows a better representation of the event (c). }
\label{fig:gmm_1}
\end{figure}

In some situations, using only one Gaussian distribution is not optimal to represent the observations, as shown in Fig. \ref{fig:gmm_1} (a). Rather, using more than one is better in representing the observations, as in Fig. \ref{fig:gmm_1} (b). Eventually, we need only one model to represent the observations; Gaussian Mixture Model (GMM) is one possibility to do that representation. GMM is basically a weighted sum of many Gaussian, as shown in Fig. \ref{fig:gmm_1} (c). For N Gaussian components, the probability density function for the Gaussian Mixture Model can be written as follows:

\begin{equation}
p_{GMM}(\boldsymbol{\mathrm{x}})=  \sum_{i=1}^{N} w_i  \mathcal{N}(\boldsymbol{\mathrm{x}}|\boldsymbol{\mu_i},\boldsymbol{\Sigma_i})
\end{equation}

where, $w_i$ represents the weight of each component $\mathcal{N}(\boldsymbol{\mathrm{x}}|\boldsymbol{\mu_i},\boldsymbol{\Sigma_i})$, and for equal components $ w_i = \frac{1}{{|N|}} $. 

The Maximum Likelihood Estimation (MLE) has no closed analytical solution to estimate the GMM parameters~\cite{nasrabadi2007pattern}. Nevertheless, another possibility to estimate GMM parameters is to use an iterative method, such as the Expectation-Maximization (EM) method. The concept behind the EM method algorithm is to introduce a latent variable $z$ and initialize the parameters that need to be estimated. Then, iterate between calculating the latent variable based on the current value of the parameters and estimating the parameters from the current value of the latent variable until it is converged, as shown in Fig. \ref{fig:em_1}.

\begin{center}
\resizebox {1.2 \textwidth} {!} {
\begin{tikzpicture}[node distance=1.5cm,
every node/.style={fill=white, font=\sffamily}, align=center]

\node (c2)     [activityStarts]              {Intialize $\boldsymbol{\mu}$ and $\boldsymbol{\Sigma}$};
\node (c3)     [nonblock, right of=c2, xshift=4cm] {};
\node (c1)     [nonblock, left of=c2, xshift=-4cm] {};

\node (c4)     [nonblock, below of=c1  ]          { };
\node (c5)     [base, below of=c2, yshift=-0.1cm ]     {Introduce a latent varriable $z^k_i$};
\node (c6)     [nonblock, below of=c3 ]          { };

\node (c7)     [nonblock, below of=c4 ]          { };
\node (c8)     [base, below of=c5, yshift=-0.1cm  ]          { $z_k^i =\frac{\mathcal{N}_k(x _i|\mu_k,\Sigma_k)}{\sum_{k=1}^{K} \mathcal{N}_k(x_i|\mu_k,\Sigma_k)}$ };
\node (c9)     [nonblock, below of=c6 ] { };

\node (c10)     [nonblock, below of=c7 ]          { };
\node (c11)     [base, below of=c8, yshift=-0.5cm ]          { $\boldsymbol{\hat{\mu_k}} = \frac{1}{z_k}\sum_{i=1}^{N} z_k^i x_i$ \\
$\boldsymbol{\hat{\Sigma_k}} = \frac{1}{z_k} \sum_{i=1}^{N} z_k^i(x_i-\hat{\mu_k})(x_i-\hat{\mu_k})^T$};
\node (c12)     [nonblock, below of=c9 ]          { };

\node (c13)     [nonblock, below of=c10]          { };
\node (c14)     [base, below of=c11, yshift=-0.5cm ]    { Stop if converged };
\node (c15)     [nonblock, below of=c12]          { };

\node (d10)     [nonblock, below of=c13 ]          { };
\node (d11)     [nonblock, below of=c14 ]          {$\boldsymbol{\hat{\mu_k}}$ , $\boldsymbol{\hat{\Sigma_k}}$ };
\node (d12)     [nonblock, below of=c15 ]          { };

\draw[->]      (c2) -- (c5);
\draw[->]      (c5) -- (c8);
\draw[->]      (c8) -- ++(-4,0) -- ++(0,-2)  -- (c11);
\draw[->]      (c11)  -- ++(4,0) -- ++(0,2)  --  (c8);
\draw[->]      (c11) -- (c14);
\draw[->]      (c14) -- (d11);

\end{tikzpicture}
}

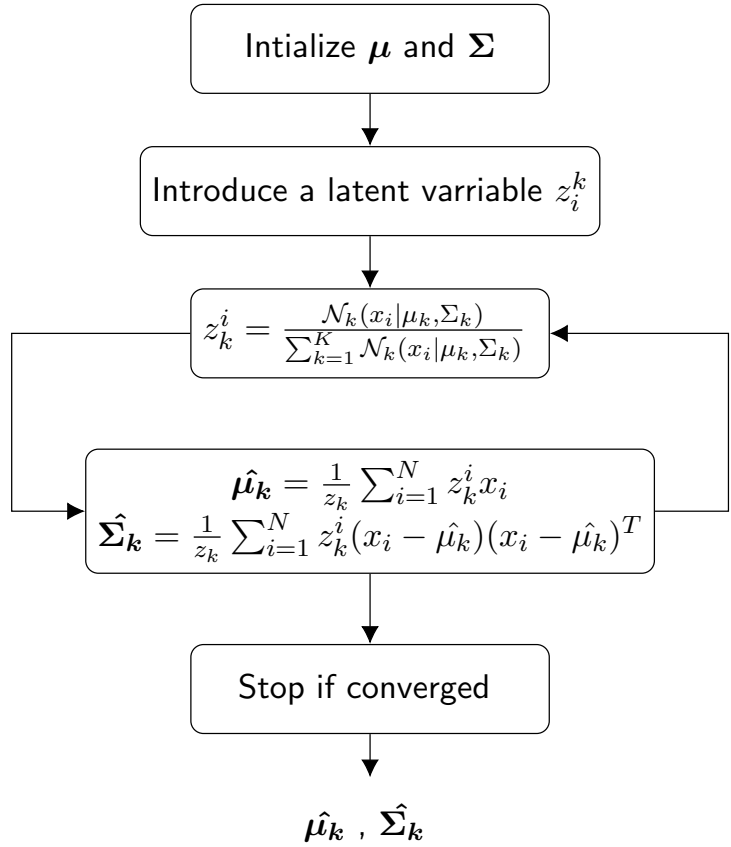
\captionof{figure}{ EM algorithm framework \\}
\label{fig:em_1}
\end{center}


\printindex

\chapter{Point Set Registration}

\begin{figure*}[h]
\centering
\includegraphics[width=0.8\textwidth]{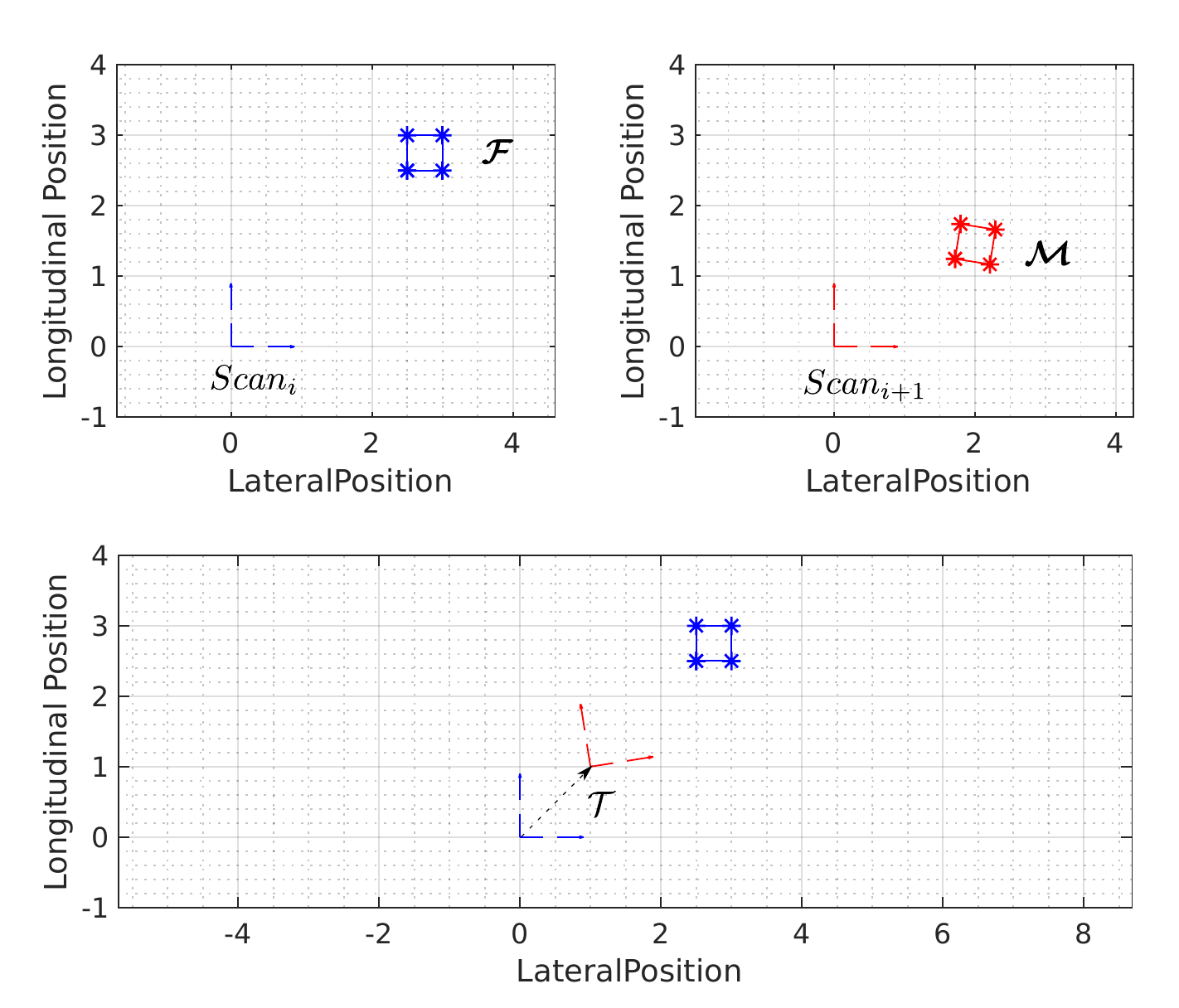}
\caption{ Simple point set registration scenario, where $\mathcal{F}$ represents the previous point set, $\mathcal{M}$ represents the current point set. The point set registration task optimizes ${T}$ to align $\mathcal{F}$ and $\mathcal{M}$ }
\label{fig:psr_1}
\end{figure*}

"Point set registration", "data association", "point matching", "finding correspondence" are different terms for the same problem depending on the chosen literature. The point set registration is a fundamental task in different applications such as computer vision, pattern recognition, and image processing. Point set registration is a process of aligning two consecutive point sets by finding the spatial transformation ~\cite{pomerleau2015review}. Versatile sensors and methods have been used to address this problem, for example, laser scanner (LIDAR) and radar. Fig. \ref{fig:psr_1} represents a simplified explicative example, where the process optimizes $\boldsymbol{{T}}$ between the two frames through aligning the two-point sets.

Mathematically, the problem can be represented as follows: Let $\{ \boldsymbol{\mathcal{M}} , \boldsymbol{\mathcal{F}}\}$ are two point sets to be aligned, where $\boldsymbol{\mathcal{M}}$ is the current point set and $\boldsymbol{ \mathcal{F}}$ is the previous point set. The registration task aims to minimize the error between the transformed current point set and the previous one to get the best estimate of the transformation matrix $\boldsymbol{{T}}$ :

\begin{equation}
{{\hat{T}}} = \argminA_{{T}} (error({{T}}({\mathcal{M}}),{\mathcal{F}})
\end{equation}

Where ${{\hat{T}}}$ indicates the best estimate for the transformation matrix. The registration methods can be sorted into two categories: The local and the global category. The local category depends on a good initial guess for $\boldsymbol{{T}}$. The global one relies on the geometrical features, which are uniquely matched~\cite{stoyanov2012point}. Besides, the transformation itself could be either rigid or non-rigid. This chapter addresses two different methods to address the point set registration: Iterative closest point \textbf{ICP} and point set registration based on the normal distribution transformation \textbf{NDT}. Moreover, the following chapter discusses a third point set registration method based on gaussian mixture model \textbf{GMM}.

 \section{Iterative Closest Point ICP }

Iterative closest point is a well-known algorithm, and it is a member of the local category. ICP introduced by Besl and Mckay in~\cite{besl1992method}. Find the correspondencesis is the main task using nearest neighbor method.
Then, compute the distance between these point pair to assemble the cost function.
In mathematical terms, for every point $\boldsymbol{{m}}$ in  $ \boldsymbol{\{ \mathcal{M}\}}$, ICP finds a correspondence point $\boldsymbol{{f}}$ in $\boldsymbol{\{\mathcal{F}\}}$ as:

\begin{equation}
 match ({T}(\mathcal{M}),\mathcal{F}) = \{({f},{m}) : {{m}} \in {\mathcal{M}} , {{f}} \in {\mathcal{F}}\}
\end{equation}
Then, the error function can be written as follows :
 \begin{equation}
error({T}(\mathcal{M}),\mathcal{F}) = \sum\limits_{(\mathrm{s},\mathrm{m})\in {match}} \mathrm{d}({{T}({m})} , {{f}})
\end{equation}

Where, $\mathrm{d}({{T}({m})} , {{f}})$ represents the distance between ${m}_i$ and ${f}_i$. ICP steps can be summarized  as shown in Fig. \ref{fig:icp_1}. However, the convergence properties of the standard ICP are feeble. Over the last decades, several researchers invested in this direction to introduce an efficient and robust ICP. Therefore, new algorithms emerged, for example, point-to-line, point-to-plane and plane-to-plane~\cite{segal2009generalized}.

\begin{center}
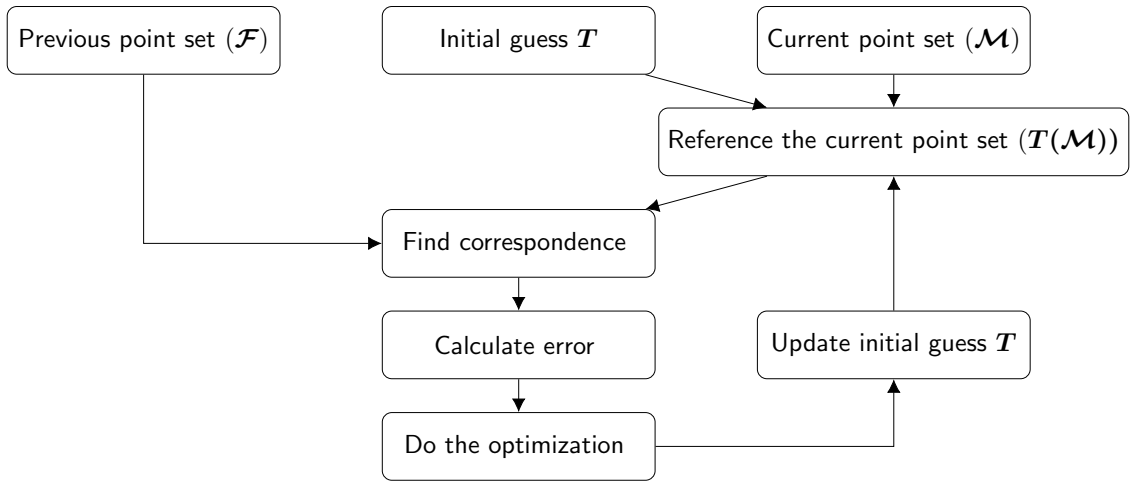

\resizebox {1\textwidth} {!} {
\begin{tikzpicture}[node distance=1.5cm,
every node/.style={fill=white, font=\sffamily}, align=center]
\node (c5)    [activityStarts]                {Initial guess $\boldsymbol {{T}}$};
\node (c6)     [base, right of=c5, xshift=4cm] {Current point set $(\boldsymbol{\mathcal{M}})$};
\node (c4)     [base, left of=c5, xshift=-4cm] {Previous point set $(\boldsymbol{\mathcal{F}})$};

\node (c7)     [nonblock, below of=c4 ]          { };
\node (c8)     [nonblock, below of=c5 ]          { };
\node (c9)     [base , below of=c6 ] {Reference the current point set $(\boldsymbol{{T}({\mathcal{M}}))}$};

\node (c13)     [nonblock, below of=c7]          { };
\node (c14)     [base, below of=c8]           { Find correspondence };
\node (c15)     [nonblock, below of=c9]          { };

\node (c19)     [nonblock, below of=c13]          { };
\node (c20)     [base,  below of=c14]           { Calculate error  };
\node (c21)     [base, below of=c15]          {Update initial guess $\boldsymbol {{T}}$};

\node (c25)     [nonblock, below of=c19]          {  };
\node (c26)     [base,  below of=c20]           { Do the optimization };
\node (c27)     [nonblock, below of=c21]          { };

\draw[->]      (c6) -- (c9);
\draw[->]      (c9) -- (c14);
\draw[->]      (c4) |- (c14);
\draw[->]      (c14) -- (c20);
\draw[->]      (c20) -- (c26);
\draw[->]      (c5) -- (c9);
\draw[->]      (c21) -- (c9);
\draw[->]      (c26) -| (c21);

\end{tikzpicture}
}
\captionof{figure}{ Basic ICP framework \\}
\label{fig:icp_1}
\end{center}

\section{Point Set Registration based on NDT }

\begin{figure*}[h]
\centering
\includegraphics[width=0.8\textwidth]{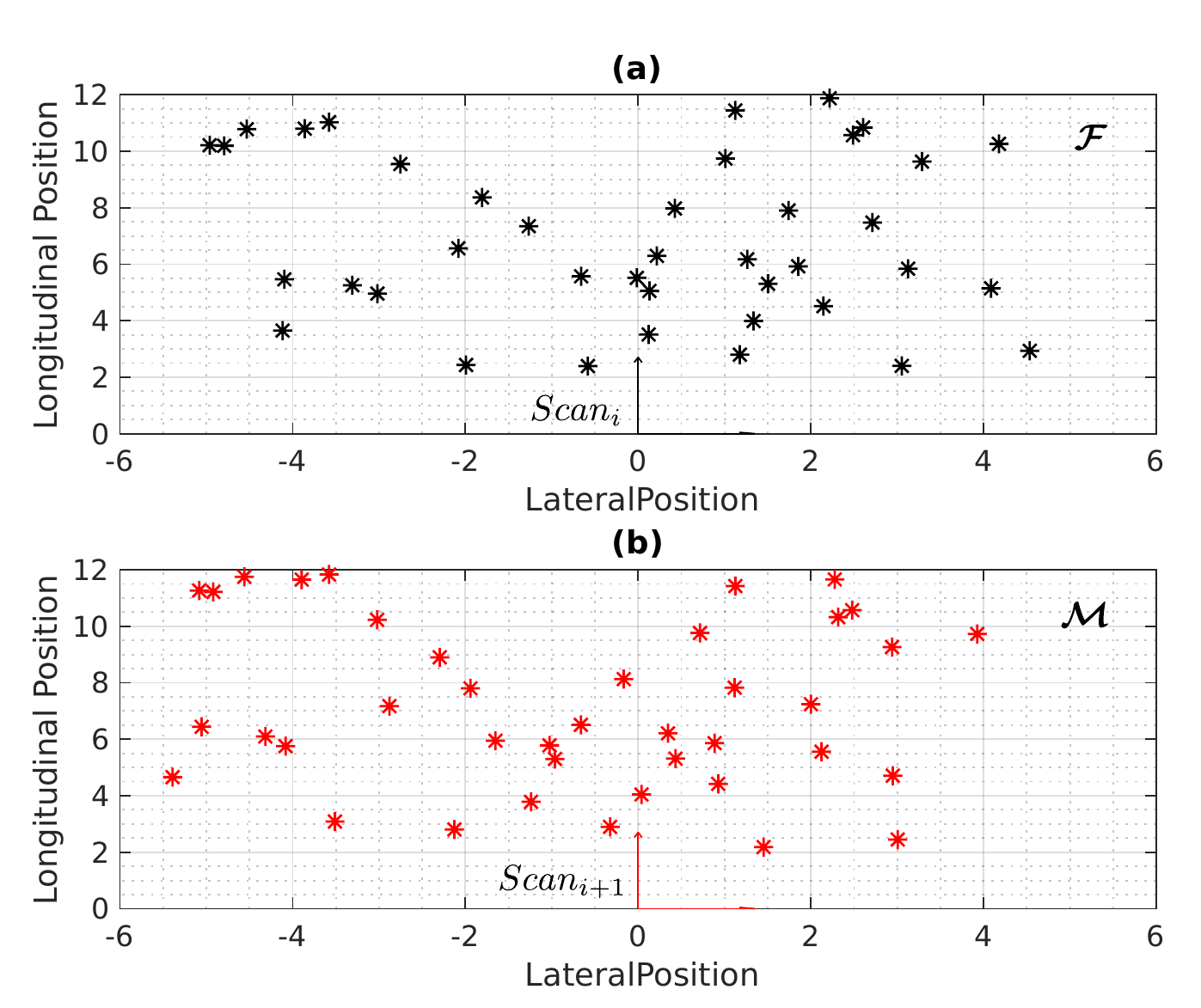}
\caption{ A synthetic scenario (a) The previous point set representation. (b) The current point set representation, which is relatively translated and rotated from the previous point set.}
\label{ndt_1}
\end{figure*}

The previous section introduces the basic ICP, where the cost function is constructed based on the euclidean distance between the point pair. However, this section presents a probabilistic approach to build the cost function. Bieber in ~\cite{biber2003normal} introduced the normal distribution transformation (NDT) as a possible way to represent the problem in a probabilistic model. NDT is a surface representation method. It is similar to an occupancy grid map, by subdividing a 2D plane into cells, where each cell has a probability distribution function depending on how many points are located in the cell. However, in an occupancy grid map, the cell is occupied or not~\cite{biber2003normal}. Two methods are introduced to achieve the point set registration based on NDT representation: Point-to-distribution and distribution-to-distribution. For a clear illustration, a synthetic scenario is created shown in Fig. \ref{ndt_1}. The scenario includes two-point sets that have a relative translation along the x-axis and rotation around the z-axis.\\

\begin{figure*}[h]
\centering
\includegraphics[width=0.8\textwidth]{{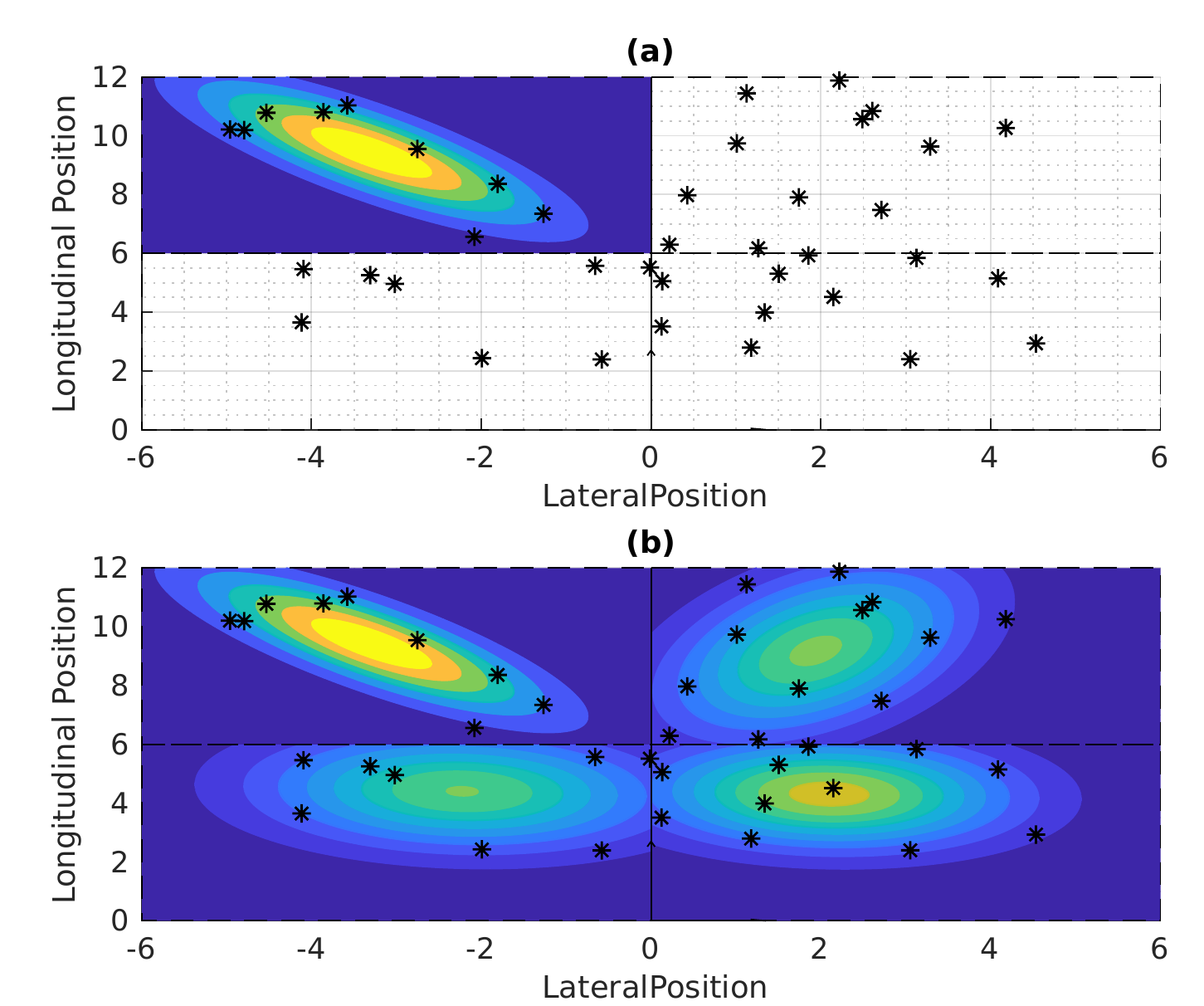}}
\caption{ A simple NDT example (a) represent the surface sectioning and build NDT for one cell. (b) The NDT exhibition for the previous point set.}
\label{ndt_2}
\end{figure*}

NDT is a representation method only, which means it can not align two-point sets by itself. Fig. \ref{ndt_2} depicts the NDT represenation for the previous point set. The main idea is, to divide the space into equal cells and build the probability distribution for each cell from the points located in the cell.\\

the mean for each cell:
\begin{equation}
\boldsymbol{\hat{\mu}} = \frac{1}{N} \sum_{i=1}^{N} \boldsymbol{\mathrm{x}_i}
\end{equation}

And, the uncertainty :
\begin{equation}
\boldsymbol{\hat{\Sigma}} =  \frac{1}{N} \sum_{i=1}^{N} (\boldsymbol{\mathrm{x}_i} - \boldsymbol{\hat{\mu}} )(\boldsymbol{\mathrm{x}_i} - \boldsymbol{\hat{\mu}} )^T
\end{equation}

Where, $\boldsymbol{\mathrm{x}_i}$ are the points located in one cell, and $N$ is the total points located in the cell.
\subsection{Point to Distribution Registration}

\begin{figure*}[h]
\centering
\includegraphics[width=0.8\textwidth]{{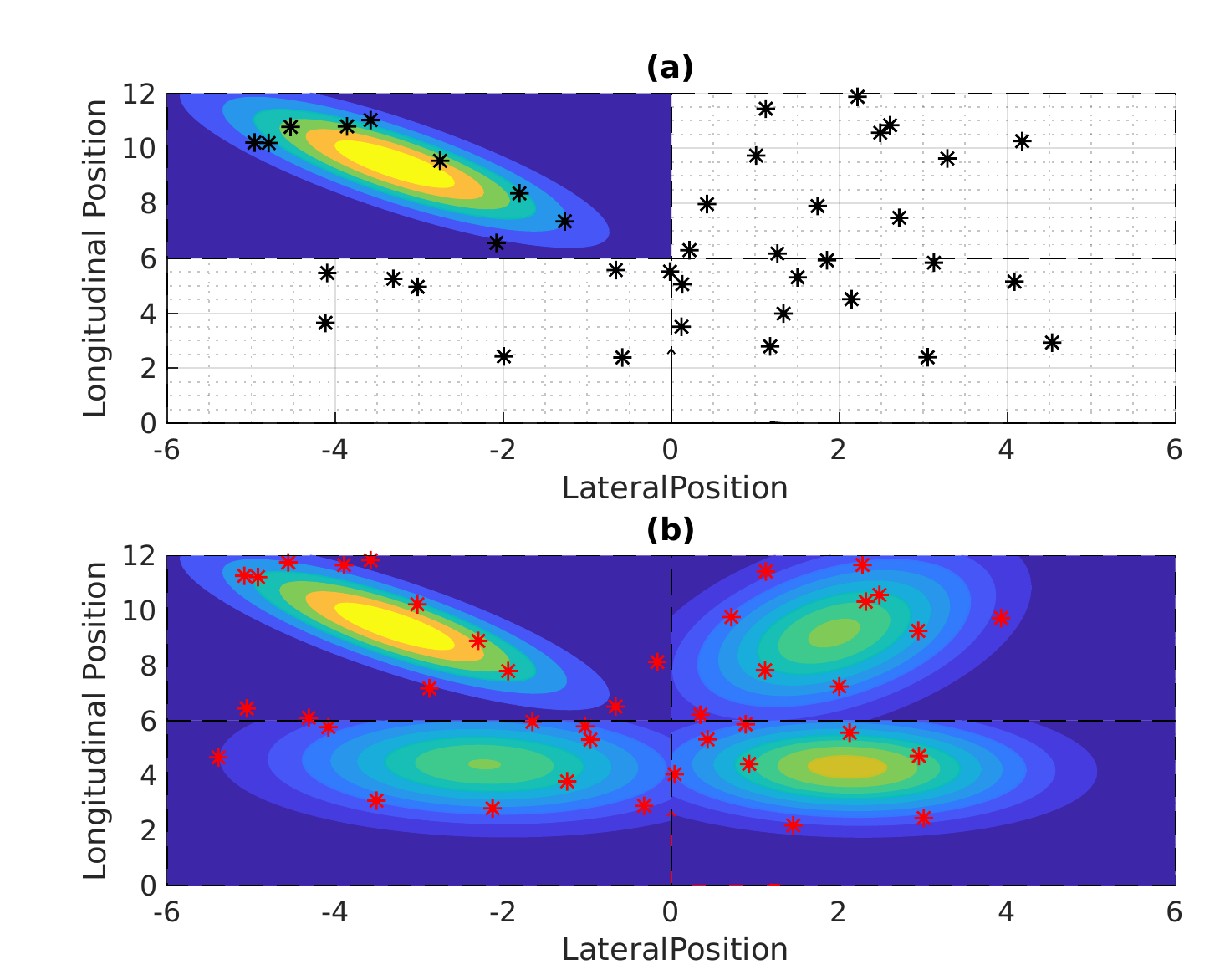}}
\caption{ A point set registration based on point to distribution (a) NDT representation for the previous point set.(b) The current point set located on the NDT for the previous point set.}
\label{fig:p2d_1}
\end{figure*}

This method builds an NDT for the previous point set only as in Fig. \ref{fig:p2d_1} (a). Then, each point in the current point set will geometrically located on the previous NDT as shown in Fig.  \ref{fig:p2d_1} (b). The cost function is constructed based on the Mahalanobis distance, where it measures the distance between each current point and corresponding previous NDT as follows:

\begin{equation}
\mathcal{N}(\boldsymbol{{T}(\mathrm{x}}) |\boldsymbol{\mu},\boldsymbol{\Sigma}) = \frac{1}{\sqrt{(2\pi)^d  det(\boldsymbol{\Sigma})}}  exp {\{- \frac{1}{2} (\boldsymbol{{T}(\mathrm{x}}) - \boldsymbol{\mu} )^T {\boldsymbol{\Sigma}}^{-1} (\boldsymbol{{T}(\mathrm{x}}) - \boldsymbol{\mu})\}}
\end{equation}

The cost function for the whole problem can be written as follows:
\begin{equation}
score({T}) = \sum_{i}^{} \mathcal{N}(\boldsymbol{{T}(\mathrm{x_i}}) |\boldsymbol{\mu_j},\boldsymbol{\Sigma_j})
\end{equation}
Where, $\boldsymbol{T}$ is the transformation matrix to be estimated, $\boldsymbol{{\mathrm{x}}_i}$ is a point form  the current point set, $(\boldsymbol{\mu}_j,\boldsymbol{\Sigma}_j)$ are the parameters for the corresponding NDT for the referenced point $\boldsymbol{{\mathrm{x}}_i}$.

The point-to-distribution algorithm steps can be summarized as follows:
\begin{center}
\resizebox {1\textwidth} {!} {
\begin{tikzpicture}[node distance=1.5cm,
every node/.style={fill=white, font=\sffamily}, align=center]
\node (c2)    [activityStarts]                {Initial guess $\boldsymbol{T}$};
\node (c3)     [base, right of=c2, xshift=4cm] {Current point set};
\node (c1)     [base, left of=c2, xshift=-4cm] {Previous point set};

\node (c7)     [base, below of=c1 ]          {Build NDT for each cell};
\node (c8)     [nonblock, below of=c2 ]          { };
\node (c9)     [base , below of=c3 ] { Reference the current point set};

\node (c13)     [nonblock, below of=c7]          { };
\node (c14)     [base, below of=c8]           { Mapping the referenced points into NDT};

\node (c19)     [nonblock, below of=c13]           { };
\node (c20)     [base,  below of=c14]           { Calculate score of $\boldsymbol{T}$ };
\node (c21)     [base, right of=c20,xshift=4cm]    {Update initial guess $\boldsymbol{T}$};

\node (c26)     [base,  below of=c20]           { Do optimization};
\node (c27)     [nonblock, below of=c21]          { };

\draw[->]      (c2) -- (c9);
\draw[->]      (c3) -- (c9);
\draw[->]      (c9) -- (c14);
\draw[->]      (c1) -- (c7);
\draw[->]      (c7) -- (c14);
\draw[->]      (c14) -- (c20);
\draw[->]      (c20) -- (c26);
\draw[->]      (c26) -| (c21);
\draw[->]      (c21) -- (c9);

\end{tikzpicture}
}
\captionof{figure}{Point set registration based on NDT representation using the point-to-distribution framework. \\}
\end{center}

\subsection{Distribution to Distribution Registration}

\begin{figure*}[h]
\centering
\includegraphics[width=0.8\textwidth]{{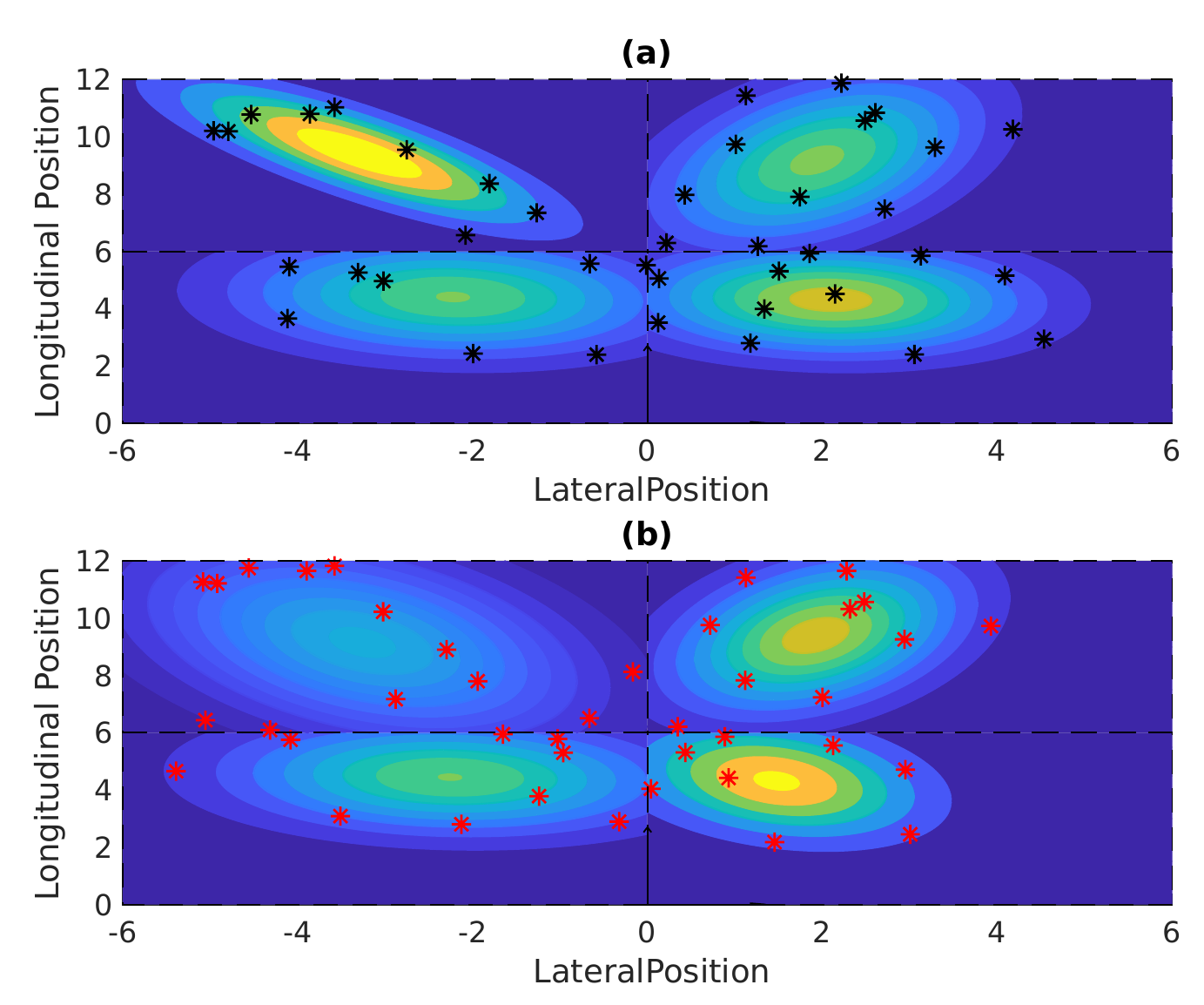}}
\caption{ A point set registration based on distribution to distribution; (a) The NDT representation for the previous point set.(b) The NDT representation for the current point set.}
\label{fig:d2d_1}
\end{figure*}

\vspace*{1cm}

This subsection illustrates the general idea of the distribution-to-distribution approach. The complete mathematical equation will be discussed in more detail in the following chapter. The main distinction between this method and the previous one is this method builds NDT for the current and the previous point set, as depicted in Fig. \ref{fig:d2d_1}. Then, the L2 metric is used to construct the cost function. The distribution-to-distribution sequence can be summarized as follows : 

\begin{center}
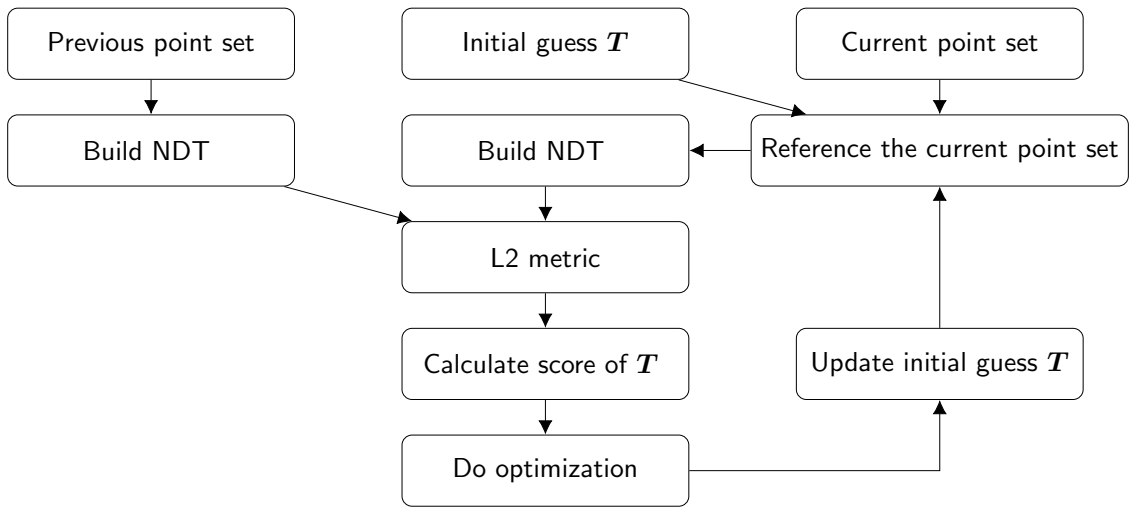

\resizebox {1\textwidth} {!} {
\begin{tikzpicture}[node distance=1.5cm,
every node/.style={fill=white, font=\sffamily}, align=center]
\node (c2)    [activityStarts]                {Initial guess $\boldsymbol{T}$};
\node (c3)     [base, right of=c2, xshift=4cm] {Current point set};
\node (c1)     [base, left of=c2, xshift=-4cm] {Previous point set};

\node (c7)     [base, below of=c1 ]       {Build NDT };
\node (c8)     [base, below of=c2 ]       { Build NDT };
\node (c9)     [base , below of=c3 ]      {Reference the current point set};

\node (c13)     [nonblock, below of=c7]          { };
\node (c14)     [base, below of=c8]           { L2 metric};

\node (c19)     [nonblock, below of=c13]           { };
\node (c20)     [base,  below of=c14]           { Calculate score of $\boldsymbol{T}$ };
\node (c21)     [base, right of=c20,xshift=4cm]    {Update initial guess $\boldsymbol{T}$};

\node (c26)     [base,  below of=c20]           { Do optimization};
\node (c27)     [nonblock, below of=c21]          { };

\draw[->]      (c2) -- (c9);
\draw[->]      (c3) -- (c9);
\draw[->]      (c9) -- (c8);
\draw[->]      (c8) -- (c14);
\draw[->]      (c1) -- (c7);
\draw[->]      (c7) -- (c14);
\draw[->]      (c14) -- (c20);
\draw[->]      (c20) -- (c26);
\draw[->]      (c26) -| (c21);
\draw[->]      (c21) -- (c9);

\end{tikzpicture}
}
\captionof{figure}{Point set registration based on NDT representation using the distribution-to-distribution framework.\\}
\end{center}


\printindex

\chapter{Cost Function}

Point set registration optimizes the relative transformation $\{ \boldsymbol{{T}}\}$ between two point sets, by minimizing the error between a transformed point set and the reference point set. The cost or error function measures that error or the distance to be minimized. However, building that function is a very critical task. Some of the algorithms use the \textit{Euclidean distance} to build the error function, such as iterative closest point algorithm (ICP). On the other hand, many researchers use the probabilities to build the error function, such as a Gaussian mixture model (GMM) or normal distribution transformation (NDT) representation. Therefore, \textit{Mahalanobis distance} or \textit{L2 metric} measure the error. This chapter presents a one-dimensional synthetic scenario for the demonstration.

\vspace*{-\baselineskip}

\section{Synthetic Scenario -1D }
\label{sec:cost_fun_1}

\begin{figure}[htb!]
\centering
\subfigure[\textbf{ The previous point set$\{\mathcal{F}\}$ }]{\label{fig:a} \includegraphics[width=70mm]{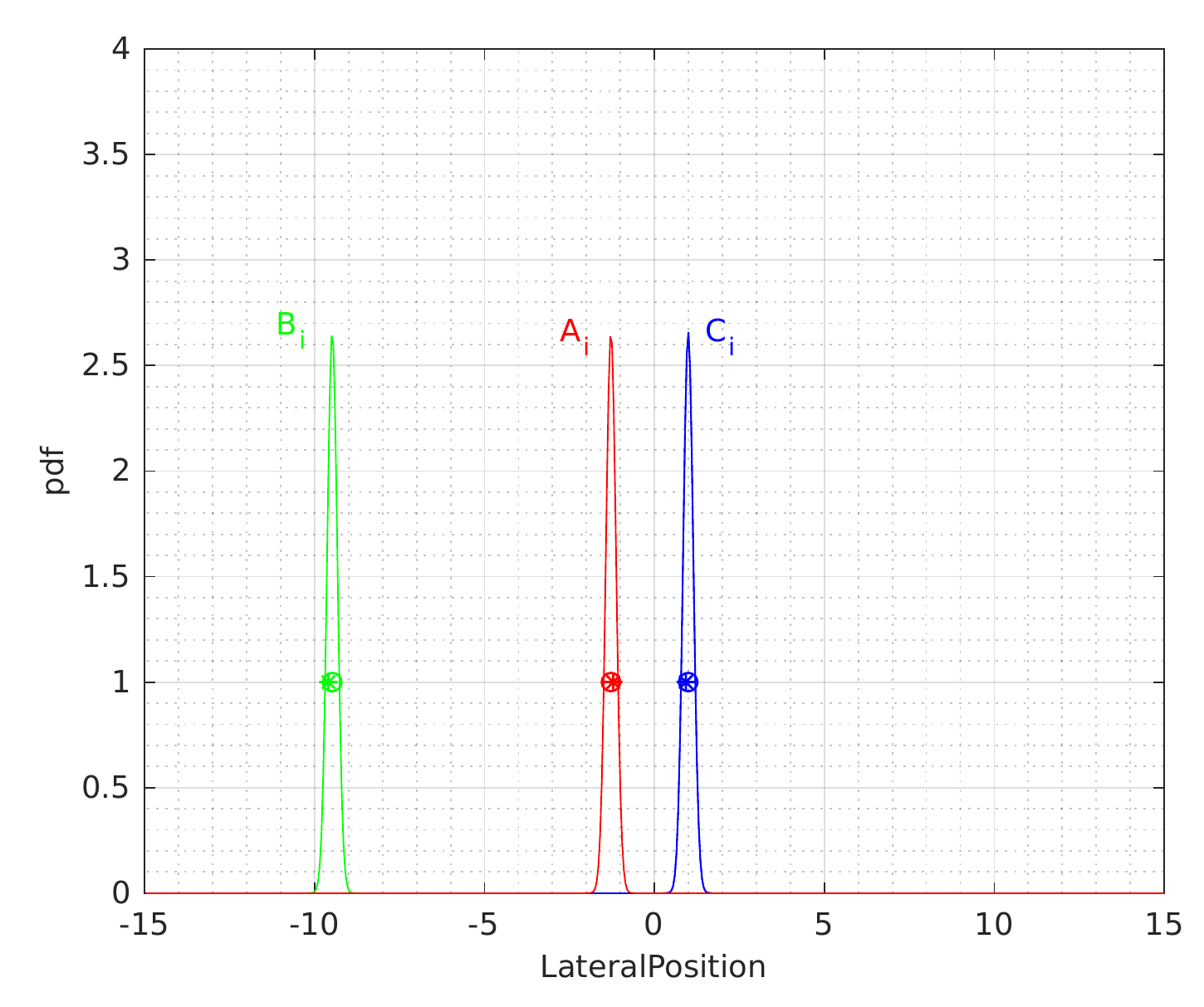}}
\subfigure[\textbf{The current point set $\{\mathcal{M}\}$}]{\label{fig:b} \includegraphics[width=70mm]{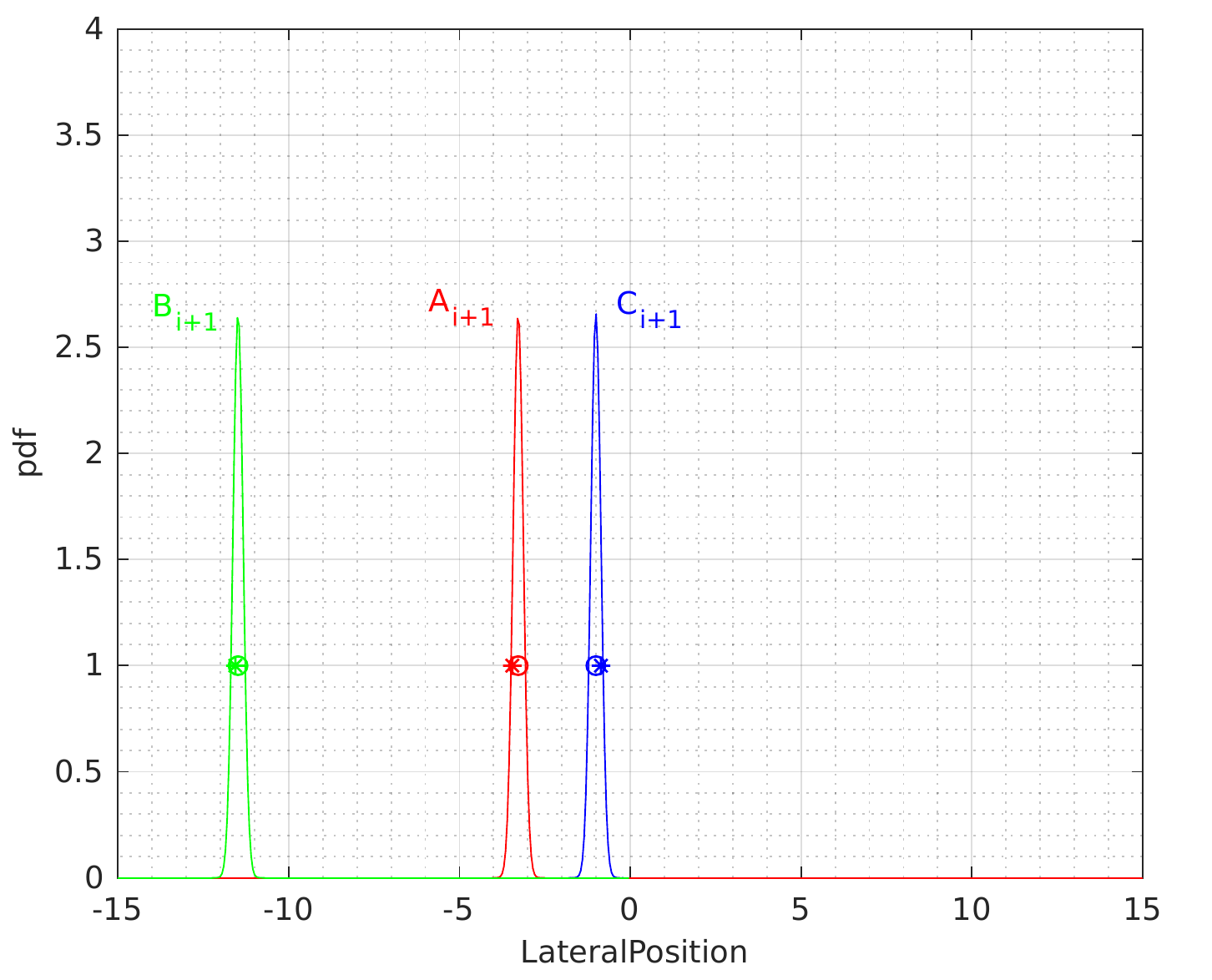}}
\caption[1-D synthetic scenario]{One dimensional synthetic scenario includes two-point sets $\{\mathcal{F},\mathcal{M}\}$, where each point set has three targets, each target has a distinct name and particular color. (a) represents the previous point set$\{\mathcal{F}\}$. (b) represents the current point set $\{\mathcal{M}\}$}
\label{fig:point_set}
\end{figure}

The scenario incorporates two-point set $\{\boldsymbol{\mathcal{F}}, \boldsymbol{\mathcal{M}}\}$, where the second point set $\{\boldsymbol{\mathcal{M}}\}$ is relatively translated from the first point set $\{\boldsymbol{\mathcal{F}}\}$; $\boldsymbol{\mathcal{M}} = \boldsymbol{T}(\boldsymbol{\mathcal{F}},\boldsymbol{\theta})$. Where, $\boldsymbol{\theta}$ represents the motion parameters, and for this scenario $\boldsymbol{\theta} = t_x = 2$ , and for 2-D scenario with 3DOF $\boldsymbol{\theta} = [t_x, t_y, \phi_z]$, where $t_x$ is the translation along the x-axis, $t_y$ is the translation along the y-axis, and $\phi_z$ represents the rotation around the z-axis. Each point set includes three targets $\{\boldsymbol{A, B, C}\}$, where each target is represented by a different color, as shown in Fig \ref{fig:point_set}. Every target is represented by a cirle and star, where the circle is the ground truth position for the target and the star represents the position after adding noise, all points have 0.15 standard deviation.\\

Each target holds two pieces of information: the target position and the standard deviation. Thus, each target can be represented as Gaussian distribution $\mathcal{N}\left(\boldsymbol{\mu}_k, \boldsymbol{\Sigma}_k \right)$ and the probability density function for target $k$ written as\footnote{ Although this chapter uses a 1-D example for the explanation, the mathematical equation is in a general form, which means for a higher dimensions.}:

\begin{equation}
\label{eq:pdf4_01}
\displaystyle p(\boldsymbol{x}|\boldsymbol{\mu}_k,\boldsymbol{\Sigma}_k)=\frac{\exp[-\frac{1}{2}(\boldsymbol{x}-\boldsymbol{\mu}_k)^{\mathrm{T}}\boldsymbol{\Sigma}^{-1}_k (\boldsymbol{x}-\boldsymbol{\mu}_k)]}{\sqrt{(2\pi)^{\delta}\vert \boldsymbol{\Sigma}_k \vert }}
\end{equation}

However, the point set is considered as a mixture of Gaussians (all targets). Consequently, the probability density function for each target can be written as :

\begin{equation}
\label{eq:pdf4_02}
p_{\mathrm{GMM}}(\boldsymbol{x} | \lambda)=\sum_{i = 1}^{\vert {\cal F}\vert } w_{i} \cdot p\left(\boldsymbol{x} | \boldsymbol{\mu}_{i}, \boldsymbol{\Sigma}_{i}\right)
\end{equation}

Where, $\lambda=\left\{w_{i}, \boldsymbol{\mu}_{i}, \boldsymbol{\Sigma}_{i}\right\}$ describes the GMM parameters and $w_{i}$ refers to mixture weights for each normal distribution (component or target). ${\vert {\cal F}\vert }$ represents the number of Gaussian components in the GMM.

\section{Distance Metrics}

This section explains two possible metrics to construct the error function: The point-to-distribution and distribution-to-distribution. Moreover, comparing both metrics showing the advantages and disadvantages.

\subsection*{Point to Distribution}

Biber in~\cite{biber2003normal} proposed this approach to achieve a 2D laser scan matching, where he used the normal distribution transformation (NDT)to represent the laser scan, then point-to-distribution approach to build the cost function. Furthermore, Magnusson in~\cite{magnusson2007scan} extends the approach; then, he proposed a 3D-NDT registration method. However, this current work uses the GMM to represent the previous point set $\{\boldsymbol{\mathcal{F}}\}$ (\ref{eq:pdf4_02}), then use the Mahalanobis distance to measure the distance between each transformed target $\boldsymbol{T}(m_k)$ in the current point set $\{\boldsymbol{\mathcal{M}}\}$ and the previous point set GMM as follows:

\begin{equation}
\centering
p_{\mathrm{GMM}}(\boldsymbol{T}(\boldsymbol{m_k})| \lambda_{\mathcal{F}}) =  \sum_{i=1}^{\vert {\cal F}\vert } w_{i} \cdot p\left(\boldsymbol{T}(\boldsymbol{\mu_k}) | \boldsymbol{\mu_{i}}, \boldsymbol{\Sigma_{i}}\right)
\end{equation}

Thus, the cost function for $\boldsymbol{\theta}$ with respect to one target $\boldsymbol{m_k}$ can be written as:

\begin{equation}
\centering
f_{\mathrm{p2d}}(\boldsymbol{\theta})=p_{\mathrm{GMM}}\left(\boldsymbol{T}\left(\boldsymbol{\mu_k}, \boldsymbol{\theta}\right) | \lambda_{\mathcal{F}}\right)
\end{equation}
Where $\{\boldsymbol{\theta}\}$ is the motion state, which needs to be optimized. Consequently, for target $\boldsymbol{B}$, the cost function can be written as:
\begin{equation}
f_{\mathrm{p2d}}(\boldsymbol{\theta})_{\text {target}(B)}= \sum_{i=1}^{\vert {\cal F}\vert } w_{i} \cdot p\left(\boldsymbol{T}(\boldsymbol{\mu}_{B}) | \boldsymbol{\mu_{i}}, \boldsymbol{\Sigma}_{i}\right)
\end{equation}

\begin{figure}[htb!]
\centering
\subfigure[\textbf{ The problem for target $\{{B}\}$}]{\label{fig:a}\includegraphics[width=69mm]{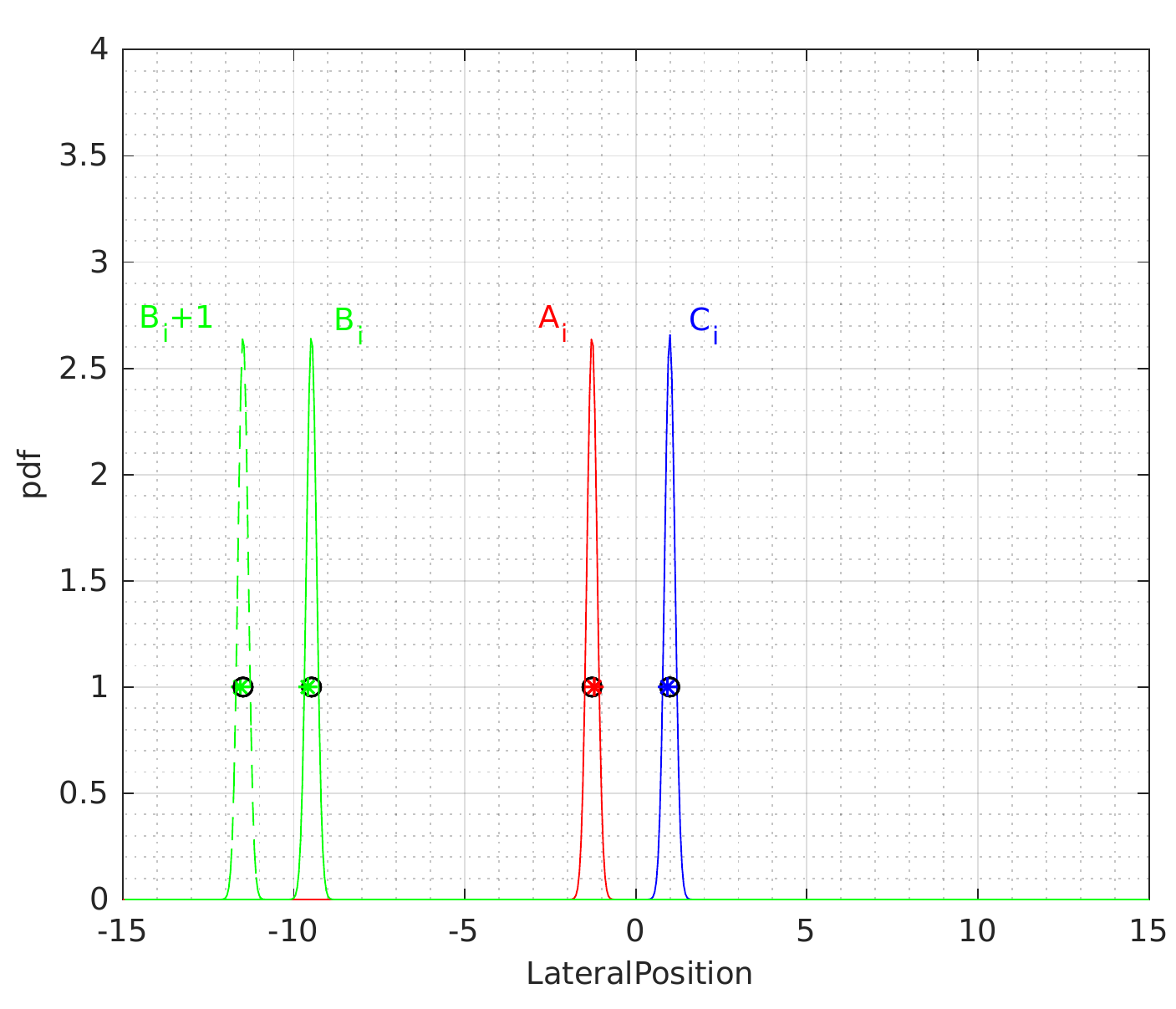}}
\subfigure[\textbf{ The cost surfacefor target $\{{B}\}$}]{\label{fig:b}\includegraphics[width=69mm]{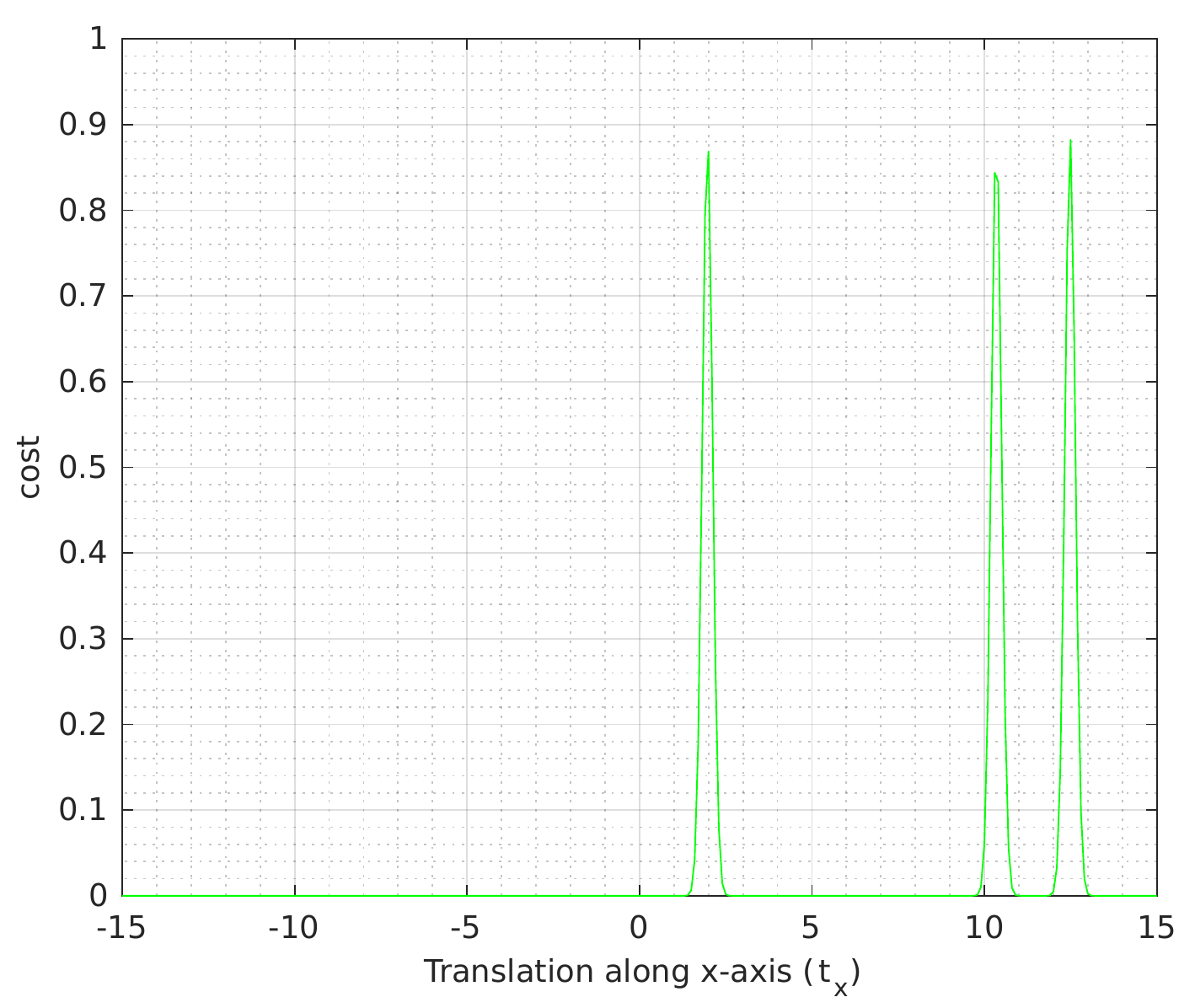}}
\caption[The registration representation for one target]{The registration problem for target $\{{B}\}$.(a) represents the aligning task for target $\{B_{i+1}\}$ and the previous point set $\{\mathcal{F}\}$. (b) represents the cost surface for that target after the aligning process.}
\label{fig:reg_B_1}
\end{figure}

The registration task translates target $\boldsymbol{B_{i+1}}$ to match all targets in the previous point set  $\{\mathcal{F}\}$, as depicted in Fig. \ref{fig:reg_B_1} (a). The previous scan has three targets. Therefore, the cost surface includes three possibilities, where at $t_x = 2$ $B_{i+1}$ matches $B_i$, $t_x = 10$  it matches $A_i$ and $t_x = 12.2$ it matches $C_i$, as shown in Fig. \ref{fig:reg_B_1}(b). By the same token, Fig. \ref{fig:reg_B_2} (b) shows the cost surfaces for all targets.

\subsection*{Distribution to Distribution}

Several metrics are proposed to measure the distance between two probability distribution, and Stoyanov in~\cite{stoyanov2012point} provides a review for most of these metrics. However, most of the recent works~\cite{stoyanov2012point}~\cite{barjenbruch2015joint} used the closed form for L2 provided by Jian in ~\cite{jian2010robust} to measure the distance between the two distributions. In comparison to the Mahalanobis distance, L2 distance uses additionally to the target position, the standard deviation of the targets in the current point set. Therefore, the distance to be minimized is the distance between each transformed target distribution and the GMM distribution for the previous point set.\\

\begin{figure}[htb!]
\centering
\subfigure[\textbf{ The cost surface for all targets p2d}]{\label{fig:b} \includegraphics[width=69mm]{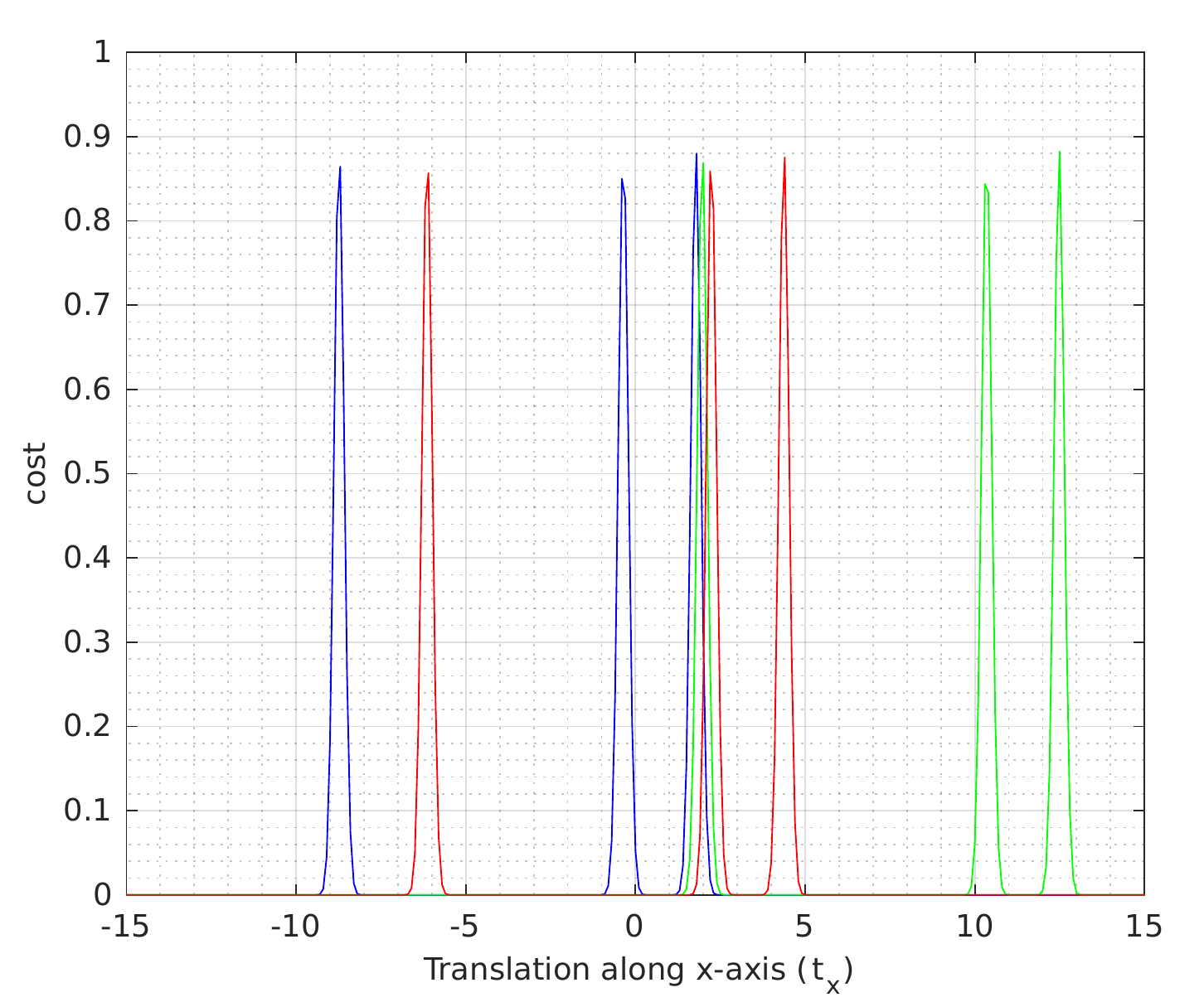}}
\subfigure[\textbf{ The cost surface for all targets d2d}]{\label{fig:b} \includegraphics[width=69mm]{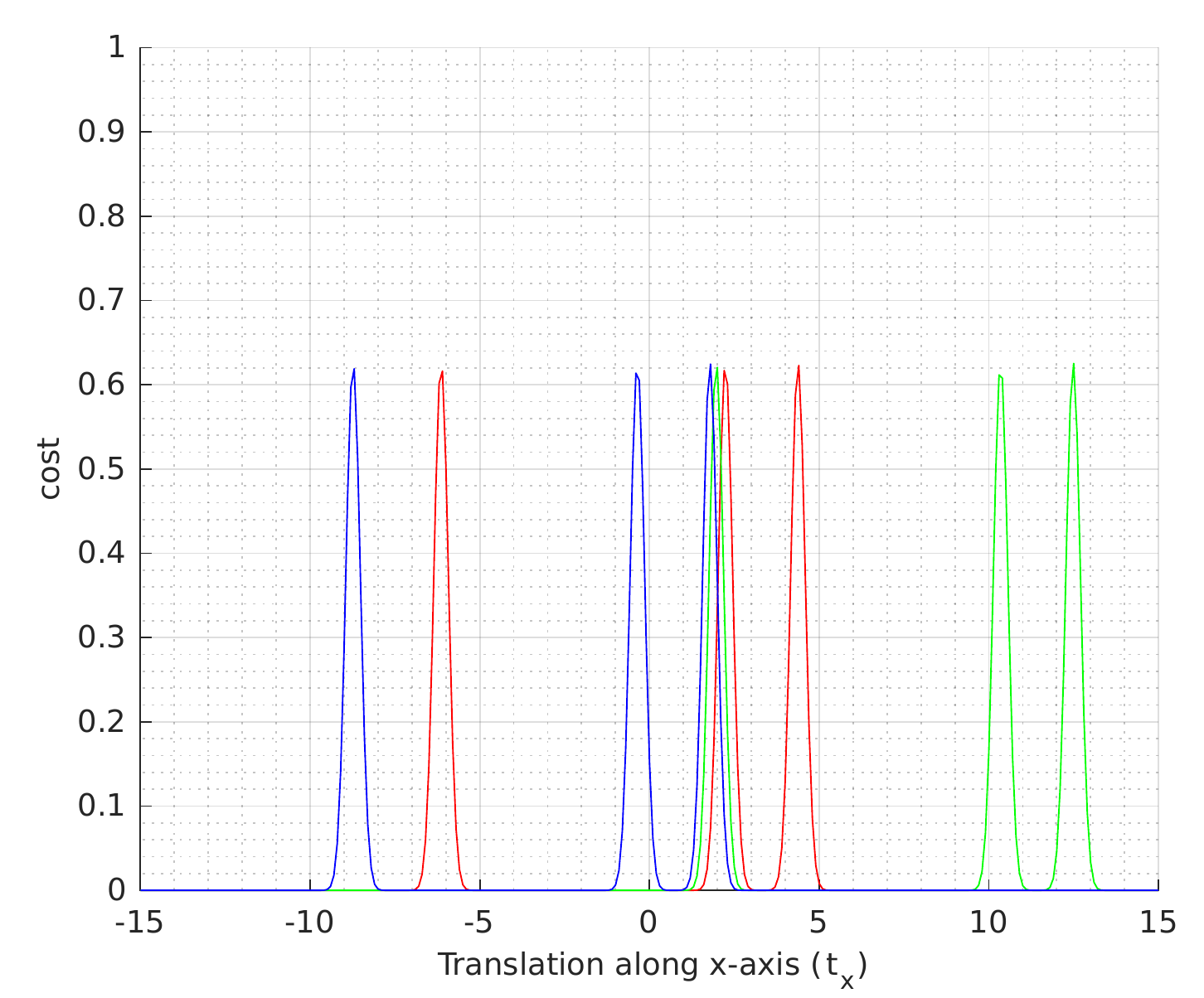}}
\caption[The cost surface based on p2d and d2d for Scenario -1D]{The cost surface for all targets, where each target in the current point set has three possibilities to be aligned with the previous point set. The cost values based on the p2d approach is greater than the d2d approach because the d2d includes the uncertainty of both points set in the cost equation.}
\label{fig:reg_B_2}
\end{figure}

Consequently, the L2 distance is used to build the cost function. Assume two distribution $\{\mathcal{N}_1(\boldsymbol{\mu_1},\boldsymbol{\Sigma_1}), \ \mathcal{N}_2(\boldsymbol{\mu_2},\boldsymbol{\Sigma_2})\}$, L2 distance can be written as :

\begin{equation}
\label{eq:l2_dis}
D_{L_{2}}(\mathcal{N}_1, \mathcal{N}_2) =  \mathcal{N}\left(0 | \boldsymbol{\mu}_{1}-\boldsymbol{\mu}_{2}, \boldsymbol{\Sigma_{1}}+\boldsymbol{\Sigma_{2}}\right)
\end{equation}

Therefore, the cost function for $\boldsymbol{\theta}$ with respect to one target $\boldsymbol{m_k}$ can be written as:

\begin{equation}
\label{eq:l2_mk}
f_{\mathrm{d2d}}(\boldsymbol{\theta}) =  \sum_{i=1}^{\vert {\cal F}\vert } w_{i} \cdot p\left( 0| (\boldsymbol{T}(\boldsymbol{\mu_k},\boldsymbol{\theta})-\boldsymbol{\mu_{i}}), (\boldsymbol{T}(\boldsymbol{\Sigma_k},\boldsymbol{\theta})+\boldsymbol{\Sigma_{i}})\right)
\end{equation}

Fig. \ref{fig:reg_B_2} (b) shows the cost surface for all current targets based on the L2 metric, where each cost surface has three possibilities. Although the cost surface behaves the same as the point-to-distribution approach, the cost values are lower than in the point-to-distribution approach, which makes sense as long as the point-to-distribution ignores the variance of the current points, but distribution-to-distribution adds both variances.

\subsection*{Discussion}

\begin{figure}[htb!]
\centering
\subfigure[\textbf{Point-to-Distribution}]{\label{fig:a} \includegraphics[width=70mm]{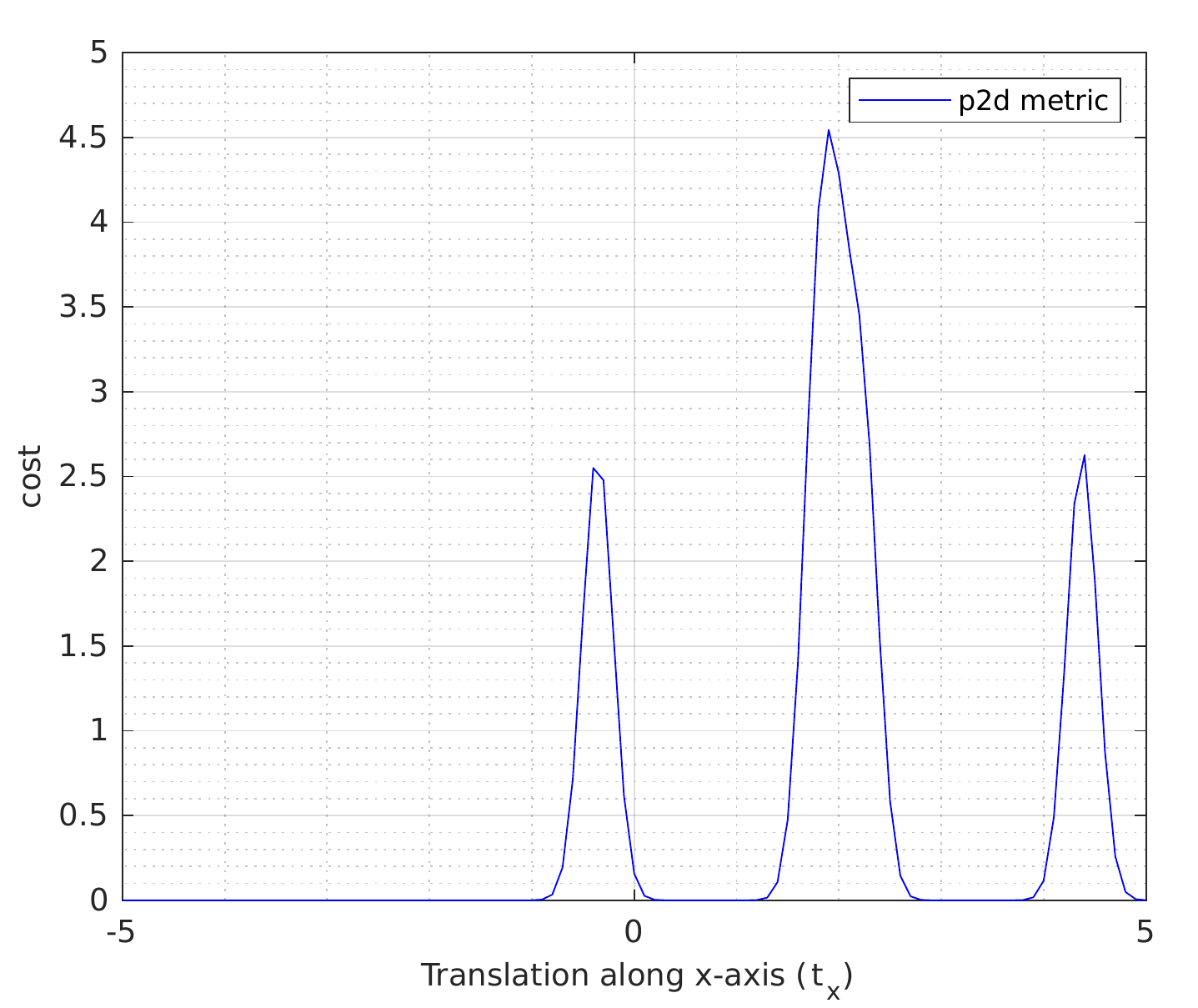}}
\subfigure[\textbf{Distribution-to-Distribution}]{\label{fig:b} \includegraphics[width=70mm]{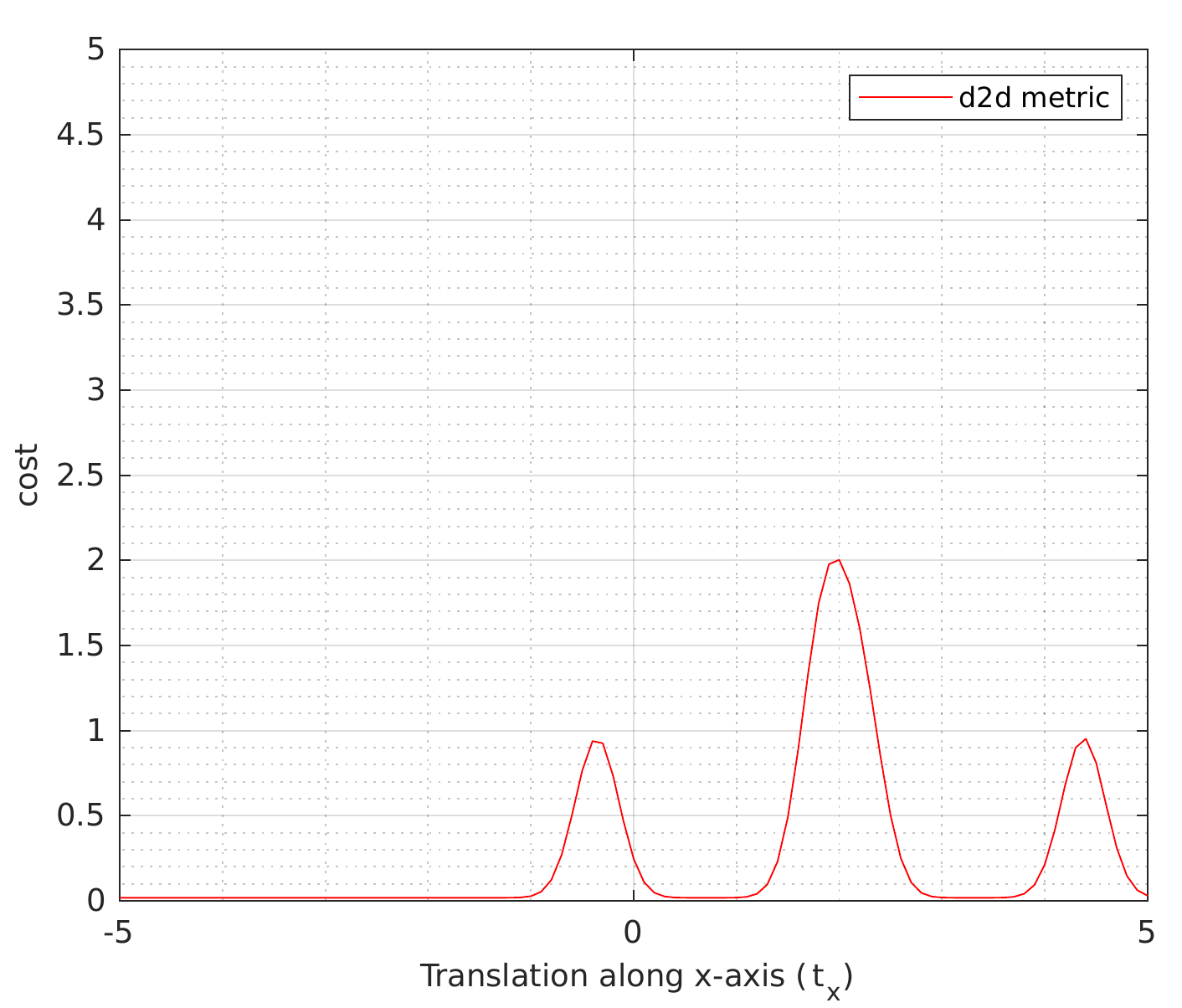}}
\caption[ The overall cost surface based on p2d and d2d.]{The overall cost surface for both approaches, where both cost surfaces have the same behave.(a) represents the cost surface based on the point-to-distribution approach. (b) represents the cost surface based on the distribution-to-distribution approach.}
\label{fig:cost_p2d_d2d_1}
\end{figure}

Fig. \ref{fig:cost_p2d_d2d_1} depicts the cost surface for the whole scenario, and the next section introduces two methods to construct the overall cost surface. The total cost surface for both approaches behave the same and have one global peak at the correct aligning. Moreover, both surfaces include local peaks, which is a well-known problem.\\

\begin{figure}[htb!]
\centering
\subfigure[\textbf{ The previous point set$\{\mathcal{F}\}$ }]{\label{fig:a}\includegraphics[width=70mm]{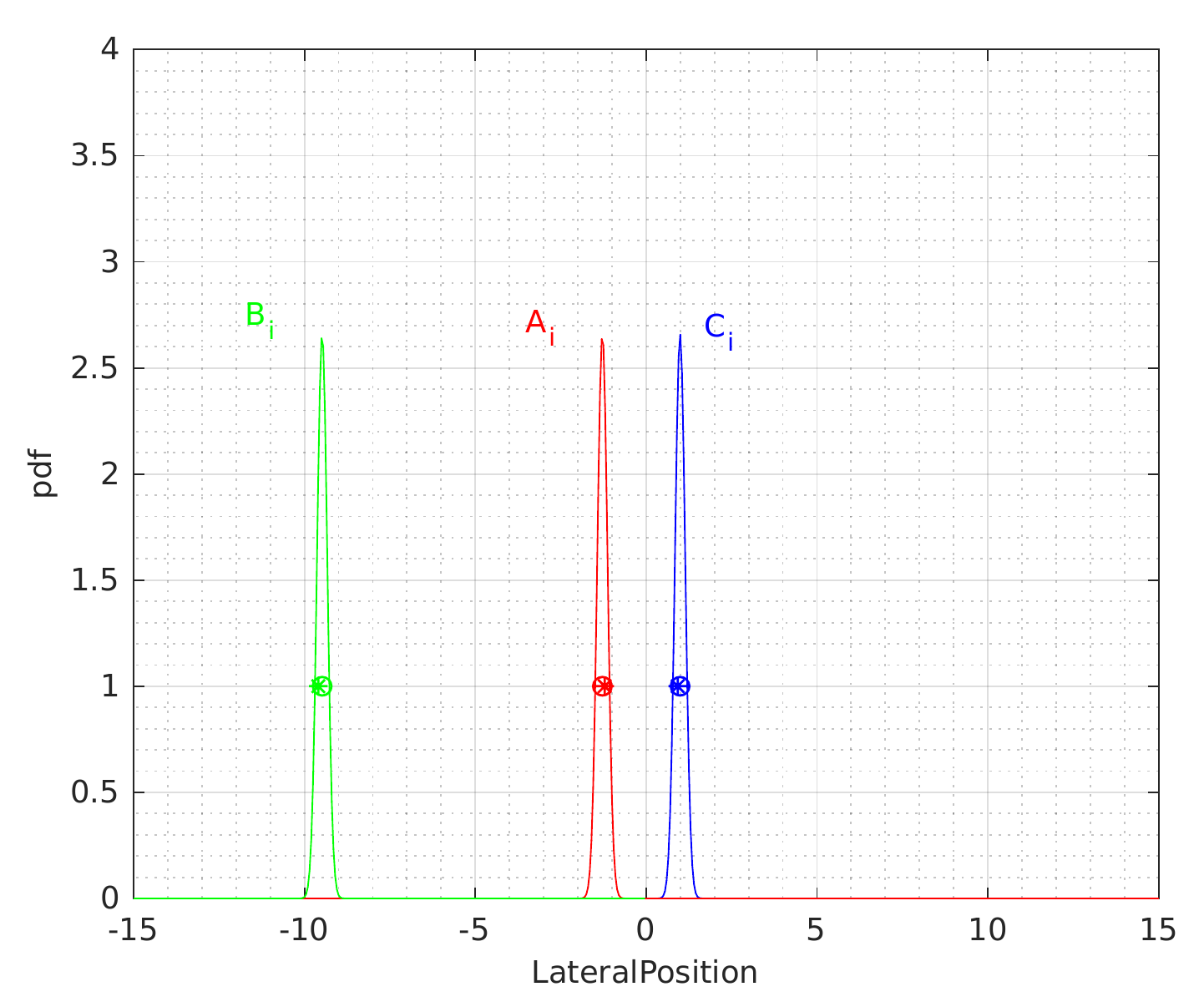}}
\subfigure[\textbf{The current point set$\{\mathcal{M}\}$ }]{\label{fig:b}\includegraphics[width=70mm]{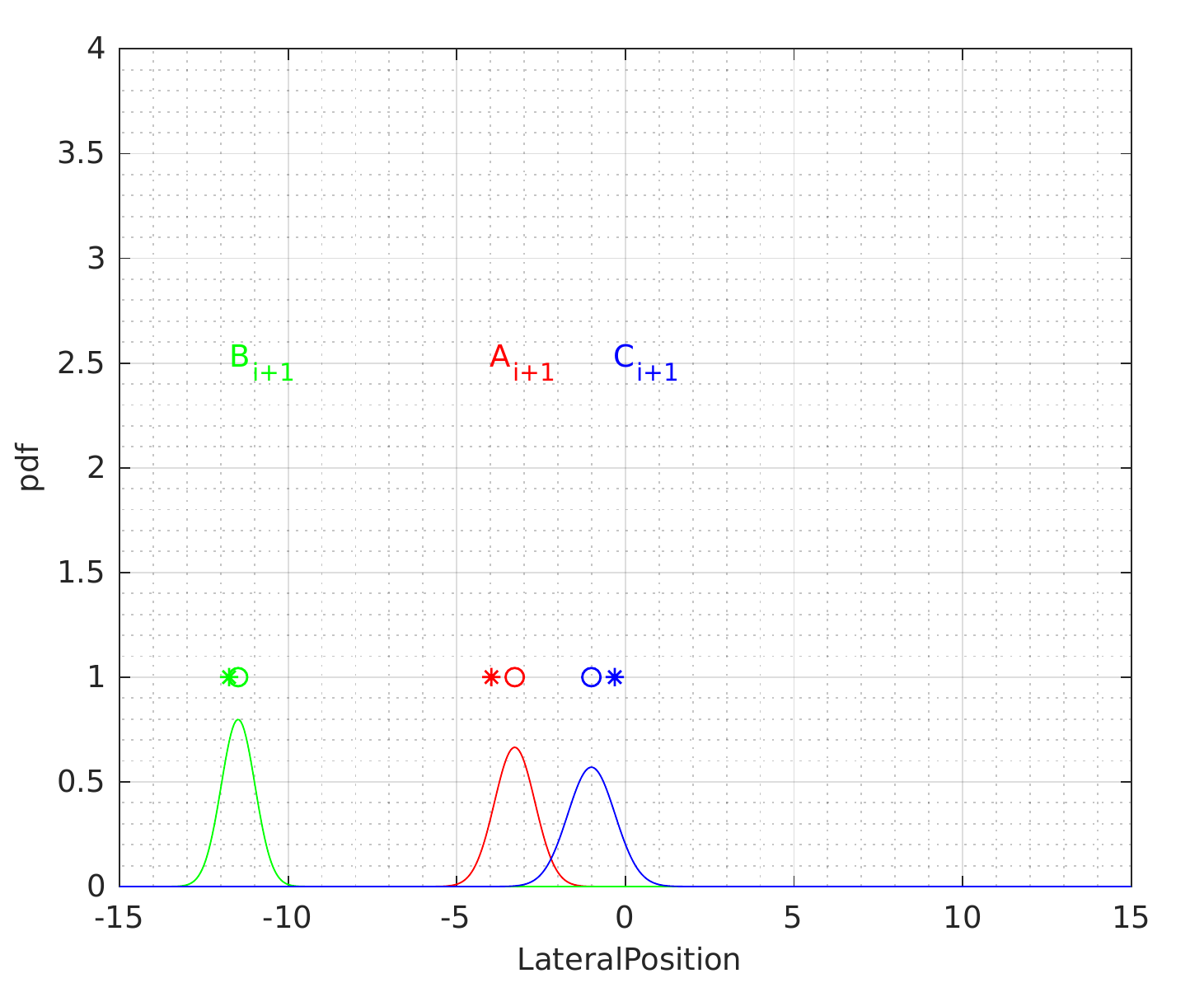}}
\caption[One dimensional synthetic scenario with different standard deviations]{One dimensional synthetic scenario, where the standard deviation for current point set $\{\mathcal{M}\}$ is larger than the standard deviation for previouse point set $\{\mathcal{F}\}$. (a) represents the previous point set $\{\mathcal{F}\}$.(b) represents the current point $\{\mathcal{M}\}$.}
\label{fig:point_set_diff4_noise_1}
\end{figure}

\begin{figure}[htb!]
\centering
\subfigure[\textbf{Point-to-Distribution}]{\label{fig:a}\includegraphics[width=70mm]{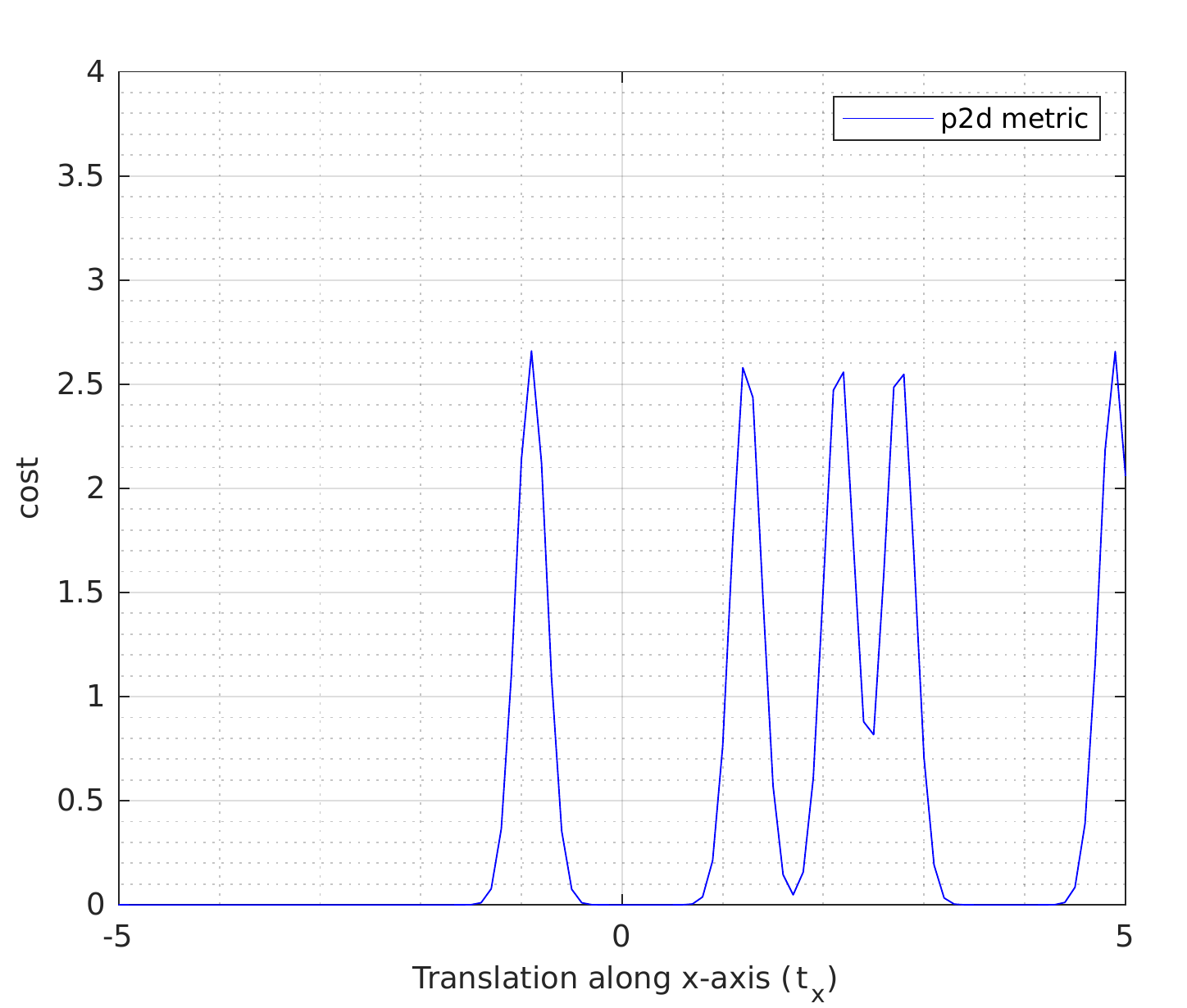}}
\subfigure[\textbf{Distribution-to-Distribution}]{\label{fig:b}\includegraphics[width=70mm]{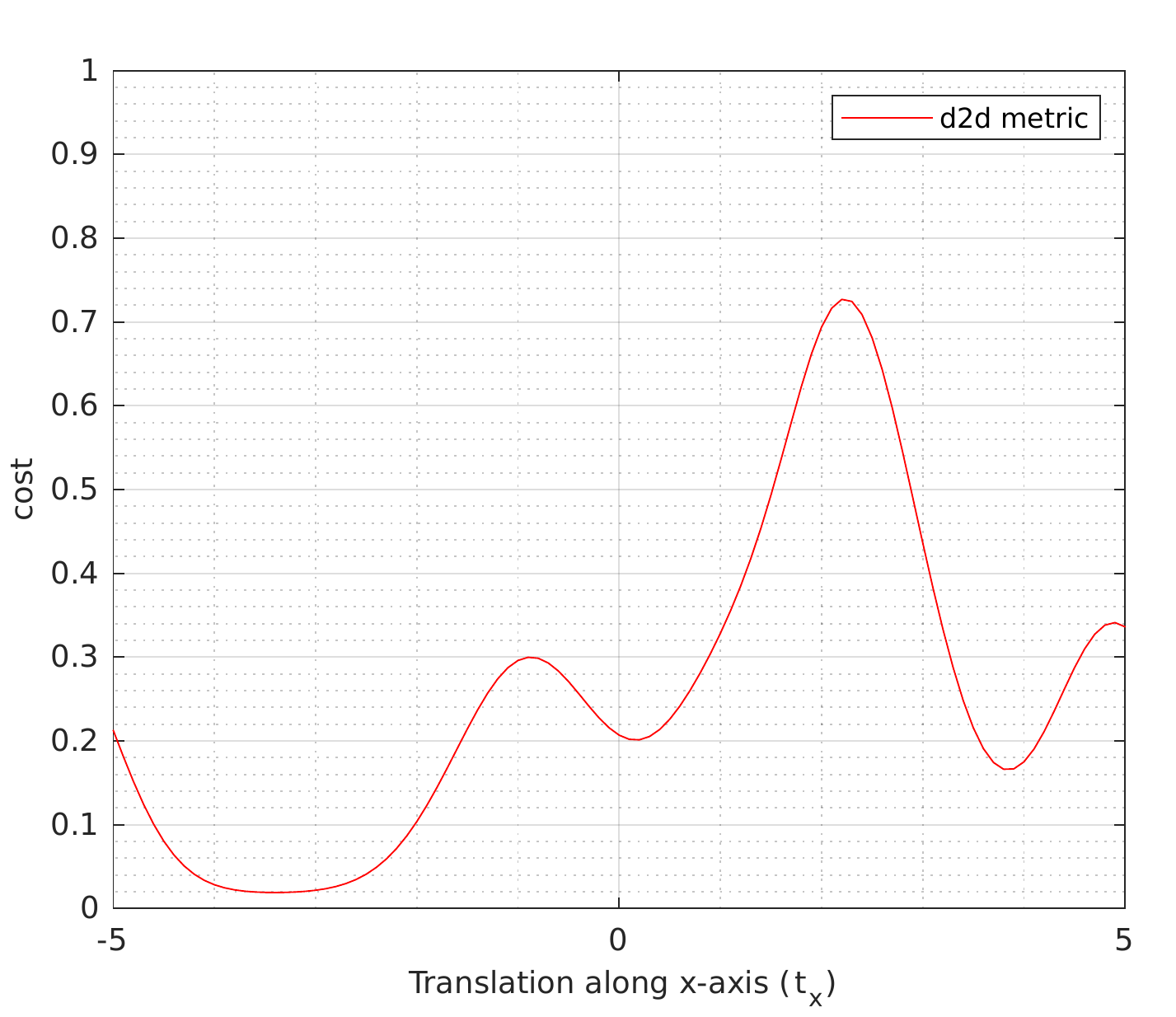}}
\caption[The overall cost surface for the scenario with different standard deviations]{ The overall cost surface for both approaches, where both cost surfaces does not have the same behave. (a) represents the cost surface based on the point-to-distribution approach.(b) represents the cost surface based on the distribution-to-distribution approach. }
\label{fig:cost_diff4_noise_2}
\end{figure}

Both cost surfaces behave the same as long as all points have the same standard deviation, which is not the case in reality. Therefore, the scenario is modified to have one point set with a small standard deviation, and the second point set has a large standard deviation, to match the reality, as shown in Fig. \ref{fig:point_set_diff4_noise_1}. From the cost surface in Fig. \ref{fig:cost_diff4_noise_2}, it is obvious that both cost surfaces do not behave the same.
Moreover, in point-to-distribution, the cost surface does not include a global peak. On the other hand, distribution-to-distribution has only one global maximum beacuse it considers the noise of the current point set in the distance metric, which is not the case in the point-to-distribution approach.\\

To sum up, the distribution-to-distribution shows a robust behavior against a high standard deviation or more noisy point set. The following section introduces two potential methods to fuse the cost surface for all targets.

\section{Fusing Distance for all Targets}

The previous section introduces two distance metrics to build the cost function for each target. Besides, the distribution-to-distribution shows a robust behavior than the point-to-distribution. Therefore, the distribution-to-distribution strategy (\ref{eq:l2_mk}) is used to build the error metric for each target. Two possible ways to implement the overall cost function: The summing method and the likelihood method. The main interest here is to explain both approaches. Furthermore, for consistent evaluation, the previous synthetic scenario is used, as shown in Fig. \ref{fig:point_set}.

\subsection*{Summing Approach }

The concept is to add or sum the cost for each target, where Barjenbruch in~\cite{barjenbruch2015joint} and Stoyanov in~\cite{stoyanov2012point}, as well as Biber~\cite{biber2004probabilistic}, used this approach in their implementation, as follows:

\begin{equation}
\label{eq:sum_cost_1}
f_{\mathrm{d2d}}(\boldsymbol{\theta})=\sum_{k = 1}^{\vert {\cal M}\vert } D_{L_{2}}(\mathcal{N}_{GMM}(\mathcal{F}), \mathcal{N}_k)
\end{equation}

\begin{equation}
\label{eq:sum_cost_2}
f_{\mathrm{d2d}}(\boldsymbol{\theta})=\sum_{k = 1}^{\vert {\cal M}\vert } \left(\sum_{i=1}^{\vert {\cal F}\vert } w_{i} \cdot p \left( 0|\boldsymbol{T}(\boldsymbol{m_k},\boldsymbol{\theta})-\boldsymbol{\mu_{i}}, \boldsymbol{T}(\boldsymbol{\Sigma_k},\boldsymbol{\theta}) + \boldsymbol{\Sigma_{i}}\right)\right)
\end{equation}\\

Fig. \ref{fig:cost_sum_like_1}(a) shows the cost surface for the three targets, where the three cost surfaces intersected at $t_x = 2$. The summing approach depicted in Fig. \ref{fig:cost_sum_like_1}(b), where the cost surface has one global maximum and many local maxima.

\subsection*{Likelihood Approach }

The idea is to joint or multiply the cost for each target (\ref{eq:l2_mk}) to form the overall cost function as follows :

\begin{equation}
\label{eq:like_cost_1}
f_{\mathrm{d2d}}(\mathcal{M}| \lambda_{\mathcal{F}}) = \prod_{k \in \mathcal{M}} D_{L_{2}}(\mathcal{N}_{GMM}(\mathcal{F}), \mathcal{N}_k)
\end{equation}

Therefore, the cost function for $\boldsymbol{\theta}$ with respect to all targets can be written as:

\begin{equation}
\label{eq:like_cost_2}
f_{\mathrm{d2d}}(\boldsymbol{\theta}) =  \prod_{k \in \mathcal{M}}\left(\sum_{i=1}^{\vert {\cal F}\vert } w_{i} \cdot p \left( 0|\boldsymbol{T}(\boldsymbol{m_k},\boldsymbol{\theta})-\boldsymbol{\mu_{i}}, \boldsymbol{T}(\boldsymbol{\Sigma_k},\boldsymbol{\theta}) + \boldsymbol{\Sigma_{i}}\right)\right)
\end{equation}

\begin{figure}[htb!]
\centering
\subfigure[\textbf{ The cost surface for all targets }]{\label{fig:a}\includegraphics[width=60mm]{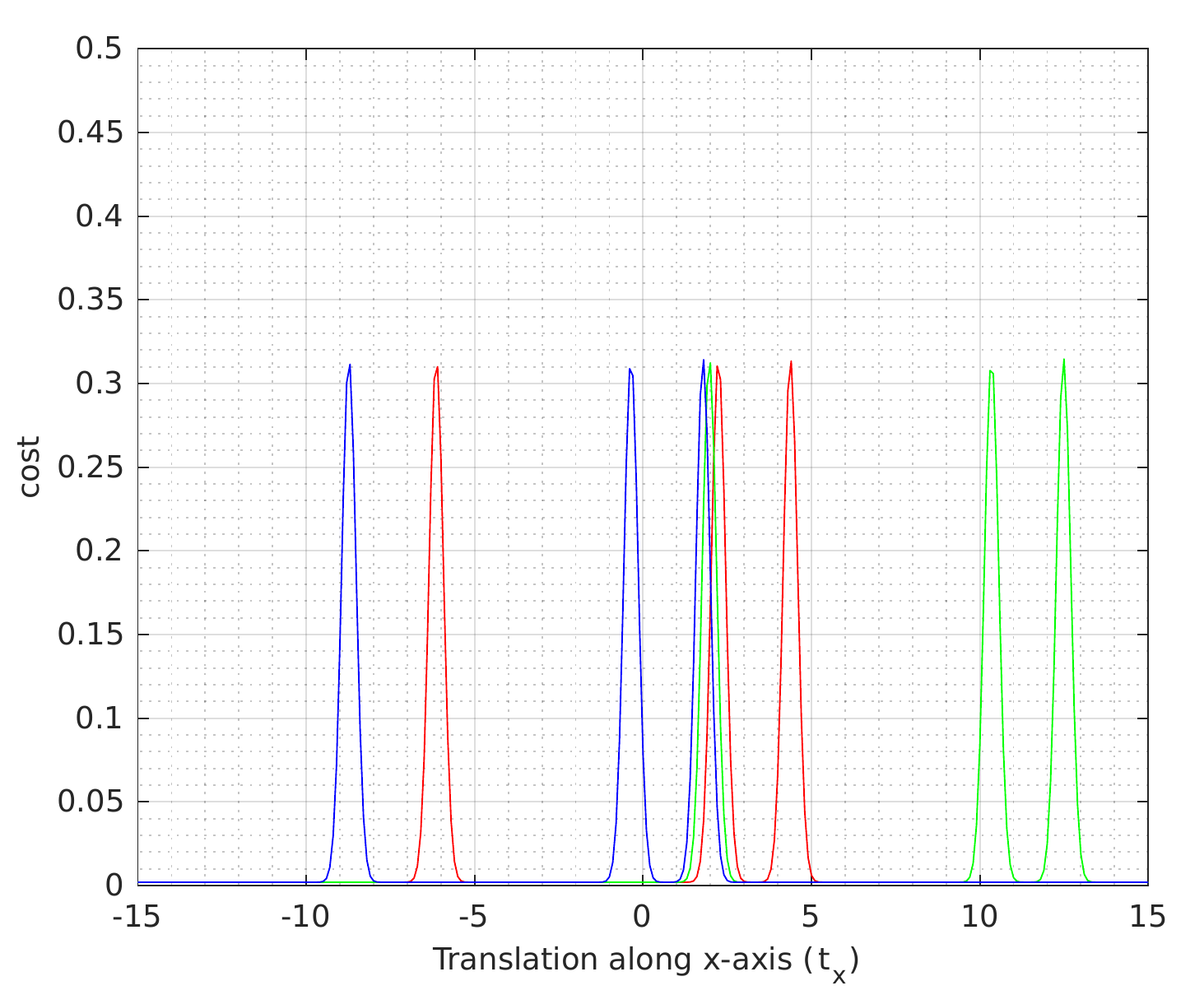}}
\subfigure[\textbf{The summing approach }]{\label{fig:b}\includegraphics[width=60mm]{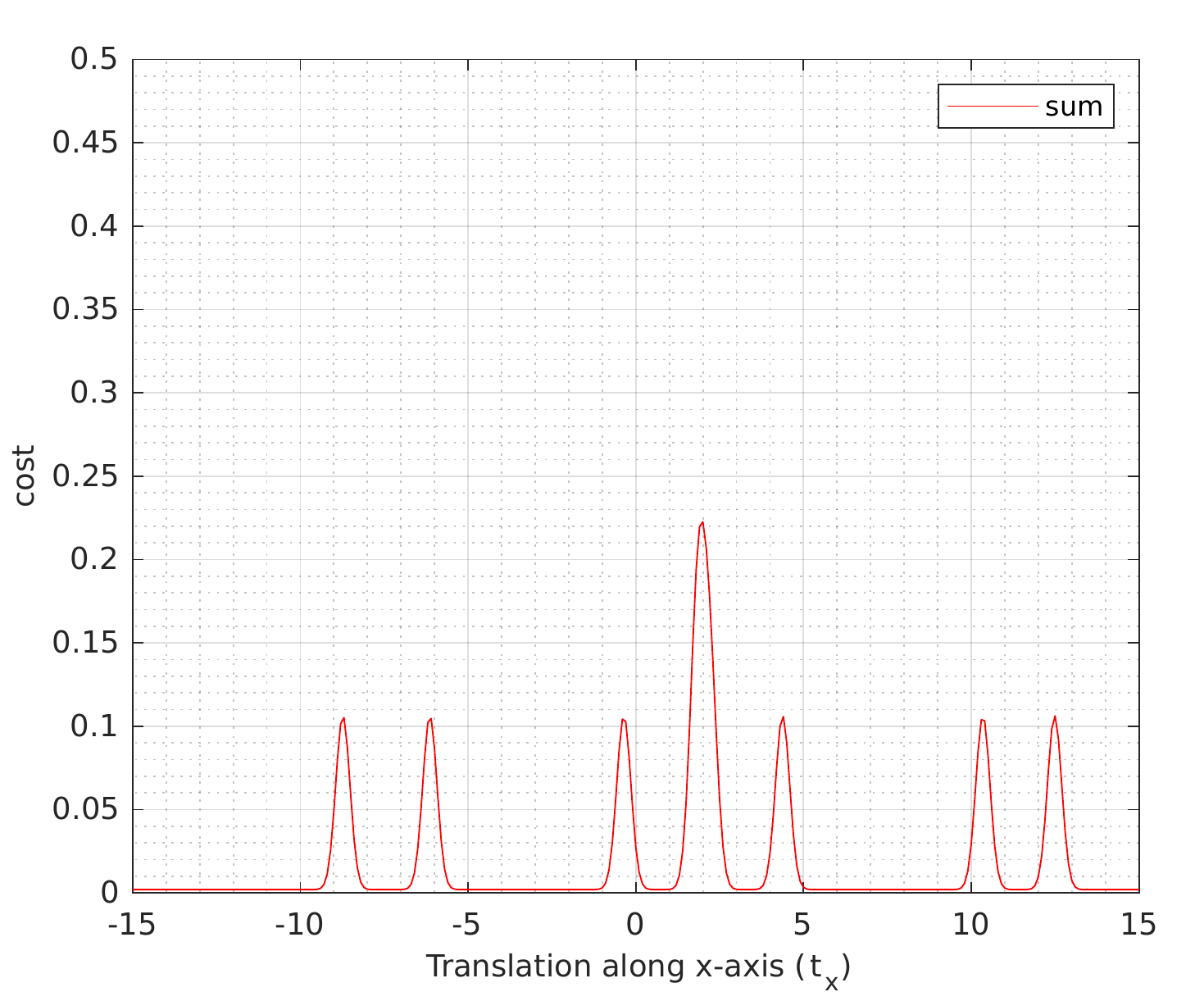}}
\subfigure[\textbf{The likelihood approach }]{\label{fig:b}\includegraphics[width=60mm]{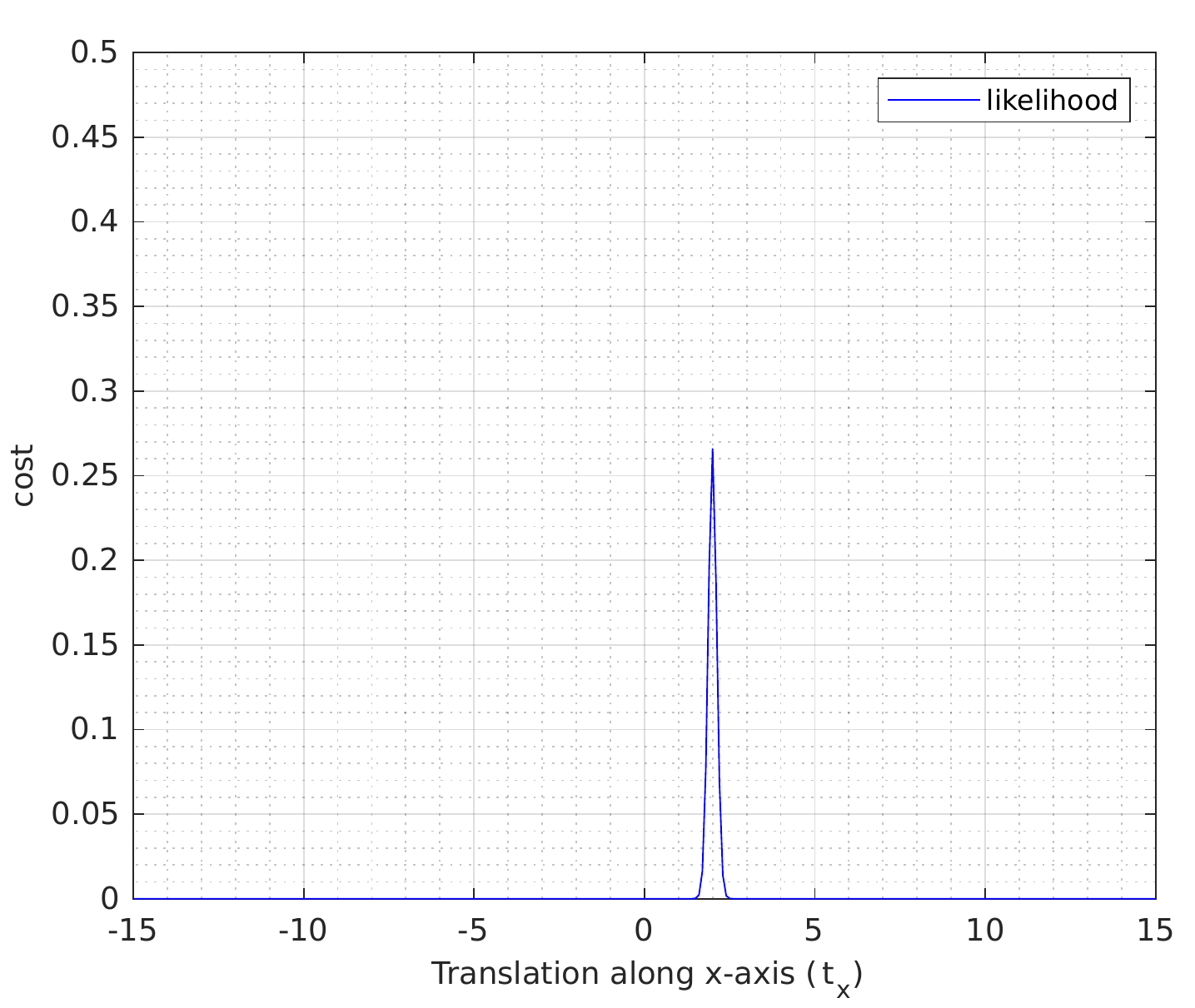}}
\caption[]{ This plot exhibits the overall cost surface  based on the summing and the likelihood approach. (a) The cost surface for each target ploted in one plot. (b) The cost surface for all targets based on the summing approach. (c) The cost surface for all targets based on the likelihood(product) approach. The cost surface based on the summing approach has many local peaks, but the likelihood has only one optmal peak.}
\label{fig:cost_sum_like_1}
\end{figure}

 Fig. \ref{fig:cost_sum_like_1}(c) exposes the likelihood approach, where the cost surface has only one global maximum.

\subsection*{Discussion}

Both approaches have one global maximum at the correct translation $ (\boldsymbol{t_x = 2})$. On the one hand, the cost surface based on the summing method is smoother than the likelihood. On the other hand, the cost surface has many local peaks, which could lead to false convergence in the optimization step. Therefore, the initial guess is a critical aspect also for the summing approach due to the local peaks.\\

Moreover, Rapp in~\cite{rapp2017probabilistic}, Barjenbruch in \cite{barjenbruch2015joint}, Stoyanov in~\cite{stoyanov2012point}, as well as Biber~\cite{biber2004probabilistic} used the summing approach to merge the cost function for all targets with various justifications. One justification is that the global maximum is smooth; thus, it should be easier for the optimizer to converge in the optimal maximum. However, the summing approach shows a robust behavior against the outlier. Hence, before discussing how the behavior is robust against the outlier, it is intuitive to explain what does it means an outlier and what is the outlier effect.

\section{Outlier}

An outlier is a data point, which remarkably deviates from the observation. In our case, if a point is in one point set and does not have a correspondence in the other set that represents an outlier, as shown in Fig. \ref{fig:outlier_prv}. This section investigates how the outlier impacts the cost surface for each target cost surface. Moreover, how the summing approach has a robust behavior against it rather than the likelihood approach, as the previous works states.

\subsection*{Outlier in Previous Point set}

\begin{figure}[htb!]
\centering
\subfigure[\textbf{The previous point set$\{\mathcal{F}\}$  }]{\label{fig:a}\includegraphics[width=60mm]{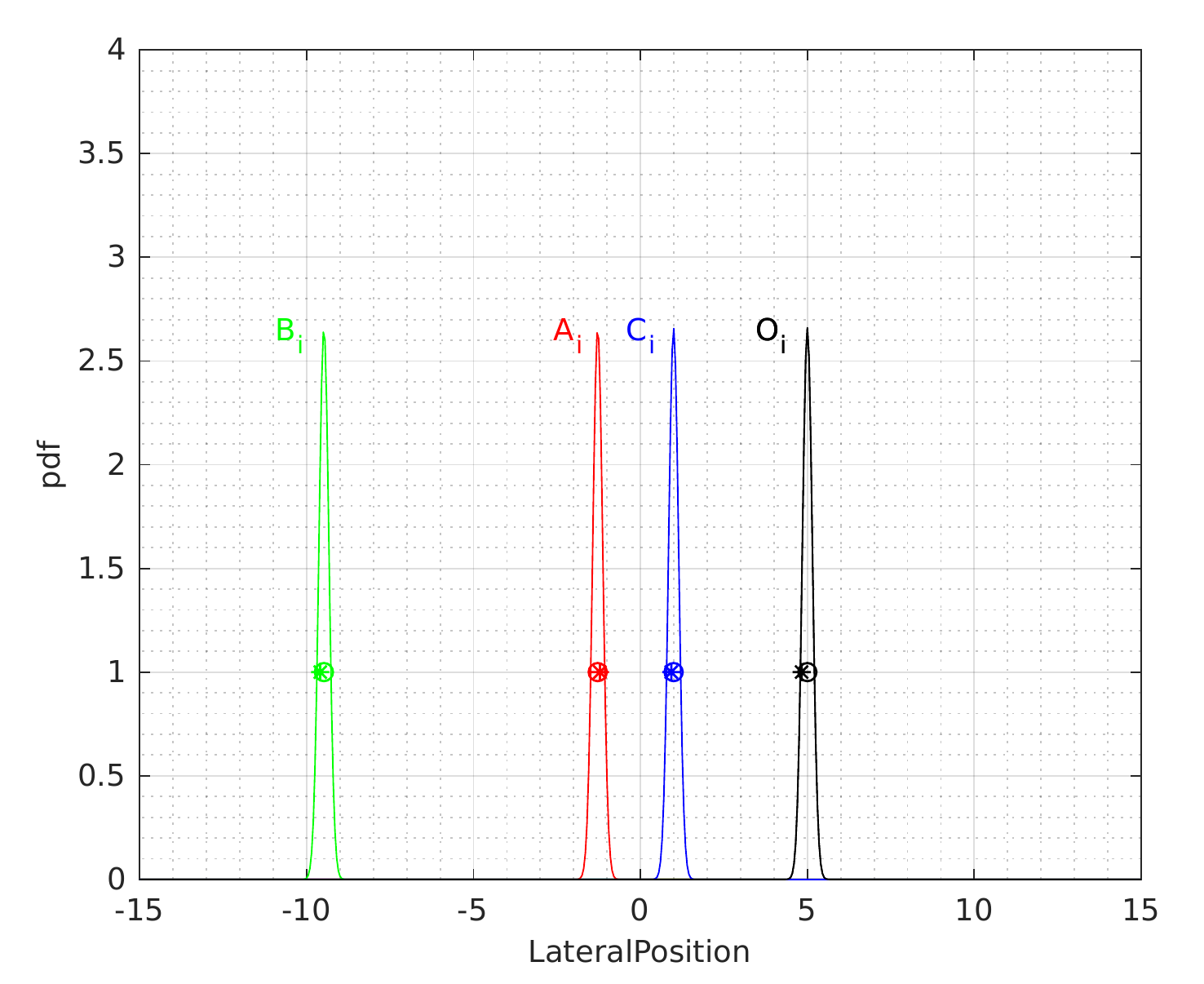}}
\subfigure[\textbf{The current point set$\{\mathcal{M}\}$ }]{\label{fig:b}\includegraphics[width=60mm]{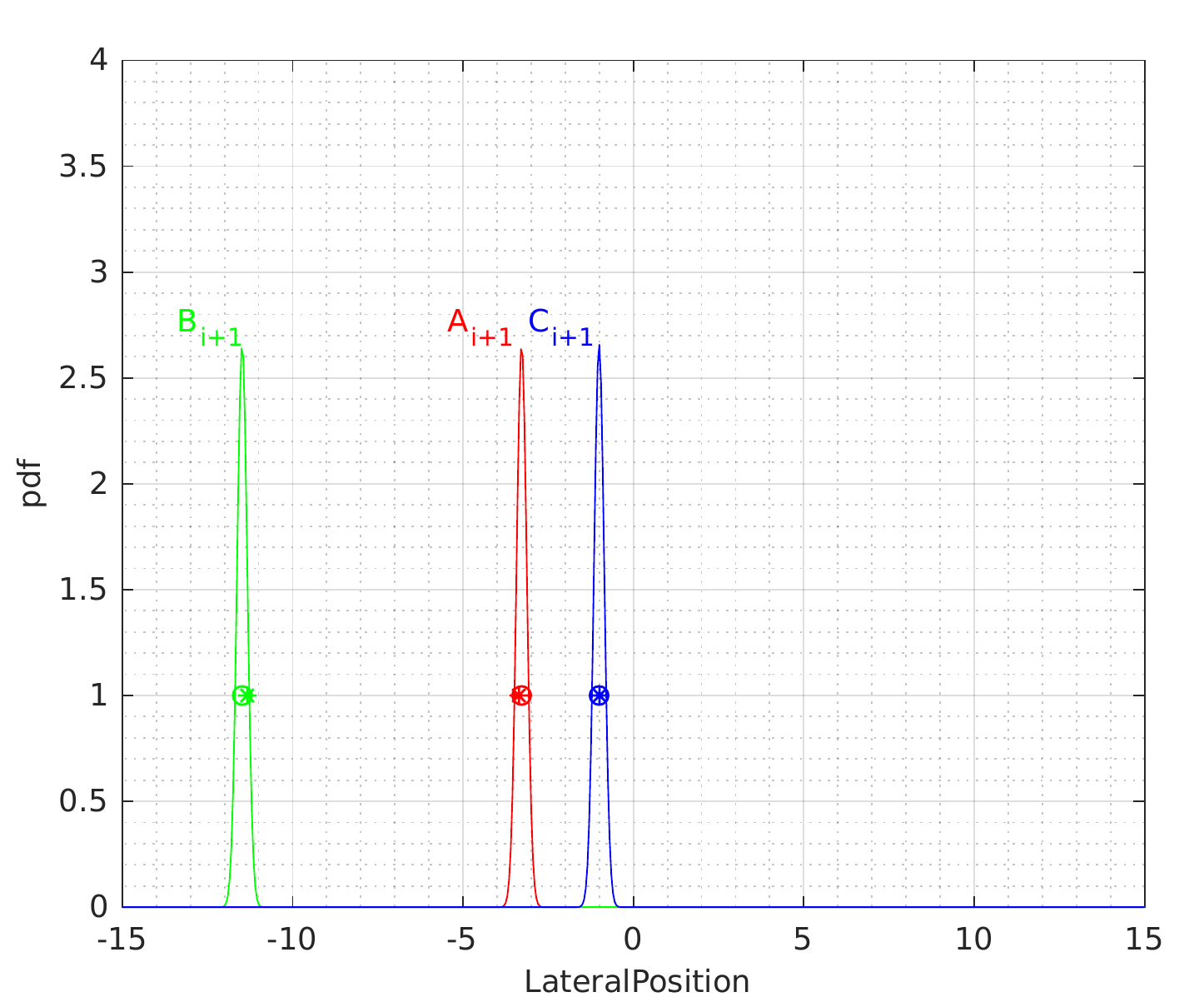}}
\subfigure[\textbf{The cost surface for all targets}]{\label{fig:a}\includegraphics[width=60mm]{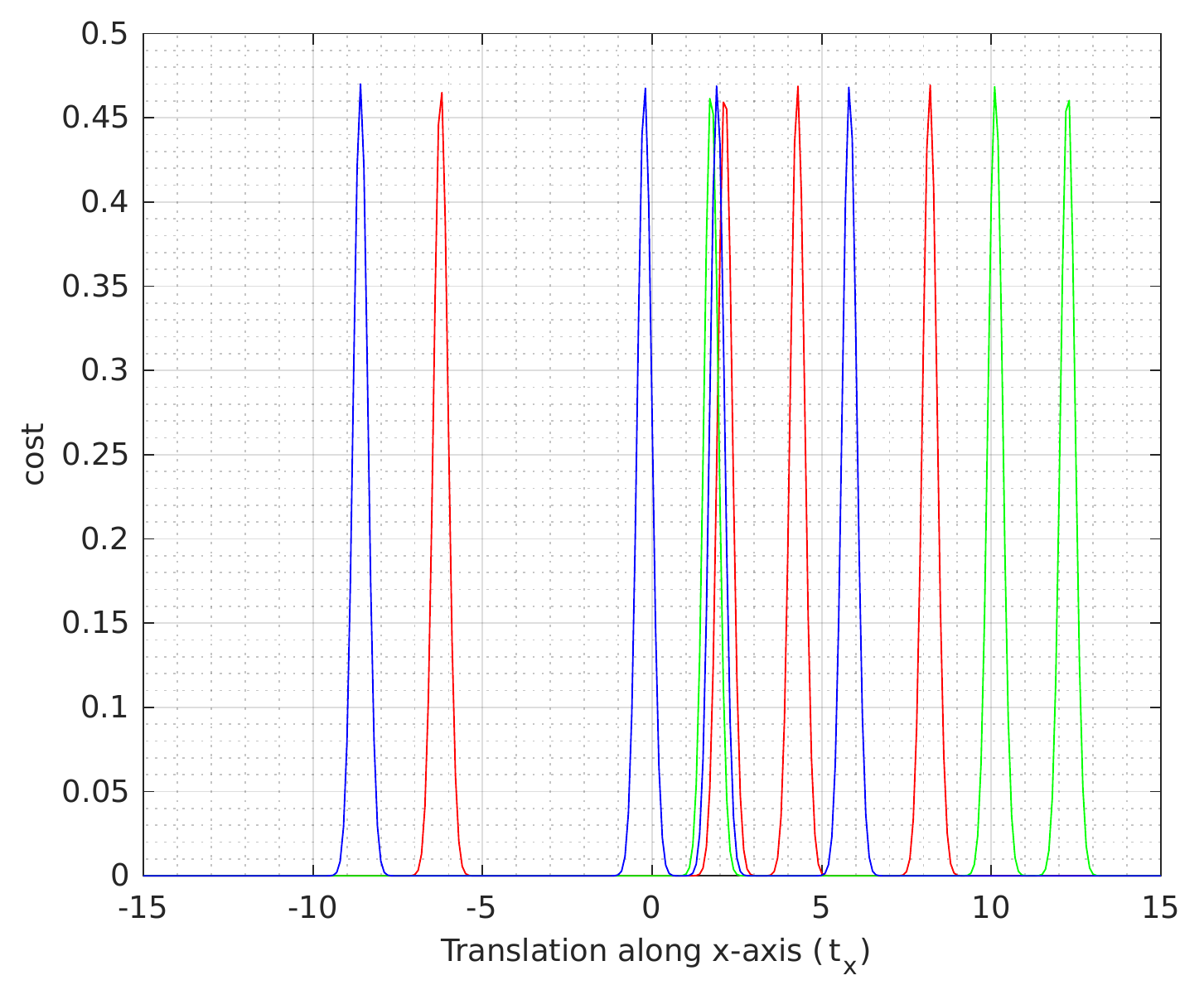}}
\caption[An outlier in previous point set]{One dimensional synthetic scenario, where both point sets have the same standard deviation. However,the previous point set $\{F\}$ have one outlier $\{ O_{i}\}$. The outlier introduces one more possibility in each target cost surface.}
\label{fig:outlier_prv}
\end{figure}

\begin{figure}[htb!]
\centering
\subfigure[\textbf{ The summing approach}]{\label{fig:b}\includegraphics[width=62mm]{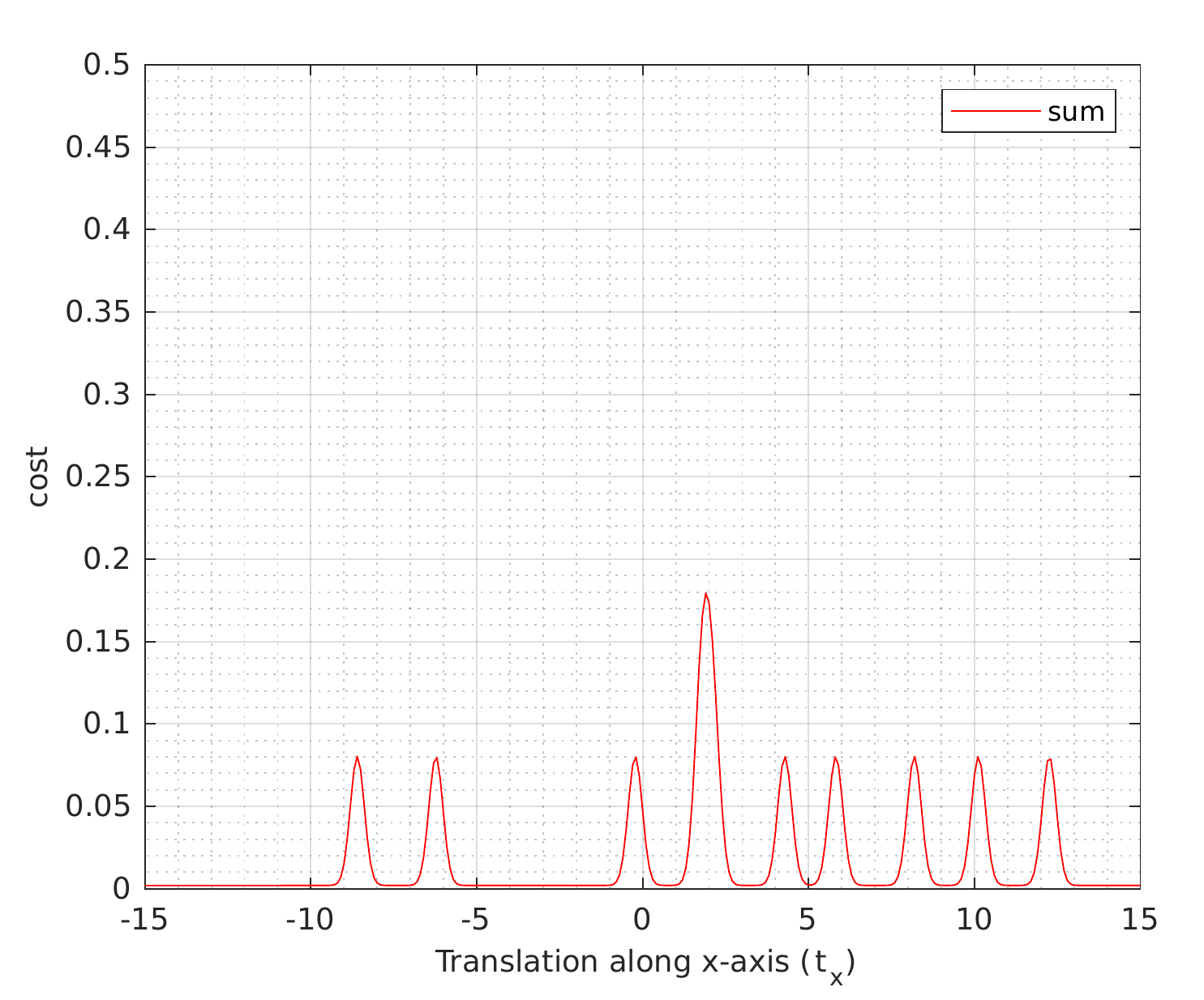}}
\subfigure[\textbf{ The likelihood approach}]{\label{fig:b}\includegraphics[width=60mm]{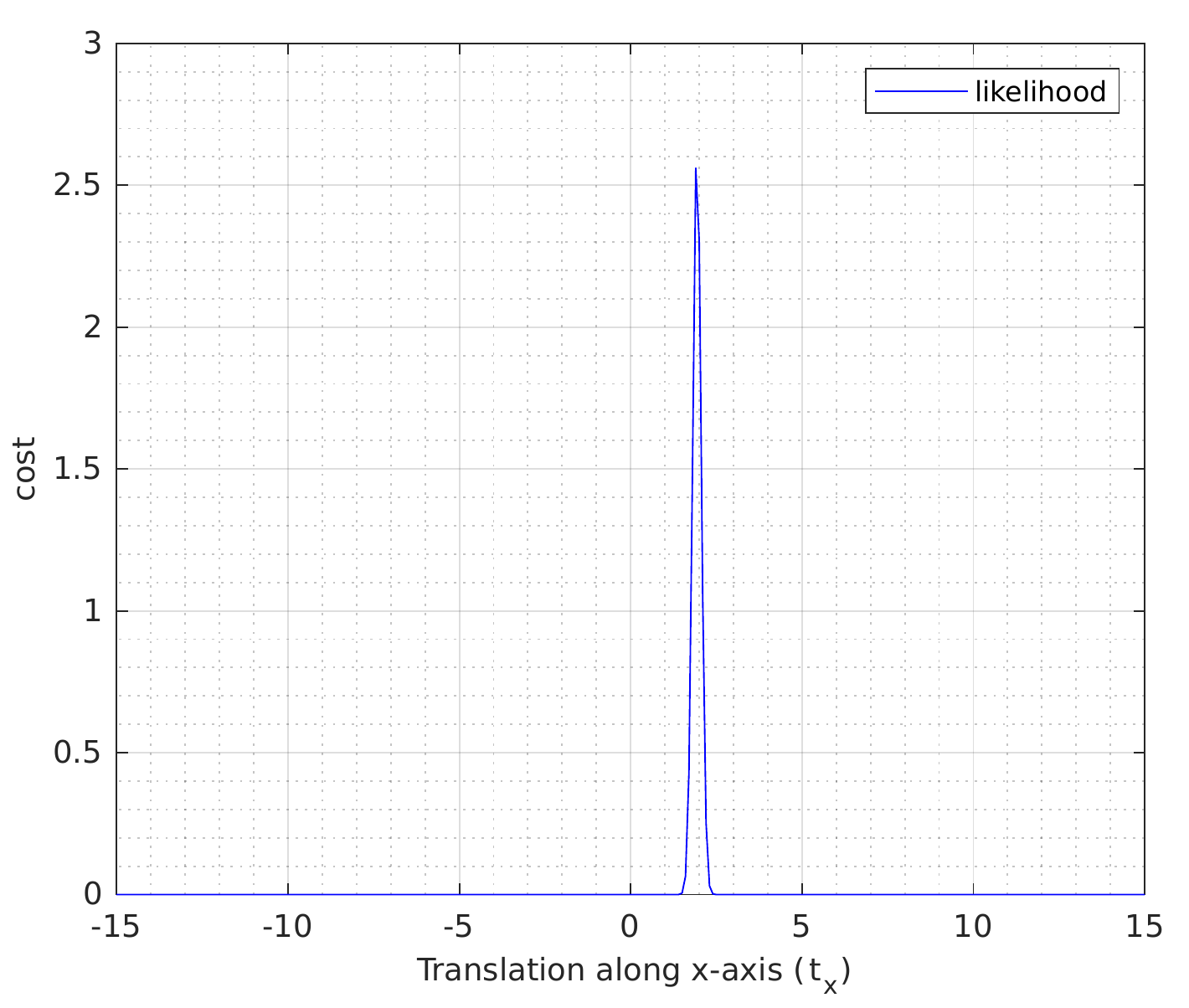}}
\caption[The cost surface based on an outlier in the previous point set]{ This plot shows the impact of the outlier in the previous point set. (a) Express the cost surface based on the summing method, where includes more local peaks as the effect of the outlier. (b) Show the cost based on the likelihood method, where it reveals a robust behavior against this type of outlier.}
\label{fig:cost_outlier_prv}
\end{figure}

One possibility is to have a point in the previous point set, and it is not in the current point set, as shown in Fig. \ref{fig:outlier_prv} (a) and (b). The cost surface for each target in the current point set will be changed because the outlier in the point set introduces the fourth possibility, as depicted in Fig. \ref{fig:outlier_prv} (c). on the one hand, the summing approach introduces more local peaks. On the other hand, the likelihood way shows a robust behavior against this type of outlier, where the likelihood cost surface has only one global peak at the correct translation, as shown in Fig. \ref{fig:cost_outlier_prv}.

\subsection*{Outlier in Current Point set}

\begin{figure}[htb!]
\centering
\subfigure[\textbf{  The previous point set$\{\mathcal{F}\}$  }]{\label{fig:a}\includegraphics[width=60mm]{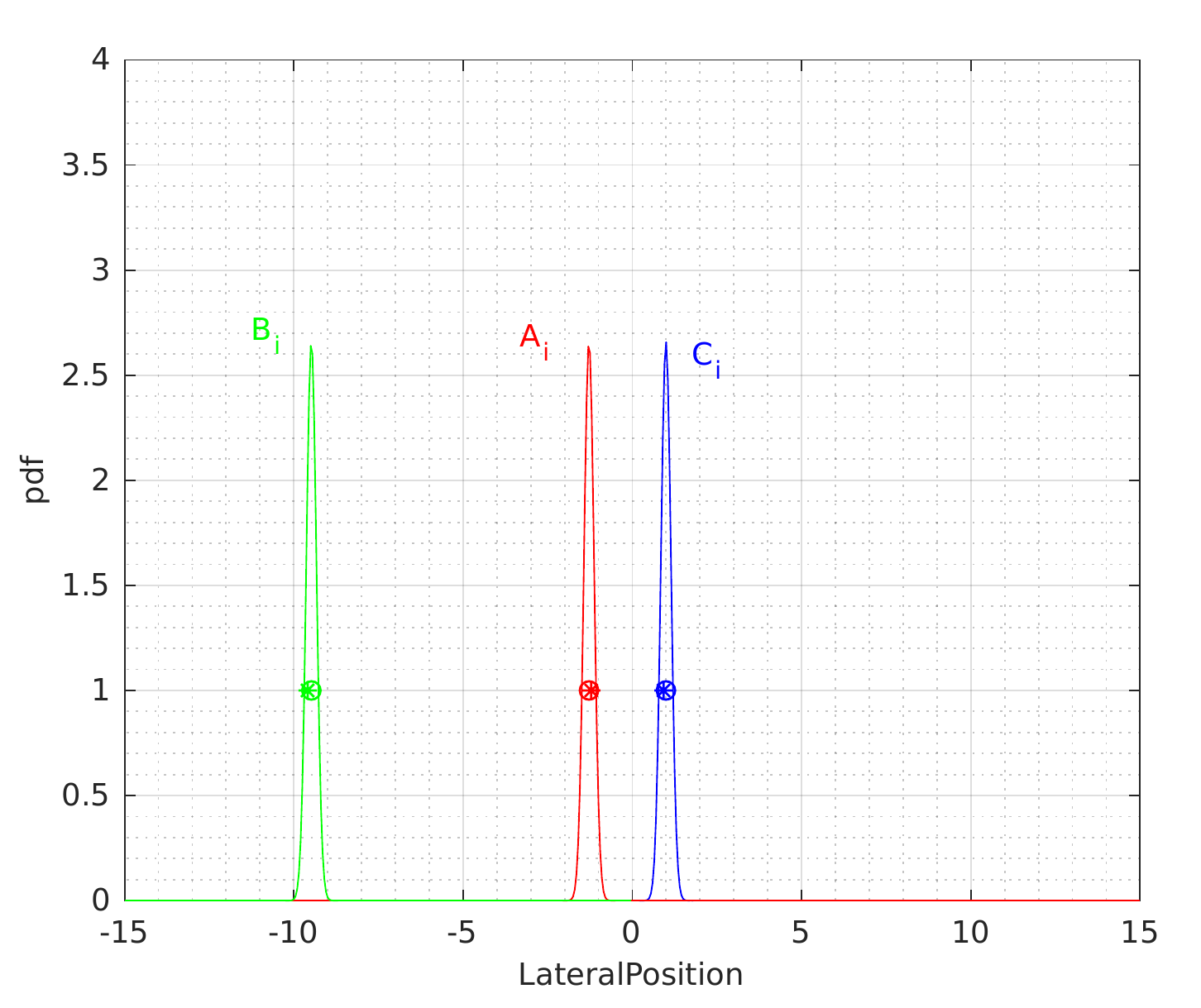}}
\subfigure[\textbf{ The current point set$\{\mathcal{M}\}$ }]{\label{fig:b}\includegraphics[width=60mm]{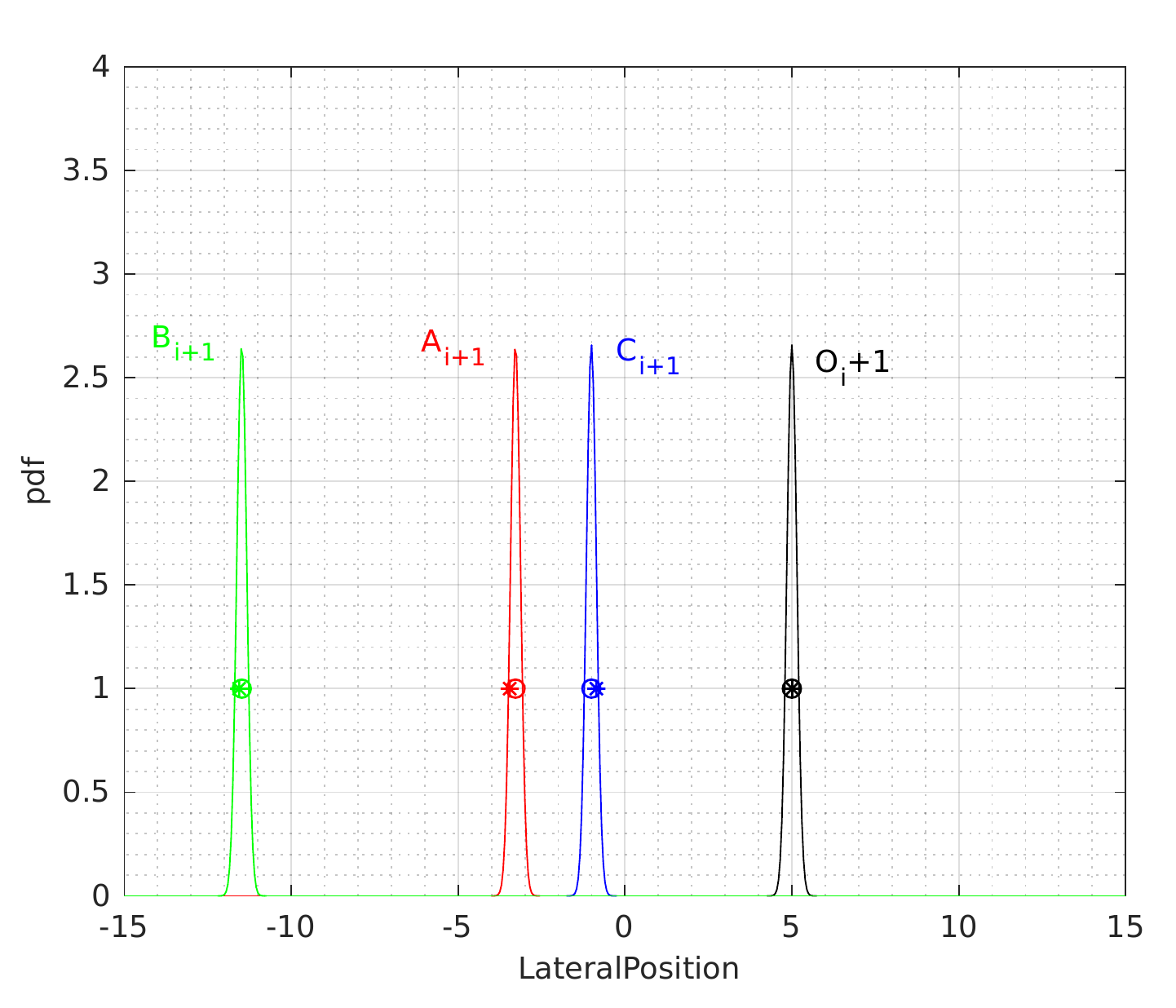}}
\subfigure[\textbf{  The cost surface for all targets  }]{\label{fig:a}\includegraphics[width=60mm]{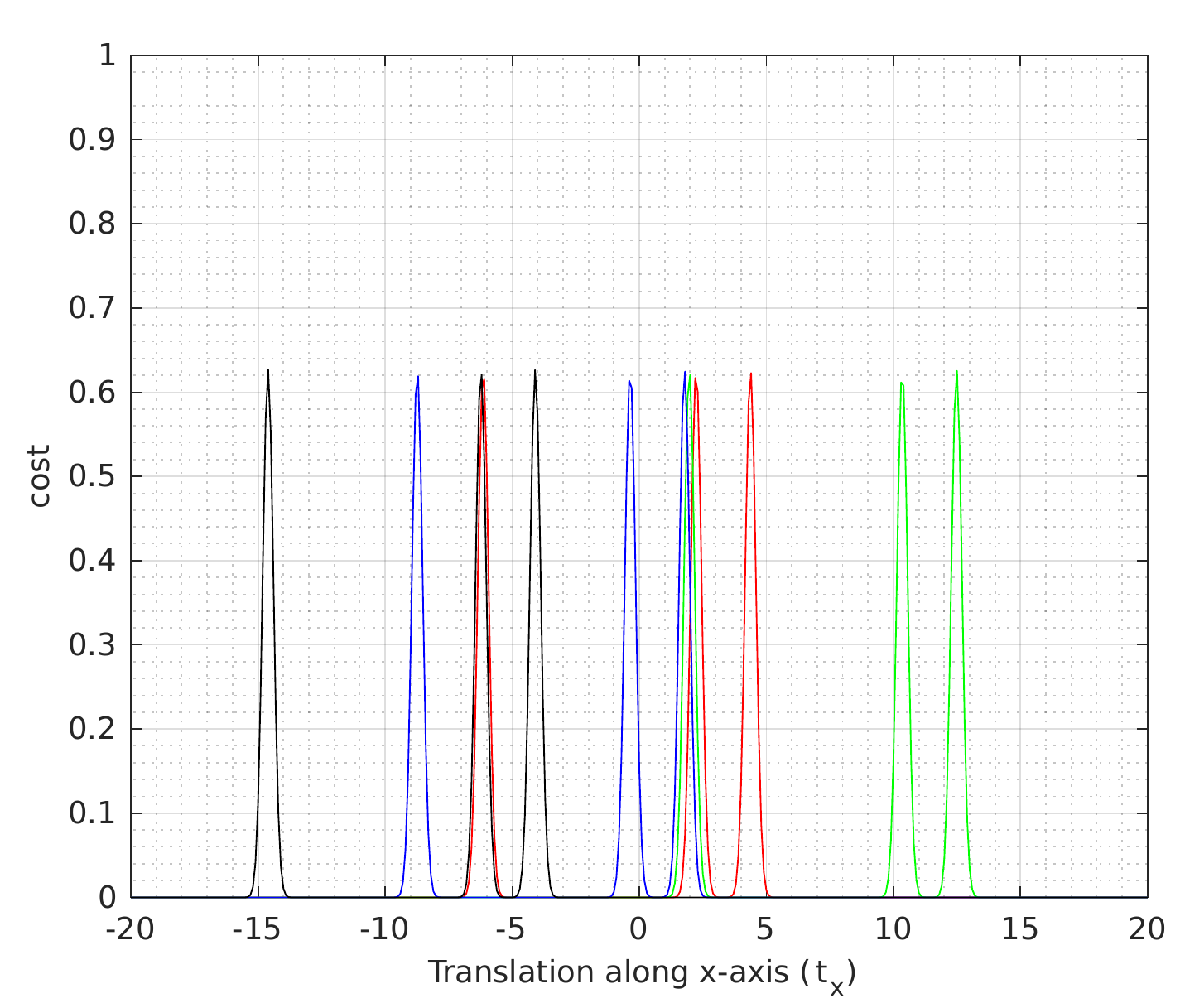}}
\caption[An outlier in the current point set]{One dimensional synthetic scenario, Where both point set have the same the standard deviation. However,the current point set $\{M\}$ icludes one outlier $\{ O_{i+1}\}$. The outlier adds one more cost surface to the overall cost surface show in black. }
\label{fig:outlier_cur}
\end{figure}

\begin{figure}[htb!]
\centering
\subfigure[\textbf{ The summing approach }]{\label{fig:b}\includegraphics[width=60mm]{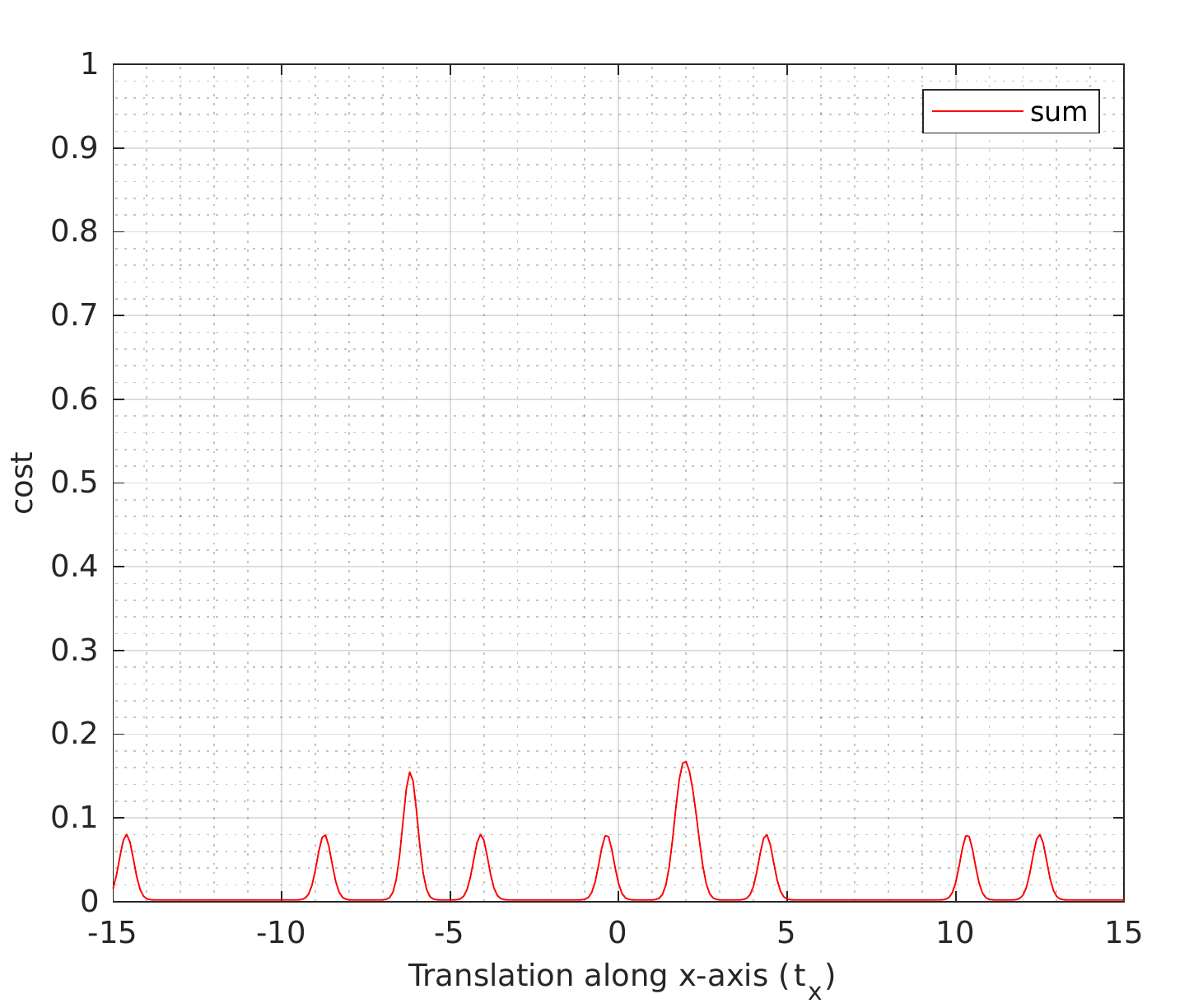}}
\subfigure[\textbf{ The likelihood approach }]{\label{fig:b}\includegraphics[width=60mm]{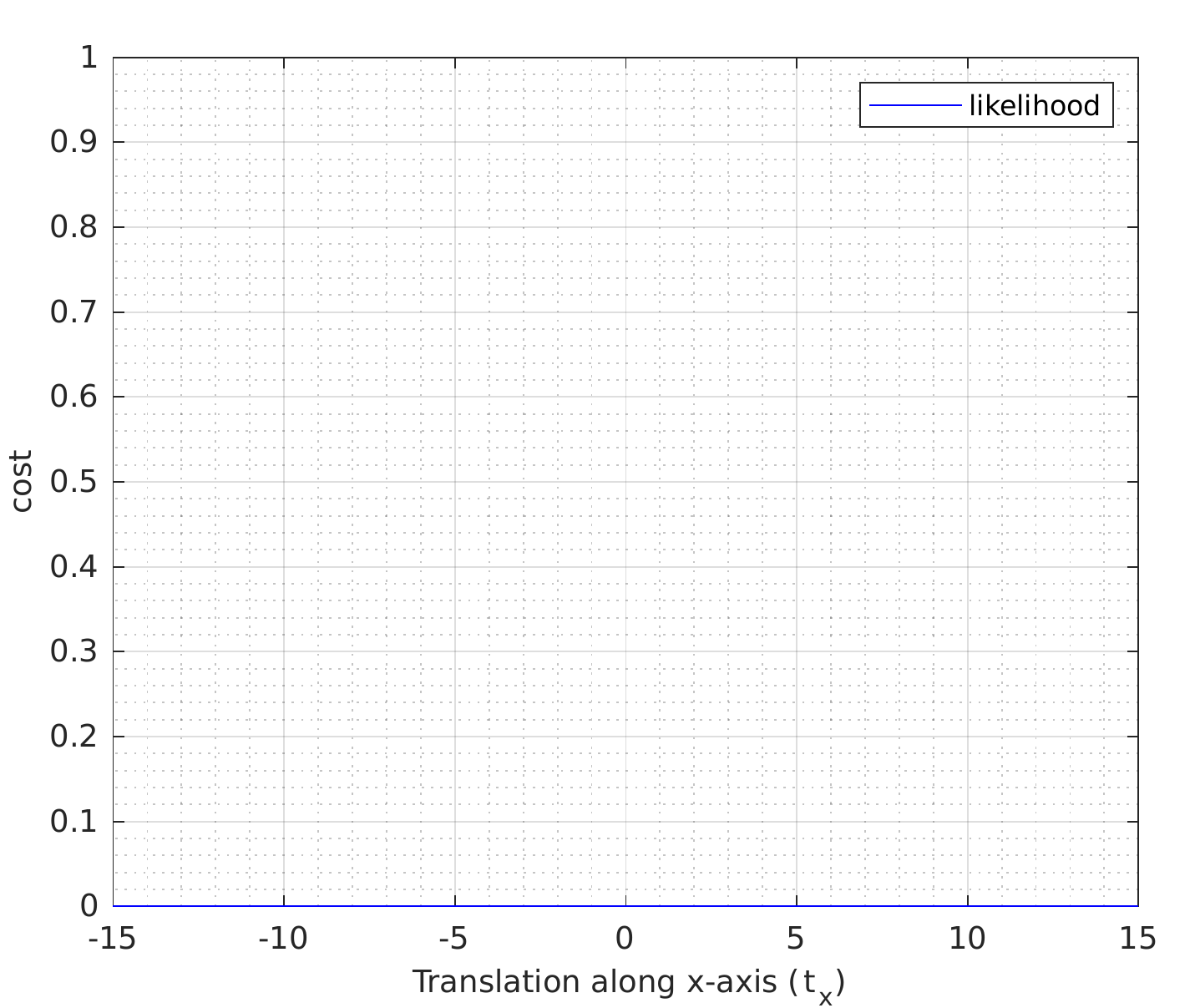}}
\caption[The cost surface based on an outlier in the current point set]{The cost surface for all targets after adding one outlier $\{ O_{i+1}\}$ to the current point set $\{M\}$. (a) Represent the cost surface for the summing approach, where more local peaks exist due to the outlier. Moreover, one local peak has a large value comparing the other peaks, but the global peak matches the correct translation. (b) Represent the likelihood approach cost surface, where it is zero. Where the outlier cost surface at the correct translation is zero, so the total product equals zero.}
\label{fig:cost_outlier_cur}
\end{figure}

A second possibility is to have a point in the current point set and not in the previous point set, as shown in Fig. \ref{fig:outlier_cur}. The cost surface for each target will not change, but the outlier introduces a new cost surface. Therefore, the outlier does not affect each target cost surface, but it affects the overall cost surface, as depicted in Fig. \ref{fig:cost_outlier_cur}.\\

In the summing approach, the outlier introduces more local peaks. Moreover, it makes one local peak has a competitive value to the global peak. However, it has more impact on the likelihood method, the overall cost surface becomes zero. The likelihood multiplies the cost surface for each target. The cost surface for the outlier has zero value at the correct transformation $(t_x = 2)$. Even so, Zisserman introduces the M-estimator concept to overcome the problem. Which models the outlier as well as the inlier in the cost function.

\section*{Robust Cost Function}

Although the outlier influenced both approaches, this work uses the likelihood approach. This section explains one possible method to overcome the outlier effect. Biber in ~\cite{biber2004probabilistic} introduced the robust cost function term. Zisserman in~\cite[Appendix. 6]{hartley2003multiple} explains the whole idea behind the robust cost function. The main idea is to model the outlier in the cost function. In other words, the idea is to make the model robust against the outlier. Zisserman introduces two ways to model the outlier: Blake–Zisserman and corrupted Gaussian. The cost function based on the distribution-to-distribution, and the likelihood is:

\begin{equation}
\label{eq:inlier_d2d}
f_{\mathrm{d2d}}(\boldsymbol{\theta}) =  \prod_{k \in \mathcal{M}}\left(\sum_{i=1}^{\vert {\cal F}\vert } w_{i} \cdot p \left( 0|\boldsymbol{T}(\boldsymbol{m_k},\boldsymbol{\theta})-\boldsymbol{\mu_{i}}, \boldsymbol{T}(\boldsymbol{\Sigma_k},\boldsymbol{\theta}) + \boldsymbol{\Sigma_{i}}\right)\right)
\end{equation}

\subsection*{Blake–Zisserman}

\begin{figure}[htb!]
\centering
\subfigure[\textbf{  The cost surface for all targets  }]{\label{fig:a}\includegraphics[width=70mm]{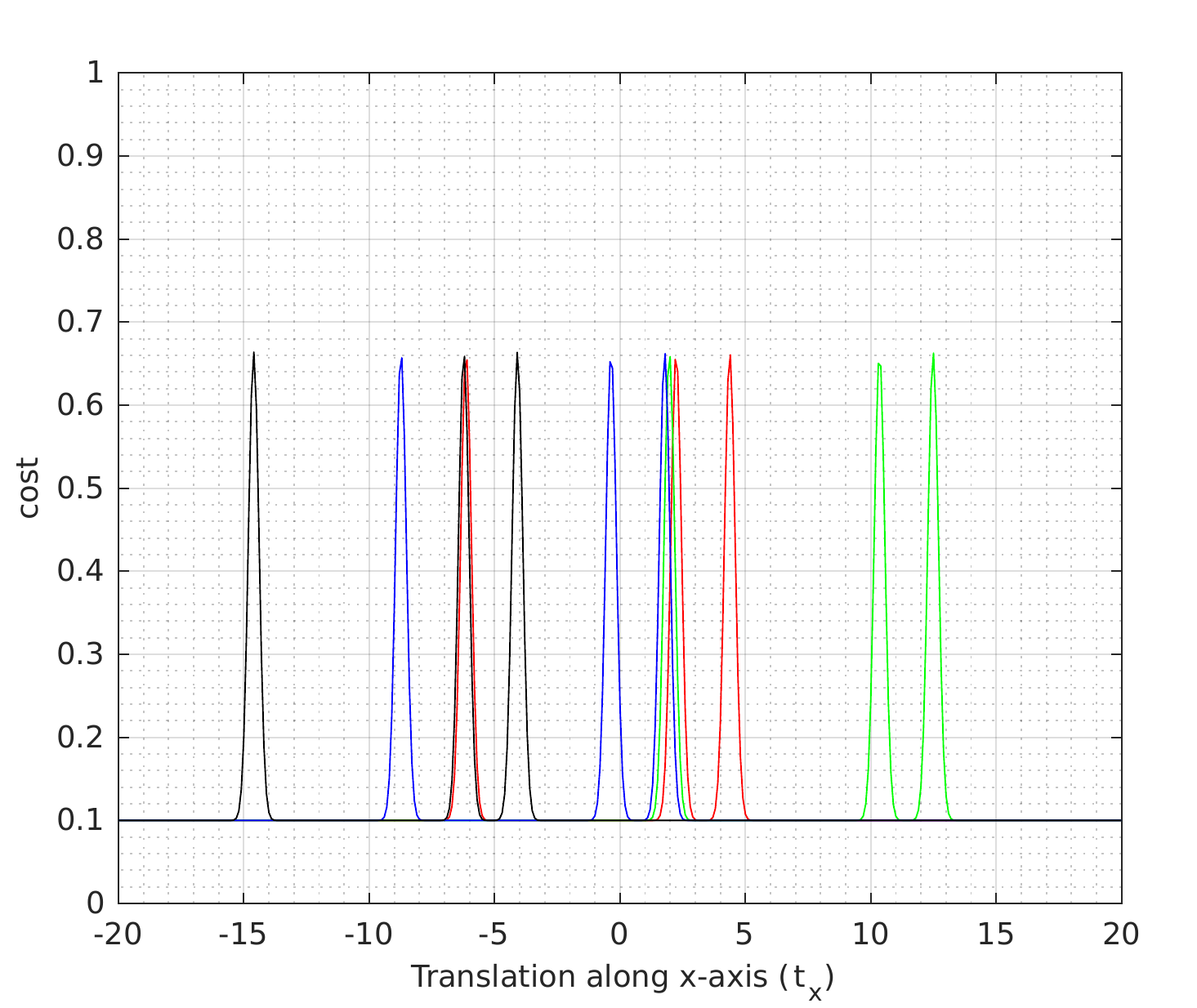}}
\subfigure[\textbf{ The likelihood approach }]{\label{fig:b}\includegraphics[width=70mm]{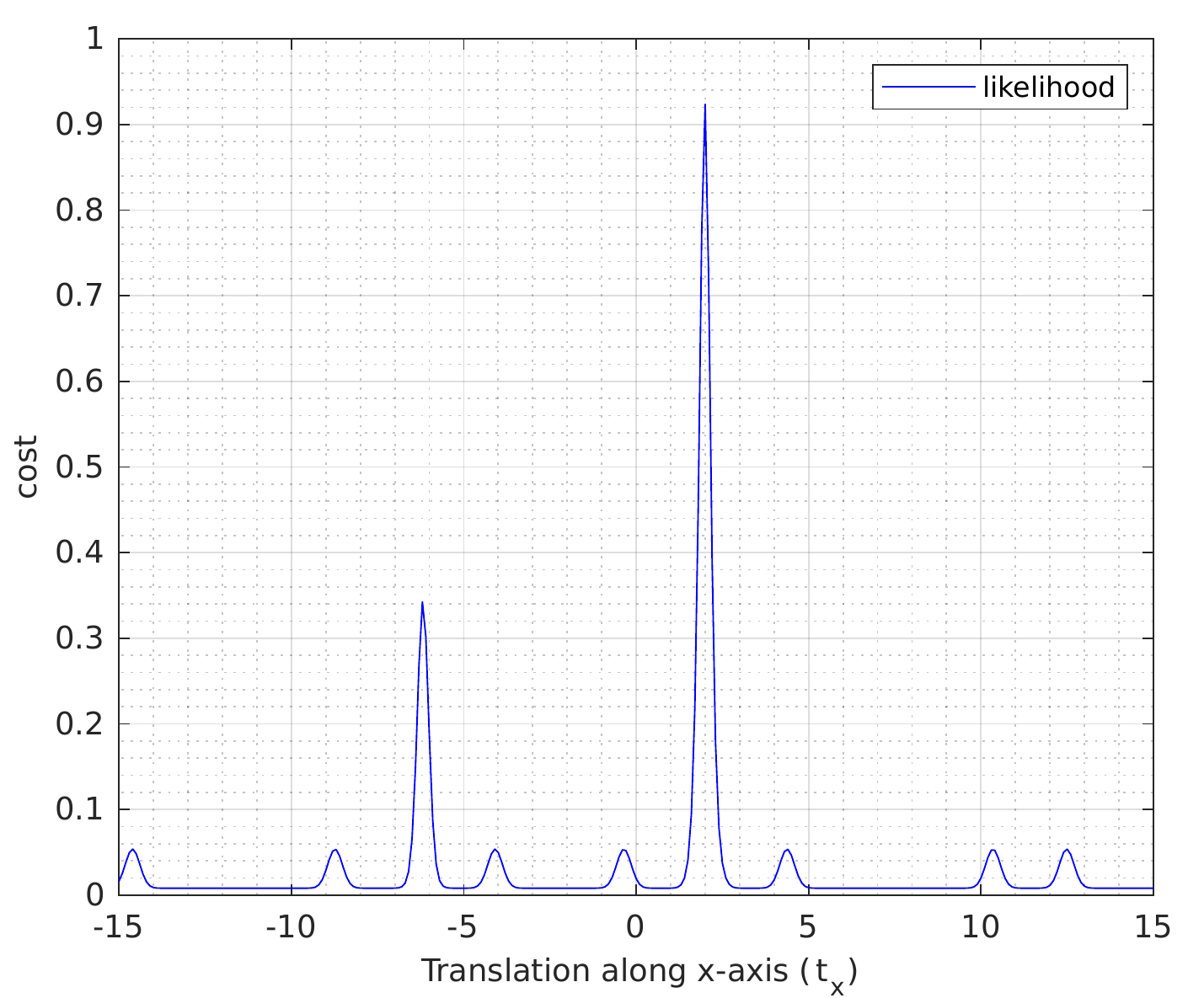}}
\caption[Model the outlier component based on Blake–Zisserman ]{The cost for the problem based on the Blake–Zisserman:(a) The cost surface for each target plotted in one plot. (b) The cost surface for all targets based on the likelihood(product) approach.}
\label{fig:cost_Blak}
\end{figure}

This method models the outlier as uniform distribution, as follows:

\begin{equation}
\label{eq:Blake_inlier}
inlier_{component} = p \left( 0|\boldsymbol{T}(\boldsymbol{m_k},\boldsymbol{\theta})-\boldsymbol{\mu_{i}}, \boldsymbol{T}(\boldsymbol{\Sigma_k},\boldsymbol{\theta}) + \boldsymbol{\Sigma_{i}}\right)
\end{equation}

\begin{equation}
\label{eq:Blake_outlier}
f_{\mathrm{d2d}}(\boldsymbol{\theta})= \prod_{k \in \mathcal{M}} \sum_{i=1}^{ | \mathcal{F|}} w_{i}\left(inlier_{component} + outlier_{component} \right)
\end{equation}

Where, the $ \boldsymbol{outlier_{component}}$ is a uniform distribution. Fig. \ref{fig:cost_Blak} exhibits the cost surface for the problem after modeling the outlier as uniform distribution, comparing to Fig. \ref{fig:cost_outlier_cur}, the likelihood approach has one global peak at the correct translation. Moreover, all local peaks are not comparable to the global one, which is not the case in the summing approach.

\subsection*{Corrupted Gaussian}
\begin{figure}[htb!]
\centering
\subfigure[\textbf{  The cost surface for all targets  }]{\label{fig:a}\includegraphics[width=70mm]{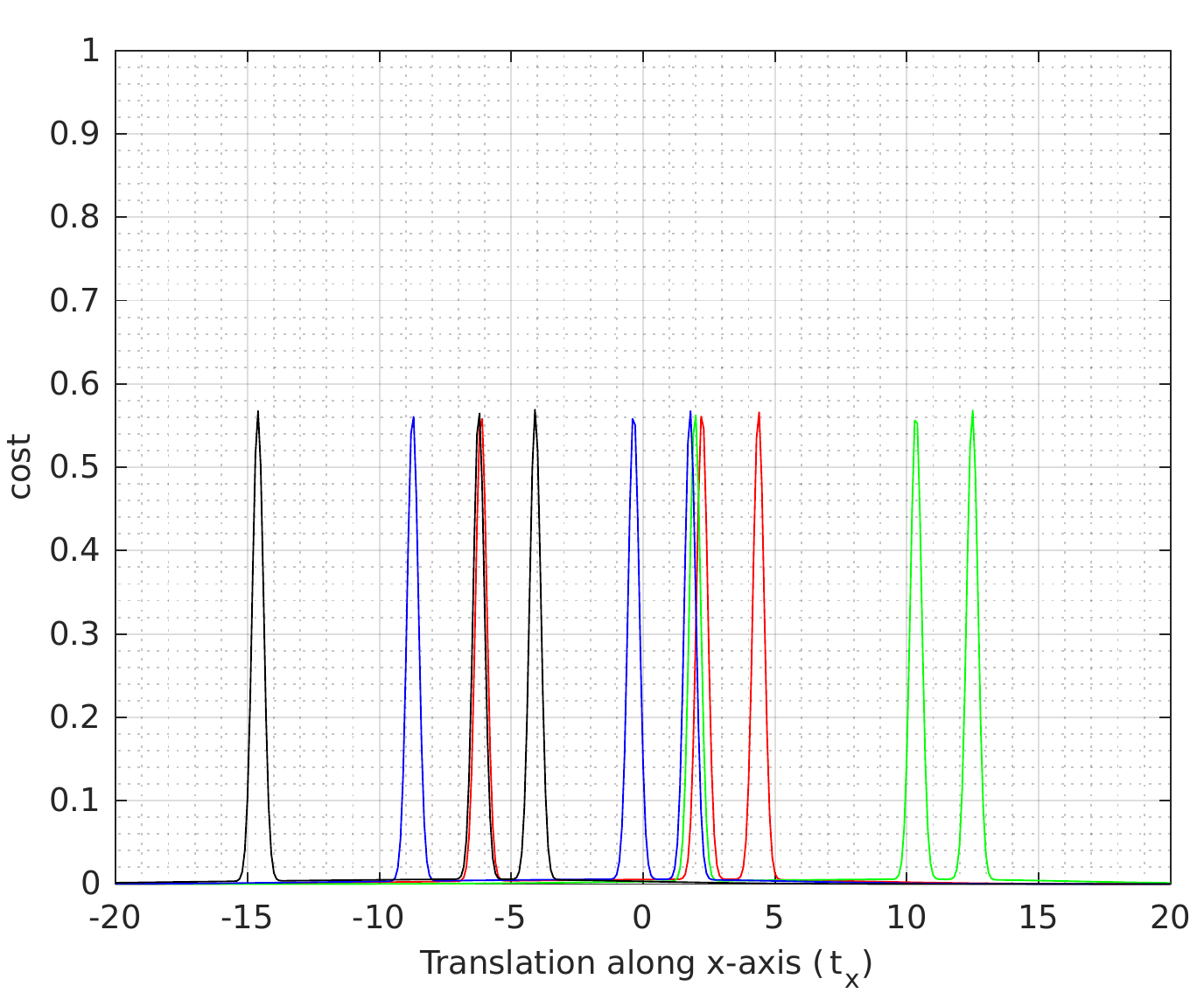}}
\subfigure[\textbf{ The likelihood approach }]{\label{fig:b}\includegraphics[width=70mm]{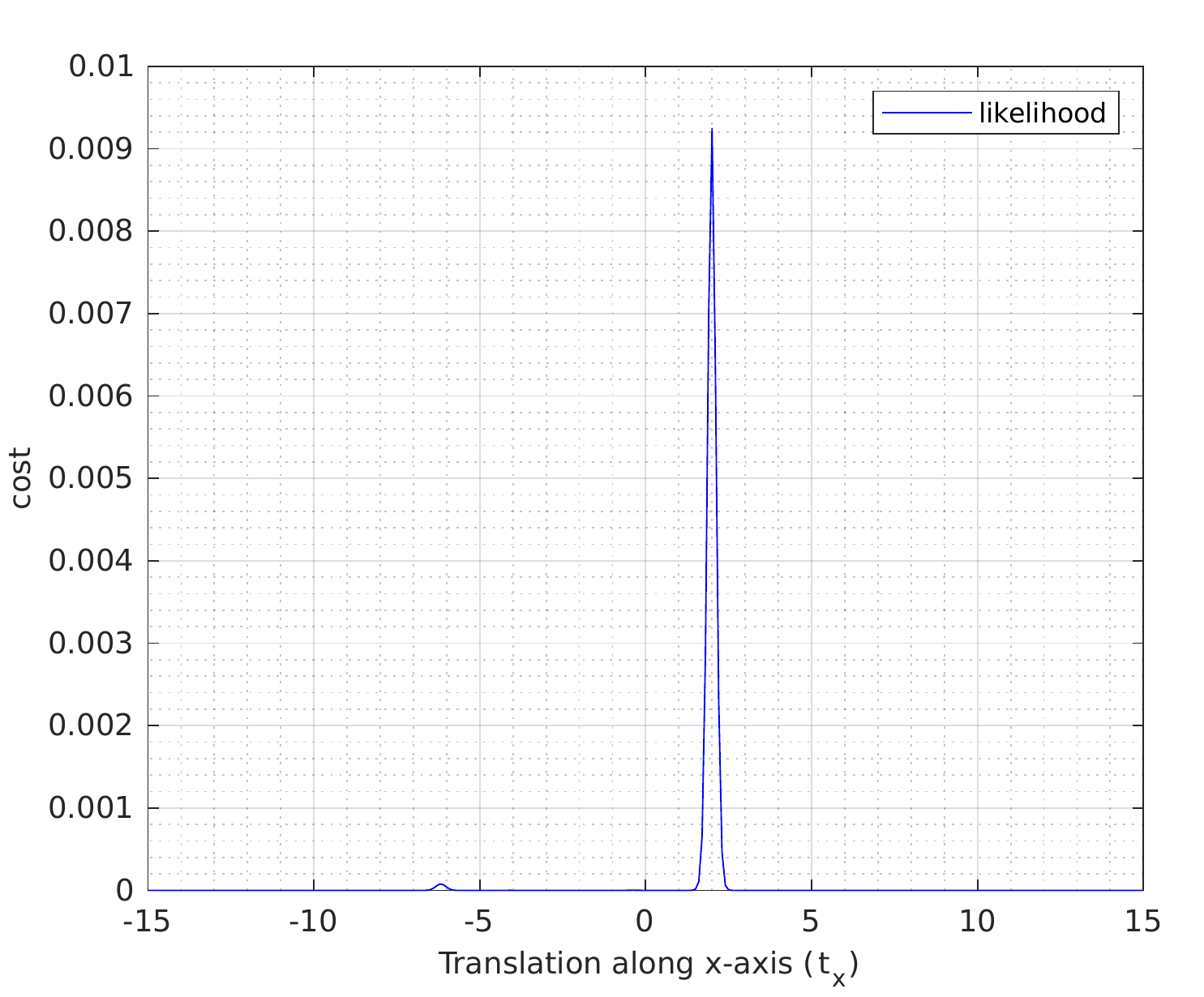}}
\caption[Model the outlier component based on Corrupted Gaussian]{The cost surface based on the corrupted Gaussian, (a) The cost surface for each target plotted in one plot. (b) The cost surface for all targets based on the likelihood(product) approach.}
\label{fig:cost_corrupted}
\end{figure}

Comparing to the previous method, this method models the outlier as a Gaussian distribution, as follows:

\begin{equation}
\label{eq:Corrupted_inlier}
outlier_{component} = p \left(\boldsymbol{T}(\boldsymbol{m_k},\boldsymbol{\theta})| \boldsymbol{\mu}_{i},\boldsymbol{\Sigma}_{Outlier}\right)
\end{equation}

\begin{equation}
\label{eq:Corrupted_outlier}
f_{\mathrm{d2d}}(\boldsymbol{\theta})= \prod_{k \in \mathcal{M}} \sum_{i=1}^{ | \mathcal{F|}} w_{i}\left( \alpha \cdot inlier_{component} + (1-\alpha)\cdot outlier_{component} \right)
\end{equation} \\

Where, $\boldsymbol{\Sigma}_{Outlier}$ represents the covariance of the outlier component, $\alpha$ is the inlier to outlier fraction ratio.\\

Fig. \ref{fig:cost_corrupted} depicts the cost surface based on the corrupted gaussian approach to model the outlier component. The first observation is, the summing approach cost surface behaves the same but with a change in the cost values. However, the cost surface based on the likelihood that all local peaks are tiny compared to the global peak. Nevertheless, modeling the outlier component's parameters $(\boldsymbol{{\Sigma}_{Outlier},\alpha})$ are very crucial as it influences the cost surface.\\

\subsection*{Discussion}

The uniform distribution to model the outlier component reveals a robust behavior against the outlier, but it violates the gaussian assumption. Which is, the area under the curve does not equal to one. However, the corrupted gaussian approach does not violate any Gaussian assumption and has a robust behavior.

\section{Clustered Points}

\begin{figure}[htb!]
\centering
\subfigure[\textbf{  The previous point set  }]{\label{fig:a}\includegraphics[width=70mm]{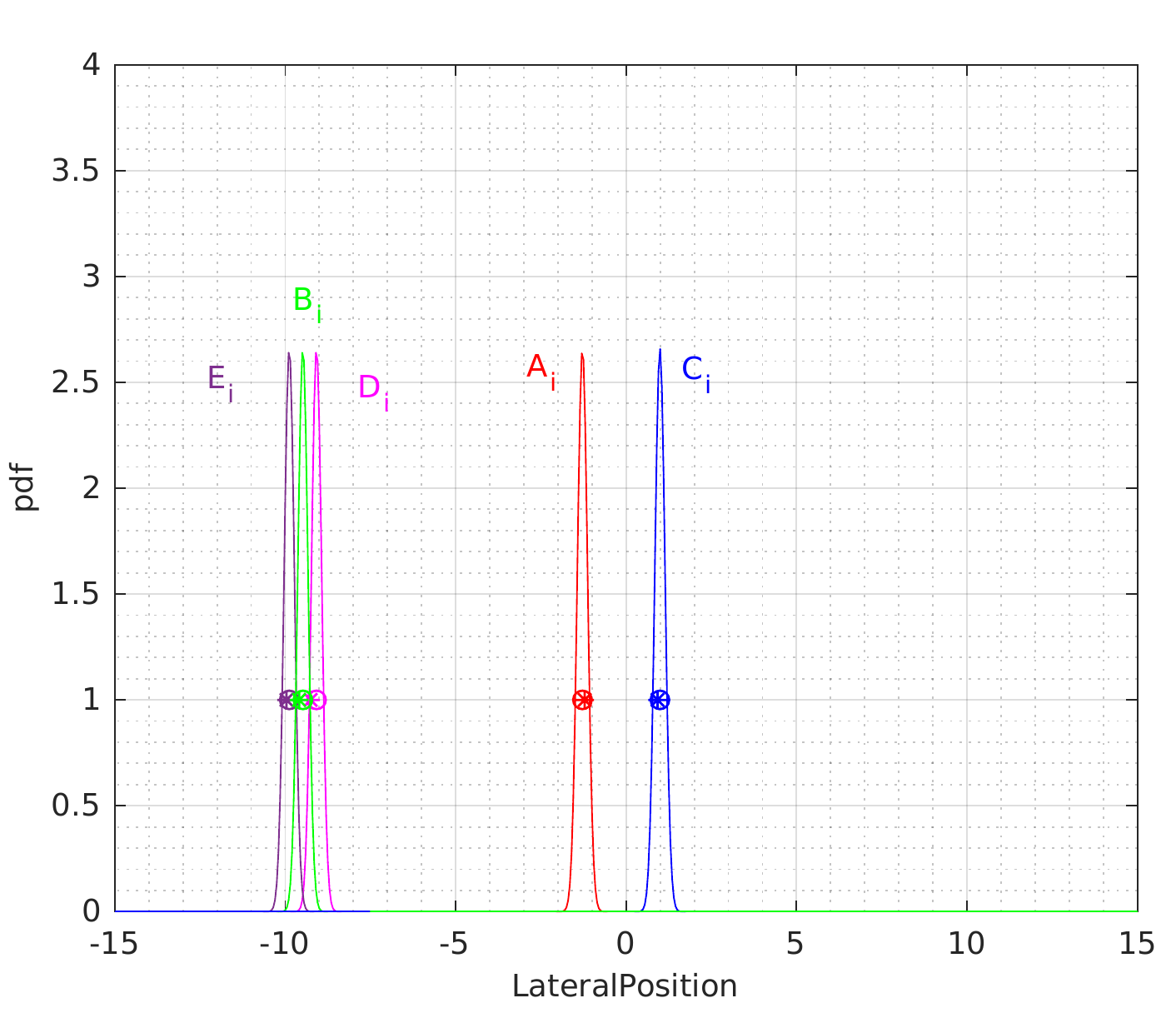}}
\subfigure[\textbf{ The current point set }]{\label{fig:b}\includegraphics[width=70mm]{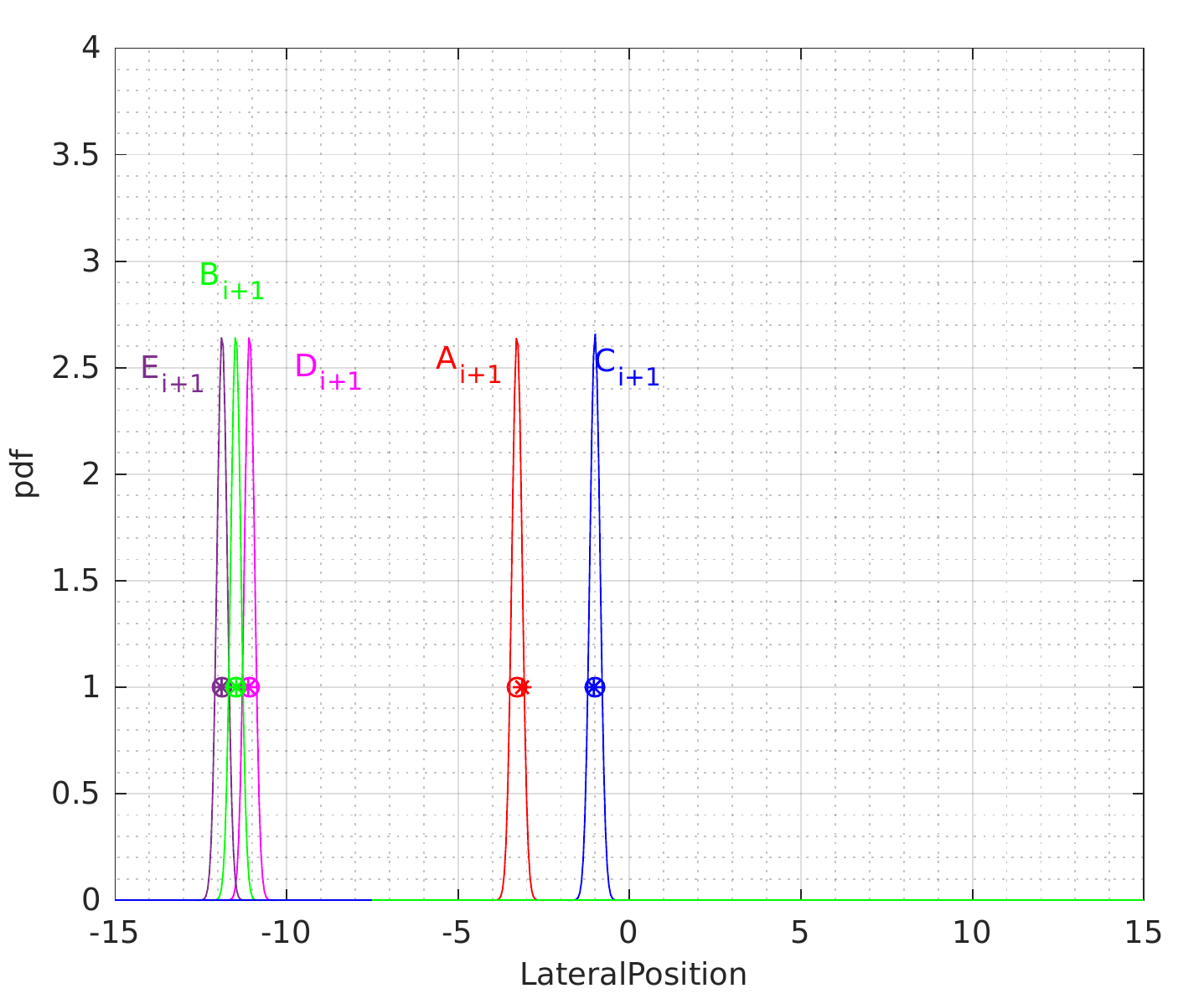}}
\caption[Very closed clustered points]{One dimensional synthetic scenario, where both point sets are fully overlapped. However, this scenario includes some points are very closed from each, which formulates a clustered scenario.}
\label{fig:clus_1}
\end{figure}

\begin{figure}[htb!]
\centering
\subfigure[\textbf{  The previous point set  }]{\label{fig:a}\includegraphics[width=70mm]{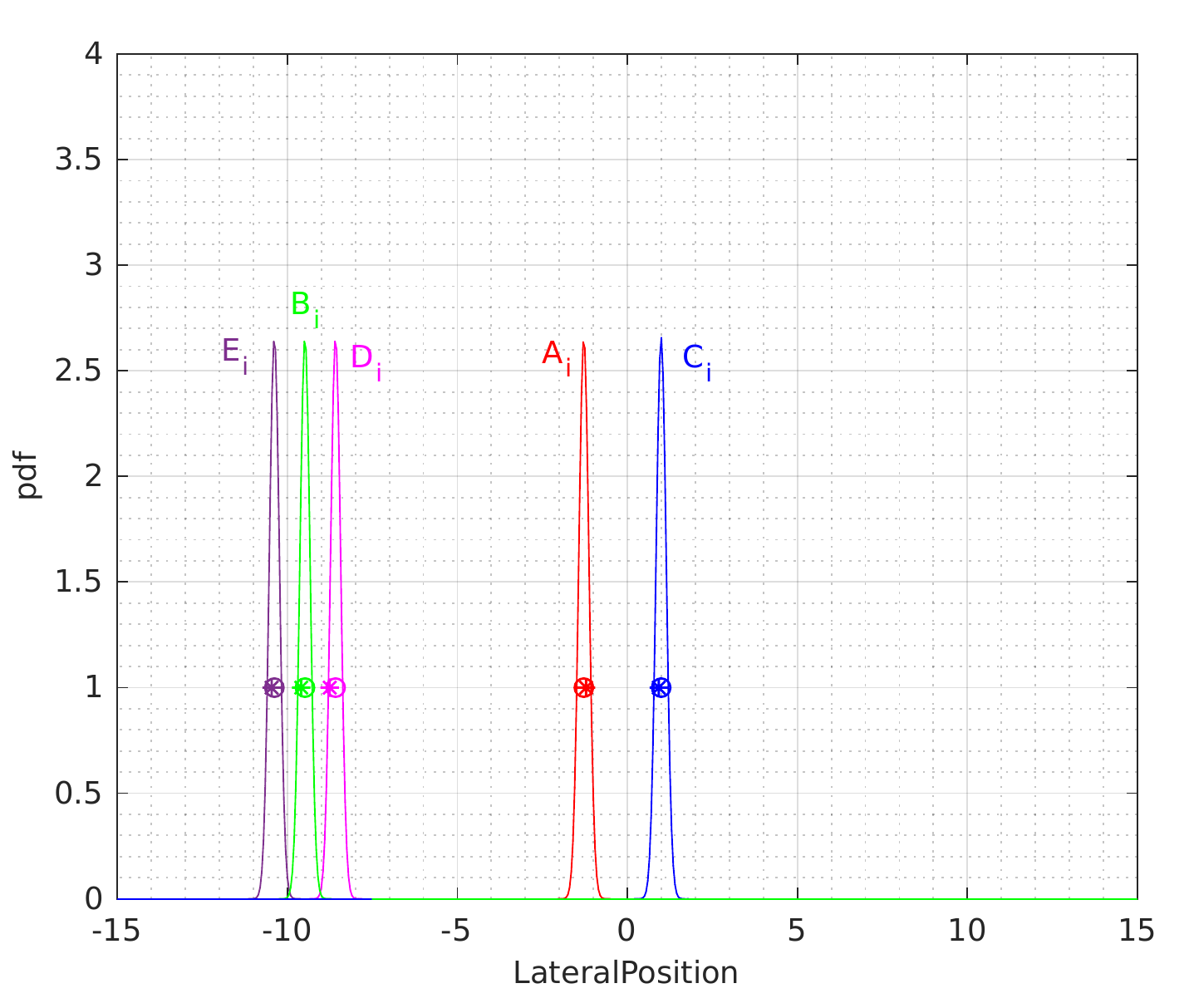}}
\subfigure[\textbf{ The current point set }]{\label{fig:b}\includegraphics[width=70mm]{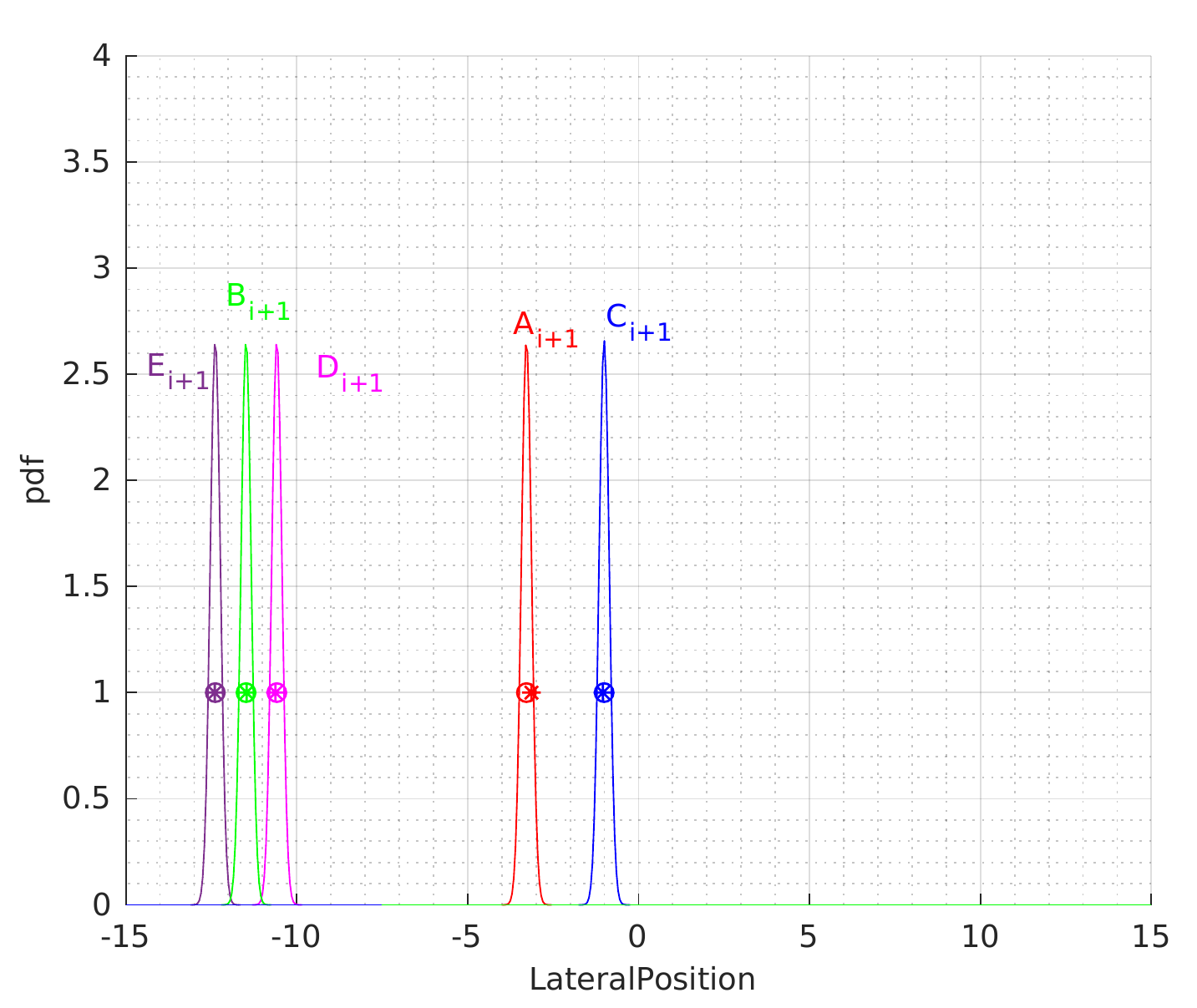}}
\caption[Closed clustered points]{One dimensional synthetic scenario, where both point sets are fully overlapped. However, this scenario includes some points are closed from each, which formulates a clustered scenario.}
\label{fig:clus_2}
\end{figure}

This is a fully overlapped scenario, but some inlier points are near to each other. The clustered points introduce multiple correct correspondences, as this section will discuss. The clustered points can be in two shapes: the points are very closed from each other, as shown in  Fig. \ref{fig:clus_1} or not completely overlapped as shown in Fig. \ref{fig:clus_2}.\\

\begin{figure}[htb!]
\centering
\subfigure[\textbf{ The cost surface for all targets  }]{\label{fig:a}\includegraphics[width=70mm]{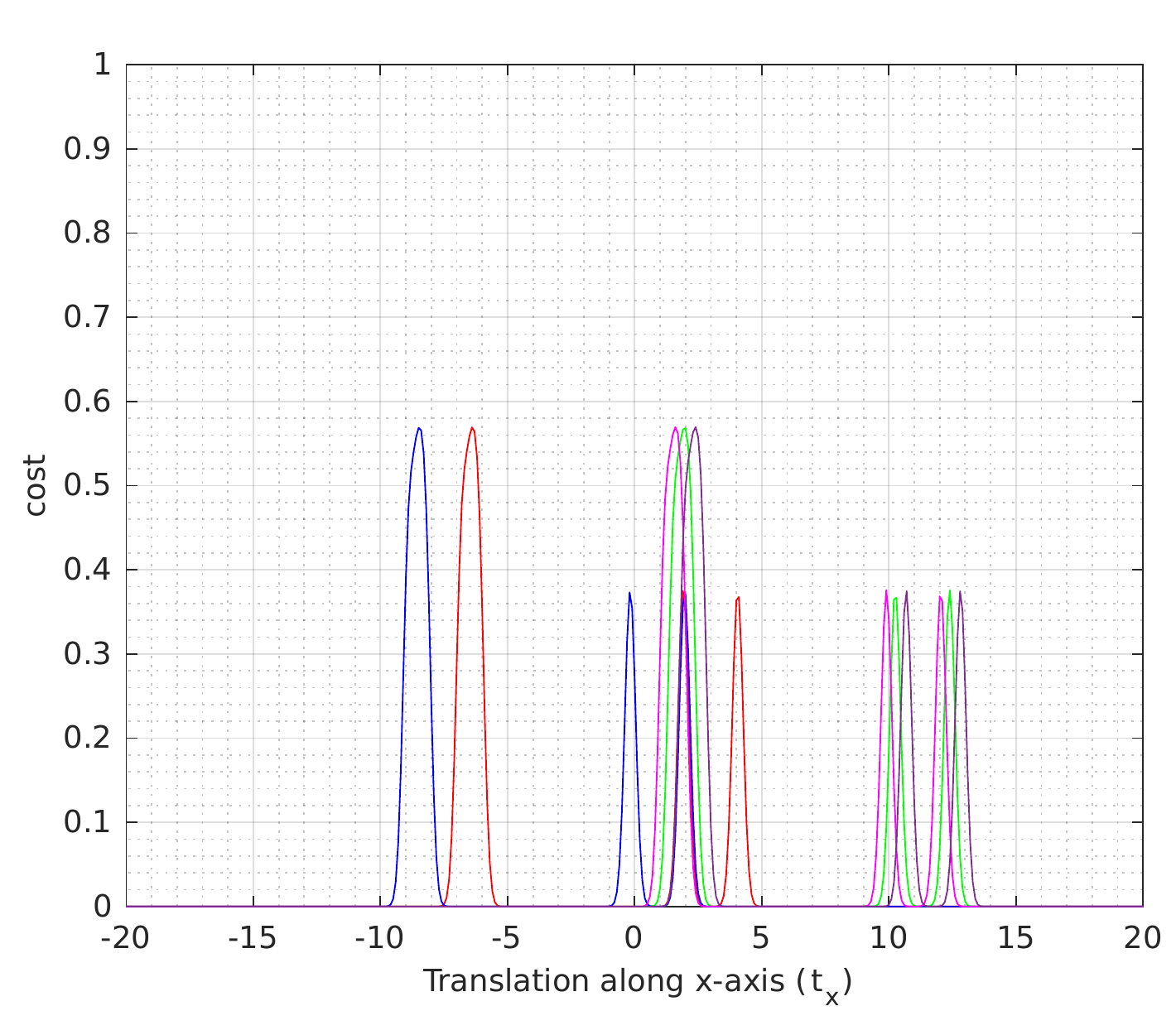}}
\subfigure[\textbf{ The summing approach }]{\label{fig:b}\includegraphics[width=70mm]{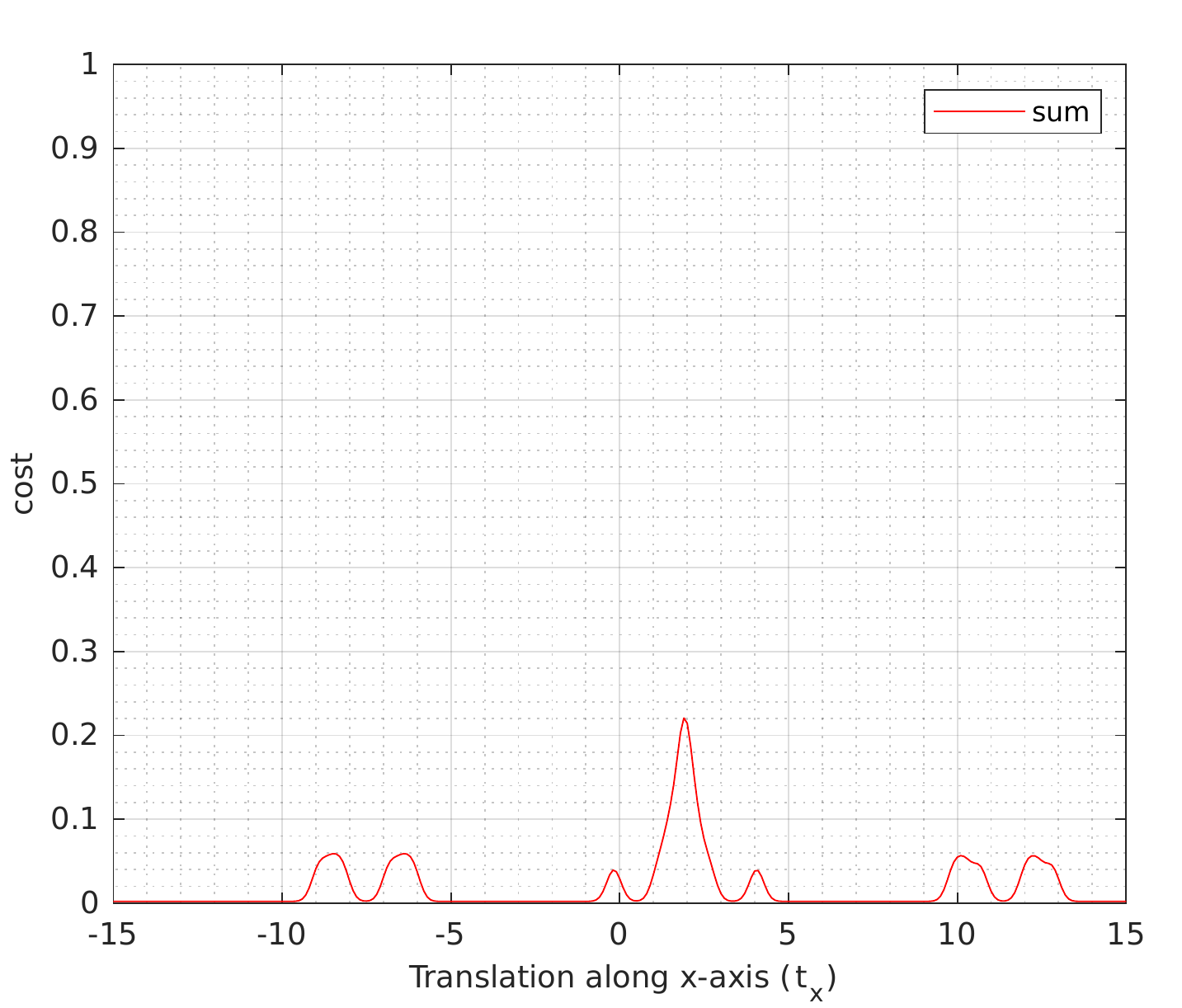}}
\subfigure[\textbf{ The likelihood approach }]{\label{fig:b}\includegraphics[width=70mm]{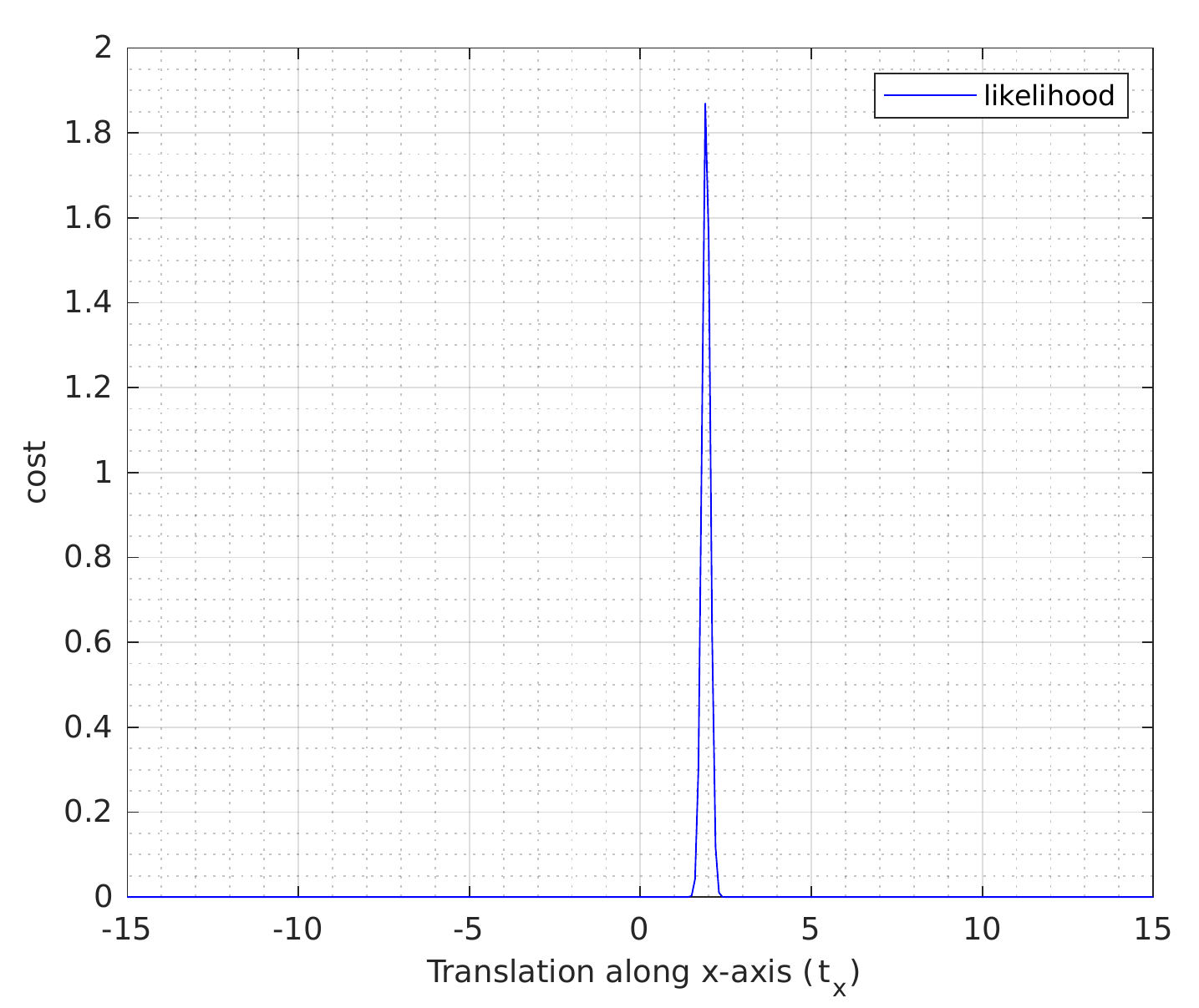}}
\caption[The cost surface for very closed clustered points]{The cost surface is based on the clustered points that are very closed from each. (a) The clustered points are considered to be one component with a significant impact. (b) The clustered points do not flaw the global peak, but it makes the cost surface more certain. (c) The likelihood surface has only one peak and gets more precise about the motion state.}
\label{fig:clus_3}
\end{figure}

\begin{figure}[htb!]
\centering
\subfigure[\textbf{ The cost surface for all targets  }]{\label{fig:a}\includegraphics[width=74mm]{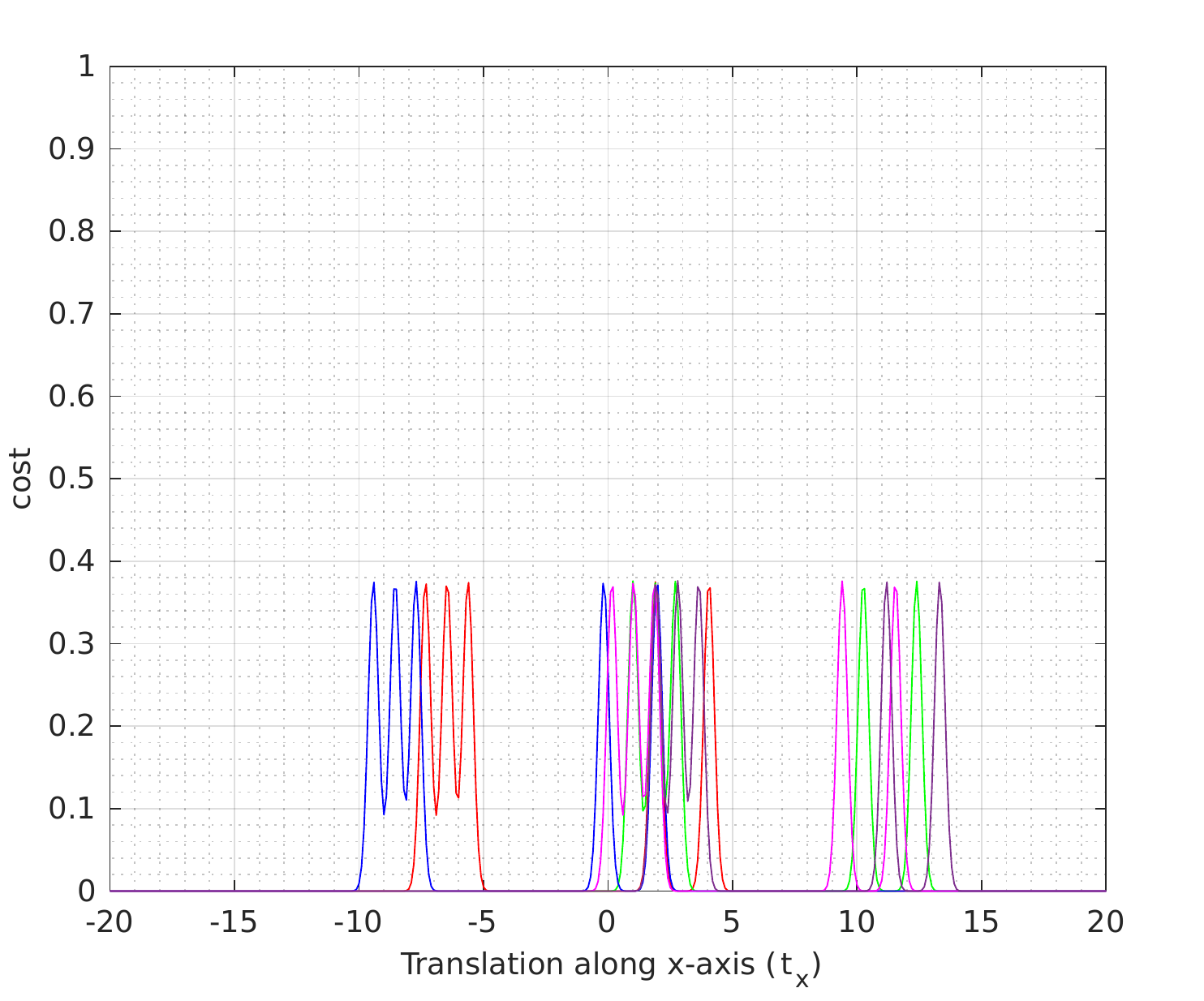}}
\subfigure[\textbf{ The summing approach }]{\label{fig:b}\includegraphics[width=70mm]{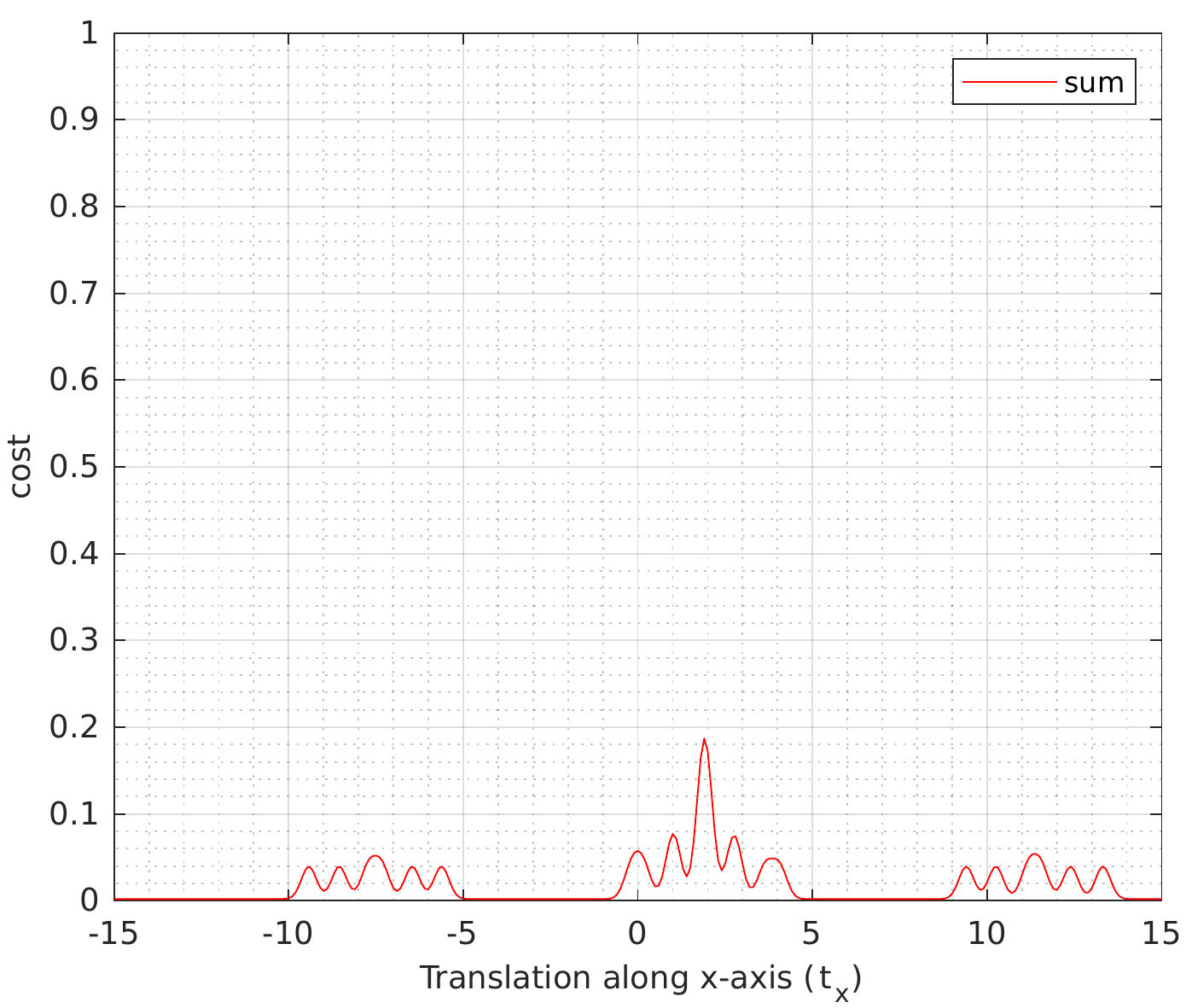}}
\subfigure[\textbf{ The likelihood approach }]{\label{fig:b}\includegraphics[width=70mm]{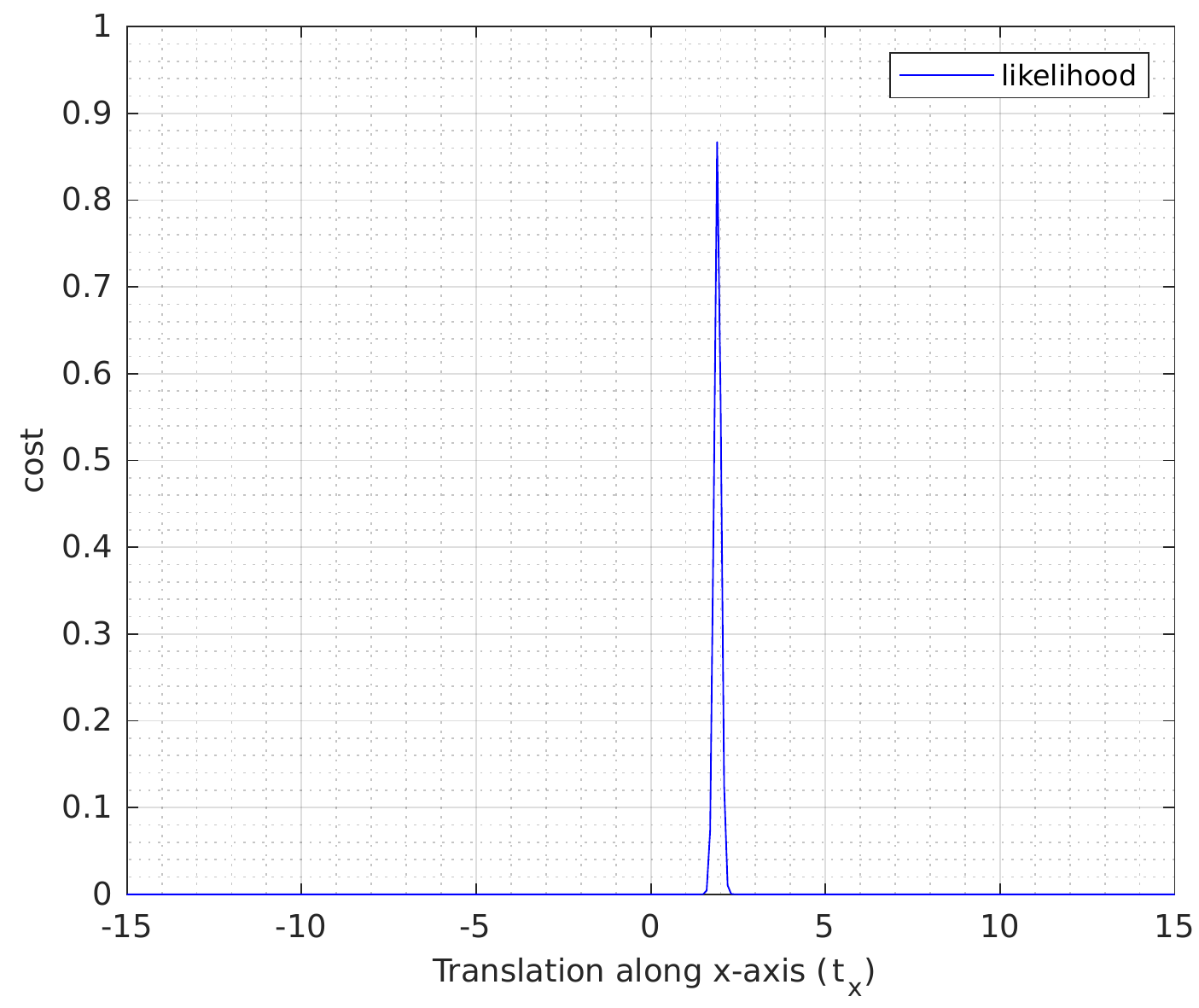}}

\caption[The cost surface for closed clustered points]{ The cost surface is based on the clustered points that are closed but not closed like the previous example. (a) The clustered points are not considered as one component anymore. (b) It introduces local peaks closed to the global peak in the summing approach, which makes it more sensitive to the initial guess. (c)It does not have a significant impact on the likelihood approach.}
\label{fig:clus_4}
\end{figure}

The impact of the clustered points on the cost surface depends on how much the points are closed to each other. Fig. \ref{fig:clus_1} the clustered points are very closed to each other; thus, the cost surface considered it as one component but with a big impact, as shown in Fig. \ref{fig:clus_3} (a). Therefore, this type of clustered points does not have a drawback influence. Moreover, the global peak gets more certain about the motion state, as shown in Fig. \ref{fig:clus_3} (b) and (c). On the other hand, when the points are not very closed, as shown in Fig. \ref{fig:clus_2}. It makes the summing approach more sensitive to the initial guess as it introduces local peaks near the global peak  Fig. \ref{fig:clus_4} (b), but it does not influence the likelihood approach as shown in (c).\\

\vspace{ 2cm}
\section*{Discussion}

This work proposed to use the L2 metric to build the cost function for each target because it includes the uncertainty for both point sets: The current and the previous point set. Moreover, it utilizes the likelihood approach to fuse the cost surface for all targets rather than the summing approach because the summing approach is more sensitive for the initial guess. Furthermore, it uses the M-estimator methodology to overcome the outlier effect on the likelihood approach. The corrupted Gaussian is used to model the outlier component as it does not violate the Gaussian assumption.


\printindex

\chapter{Estimation}

The previous chapter introduces how to build a robust cost function $\boldsymbol{f_{d2d}}$ to estimate the motion state $\boldsymbol{\theta}$. However, how does the estimation process works? That is the focus of this chapter. The estimation process should optimize the state as well as the uncertainty of the optimized value itself. The basic concept is, an algorithm $\boldsymbol{A}$ aims to find the best estimate $\boldsymbol{\hat{\theta}}$ for unknown paramter $\boldsymbol{\theta}$ based on some inputs $\boldsymbol{z}$, by minmizing the cost function $\boldsymbol{f_{d2d}}$, $\hat{\boldsymbol{\theta}}=\boldsymbol{A}(\check{\boldsymbol{z}})=\arg \min _{\boldsymbol{\theta}} \boldsymbol{f_{d2d}}(\check{\boldsymbol{z}}, \boldsymbol{\theta})$. Then, what is the uncertainty $\boldsymbol{\Sigma_{\check{\boldsymbol{\theta}}}} $ for the estimated value $\check{\boldsymbol{\theta}}$ when $\boldsymbol{\Sigma_{z}}$ is the uncertainty of the input $\boldsymbol{z}$.
\section{Motion Estimation}

The optimization problem refers always to find unknown parameters, which minimize or maximize a function. In our case, the function refers to the cost function $\boldsymbol{f_{d2d}}$ and the unknown parameters are the motion state $\boldsymbol{\theta}$.

\begin{equation}
\label{eq: optimization_1}
\hat{\boldsymbol{\theta}} =\arg \min _{\boldsymbol{\theta}} \boldsymbol{f_{d2d}}(\check{\boldsymbol{z}}, \boldsymbol{\theta})
\end{equation}

Thus, the optimization process interested on the value of $\boldsymbol{\theta}$ that minimize the value of the $\boldsymbol{f_{d2d}}$. The optimization problems could be categorized as follows: global or local, convex or non-convex, constrained or unconstrained, continuous or discrete, and linear or nonlinear. Sünderhauf provides a clear explanation regarding this categorization in ~\cite{sunderhauf2012robust}.\\

For simplification and generalization, assume the objective function is $\boldsymbol{\mathrm{F(x)}}$ depends on $\boldsymbol{\mathrm{x}}$, which need to be optimized as follows
\begin{equation}
\label{eq: least_square_1}
\hat{\boldsymbol{\mathrm{x}}} =\arg \min _{\boldsymbol{\mathrm{x}}} \ \boldsymbol{\mathrm{F(x)}}
\end{equation}

For the least square problem, $\boldsymbol{\mathrm{F(x)}}$ is represented as a sum over squared terms as
\begin{equation}
\label{eq: least_square_2}
\boldsymbol{\mathrm{F(x)}} = \frac{1}{2} \boldsymbol{f(\mathrm{x})}  \boldsymbol{f(\mathrm{x})}^T
\end{equation}
So, the problem expressed as
\begin{equation}
\label{eq: least_square_3}
\hat{\boldsymbol{\mathrm{x}}} =\arg \min _{\boldsymbol{\mathrm{x}}} \ \frac{1}{2} \boldsymbol{f(\mathrm{x})}  \boldsymbol{f(\mathrm{x})}^T
\end{equation}

The dependency between $\boldsymbol{f(\mathrm{x})}$ and $\boldsymbol{\mathrm{x}}$ defines wether the problem is a linear or a nonlinear.
\vspace{10mm}

For the linear case, $\boldsymbol{f(\mathrm{x})} = \boldsymbol{A} \boldsymbol{\mathrm{x}} - \boldsymbol{b}$ where  $\boldsymbol{f(\mathrm{x})}$ is linearly depends on $\boldsymbol{\mathrm{x}}$. Then, the problem solved by finding the points $\boldsymbol{\hat{\mathrm{x}}}$ which makes the first drevative equal to zero as follows:
\begin{equation}
\label{eq: linear_least_square_1}
\begin{split}
{\boldsymbol{\frac{\partial \boldsymbol{\mathrm{F(x)}}}{\partial \mathrm{x}}}}  & = \boldsymbol{f(\mathrm{x})}^T \frac{\partial \boldsymbol{f(\mathrm{x})}}{\partial \mathrm{x}}
\end{split}
\end{equation}

\begin{equation}
\label{eq: linear_least_square_2}
\frac{\partial \boldsymbol{f(\mathrm{x})}}{\partial \boldsymbol{\mathrm{x}}} = \boldsymbol{A}
\end{equation}

\begin{equation}
\label{eq: linear_least_square_3}
{\boldsymbol{\frac{\partial \boldsymbol{\mathrm{F(x)}}}{\partial \mathrm{x}}}} =
(\boldsymbol{A} \boldsymbol{\mathrm{x}} - \boldsymbol{b})^T \cdot{\boldsymbol{A}}
\end{equation}

Which equal to zero at the optimal value $\boldsymbol{\hat{\mathrm{x}}}$
\begin{equation}
\label{eq: linear_least_square_4}
(\boldsymbol{A} \boldsymbol{\mathrm{x}} - \boldsymbol{b})^T \cdot{\boldsymbol{A}} = 0
\end{equation}

Which can be written as a normal equation as:
\begin{equation}
\label{eq: linear_least_square_5}
\boldsymbol{A} \boldsymbol{A}^T \boldsymbol{\hat{\mathrm{x}}} = {\boldsymbol{A}}^T \boldsymbol{b}
\end{equation}

Various methods can solve the ordinary equation such as, directly if $\boldsymbol{A}$ is inavertible, QR decomposition, Cholesky decomposition, or singular value decomposition.\\

For the nonlinear case, the relation between $\boldsymbol{\mathrm{x}}$ and $\boldsymbol{f(\mathrm{x})}$ is nonlinear. Thus, the normal equation can not be used to solve the problem directly. Nevertheless, the iterative strategies are used instead to solve it. These methods start from an initial guess then converge at the optimal minimum. The literature offers different methods as an iterative Strategy, such as gradient descent, Newton's Method, and Gauss-Newton. The following part introduces the gradient descent method as a basic approach. Then, it compares distinct philosophies for various iterative strategies.

\subsection*{Gradient Descent}

Gradient descent is a well-known technique for solving the nonlinear least square problem iteratively. The idea is to start from an initial guess $\boldsymbol{\mathrm{x}_0}$. Then, move towards the minimum by the negative of the gradient $\boldsymbol{\mathrm{J}}_{\mathrm{F}}$ computed at that point. The algorithm computes the new point  $ \boldsymbol{x_i}$ using the initial guess and the negative of the Jacobian as follows:

\begin{equation}
	\delta \mathrm{x} = - \alpha  \mathrm{J_F} |_{\mathrm{x} = \mathrm{x_i}}
\end{equation}

Where, $ \alpha$ represents the step size, and it has a big rule to prevent $\mathrm{F}(\mathrm{x_{i+1}})$ gets bigger than $\mathrm{F}(\mathrm{x_{i}})$. Which avoids to miss the local minimum point in the next iteration, thus the new evaluation point is:
\begin{equation}
	\mathrm{x_{i+1}} = \mathrm{x_{i}} - \delta \mathrm{x}
\end{equation}

\begin{figure*}[h!]
\centering
\includegraphics[width=0.7\textwidth]{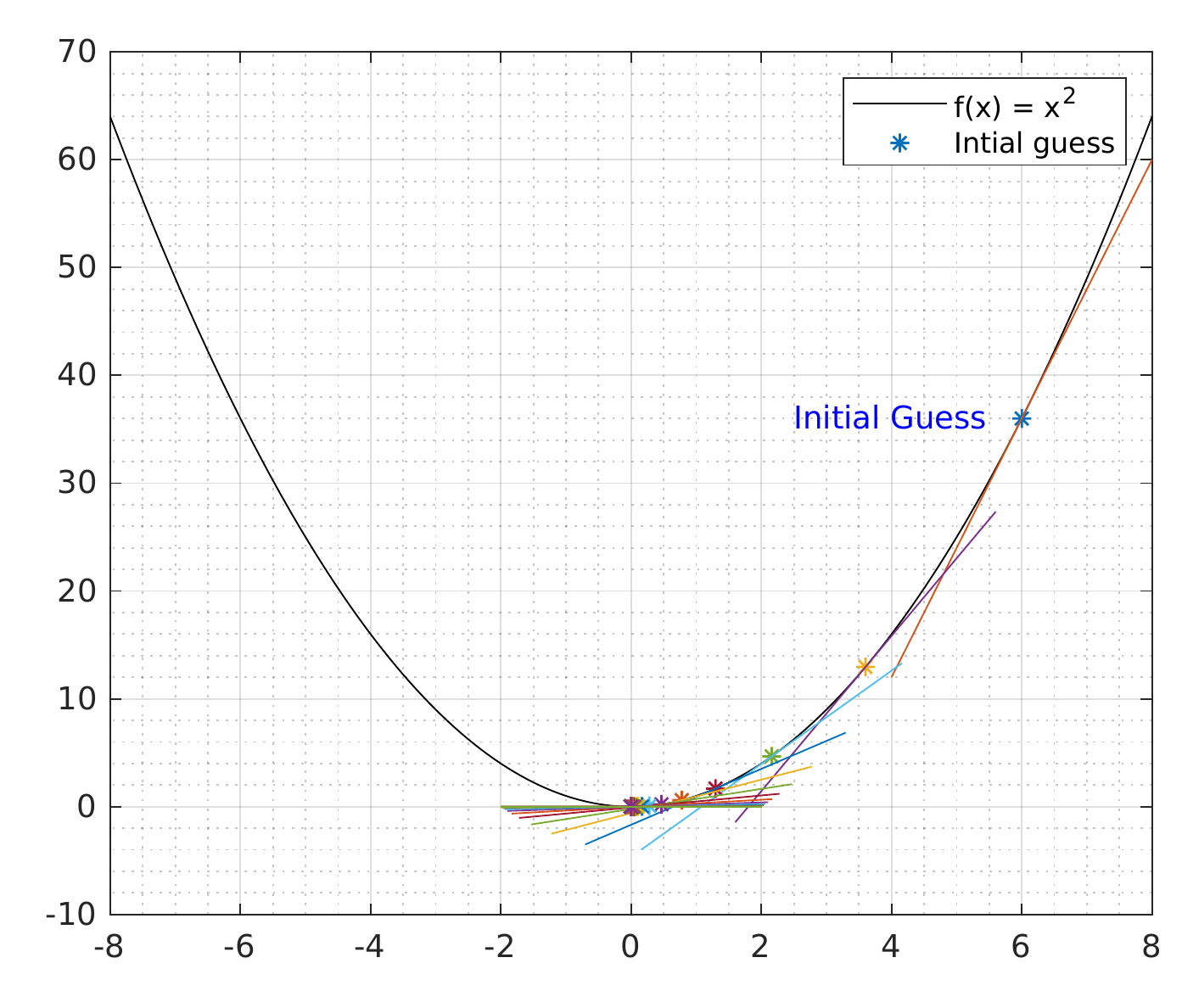}
\caption{ an simple example explains the gradient descent approach, where the algorithem start from intial guess at $\boldsymbol{x_0 = 6}$, then it reach the minmal using the negative jacobian and the step size equal 0.2.}
\label{fig:gradient_descent_1}
\end{figure*}

Fig. \ref{fig:gradient_descent_1} illustrates how the gradient descent is working in a simple example. Where the algorithm starts from an initial guess, then move down based on the minus of the Jacobian till it reaches the local minimum. The main difference between the gradient descent strategy and other strategy is how fast the algorithm reaches minimal value and converges; in other words, how to calculate the next evaluation point.

\textbf{Newton's method} locates minimal by analyzing the curvature of the cost function, Where the second derivative of the function incorporates that information. Thus, the step size depends on the second dervative as follows:

\begin{equation}
	\delta \mathrm{x} = \frac{\mathrm{F'}(\mathrm{x_i})}{\mathrm{F''}(\mathrm{x_i})}
\end{equation}

Where $\boldsymbol{\delta \mathrm{x}}$ called the Newton step. However, to apply this method, the second derivative should be computable.  \textbf{Gauss-Newton} is a variant approach to puzzle out the problem. The hypothesis is to linearise the problem at $\boldsymbol{\mathrm{x_i}}$ using first-order Taylor expansion. Then solve the problem as a normal linear equation. The last method to be reviewed is \textbf{Levenberg-Marquardt}. It is a hybrid between Gradient descent and Gauss-Newton. Where the step size depends on a damping factor. The algorithm reacts like gradient descent with a large damping factor. However, the algorithm acts as Gauss-Newton with a small value~\cite{sunderhauf2012robust}. Each method tries to get to the minimal value fast through modify the step size $\boldsymbol{\delta \mathrm{x}}$ as depicted in Table \ref{table:1}.\\
\begin{table}[h!]
\centering
\begin{tabular}{|c c c|}
\hline
$\boldsymbol{Method}$ &  & $ {\boldsymbol{Step}} \Delta \boldsymbol{\mathrm{x}}$ \\
\hline
$\text { Gradient Descent }$ &  & ${\Delta \mathrm{x}=-\alpha \mathrm{J}^{\mathrm{T}} \mathrm{f}(\mathrm{x})}$ \\
\hline
$\text { Newton's Method }$ &  & ${\left(\mathrm{H}^{\top} \mathrm{f}+\mathrm{J}^{\top} \mathrm{J}\right) \Delta \mathrm{x}=-\mathrm{J}^{\top} \mathrm{f}(\mathrm{x})}$ \\
\hline
$\text { Gauss-Newton }$ & &  ${\mathrm{J}^{\top} \mathrm{J} \Delta \mathrm{x}=-\mathrm{J}^{\top} \mathrm{f}(\mathrm{x})} $\\
\hline
$\text { Levenberg-Marquardt}$ & & ${\left(\mathrm{J}^{\top} \mathrm{J}+\lambda \mathrm{I}\right) \Delta \mathrm{x}=-\mathrm{J} {\top} \mathrm{f}(\mathrm{x})}$ \\
\hline
\end{tabular}
 \caption{ This table provides some algorithms and the corresponding step size calculation \cite{sunderhauf2012robust}}
\label{table:1}
\end{table}

\section{Covariance Estimation}

The previous chapter introduced a robust cost function model. Nevertheless, the optimal peak does not ensure the perfect estimate, but it represents the best estimate. Usually, there is a deviation between the optimized pose and the ground truth. Thus, for a consistent estimation, optimizing the uncertainty of the optimized pose is as crucial as optimizing the relative pose itself. Moreover, it is not possible to achieve sensor fusion without covariance. Lots of research is performed in this focus, although the proposed methods are weather inaccurate or too expensive in the calculation. Censi in ~\cite{censi2007accurate} reviews various existing approaches to estimate the covariance as well as demonstrates the advantage and the disadvantage of each. This section focuses on optimizing the covariance based on the error propagation model and based on the fisher information model.

\subsection*{Error Propagation}

This method estimates the variance of the output by propagating the uncertainty of the input. However, the function type impacts the propagation model, where the function could be in a different form, such as a linear or a nonlinear, and sometimes there no explicit access for the function itself.

\begin{problem}
 let $\boldsymbol{x}$ be the input for a function $\boldsymbol{f(\mathrm{x})}$, then the function output: $\boldsymbol{y = f(\mathrm{x})}$. The question is how to map the input uncertainty $\boldsymbol{\Sigma_{\mathrm{x}}}$ to the output uncertainty $\boldsymbol{\Sigma_y}$.
\end{problem}

\subsection*{Linear Function}

\begin{figure*}[h!]
\centering
\includegraphics[width=0.7\textwidth]{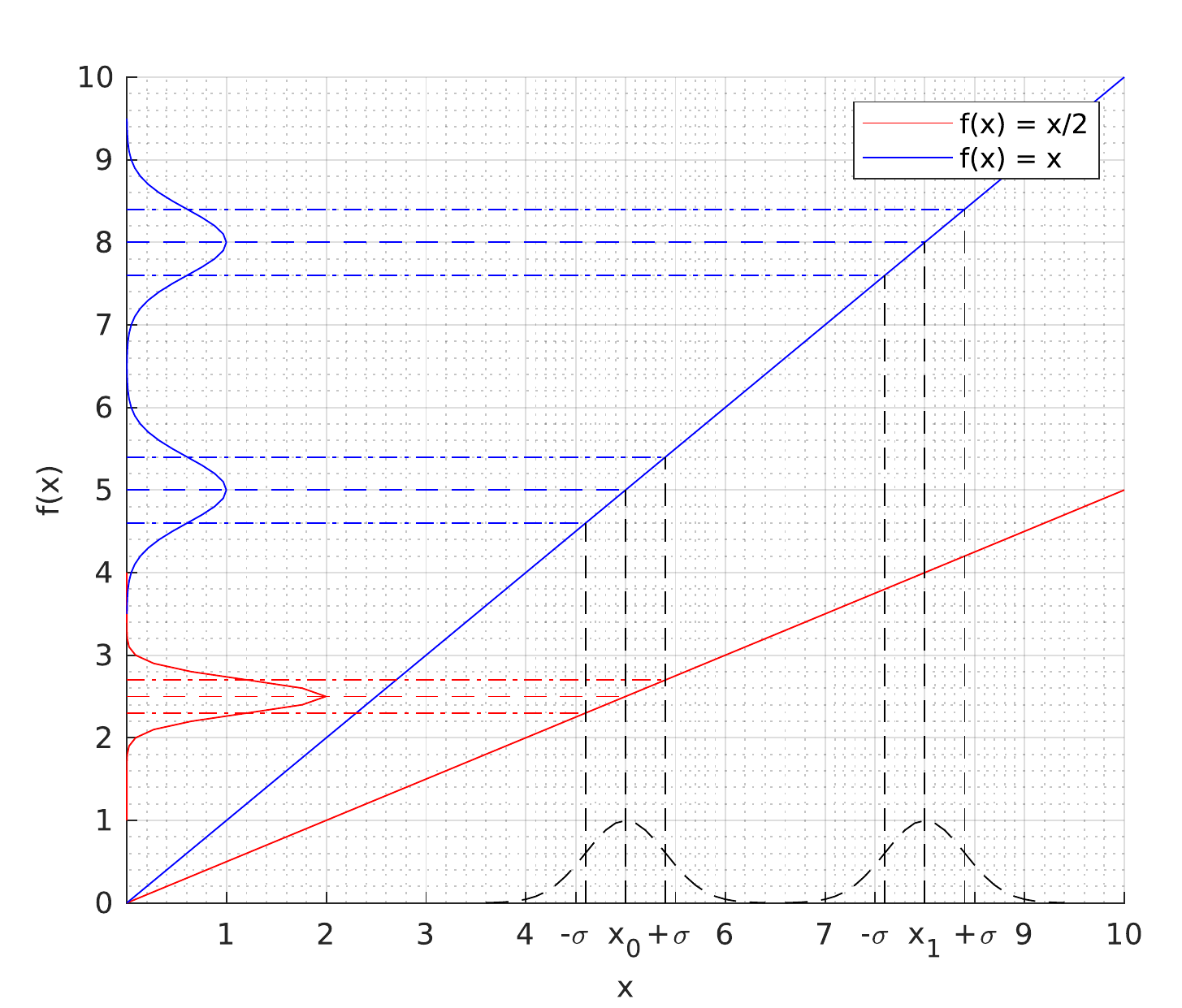}
\caption{This example shows the error propagation for a linear function. Where, it shows the output variance changes with the line slop as shown at $\boldsymbol{x}_0$, in addition to, the slope of the line is constant the output does not change as shown in $\boldsymbol{x}_1$ in the blue line.}
\label{fig:linear_fun}
\end{figure*}

For a linear function, like, $\boldsymbol{f(\mathrm{x})} = b(\boldsymbol{\mathrm{x}})$, where b is a constant coefficient, Fig. \ref{fig:linear_fun} shows how to propagate the input uncertainty. Where the uncertainty changes based on the line slope as depicted at $\boldsymbol{\mathrm{x}}_0$. Moreover, the error propagtion does not change as long as the function is linear as shown at $\boldsymbol{\mathrm{x}}_1$. The line slope represents the first derivative of the function with respect to the input.\\

\subsection*{Nonlinear Function}

For a nonlinear function such as a quadratic one, $\boldsymbol{f(\boldsymbol{\mathrm{x}})} = b(\boldsymbol{\boldsymbol{\mathrm{x}}^2})$, where b is a constant coefficient. The error propagation process is done in two steps: Firstly, linearise the function at a specific value $\boldsymbol{\mathrm{x}}_0$. Secondly, propagate the uncertainty through the linearized model. The first-order Taylor series expansion~\cite{thomas2010uncertainty} is used to linearize the function at a given value $\boldsymbol{\mathrm{x}}_0$ as follows:

\begin{equation}
\label{eq: taylor_1}
\boldsymbol{
\mathrm{y}_{i}=f\left(\mathrm{x}_{i}\right)=\underbrace{f\left(\mathrm{x}_{0}\right)+\left(\frac{\mathrm{d} \mathrm{y}}{\mathrm{d} \mathrm{x}}\right)_{0}\left(\mathrm{x}_{i}-\mathrm{x}_{0}\right)}_{\text {linear terms }}+\frac{1}{2}\left(\frac{\mathrm{d} \mathrm{y}}{\mathrm{d} \mathrm{x}}\right)_{0}^{2}\left(\mathrm{x}_{i}-\mathrm{x}_{0}\right)^{2}+\cdots
}
\end{equation}

Most of the literature uses a general formula $\boldsymbol{(J \Sigma J^T)}$ for a nonlinear function, which is basically derived from (\ref{eq: taylor_1}) as follows:

\begin{equation*}
\boldsymbol{
f\left(\mathrm{x}_{i}\right) - f\left(\mathrm{x}_{0}\right) = \left(\frac{\mathrm{d} \mathrm{y}}{\mathrm{d} x}\right)_{0}\left(\mathrm{x}_{i}-\mathrm{x}_{0}\right)
}
\end{equation*}

Where, $\boldsymbol{f\left(\mathrm{x}_{i}\right) - f\left(\mathrm{x}_{0}\right)}$ represents the output deviation at tha specific point $\boldsymbol{x}_0$, square that deviation gives the uncertainty as follows:

\begin{equation}
\boldsymbol{
\begin{split}
\Sigma_{\mathrm{y}} & = (f\left(\mathrm{x}_{i}\right) - f\left(\mathrm{x}_{0}\right))(f\left(\mathrm{x}_{i}\right) - f\left(\mathrm{x}_{0}\right))^{\mathrm{T}} \\
& = \left(\left(\frac{\mathrm{d} y}{\mathrm{d} \mathrm{x}}\right)_{0}\left(\mathrm{x}_{i}-\mathrm{x}_{0}\right)\right)\left(\left(\frac{\mathrm{d} \mathrm{y}}{\mathrm{d} \mathrm{x}}\right)_{0}\left(\mathrm{x}_{i}-\mathrm{x}_{0}\right)\right)^{\mathrm{T}} \\
& = \left(\frac{\mathrm{d} \mathrm{y}}{\mathrm{d} \mathrm{x}}\right)_{0}\left(\mathrm{x}_{i}-\mathrm{x}_{0}\right) \left(\mathrm{x}_{i}-\mathrm{x}_{0}\right)^{\mathrm{T}} \left(\frac{\mathrm{d} \mathrm{y}}{\mathrm{d} \mathrm{x}}\right)_{0}^{\mathrm{T}}
\end{split}
}
\end{equation}

Therefore, the known formula for a nonlinear function written as follows:

\begin{equation}
\label{eq:general_prop_cov}
\boldsymbol{
\boldsymbol{\Sigma}_\mathrm{y} = \left(\frac{\mathrm{d} f}{\mathrm{d} {\mathrm{x}}}\right) \ \boldsymbol{\Sigma}_{\mathrm{x}}  \ \left(\frac{\mathrm{d} f}{\mathrm{d} \mathrm{x}}\right)^{\mathrm{T}}
}
\end{equation}

where $\boldsymbol{\frac{\mathrm{d} f}{\mathrm{d} x}}$ represents the Jacobian of $\boldsymbol{f}$ with respect to inputs $\boldsymbol{\mathrm{x}}$.

\begin{figure*}[h!]
\centering
\includegraphics[width=0.8\textwidth]{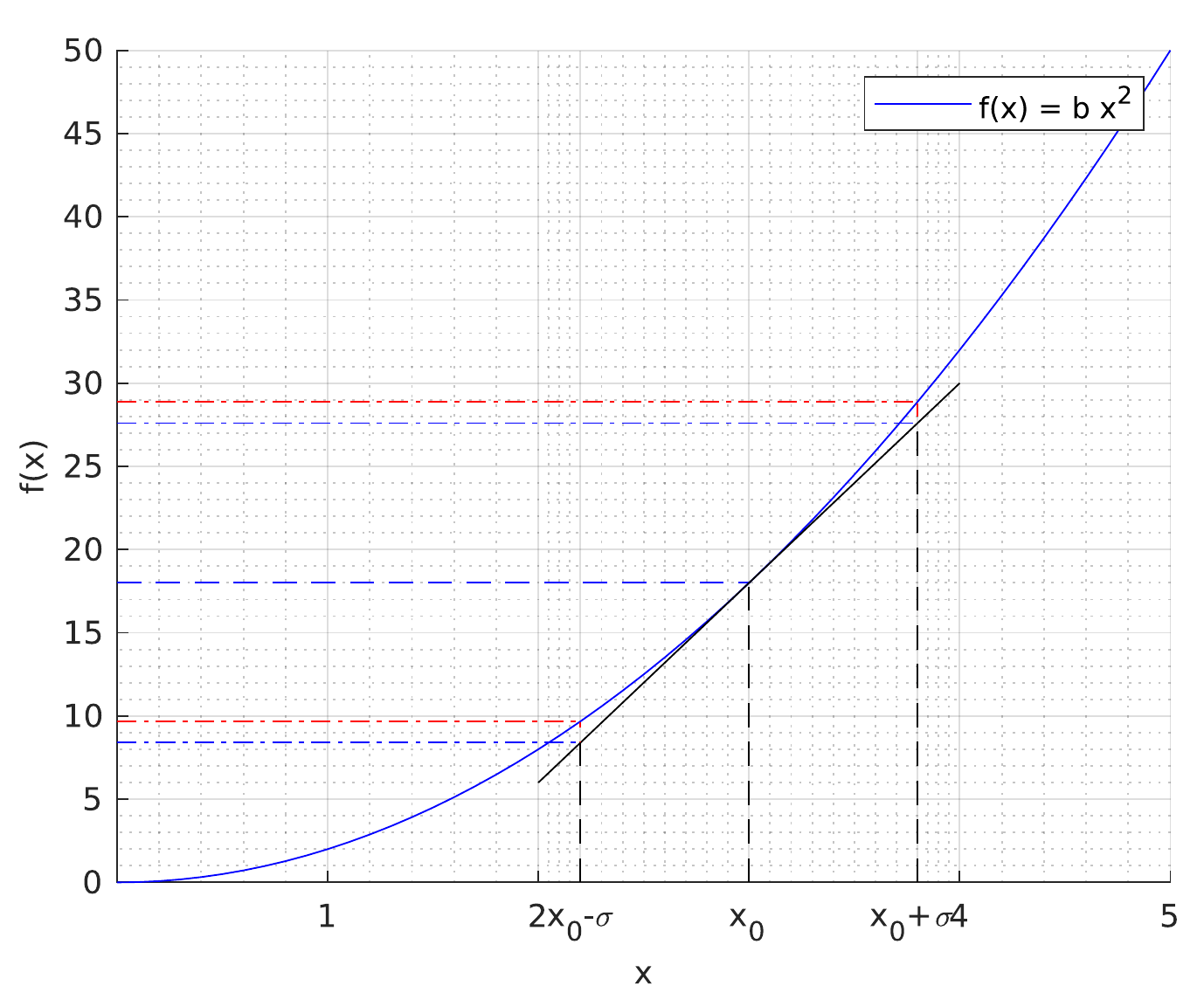}
\caption{This example shows the error propagation for a nonlinear function, where The first-order Taylor series expansion is used to linearze the function at $\boldsymbol{x}_0$, then propagate the input covariance to get the output covariance.}
\label{fig:linear}
\end{figure*}

After linearizing the problem, the linearized form is used to propagate the uncertainty, as depicted in Fig. \ref{fig:linear}.\\

\subsection*{Implicit Function Theorem}
The main concern of this section is to introduce a common method to define the Jacobian of function $\boldsymbol{f(\mathrm{x})}$, if there is no explicit expression ($\boldsymbol {\mathrm{y} = f(\mathrm{x})}$).\\

\begin{theorem}
Let $\Phi: \mathcal{S} \rightarrow  \mathbb{R}^h$ is continuous differentiable ($C^1$). Assuming a point $\boldsymbol{(x_0, y_0)} \in \mathcal{S}$, where $\Phi(x_0, y_0) = 0$ and $\frac{\partial{\Phi}}{\partial{y}} \neq 0$. Then, there is open neighbourhood $\mathrm{X} \subset \mathbb{R}^n$ of $\boldsymbol{x_0}$ and neighbourhood $\mathrm{Y} \subset \mathbb{R}^m$ of $\boldsymbol{y_0}$, and a unique function $f: \mathrm{X} \rightarrow  \mathrm{Y}$, then $\Phi$ can be written as:
\begin{equation}
		\boldsymbol{\Phi(\mathrm{x}, f(\mathrm{x})) = 0}
\end{equation}

Then,the derivative of $\boldsymbol{\Phi}$ with respect to $\boldsymbol{\mathrm{x}}$ witten as:
\begin{equation}
		\boldsymbol{\frac{\partial{\Phi}}{\partial{\mathrm{x}}}} +  \boldsymbol{\frac{\partial{\Phi}}{\partial{\mathrm{f}}}} \boldsymbol{\frac{\mathrm{d} f}{\mathrm{d}\mathrm{x}}}  = \boldsymbol 0
\end{equation}

Then, the derivative of $f$ with respect to $\mathrm{x}$:

\begin{equation}
\begin{split}
\boldsymbol{\frac{\mathrm{d} f}{\mathrm{d}\mathrm{x}}} & = - {\boldsymbol{\frac{\partial{\Phi}}{\partial{\mathrm{x}}}}}  /{\boldsymbol{\frac{\partial{\Phi}}{\partial{\mathrm{f}}}}}
\end{split}
\end{equation}

the equation can be rephrased as:
\begin{equation}
\boldsymbol{\frac{\mathrm{d} f}{\mathrm{d}\mathrm{x}}} = - (\boldsymbol{\frac{\partial{\Phi}}{\partial{\mathrm{f}}}})^{-1} (\boldsymbol{\frac{\partial{\Phi}}{\partial{\mathrm{x}}}})
\end{equation}

\end{theorem}

\begin{figure*}[h!]
\centering
\includegraphics[width=0.8\textwidth]{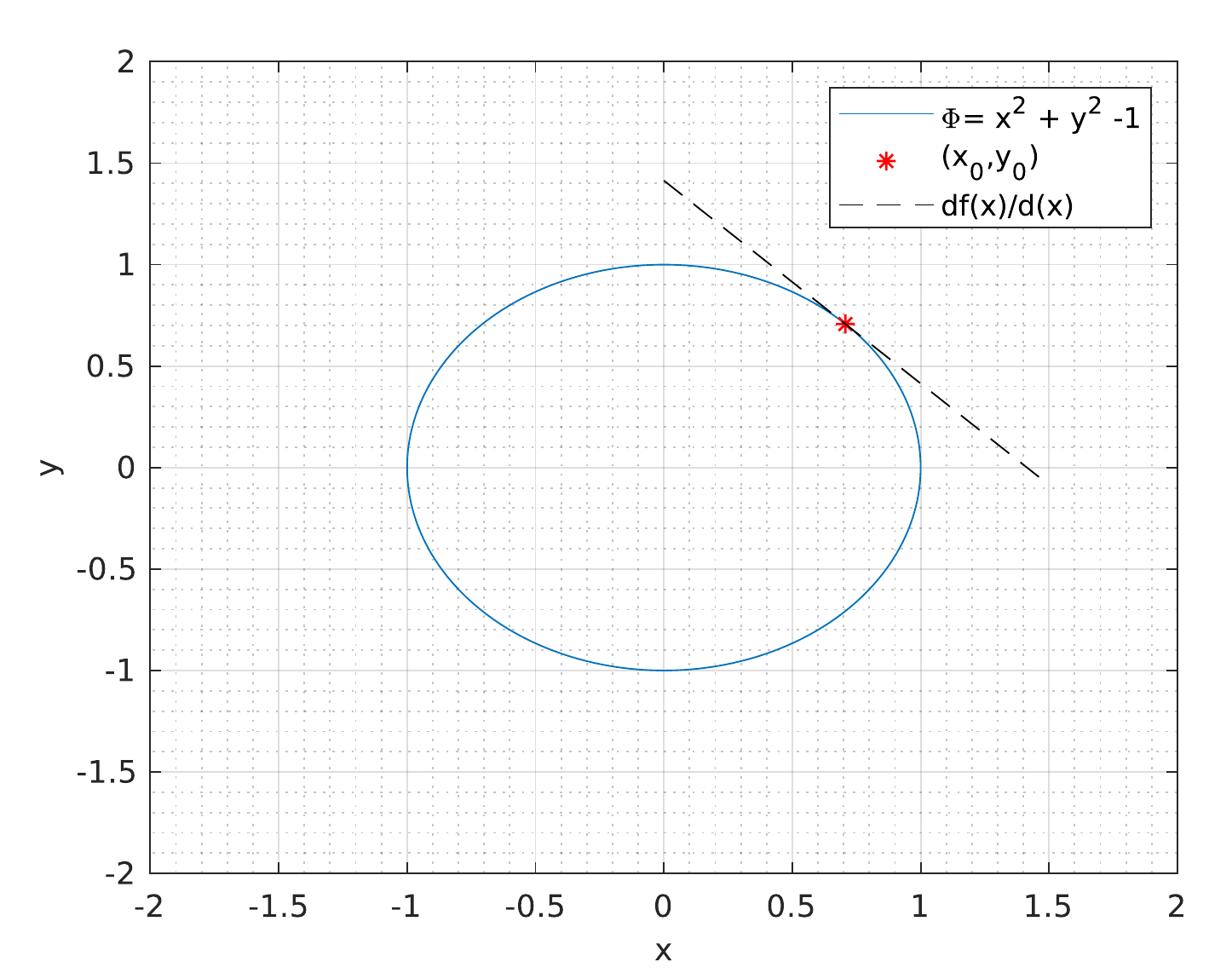}
\caption{This Figure shows a unit circle as an example for implicit function, where the tangent for the circle is computed at $(1/ \sqrt{2},1/ \sqrt{2})$.}
\label{fig:implicit}
\end{figure*}

A simple example could be the easiest way to get the impression behind the theorem, like, the tangent to a unit circle as depicted in Fig. \ref{fig:implicit}.
\begin{example}
The equation for a unit circle is $\boldsymbol{x^2 + y^2 =1}$, and it could be written as $\boldsymbol{\Phi} = \boldsymbol{x^2+y^2-1}$, calculate the tanget at $(\frac{1}{\sqrt{2}},\frac{1}{\sqrt{2}})$. \\

\textbf{Checks:}
\begin{itemize}
  \item $\Phi(x_0,y_0) = 0 \Rightarrow ((\frac{1}{\sqrt{2}})^2 + (\frac{1}{\sqrt{2}})^2 -1 = 0)$ valid condition.
  \item $\frac{\partial{\Phi}}{\partial{y}} \neq 0 \Rightarrow \frac{\partial{\Phi}}{\partial{y}}|_{y_0 = \frac{1}{\sqrt{2}}} =\frac{2}{\sqrt{2}} \neq 0 $
\end{itemize}

The two conditions are valid. Therefore, the theorem claims that function $\boldsymbol{\Phi(x,y)}$ is implicitly defining $y = f(x)$ at the neighborhood of $\boldsymbol{(x_0, y_0)}$, Then the derivative of $\boldsymbol{f}$ with respect to $\boldsymbol{\mathrm{x}}$ written as:

\begin{equation}
	\boldsymbol{\frac{\mathrm{d} f}{\mathrm{d}\mathrm{x}}} = - {\boldsymbol{\frac{\partial{\Phi}}{\partial{\mathrm{x}}}}}  /{\boldsymbol{\frac{\partial{\Phi}}{\partial{\mathrm{f}}}}} = -  \boldsymbol{\frac{2 \mathrm{x}}{2 \mathrm{y}}} = -1
\end{equation}

\end{example}

\subsection{Error Propagation Model}

Censi in~\cite{censi2007accurate} used the error propagation approach to estimate the covariance for the estimated pose based on \textbf{ICP} approach. Furthermore, Stoyanov in ~\cite{stoyanov2012fast} follows the same way to optimize the covariance for the optimized pose based on the  \textbf{NDT} approach.\\

The problem $\boldsymbol{(A)}$ is nonlinear. Hence, the first-order Taylor approximation is used to propagate the covariance. Although $\boldsymbol{(A)}$ is not in a closed form, the $\boldsymbol {\frac{\partial A}{ \partial z}}$ still can be computed. With one condition, find an implicit function defines $\boldsymbol{A} $ and $\boldsymbol{z}$, like, $\boldsymbol {\Phi(\theta, z) = 0}$. The gradient of the cost function at the optimized value $\boldsymbol{\hat{\theta}}$ should be zero, which forms an implicit function $ \boldsymbol{\Phi} = \boldsymbol{\frac{\partial{f_{d2d}}}{\partial{\theta}}}$. Hence, the Jacobian $\boldsymbol {\frac{\partial A}{ \partial z}}$ written as  follows:

\begin{equation}
\boldsymbol{ \frac{\partial A(z)}{\partial z} = -(\frac{\partial f_{d2d}}{\partial \mathrm{\theta}^2})^{-1} (\frac{\partial f_{d2d}}{\partial {z} \partial \mathrm{\theta}})}
\end{equation}

Then, the uncertainty of the estimated value $\boldsymbol{\hat{\theta}}$ based on the error ptopagation  approach (\ref{eq:general_prop_cov}) is:

\begin{equation*}
\boldsymbol{ \operatorname{cov}({\boldsymbol{\hat{\theta}}}) \simeq \left(\frac{\partial^{2} f_{d2d}}{\partial \theta^{2}}\right)^{-1} \frac{\partial^{2} f_{d2d}}{\partial z \partial \theta} \operatorname{cov}(\boldsymbol{z})
\frac{\partial^{2} f_{d2d}}{\partial z \partial \theta} \left(\frac{\partial^{2} f_{d2d}}{\partial \theta^{2}}\right)^{-1}}
\end{equation*}

\subsection{Fisher Information Model}

This approach estimates the uncertainty based on the Fisher Information Matrix \textbf{FIM}. One justification for that method is, FIM measures how much information about the unknown parameter is included in the observable random variable~\cite{pei2009fisher}. Moreover, for an unbiased estimator, Cramer-Rao inequality provides the lower bound of the variance, which is based on the FIM. For an estimation $\boldsymbol{\hat{\theta}}$ the inequality can be written as follows:

\begin{equation}
\boldsymbol{ E\left[(\hat{\theta}- \theta)(\hat{\theta}- \theta)^{\prime}\right] \geq F^{-1}}
\end{equation}

Where $\boldsymbol{F}$ represents the Fisher Information Matrix(FIM) and $ \boldsymbol{E\left[(\hat{\theta}-\theta)(\hat{\theta}-\theta)^{\prime}\right] }$ is the mean square error. Consequently, a common approach is to use the inverse of the Fisher information matrix as the covariance matrix for the maximum likelihood estimator.\\

In practice, this approach analyses the shape of the cost function to estimate the covariance, where Censi, in his work, called it \textit{the black box method}. The idea is as long as the cost function $\boldsymbol{f_{d2d}}$ is quadratic, so the optimal least-squares would be $\boldsymbol{\hat{\theta}}$ with covariance as follows:

\begin{equation*}
\boldsymbol{\operatorname{cov}({\boldsymbol{\hat{\theta}}}) \simeq \left(\frac{\partial^{2}}{\partial \boldsymbol{\theta}^{2}} f_{d2d}({\boldsymbol{z}}, \boldsymbol{\theta})\right)^{-1}}
\end{equation*}

Where it is basically the inverse of Hessian.

\subsection*{Discussion}

Stoyanov, in his work, noted:“[...] the shape of the error function close to the solution is not the only factor that influences the covariance of the result. In fact, it is important to also account for the variance of the objective function, with respect to the noise in the input point scan data”~\cite{stoyanov2012fast}.\\

Stoyanov did not include the input noise in his cost function, but he builds a covariance for each cell based on the target inside this cell, as well as, Censi error function does not include the input error. However, the proposed cost function in this work is constructed based on the distribution-to-distribution approach, which includes the noise for both point sets in the cost function, where it shows that the noise of the input points affects the cost surface.  Therefore, analysis of the cost surface to estimate the covariance should be enough as long as the input noise is included in the cost function and no need to use the error propagation model.  Moreover, the likelihood approach is used to fuse the cost surface for all targets, so the proposed method did not violate any Gaussian assumption like the summing one. The interesting point now is how to evaluate the complete estimation process? Which will be discussed in the next chapter.

\printindex

\chapter{Evaluation}

Earlier,  this work introduced the point set registration problem, where the selection of the error function, which fits the problem, is a crucial task. Then, it introduces the concept of a robust cost function based on the M-estimator philosophy.  Following that, it presents the least-squares problem and different algorithms to solve it. Besides, it explains two different ways to estimate the uncertainty for the estimation. This work introduced and compared two approaches for the cost function: The summing and the likelihood approach.

\begin{figure*}[htb!]
\centering
\includegraphics[width= 0.9\textwidth]{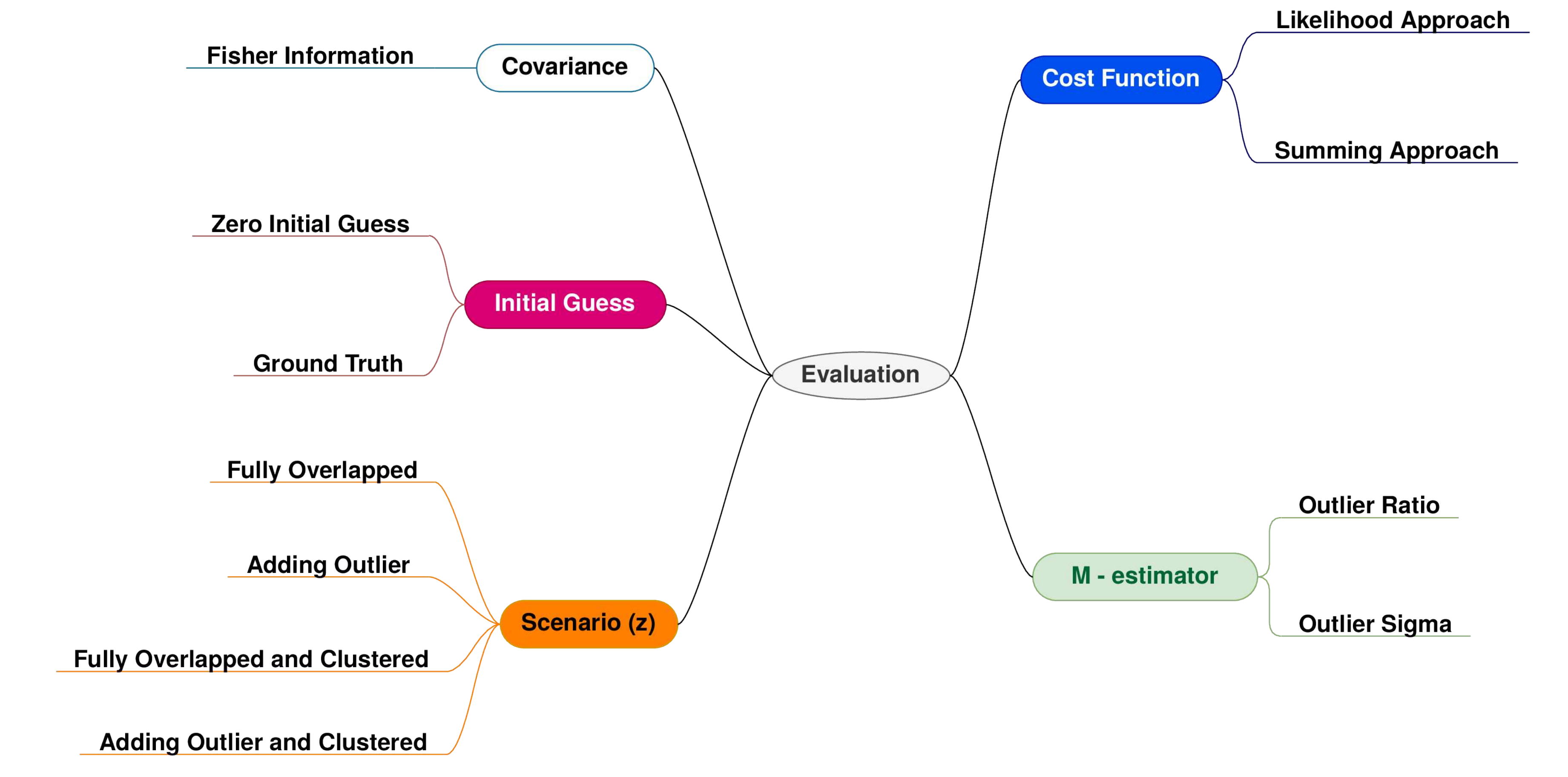}
\caption{The complete evaluation map and all aspects for each evaluation point.}
\label{fig:eval_tax_1}
\end{figure*}

The summing approach: 
\begin{equation}
\label{eq:sum_cost_62}
f_{\mathrm{d2d}}(\boldsymbol{\theta})=\sum_{k = 1}^{\vert {\cal M}\vert } \left(\sum_{i=1}^{\vert {\cal F}\vert } w_{i} \cdot p \left( 0|\boldsymbol{T}(m_k,\boldsymbol{\theta})-\mu_{i}, \boldsymbol{T}(\boldsymbol{\Sigma_k},\boldsymbol{\theta}) + \boldsymbol{\Sigma_{i}}\right)\right)
\end{equation}

The likelihood approach: 
\begin{equation}
\label{eq:like_cost_62}
f_{\mathrm{d2d}}(\boldsymbol{\theta}) =  \prod_{k \in \mathcal{M}}\left(\sum_{i=1}^{\vert {\cal F}\vert } w_{i} \cdot p \left( 0|\boldsymbol{T}(\boldsymbol{m_k},\boldsymbol{\theta})-\boldsymbol{\mu_{i}}, \boldsymbol{T}(\boldsymbol{\Sigma_k},\boldsymbol{\theta}) + \boldsymbol{\Sigma_{i}}\right)\right)
\end{equation}

For the evaluation process, the cost function can be written as follows:

\begin{equation}
f_{\mathrm{d2d}}(\boldsymbol{ \textcolor{orange}{\check{\boldsymbol{z}}}, \theta})= \underbrace{ \textcolor{blue}{Fuse \ all \ targets}}_{ Likelihood \ or \  Summing } \  \underbrace{\sum}_{P_{GMM} \ (\boldsymbol{\mathcal{F}})} \ \underbrace{\left( \textcolor{green}{(1-\alpha)} * p_{\text {inlier}}  + \textcolor{green}{\alpha} \ * \ \textcolor{green}{p_{outlier}} \right)}_{\text {each target cost }}
\label{eq:eval_eq_1}
\end{equation}

This work proposed to model the outlier component by Gaussian distribution, centered around the targets in the previous scan with large covariance, as follows:

\begin{equation}
	\textcolor{green}{p_{outlier}} = \sum_{i=1}^{|\mathcal{F}|} p(\mathcal{T}(m_k,\boldsymbol{\theta}) | \mu_i,\Sigma_{outlier})
\label{eq:out1_f_2}
\end{equation}
Thus, the optimization problem written as:

\begin{equation}
\hat{\boldsymbol{\theta}} \ = \ \arg \min _{\boldsymbol{\theta}} \boldsymbol{f}_{d 2 d}(\textcolor{orange}{\check{\boldsymbol{z}}}, \boldsymbol{\textcolor{magenta}{{\theta}_{initial}}})
\end{equation}

This chapter evaluates the summing and likelihood approach for merging the cost function for each target where the evaluation process has different aspects, as depicted in Fig. \ref{fig:eval_tax_1}, all the evaluation aspects colored by the same color in the equation and the evaluation taxonomy. Firstly, this chapter evaluates the dependencey between the good initial guess, and the optimizer converges in the optimal minimum, colored by red. Secondly, it examines the impact of the outlier in both approaches, colored by orange, as well as modeling the outlier component in the cost function to suppress the outlier effect. Lastly, it shows what will happen when some points are closed to each other. \\

\section{Synthetic Scenario - 2D }

\begin{figure*}[htb!]
\centering
\includegraphics[width=0.9\textwidth]{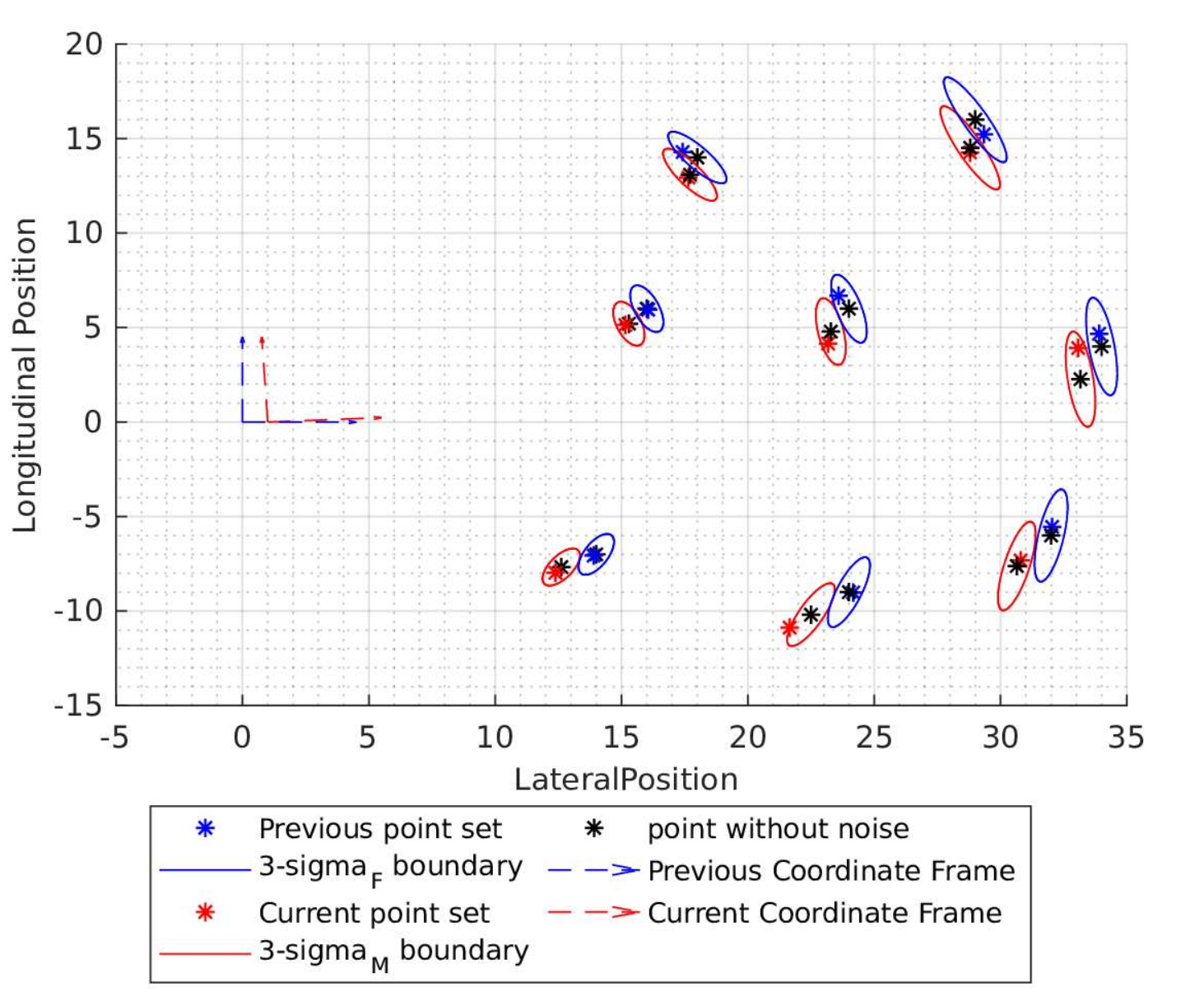}
\caption{  The synthetic scenario includes two-point sets. Where the blue is the previous point set $\boldsymbol{\mathcal{F}}$, and the red is the current point set $\boldsymbol{\mathcal{M}}$. Besides, each point set comprises of 8 points. The error ellipses are in the 3 sigma boundary. The current point set $\boldsymbol{\mathcal{M}}$ is relatively transformed from the previous point set $\boldsymbol{\mathcal{F}}$ by $\boldsymbol{\mathcal{\theta}_g}$.}
\label{fig:point_set_1}
\end{figure*}

For the evaluation process, a synthetic scenario is created, which includes two-point sets $\{\boldsymbol{\mathcal{F}, \mathcal{M}}\}$, where $ \boldsymbol{\mathcal{M}} = \boldsymbol{T}(\boldsymbol{\mathcal{F},\theta_g})$, where at least 8 points from one set have a correspondence in the second set, but, the size of the point set changes based on the evaluation point. Generating the scenario can be summarized in a few steps as follows: Firstly, we have chosen a well-distributed 8 point to represent the seed for the scenario, and if the scenario evaluates the clustering effect, so more points are added close to some points from the 8 points, that build the first point set $\boldsymbol{\mathcal{F}}$. Secondly, rotating and translating $\boldsymbol{\mathcal{F}}$ (blue) by $\boldsymbol{\theta_g}$ to get the second point set $\boldsymbol{\mathcal{M}}$ (red). Then, if the scenario evaluates the outlier effect, random points are added in both point sets, and these point does not have a correspondence to both point sets, where it could be close to any point from the 8 points or far from the whole set. This work evaluates a robust point set registration applied to an automotive Doppler radar, thus both point set $\{\boldsymbol{\mathcal{F}},\boldsymbol{\mathcal{M}}\}$ is converted from the cartesian space to polar space ($[r_k, \phi_k]'$, where $r_k$ is the relative distance and the $\phi_k$ is relative angel between)\footnote{ The radar gives a relative measurement between the detected targets and the target positions.}. Lastly, Both point sets are subjected to noise, to simulate the reality; for example, the fully overlapped scenario is depicted in Fig. \ref{fig:point_set_1}.\\

Lastly, in each evaluation scenario, the test is run 1000 times with the same condition, to check the result consistency. The ground truth $\boldsymbol{\theta_g}$ for the scenario is, the second point set is translated along the $\boldsymbol{x-axis}$ by $\boldsymbol{5 m}$ and rotated around the $\boldsymbol{z-axis}$ by $\boldsymbol{15^{\circ}}$, assuming the time difference between the two scans is 200 msec. All the points in point set have the same standard deviation $(\sigma_{rk} = 0.200, \sigma_{\phi} = 0.0300)$, these standard deviations in this scenario are the most common standard deviation we received from the radar in one real data set.

\section{Credibility Test}

Credibility test is a term used to assess the estimation error, where the assessment includes the error in the estimated state and the estimated uncertainty. Different metrics are introduced to perform the credibility assessment, and Ivanov gives a review for most of these methods in ~\cite{ivanov2014evaluating}. However, the normalized state estimation error squared \textbf{(NEES)} is a well-known approach for assessing the estimation, which introduced by Bar-Shalom and Li in ~\cite{bar2004estimation}. Nevertheless,  the NEES test uses the arithmetic average, which could give a miss leading results, as  Rong Li and Zhao criticize in \cite{li2006measuring}. Thus, they used the geometric average to perform the assessment by introducing the non-credibility index\textbf{(NCI)} and the inclination indicator \textbf{(I$^2$)} as an alternative method. The credibility test follows the Monte Carlo approach\footnote{ This method is used to solve a mathematical or statistical problem. The result is obtained by repeating the random sampling and statistical analysis~\cite{kroese2013handbook}.}.

\subsection*{NESS}
This method computes the estimation error in each Monte Carlo run:

\begin{equation}
\label{eq:nees_01}
\epsilon_i = (\boldsymbol{\hat{\theta}} - \boldsymbol{\theta_{g}}) \ \ \boldsymbol{\hat{\Sigma_{\theta}}} \ \ (\boldsymbol{\hat{\theta}} - \boldsymbol{\theta_{g}})^T
\end{equation}

Then normalize the error:
\begin{equation}
\label{eq:nees_02}
\bar{\epsilon} = \frac{1}{N} \sum_{i=0}^{N} \ \ \epsilon_i
\end{equation}

Then validates the error against the chi-square test with a 95 \% probability on both sides. Firstly,  compute chi-square density with $ \boldsymbol{N n_{\theta}}$ degree of freedom, where $\boldsymbol{N}$  is the total number of Monte Carlo run and $\boldsymbol{n_{\theta}}$ is the degree for the problem. Then, the test is valid when:

\begin{equation}
\label{eq:nees_03}
N \ \bar{\epsilon} \in [l_1,l_2] 
\end{equation}\\

Where for  95\% probability region   $l_1 = \chi^2_{N n_{\theta}}(0.05)$ and $l_2 =\chi^2_{N n_{\theta}}(0.95)$.

\subsection*{NCI}

This approach computes the bias $\tilde{\boldsymbol{\theta}}$ in each Monte Carlo run, then based on the $\boldsymbol{N}$ bias, it calculates the actual uncertainty $\boldsymbol{\Sigma}^{*}$, as follows:

\begin{equation}
\label{eq:nci_01}
\tilde{\boldsymbol{\theta}_i} = (\boldsymbol{\theta}_i - \boldsymbol{\theta}_g)
\end{equation}

Then, compute the actual uncertainty:
\begin{equation}
\label{eq:nci_02}
\boldsymbol{\Sigma}^{*} = \frac{1}{N} \sum_{i=1}^{N} \tilde{\boldsymbol{\theta}_i} \tilde{\boldsymbol{\theta}_i}^T
\end{equation}

Rong Li and Zhao introduce a new term \textbf{ \textit{credibility ratio}}  $\rho^{*}$, it represents the ratio between the estimated uncertainty $\boldsymbol{\Sigma_{\theta}} $ to the actual uncertainty $\boldsymbol{\Sigma}^{*}$:
\begin{equation}
\label{eq:nci_03}
\rho^{*} = \frac{\tilde{\boldsymbol{\theta}_i} \ (\boldsymbol{\Sigma_{\theta}})^{-1} \ \tilde{\boldsymbol{\theta}_i}^T}{\tilde{\boldsymbol{\theta}_i} \ (\boldsymbol{\Sigma^{*}})^{-1} \ \tilde{\boldsymbol{\theta}_i}^T}
\end{equation}

One of the criticises for the \textbf{NEES} is the arthematice average is strongly dependent on $\boldsymbol{\hat{\theta}}$, which mean if some iterations $ \epsilon_i > l_2$ and some other iterations $ \epsilon_i < l_1$ can compensate each others, then the test is valid, but it is a misleading results. Therefore, Rong Li and Zhao instead use the geometric average to minimize that dependency. Consequently, they define the \textbf{\textit{noncredibilty index}} \textbf{NCI} $(\boldsymbol{\gamma})$ as follows:
\begin{equation}
\label{eq:nci_04}
\gamma = \frac{10}{N} \sum_{i=1}^{N} | \log_{10} \rho_i|
\end{equation}

Then, the \textbf{\textit{inclination indicator}} \textbf{(I$^2$)} $(\nu)$:

\begin{equation}
\label{eq:nci_05}
\nu = \frac{10}{N} \sum_{i=1}^{N} \log_{10} \rho_i
\end{equation}

\vspace*{2cm}

\section{The Evaluation}

Each section begins with an evaluation map, and the plot depicted the scenario. The evaluation map represents the evaluation points in that section. Then, it ends with the results and a discussion. 

\subsection*{Overlapped Scenario}

\begin{figure*}[htb!]
\centering
\includegraphics[width= 0.8 \textwidth]{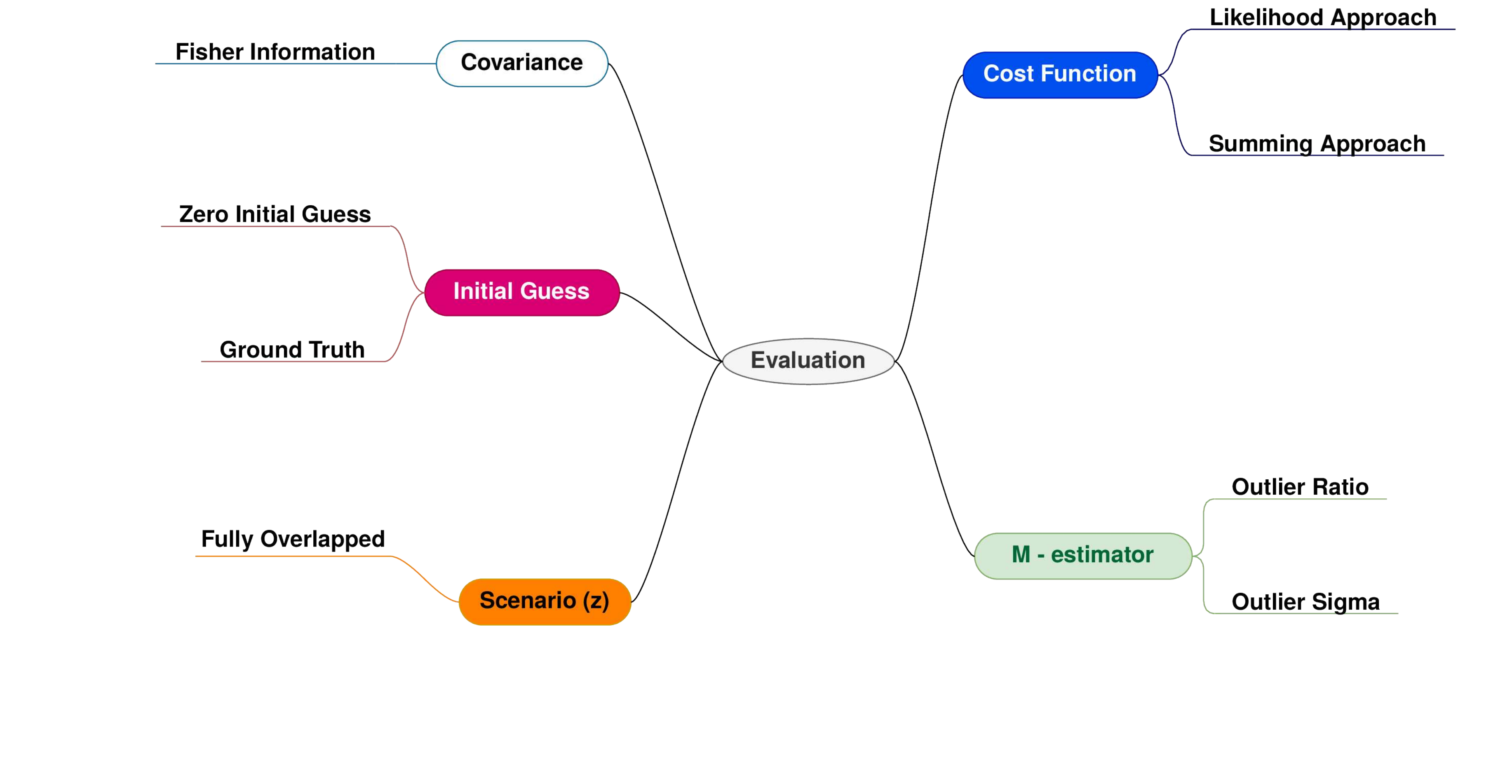}
\caption{The evaluation map for the full overlapped scenario.}
\label{fig:eval_1}
\end{figure*}

In this scenario, each point set includes 8 points, and each point from one set has a correspondence in the other set, which means a fully overlapped scenario, as depicted in Fig. \ref{fig:point_set_1}. The evaluation point here is, which approach is more sensitive to the initial guess. Besides, analyze the estimated covariance. Moreover, this section evaluates the impact of the M-estimator component on the estimated covariance based on the likelihood approach, to show the effect of the outlier on the estimated covariance, as the evaluation map shows in Fig. \ref{fig:eval_1}, where the outlier component parameter are $ \alpha = 0.2$ and $\Sigma_{outlier} = [ \begin{array}{cc} 100 & 0 \\ 0 & 100  \end{array}]$.

\begin{figure}[htb!]
\centering
\subfigure[\textbf{The summing approach}]{\label{fig:a}\includegraphics[width=70mm]{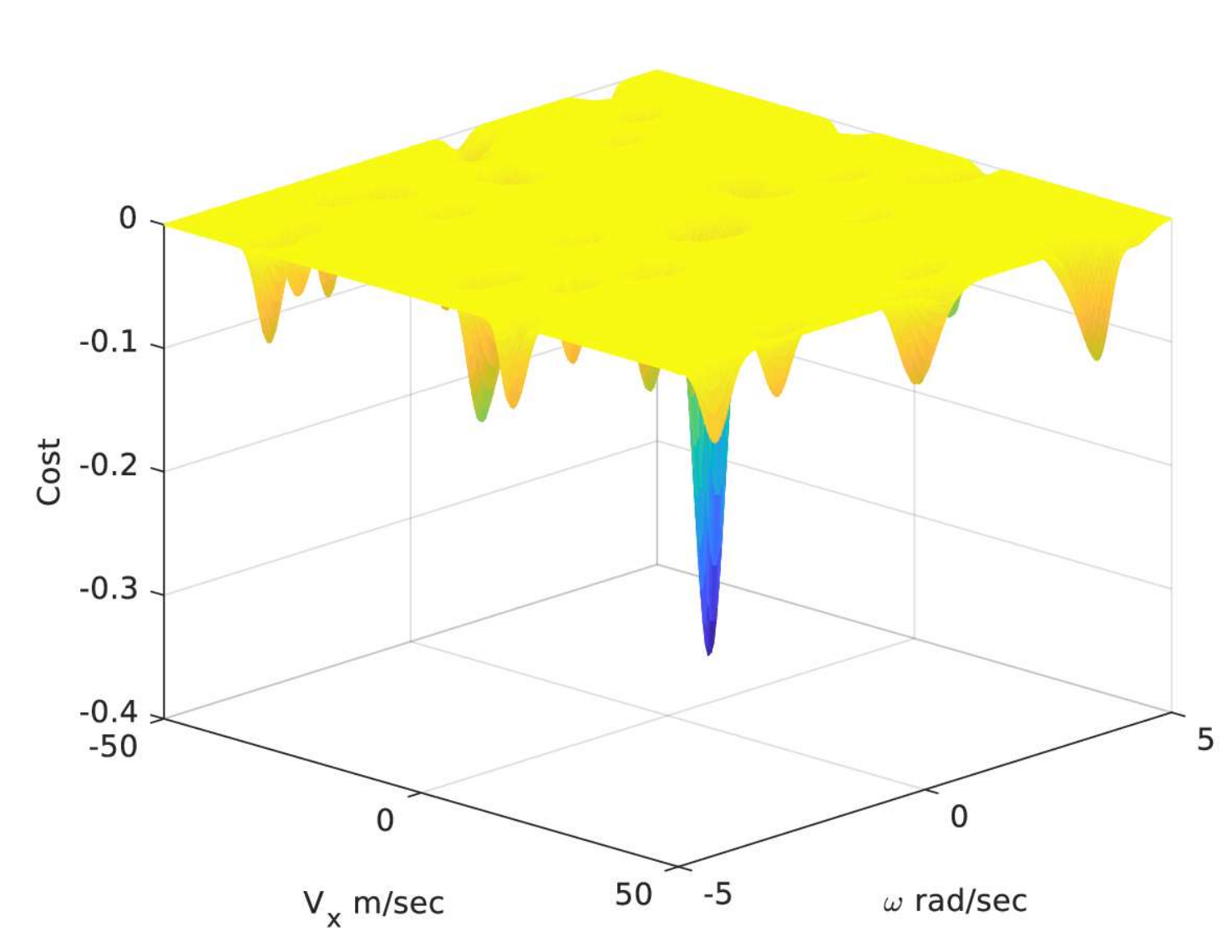}}
\subfigure[\textbf{The likelihood approach}]{\label{fig:b}\includegraphics[width=70mm]{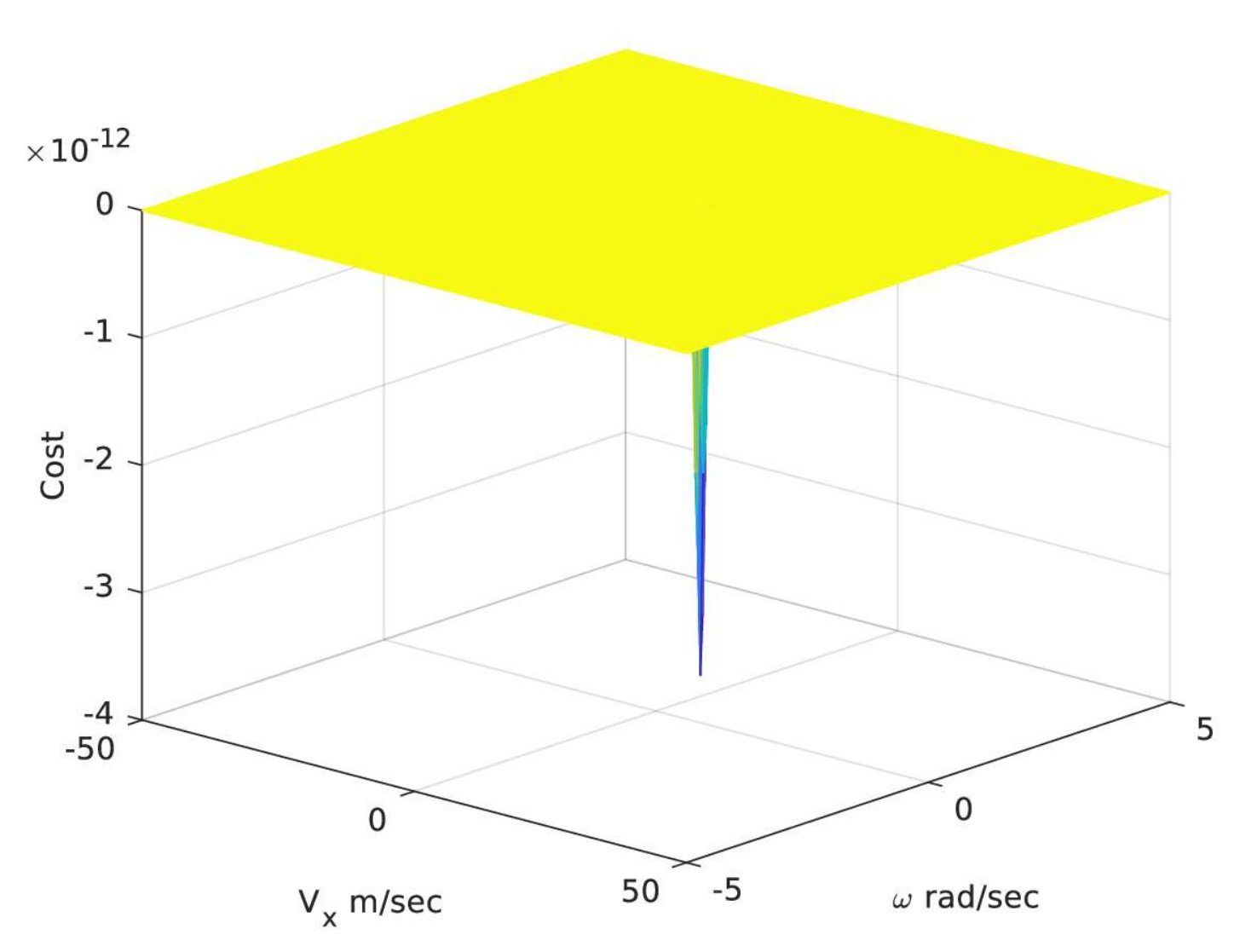}}
\caption{The cost surface for both approaches based on the fully overlapped scenario, where the summing cost surface includes many local minimums. But, the likelihood has only one global minimum at $\boldsymbol{\theta_g}$.}
\label{fig:sum_like_1}
\end{figure}

\begin{figure}[htb!]
\centering
\subfigure[\textbf{For $\boldsymbol{{\theta_g}}$ as initial guess, the estimation error in the translation}]{\label{fig:a}\includegraphics[width=65mm,height=50mm]{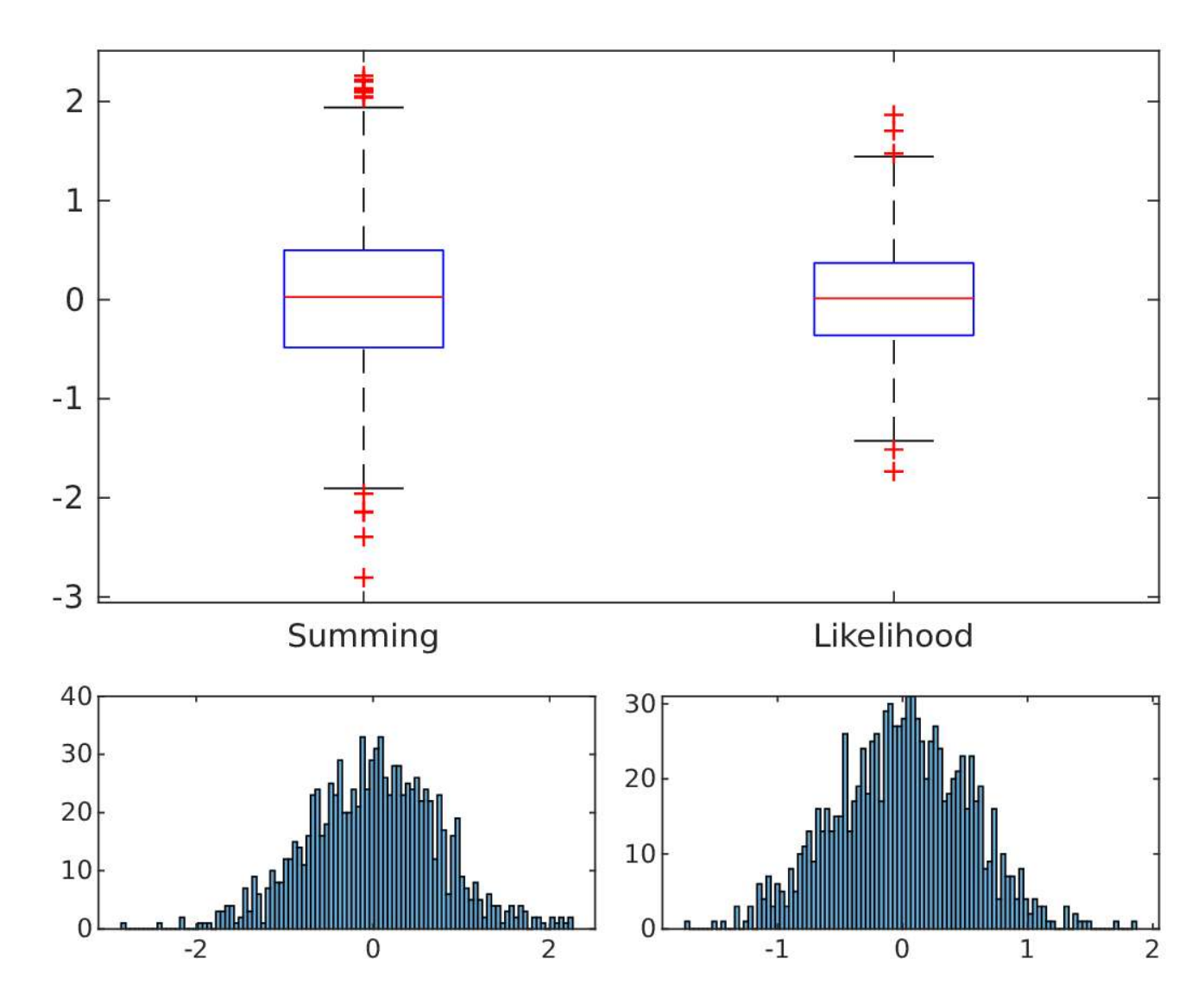}}
\subfigure[\textbf{For $\boldsymbol{{\theta_g}}$ as initial guess, the estimation error in the rotation}]{\label{fig:b}\includegraphics[width=65mm,height=50mm]{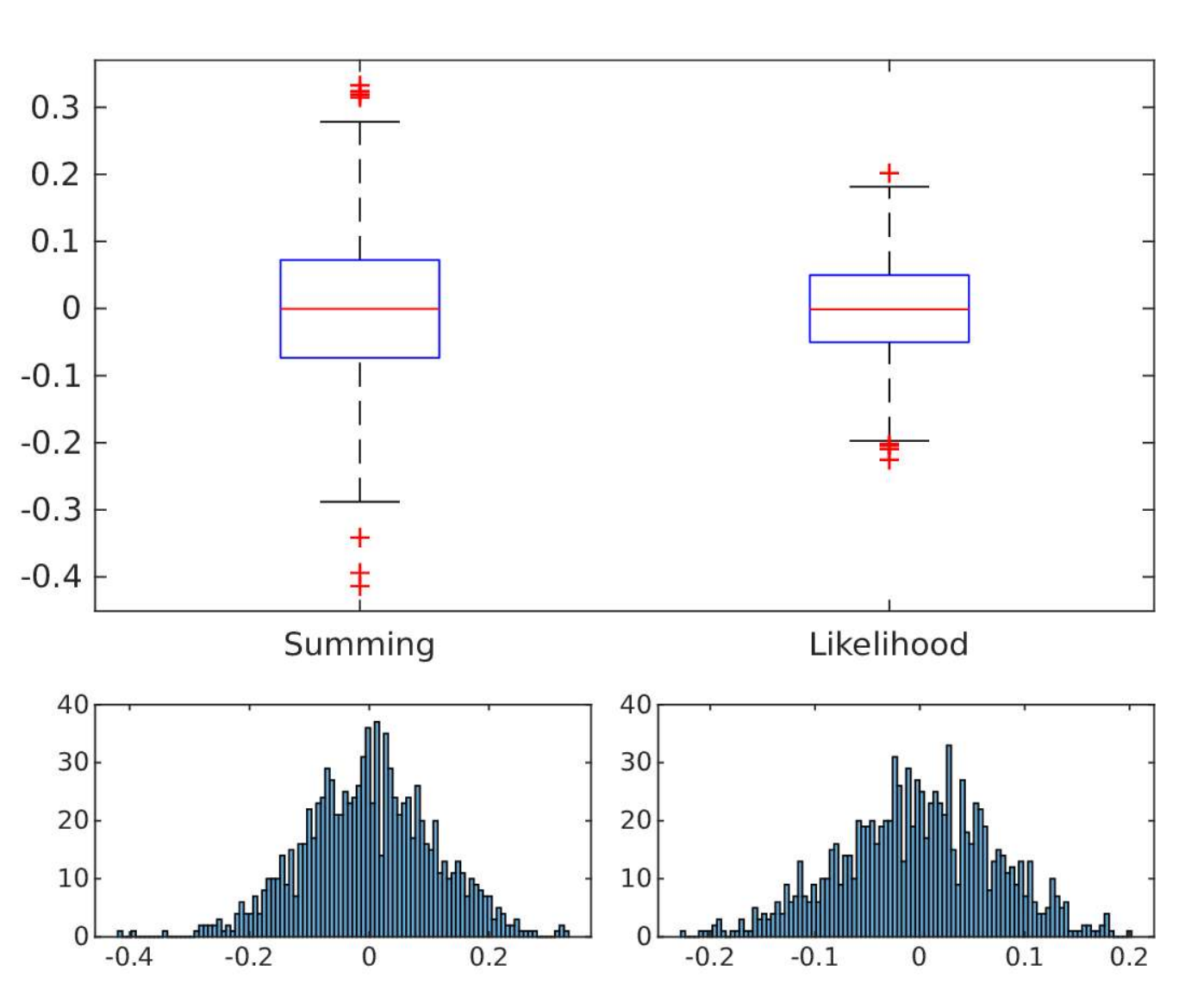}}
\subfigure[\textbf{For zero as initial guess, the estimation error in the translation}]{\label{fig:a}\includegraphics[width=65mm,height=50mm]{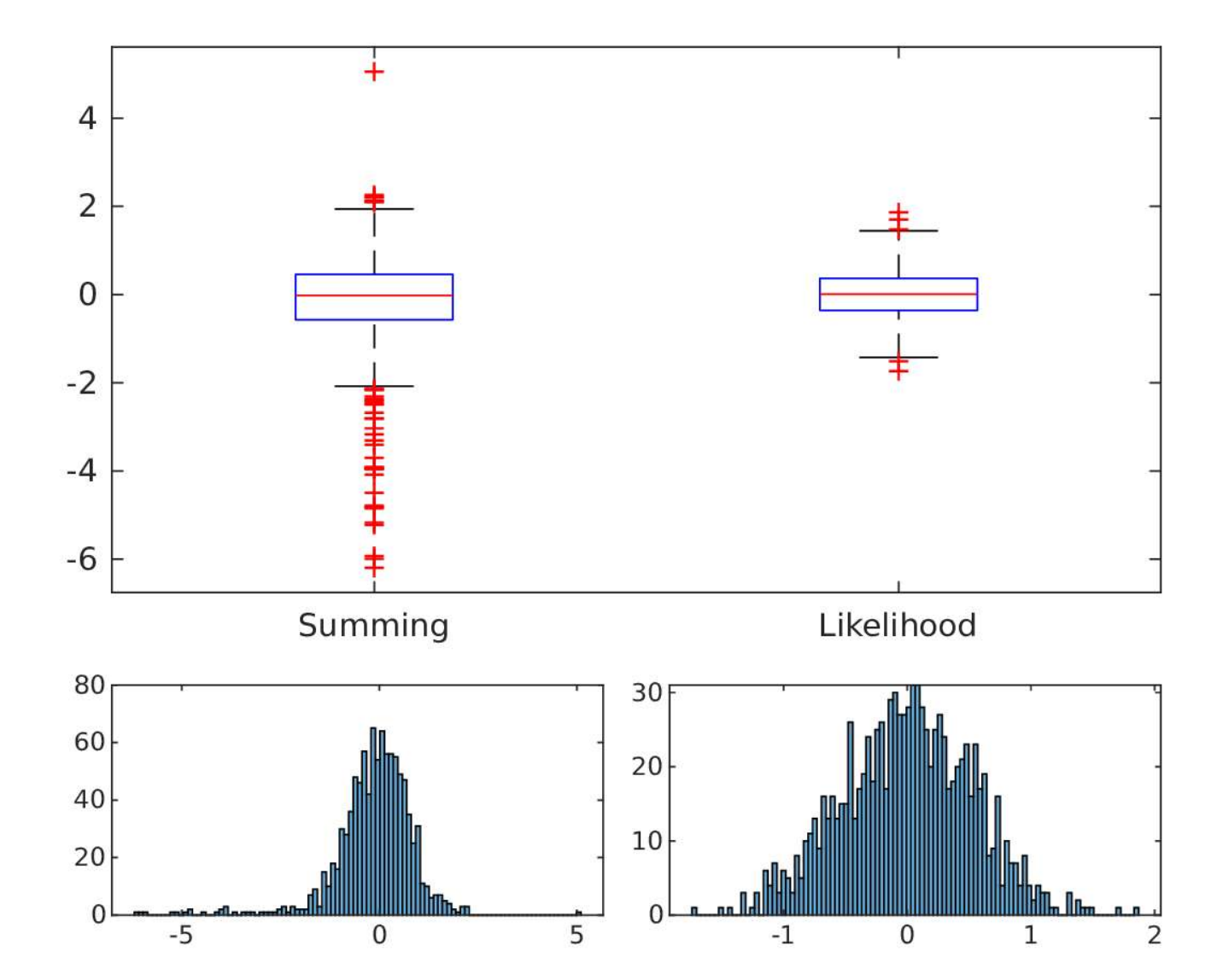}}
\subfigure[\textbf{For zero as initial guess, the estimation error in the rotation}]{\label{fig:b}\includegraphics[width=65mm,height=50mm]{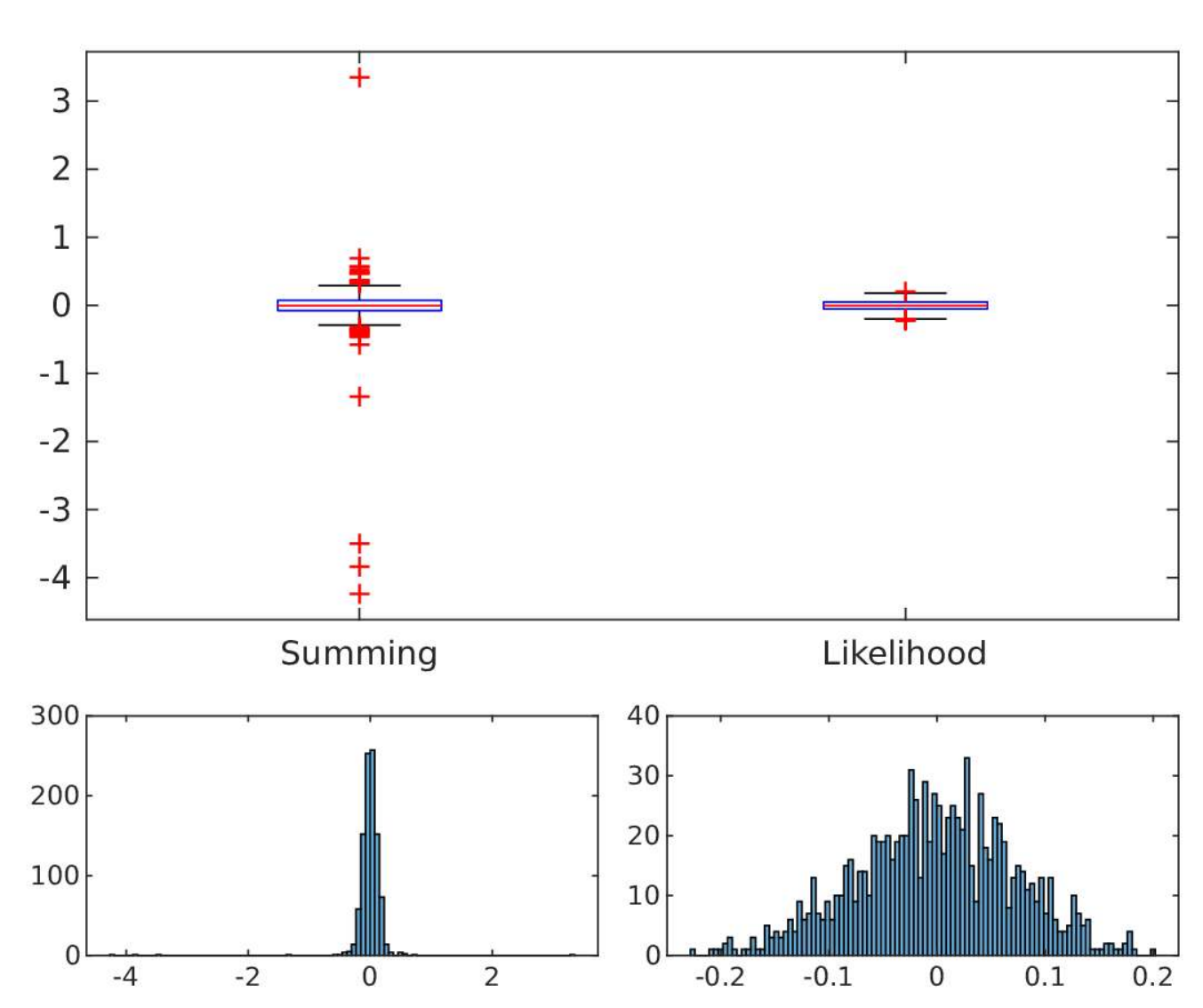}}

\caption{The impact of the goodness of the initial guess on the estimation error for the summing and the likelihood approach. Where the summing approach is more sensitive to the initial guess than the likelihood due to the local minimums.}
\label{fig:error_1}
\end{figure}

\begin{figure}[htb!]
\centering
\subfigure[\textbf{ The summing approach}]{\label{fig:b}\includegraphics[width=65mm,height=45mm]{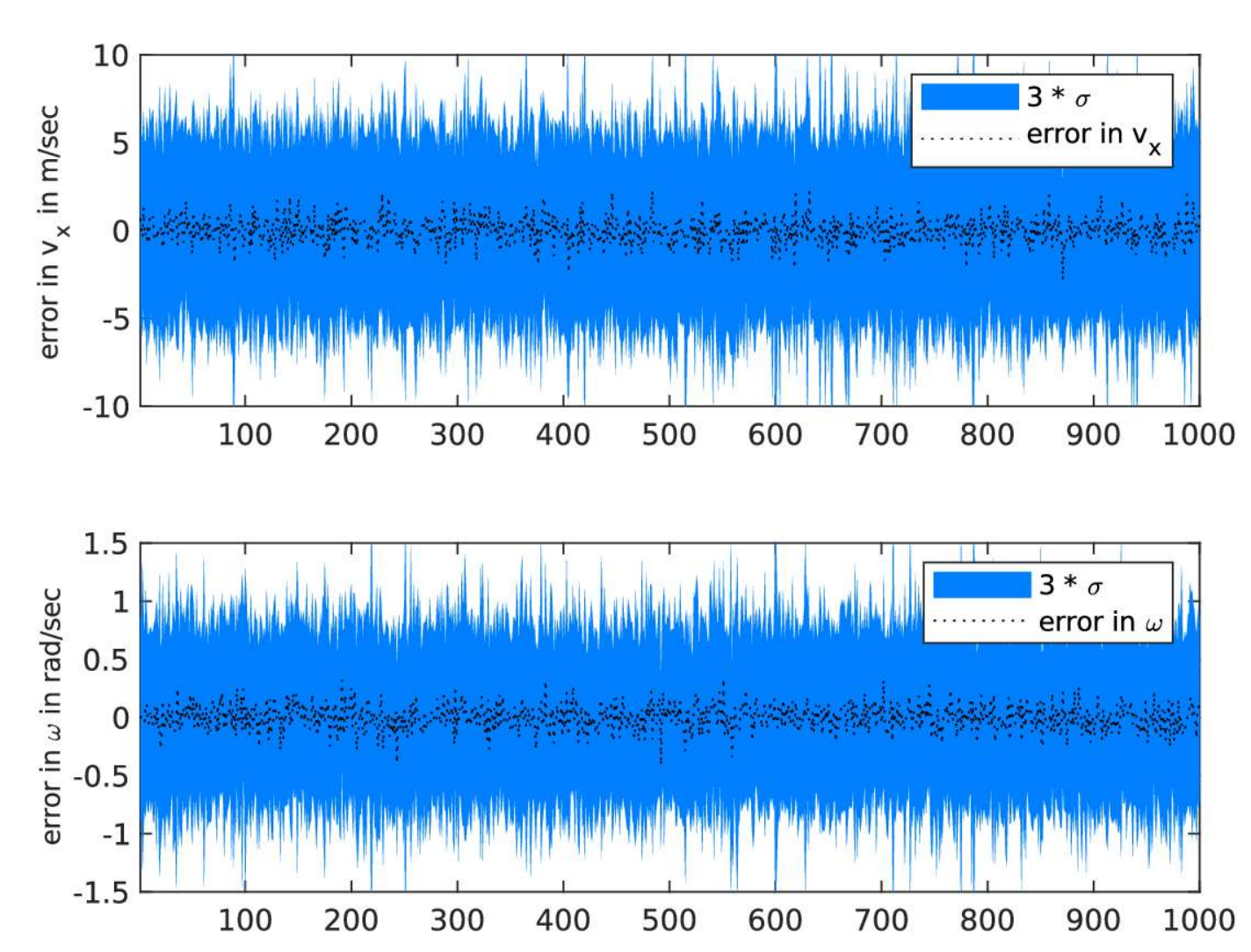}}
\subfigure[\textbf{ The likelihood approach}]{\label{fig:b}\includegraphics[width=65mm,height=45mm]{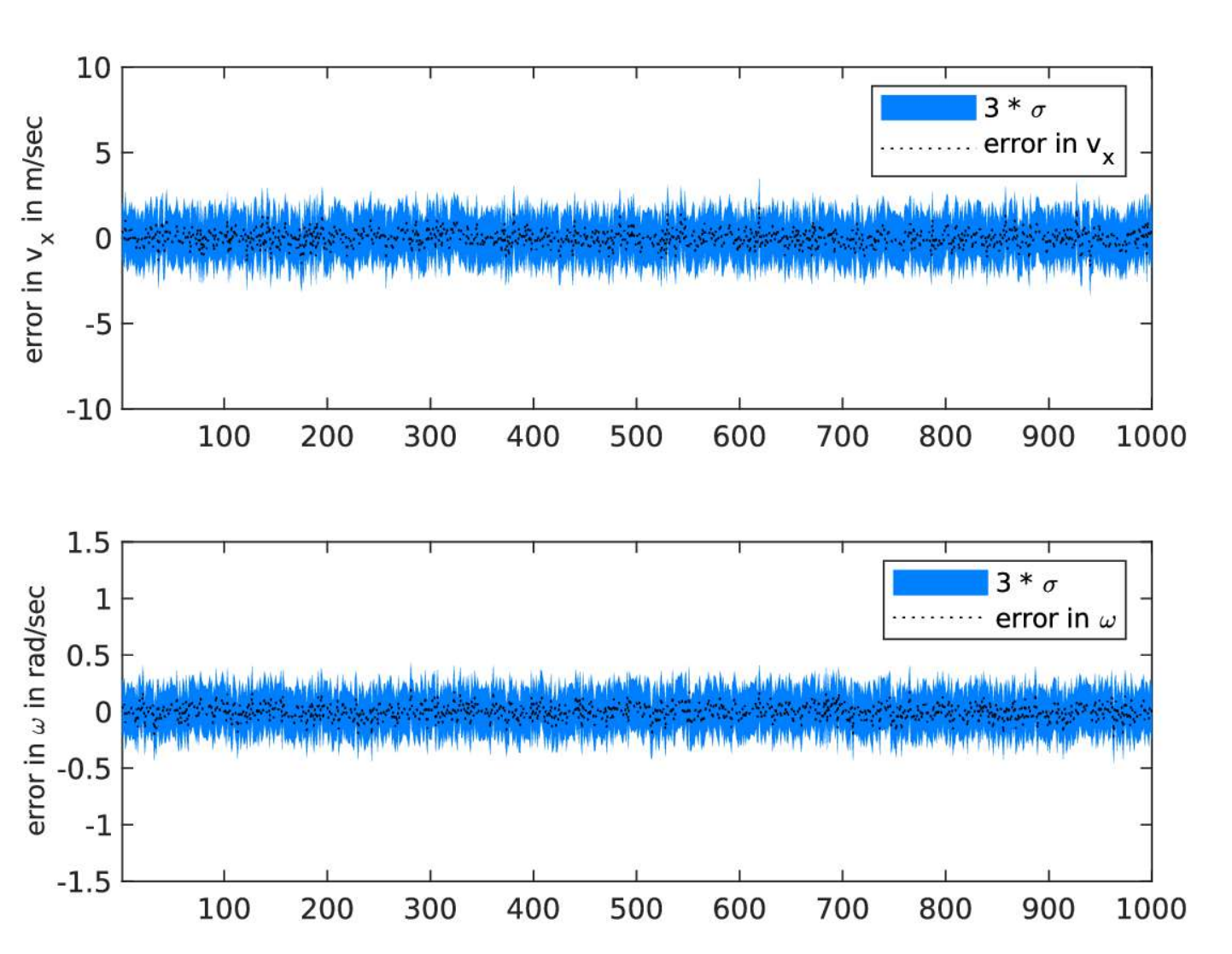}}
\caption{ This figure shows the estimation error (black) and the three-sigma bound of the estimated covariance in each dimension (light blue). The estimated covariance based on the summation approach is bigger than the likelihood one because the optimal minimum in the summing approach has more curvature than the likelihood.}

\label{fig: cov_check_1}
\end{figure}

\subsubsection*{Results and Discussion }

Firstly, Fig. \ref{fig:sum_like_1} shows the cost surface for the scenario based on the summing and the likelihood approach, where both approaches have one optimal minimum at $\boldsymbol{\theta_g}$. The summing approach shows a typical problem for multiple locals minimal, but the likelihood cost surface has only one minimum. Moreover, the estimation based on the likelihood approach Fig. \ref{fig:error_1}(a) is smaller than the estimation error based on the summing approach Fig. \ref{fig:error_1}(b). Furthermore, change the initial guess impacts more the summing approach, where the zero initial estimation error is greater than the ground truth, but the likelihood approach shows a consistent result against the change in the initial guess. Secondly, the optimal minimum in the summing approach has more curvature than the likelihood approach; thus, the estimated covariance based on the summing approach is overestimating the error compared to the likelihood approach, as depicted in Fig. \ref{fig: cov_check_1}. \\

\begin{figure}[h!]
\centering
\subfigure[\textbf{Summing approach with $\boldsymbol{{\theta_g}}$ as initial guess }]{\label{fig:a}\includegraphics[width=68mm]{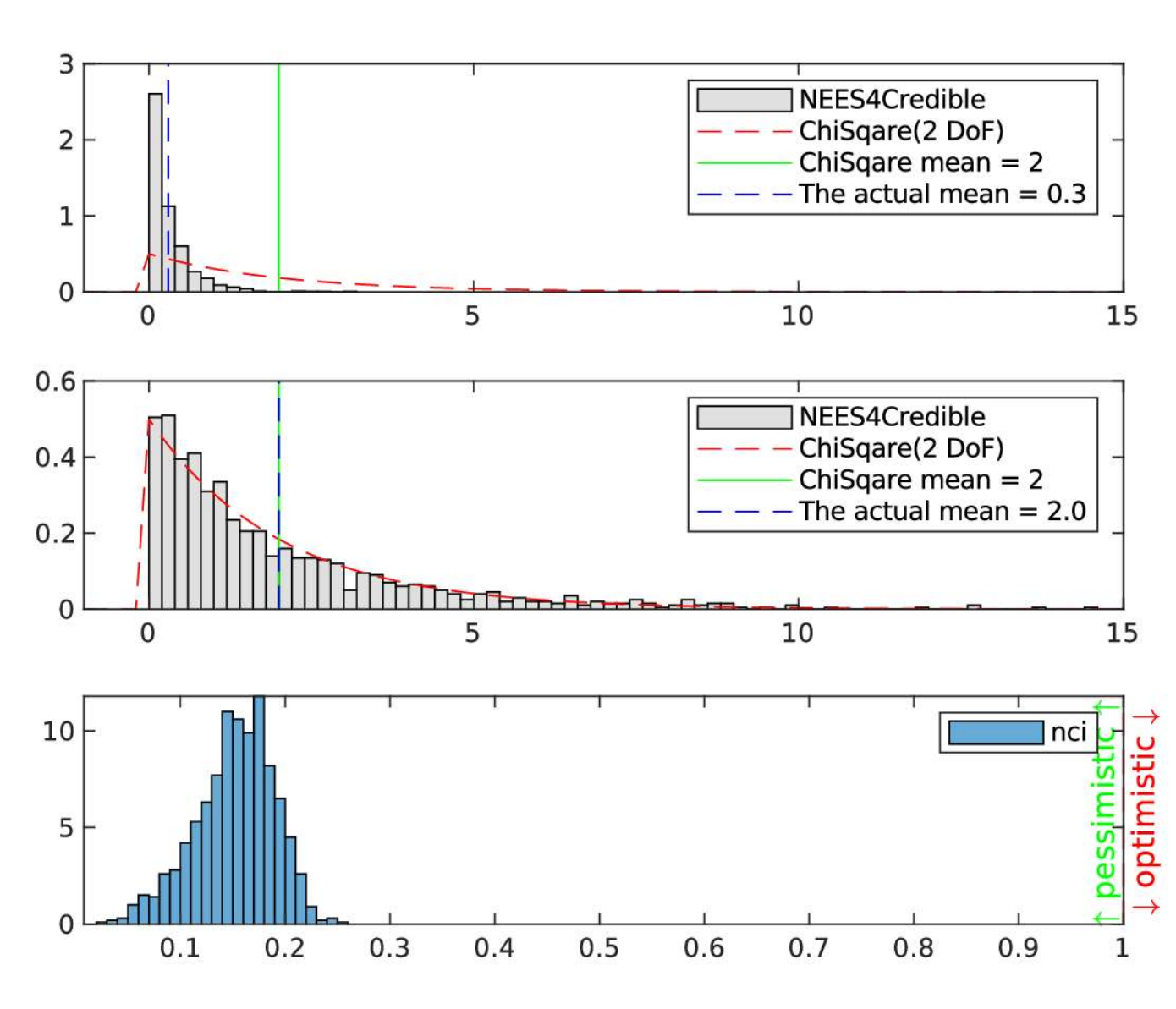}}
\subfigure[\textbf{Likelihood approach with $\boldsymbol{{\theta_g}}$ as initial guess}]{\label{fig:b}\includegraphics[width=68mm]{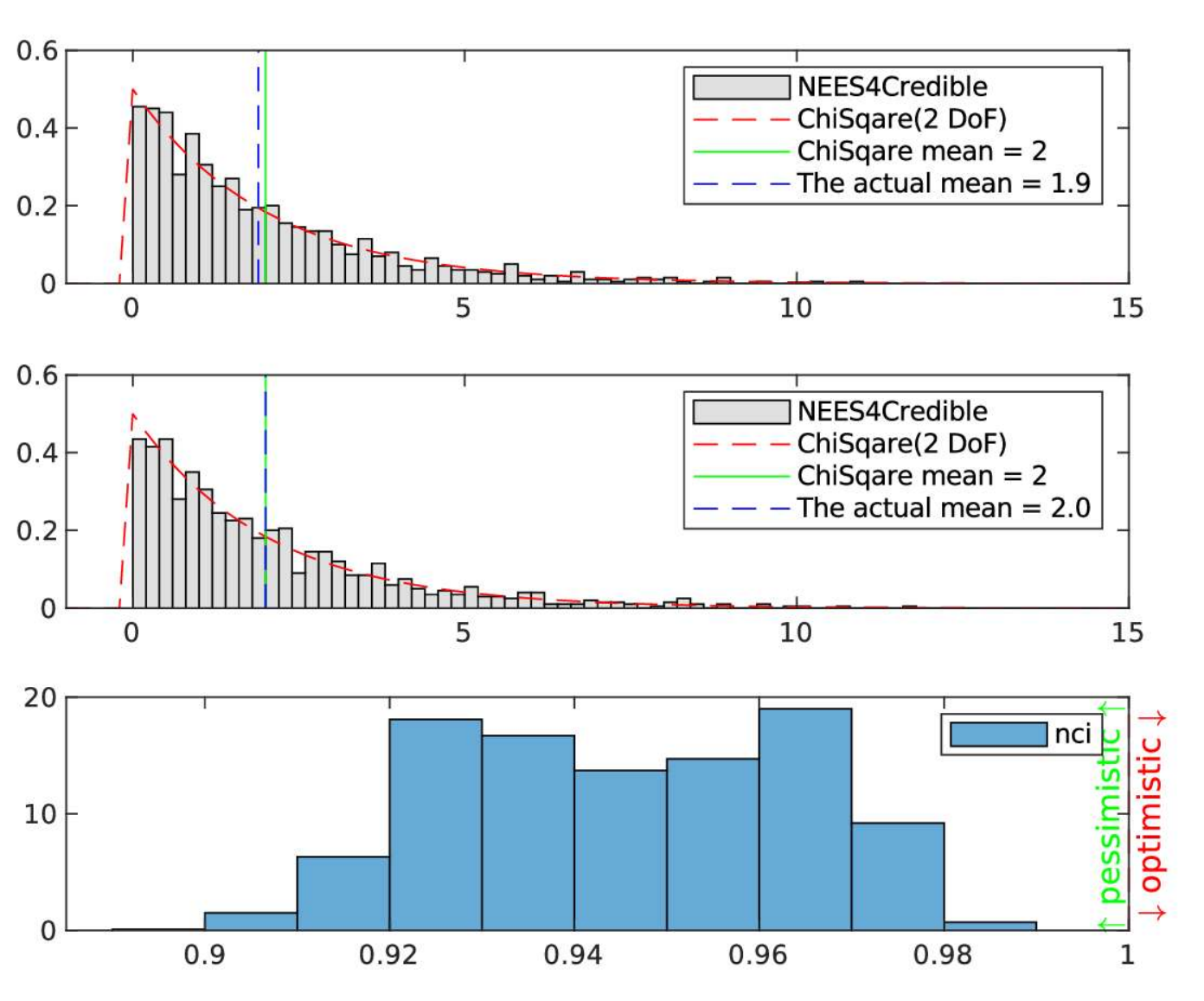}}
\subfigure[\textbf{Adding the outlier component to likelihood approach}]{\label{fig:b}\includegraphics[width=68mm]{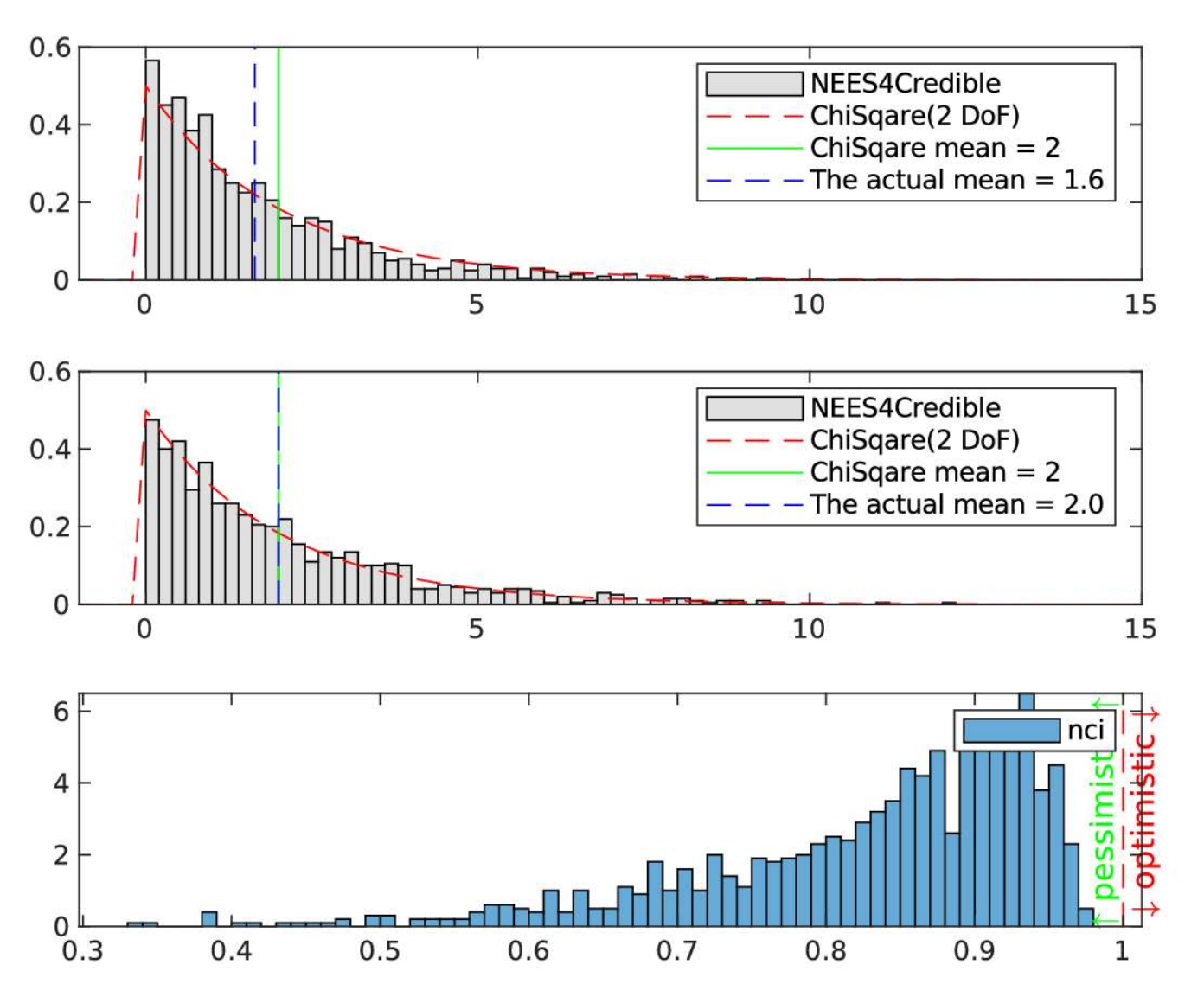}}
\caption{ The credibility test results for the full overlapped scenario using the ground truth $\boldsymbol{{\theta g}}$ as the initial guess. The summing approach shows more pessimistic results than the likelihood because the summing is overestimating the covariance. Moreover, adding the outlier component adds curvature to the optimal local minimum and that makes the covariance more uncertain.}
\label{fig: credibility_test_1}
\end{figure}

The credibility of this scenario is depicted in Fig. \ref{fig: credibility_test_1}, where each subfigure has three plots: The top one represents the NEES for the estimator, and the middle represents the NESS for the credible estimator, and the bottom represents the credibility ratio. NCI for the summing approach Fig. \ref{fig: credibility_test_1} (a) shows more pessimistic results than the likelihood approach Fig. \ref{fig: credibility_test_1} (b). On the one hand, NCI for the summing approach is more pessimistic because it overestimates the error, which is clear in the NEES for the estimator subplot, which is completely different from the NEES for credible estimator subplot. On the other hand, NEES values for the credible estimator and the actual estimator are close to each other, thus NCI for the likelihood approach close to one.\\

Lastly, adding model the outlier component in the cost function does not affect the estimation error, but it affects the uncertainty, where the outlier component adds curvature to the optimal minimum; thus, NCI becomes more pessimistic as shown in Fig. \ref{fig: credibility_test_1} (c).

\subsection*{Outlier Scenario}

\begin{figure*}[htb!]
\centering
\includegraphics[width=0.9\textwidth]{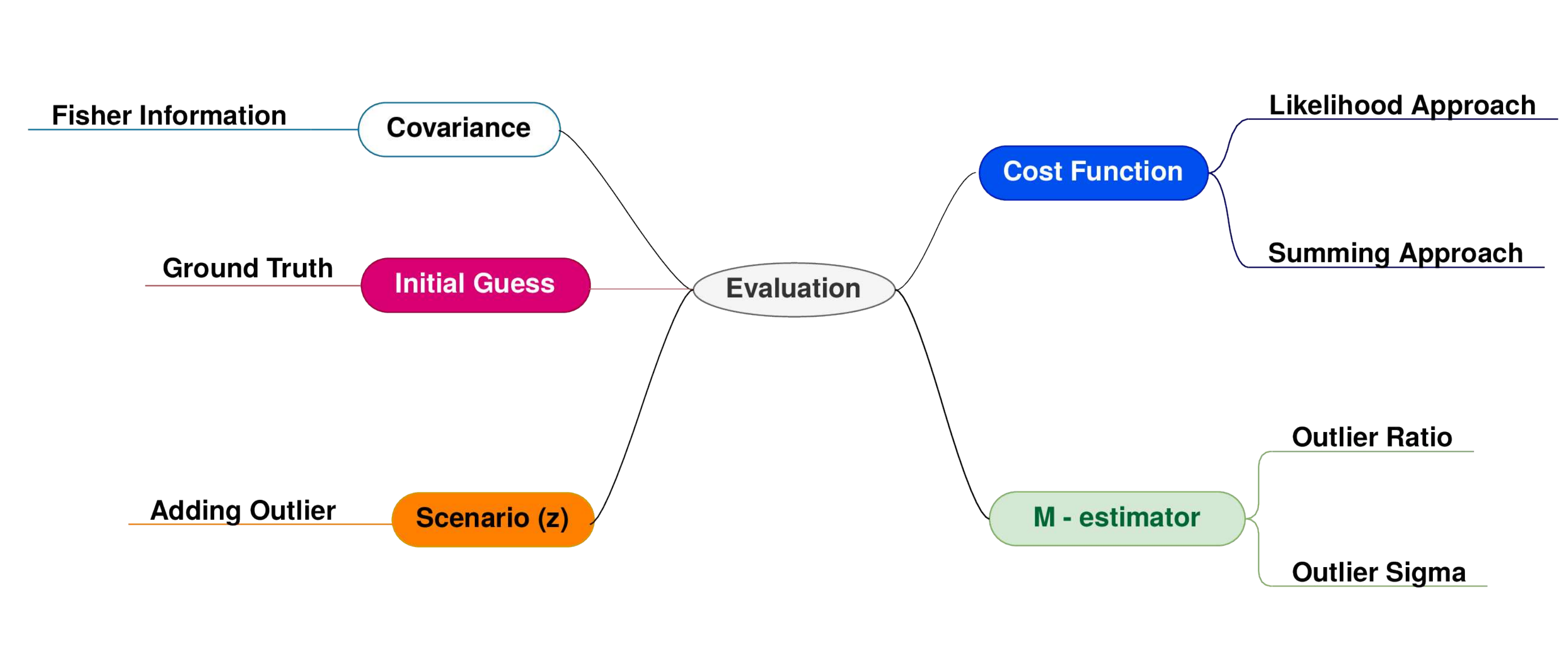}
\caption{ The evaluation map of the outlier scenario.}
\label{fig:eval_2}
\end{figure*}

\begin{figure*}[htb!]
\centering
\includegraphics[width=0.7\textwidth]{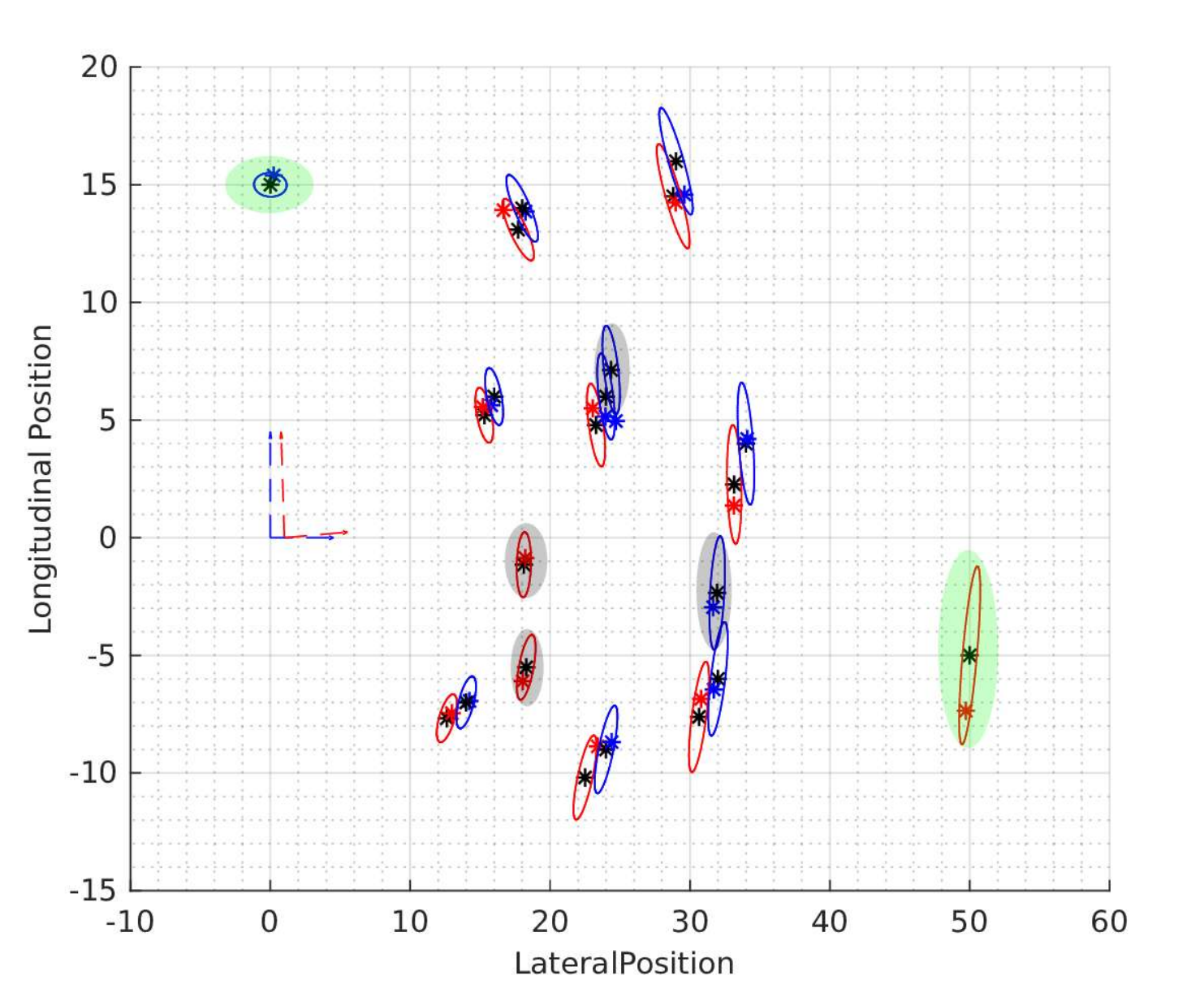}
\caption{This figure represents a point set registration problem with outliers. Where the scenario includes two types of an outlier: Close to the points as marked by gray and outrange outlier as marked by green.}
\label{fig:point_set_2}
\end{figure*}

This scenario examines the impact of the existing outlier in the point set, which means not each point in one point set has a correspondence in the second point set. Where, the outlier could be in different forms, such as points closed to the inliers or a point far from the inlier points, as shown in Fig. \ref{fig:point_set_2}. This scenario evaluates the outlier impact on the summing, likelihood, and credibility test. This section is not interested in evaluating the initial guess. Thus, the optimizer uses the ground truth as the initial guess. Besides,  evaluating the robust cost function implementation to suppress the outlier effect, the complete evaluation aspects depicted in \ref{fig:eval_2}.


\begin{figure}[htb!]
\centering
\subfigure[\textbf{The summing approach}]{\label{fig:a}\includegraphics[width=70mm]{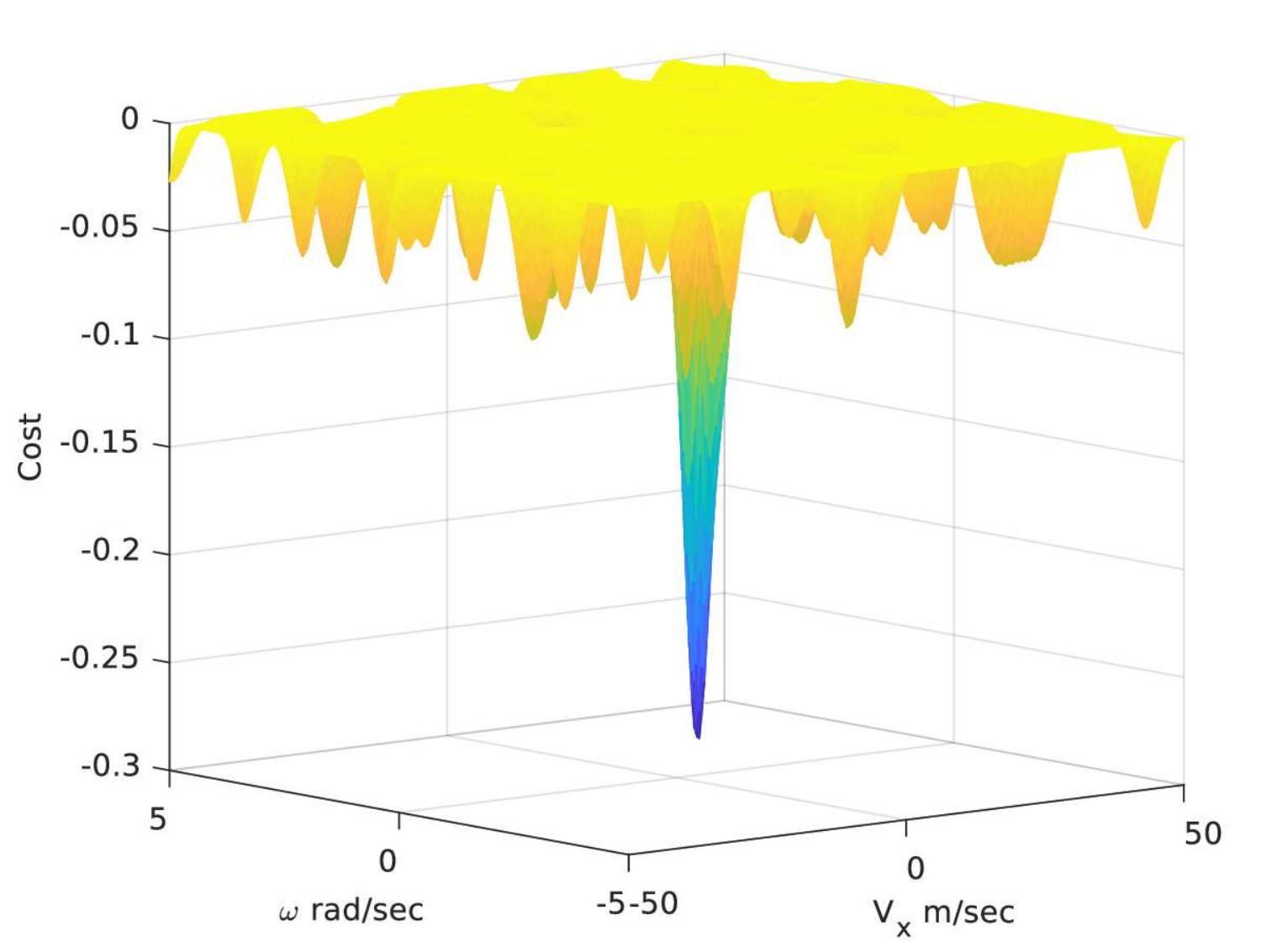}}
\subfigure[\textbf{The likelihood approach}]{\label{fig:b}\includegraphics[width=70mm]{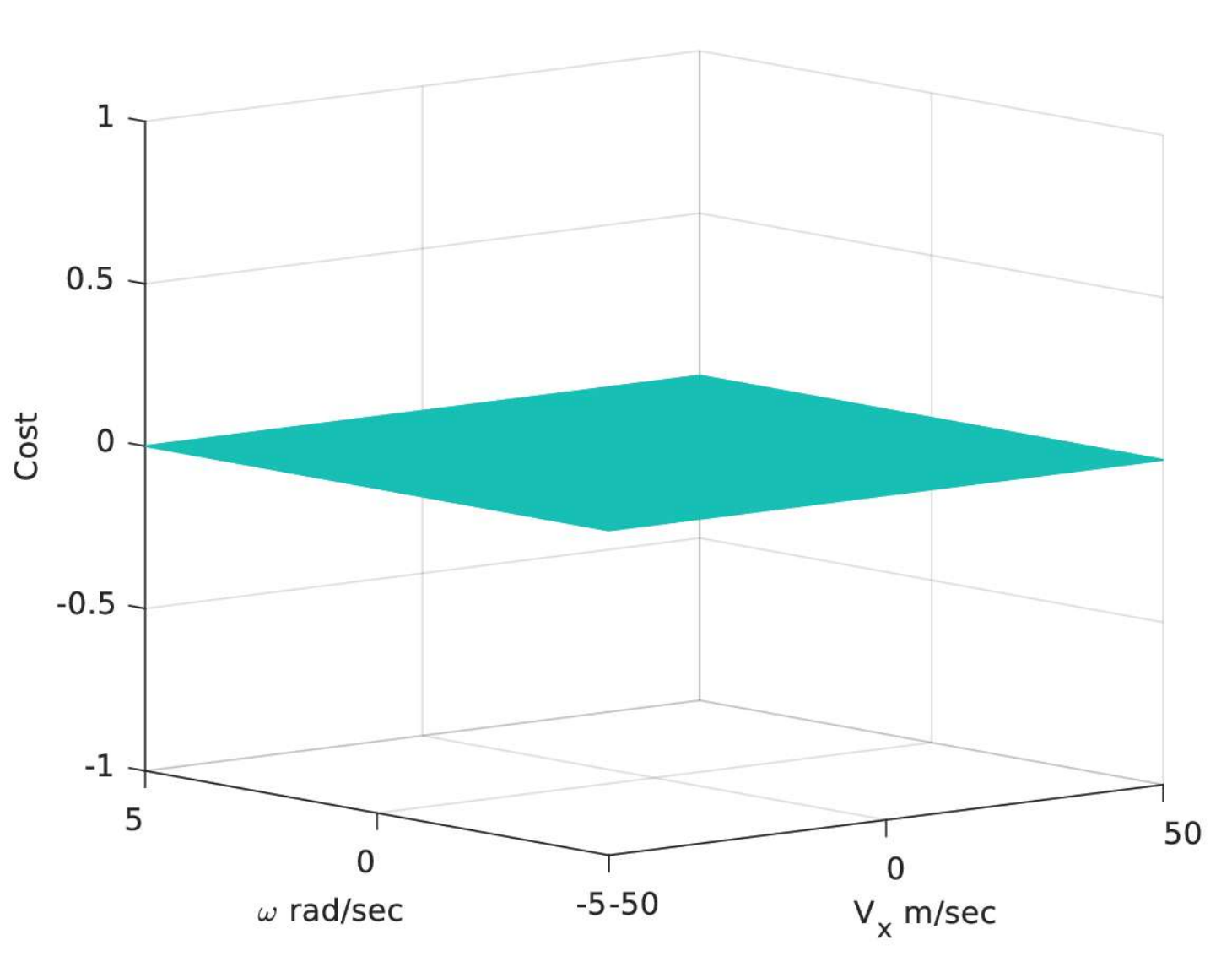}}
\subfigure[\textbf{The robust cost function approach}]{\label{fig:b}\includegraphics[width=70mm]{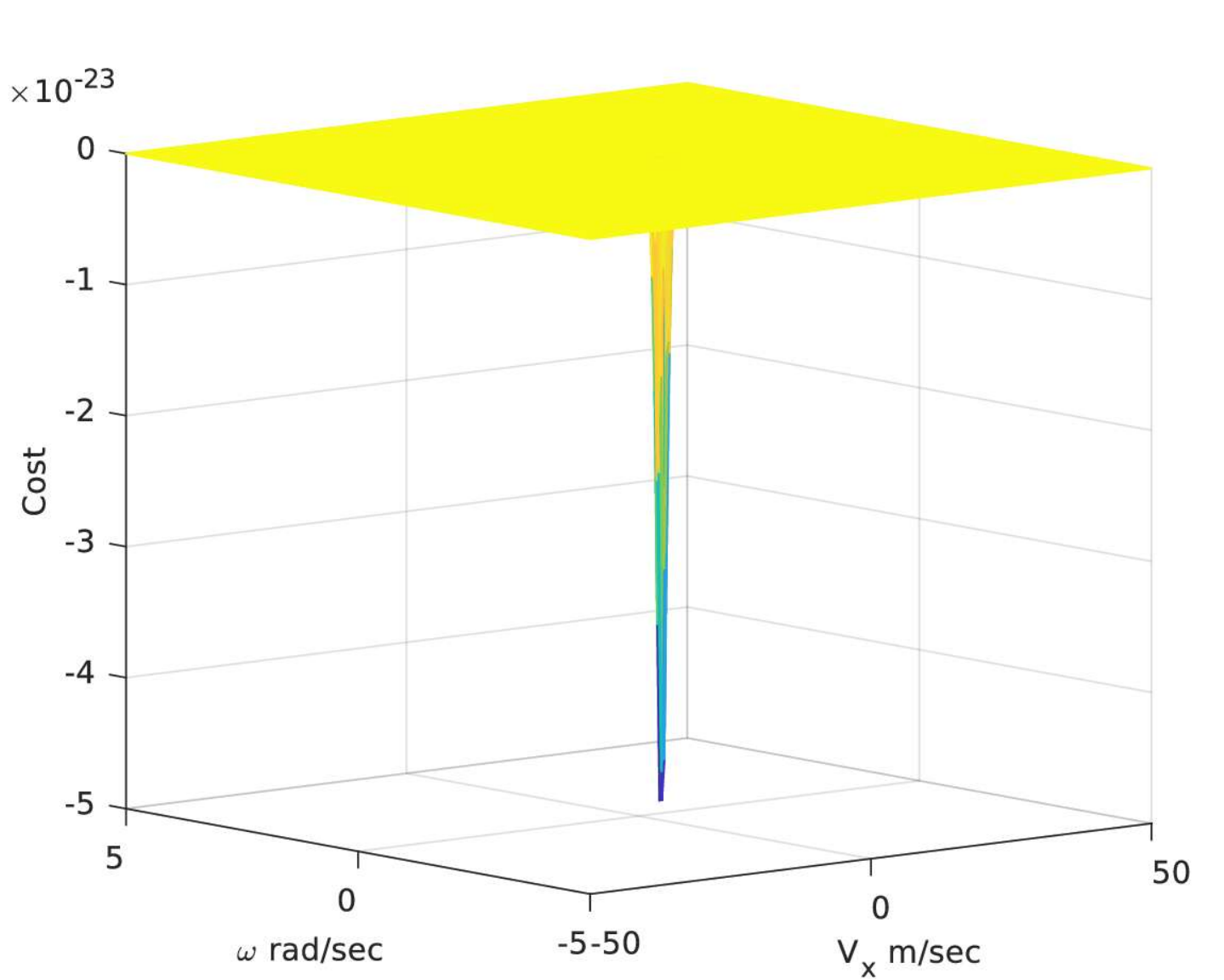}}
\caption{The cost surface for both approaches based on the outlier scenario, where the summing cost surface includes many local minimums and one global minimum at $\boldsymbol{\theta_g}$. But, the likelihood cost surface gets zero, because of the product of the outlier cost surface. Nevertheless, adding the outlier component in the likelihood approach gives one global minimum at $\boldsymbol{\theta_g}$.}
\label{fig:cost_2}
\end{figure}

\subsubsection*{Results and Discussion}

\begin{figure}[htb!]
\centering
\subfigure[\textbf{The estimation error in the translation }]{\label{fig:a}\includegraphics[width=70mm]{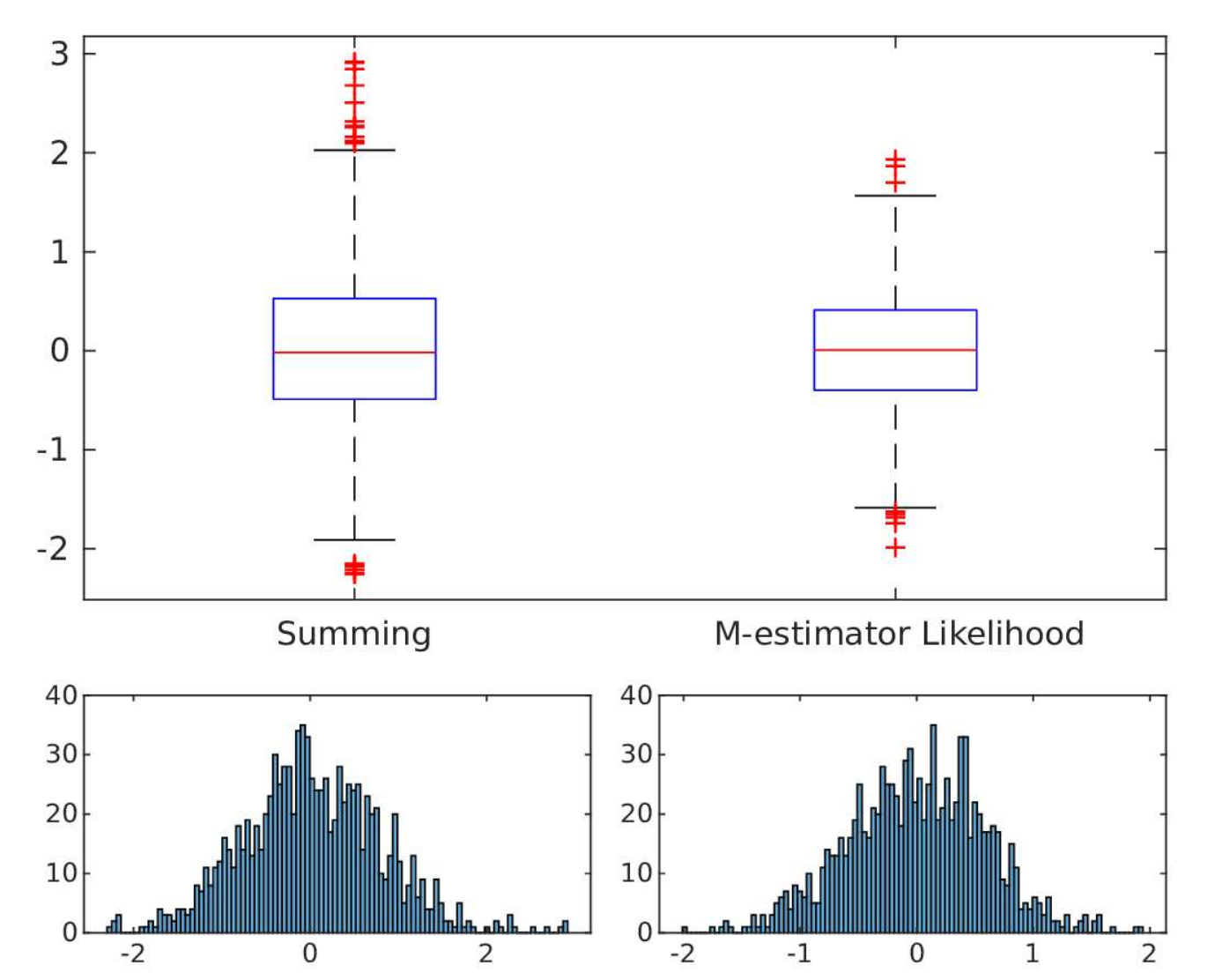}}
\subfigure[\textbf{The estimmation error  in rotation }]{\label{fig:b}\includegraphics[width=70mm]{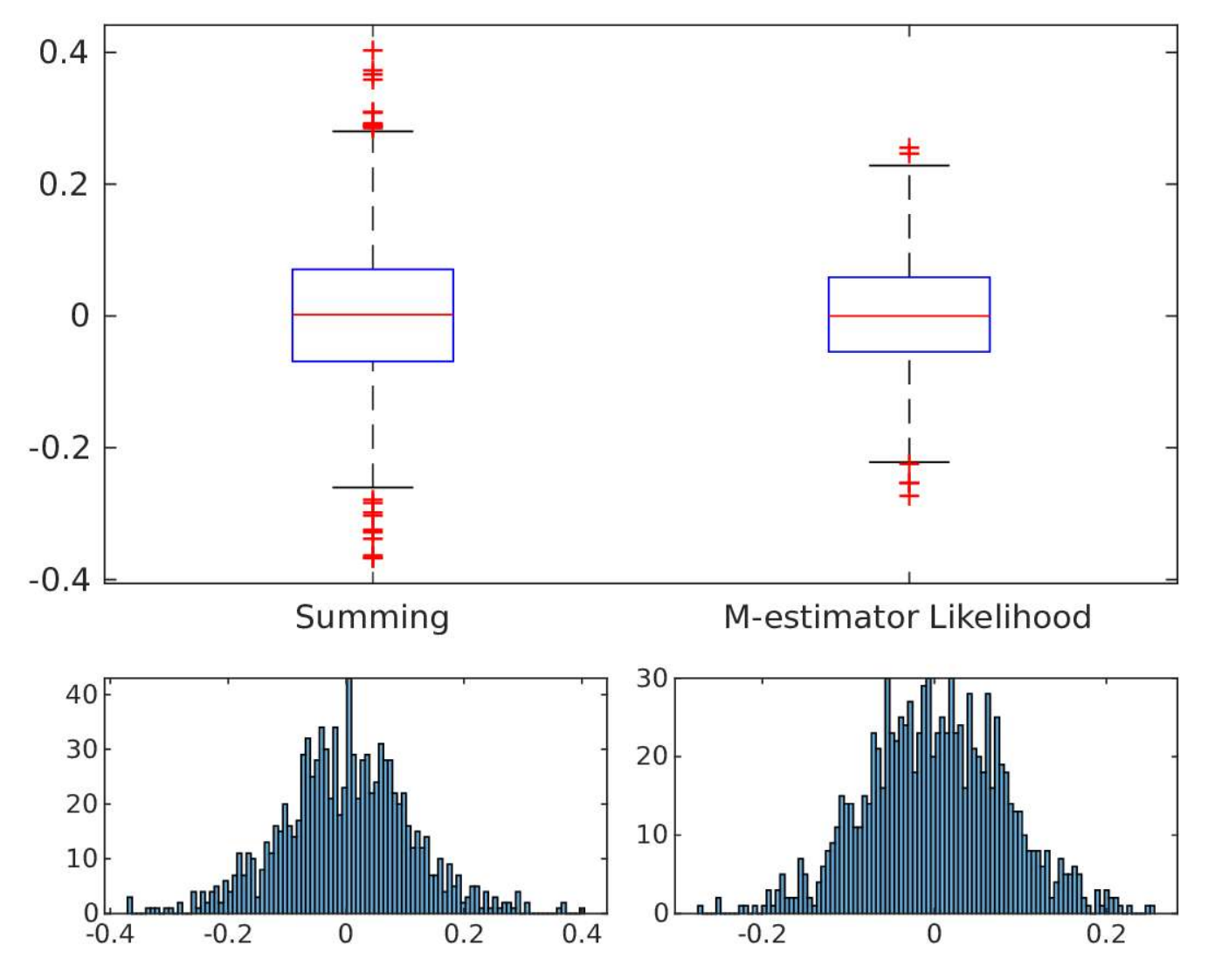}}
\caption{ The estimation error in the summing approach and in the likelihood approach after adding the doppler component to the cost function.}
\label{fig:error_3}
\end{figure}

Firstly, Fig. \ref{fig:cost_2} shows the cost surface for the summing and the likelihood approach, where the summing has one global minimum Fig. \ref{fig:cost_2} (a), but the cost surface based on the likelihood gets zero. Therefore, Fig. \ref{fig:cost_2} (c) shows the behaviour of the like hood approach after modelling the outlier component into the cost function following the corrupted Gaussian approach to build a robust cost function, where he outlier ratio $(\alpha = 0.2)$ and the outlier covariance is $\Sigma_{outlier} = [ \begin{array}{cc} 100 & 0 \\ 0 & 100 \end{array}]$. Modeling the outlier in the cost function suppresses the outlier effect; thus, the cost surface has only one global minimum. However, modeling the outlier component is a challenging task in the real scenario, but in the synthetic scenario, we know this ratio and the distribution of the points beforehand. Secondly, the estimation error based on the summing approach is greater than the estimation error based on the M-estimator likelihood approach, as depicted in Fig. \ref{fig:error_3}. \\

\begin{figure}[htb!]
\centering
\subfigure[\textbf{The summing approach}]{\label{fig:a}\includegraphics[width=70mm,height=60mm]{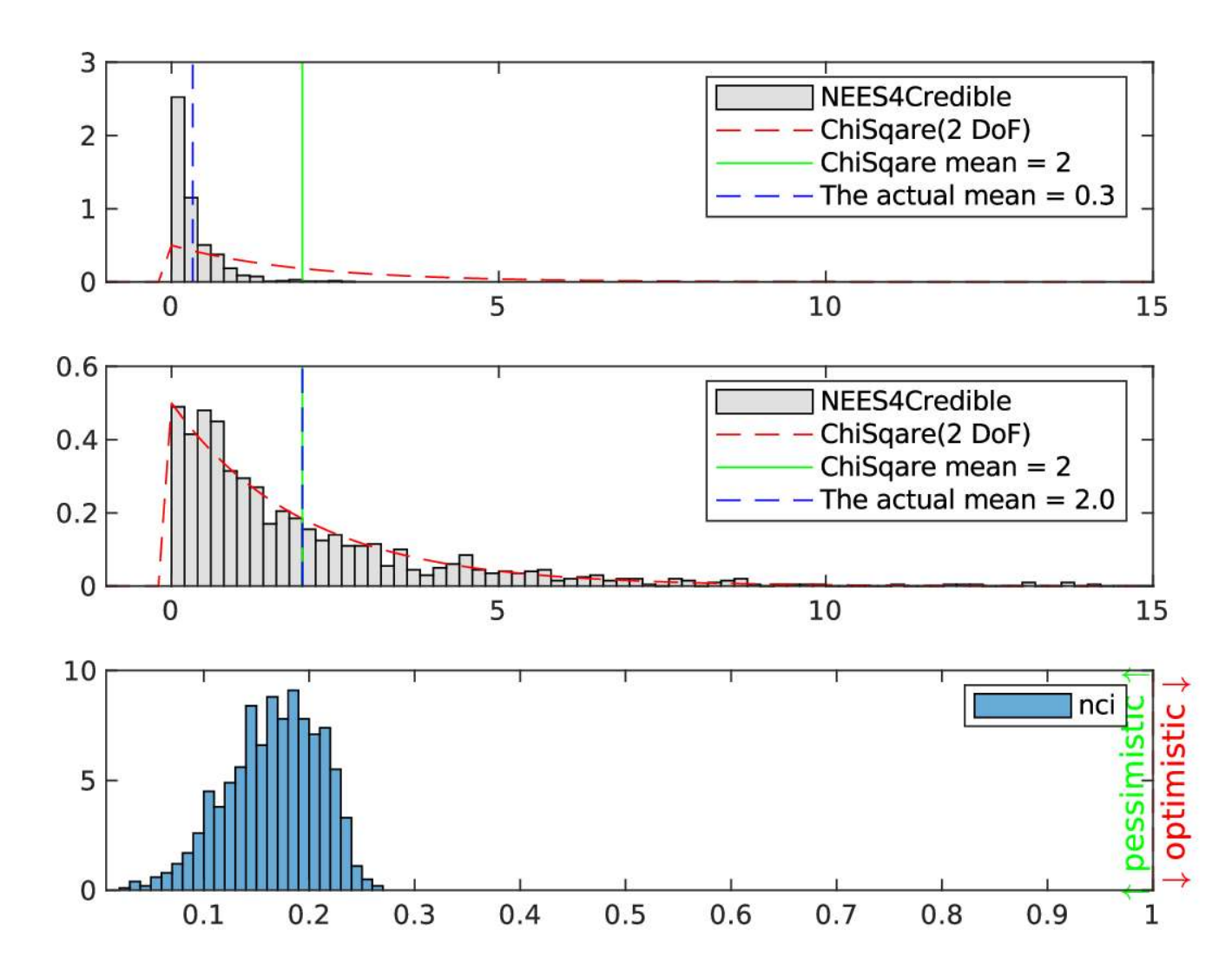}}
\subfigure[\textbf{The likelihood approach}]{\label{fig:b}\includegraphics[width=70mm,height=60mm]{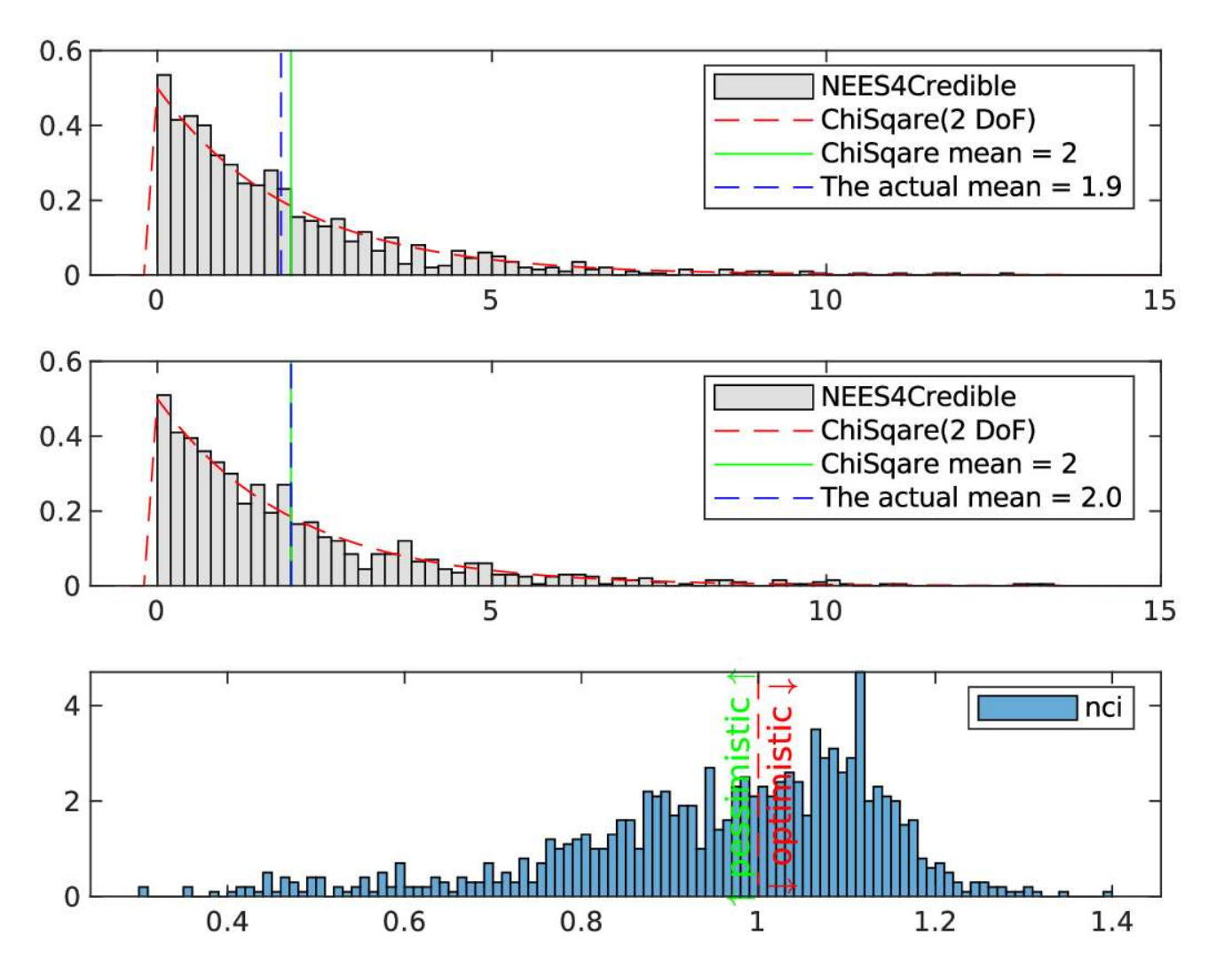}}
\caption{The credibility test results for the outlier scenario, where (a) represents the summing approach results and (b) the likelihood results after including the outlier component into the cost function. The summing approach gives more pessimistic results than the likelihood.}
\label{fig:credibility_test_2}
\end{figure}

The credibility test for the summing approach in this scenario is almost the same as the fully overlapped scenario. However, the likelihood has two components: The pessimistic one and the optimistic component, as shown in FIg.  \ref{fig:credibility_test_2}. One possible clarification for the pessimistic part is, adding the outlier component adds more curvature to the global minimum, and for the optimistic is, each point set has more points (more information) even if it is an outlier. 

\clearpage
\subsection*{Clustering Scenario}

\begin{figure*}[htb!]
\centering
\includegraphics[width=0.7\textwidth]{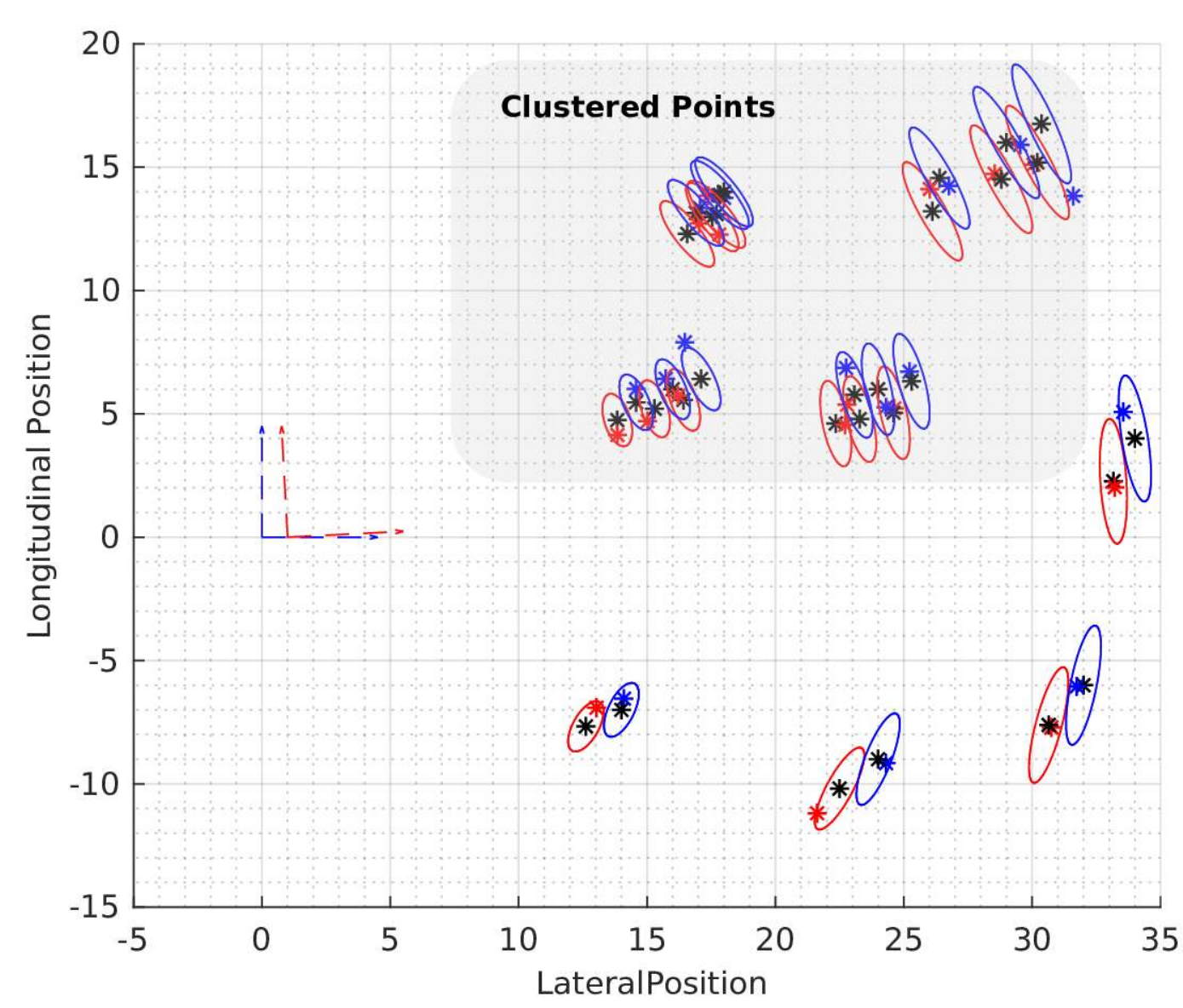}
\caption{A fully overlapped scenario, where each point in one point set has a correspondence in the second point set. Where the scenario introduces the clustered points in the overlapped scenario.}
\label{fig:point_set_3}
\end{figure*}

This section is an extension of the fully overlapped scenario but with adding more inlier points but clustered with some existing target, as shown in Fig. \ref{fig:point_set_3}. Therefore, the evaluation point is the impact of these clustered points on the global minimum. The evaluation of the initial guess is not the focus of the section, the evaluation points depicted in Fig. \ref{fig:eval_3}. As long as this is a fully overlapped scenario with extra points, the expected results should be the same or even better than the fully overlapped scenario.

\begin{figure*}[htb!]
\centering
\includegraphics[width=0.9 \textwidth]{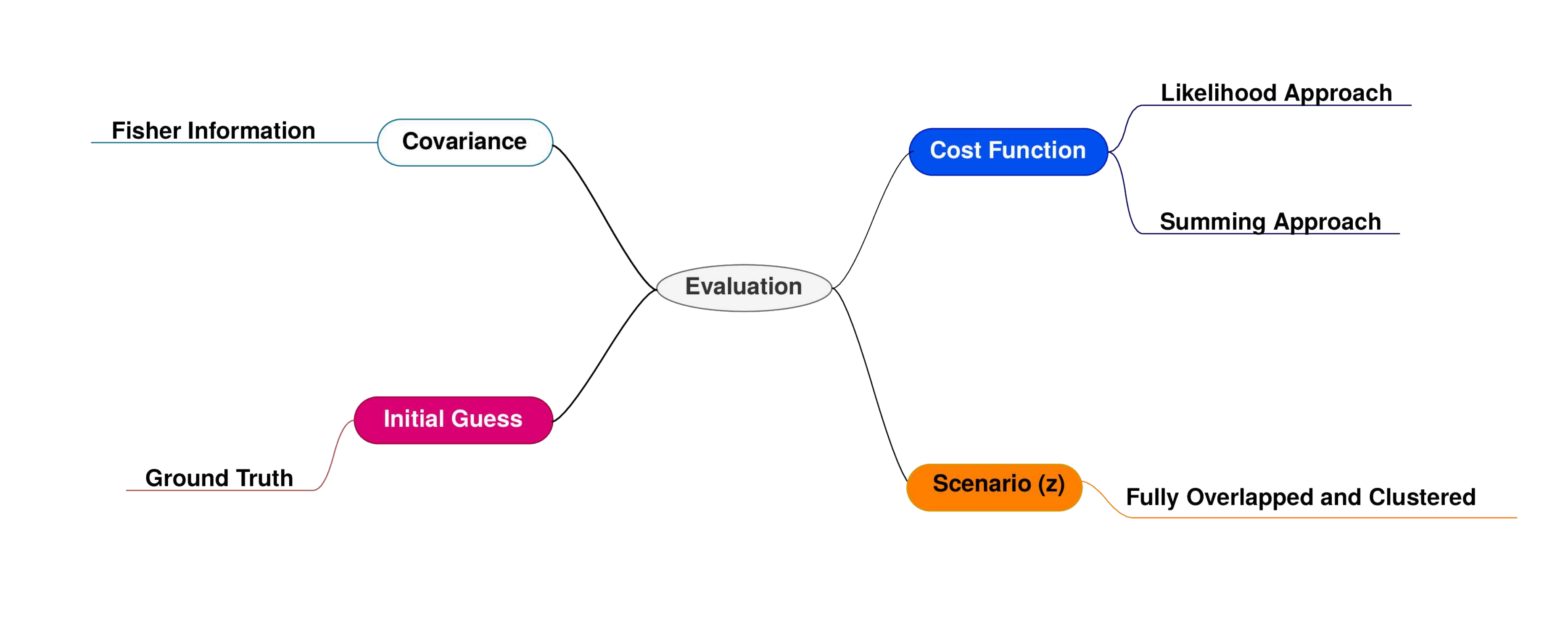}
\caption{ The evaluation map for the clustering points scenario.}
\label{fig:eval_3}
\end{figure*}

\subsubsection*{Results and Discussion}

\begin{figure}[htb!]
\centering
\subfigure[\textbf{The likelihood cost surface for the clustered scenario. }]{\label{fig:a}\includegraphics[width=70mm]{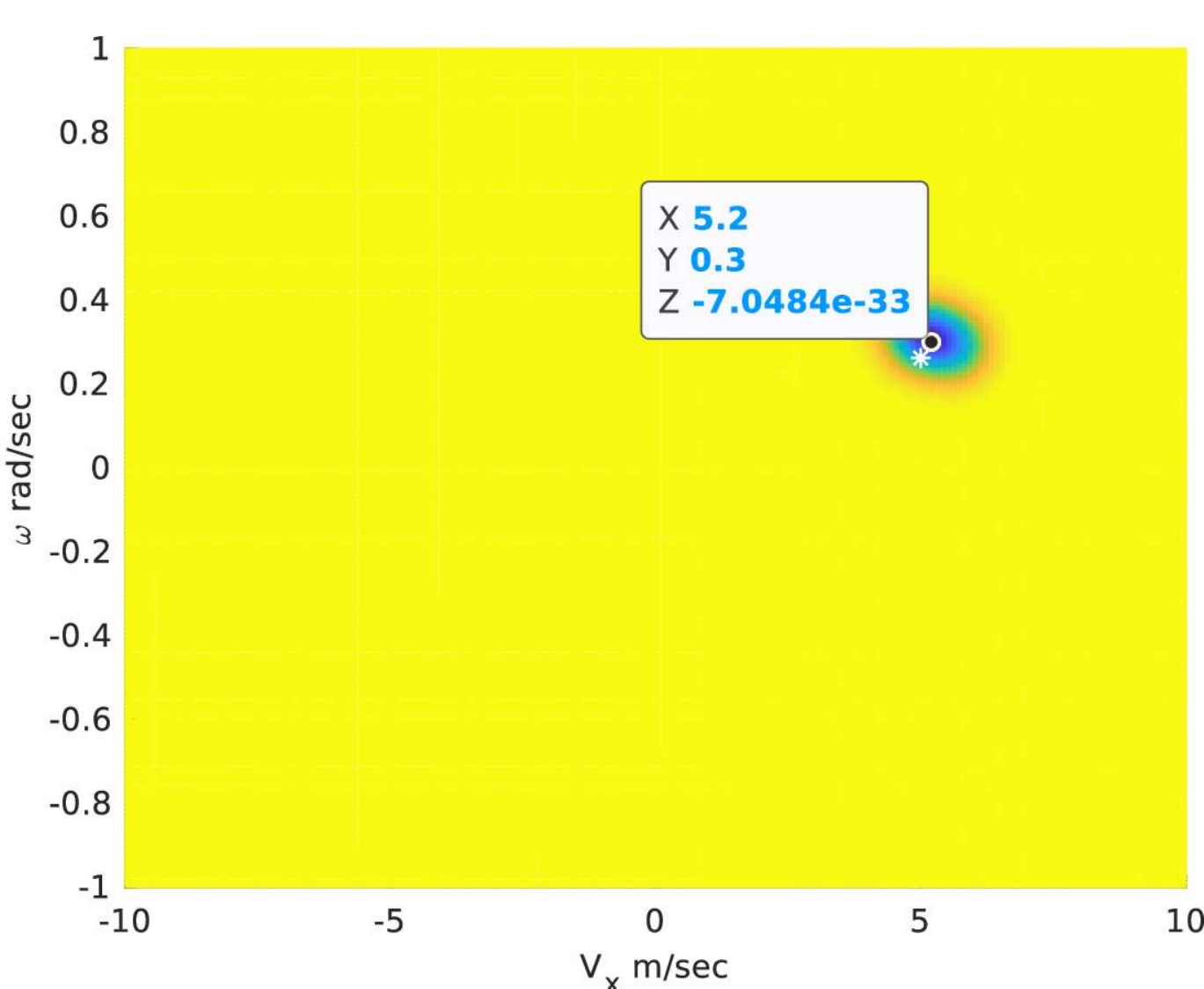}}
\subfigure[\textbf{The summing cost surface for the clustered scenario.}]{\label{fig:b}\includegraphics[width=70mm]{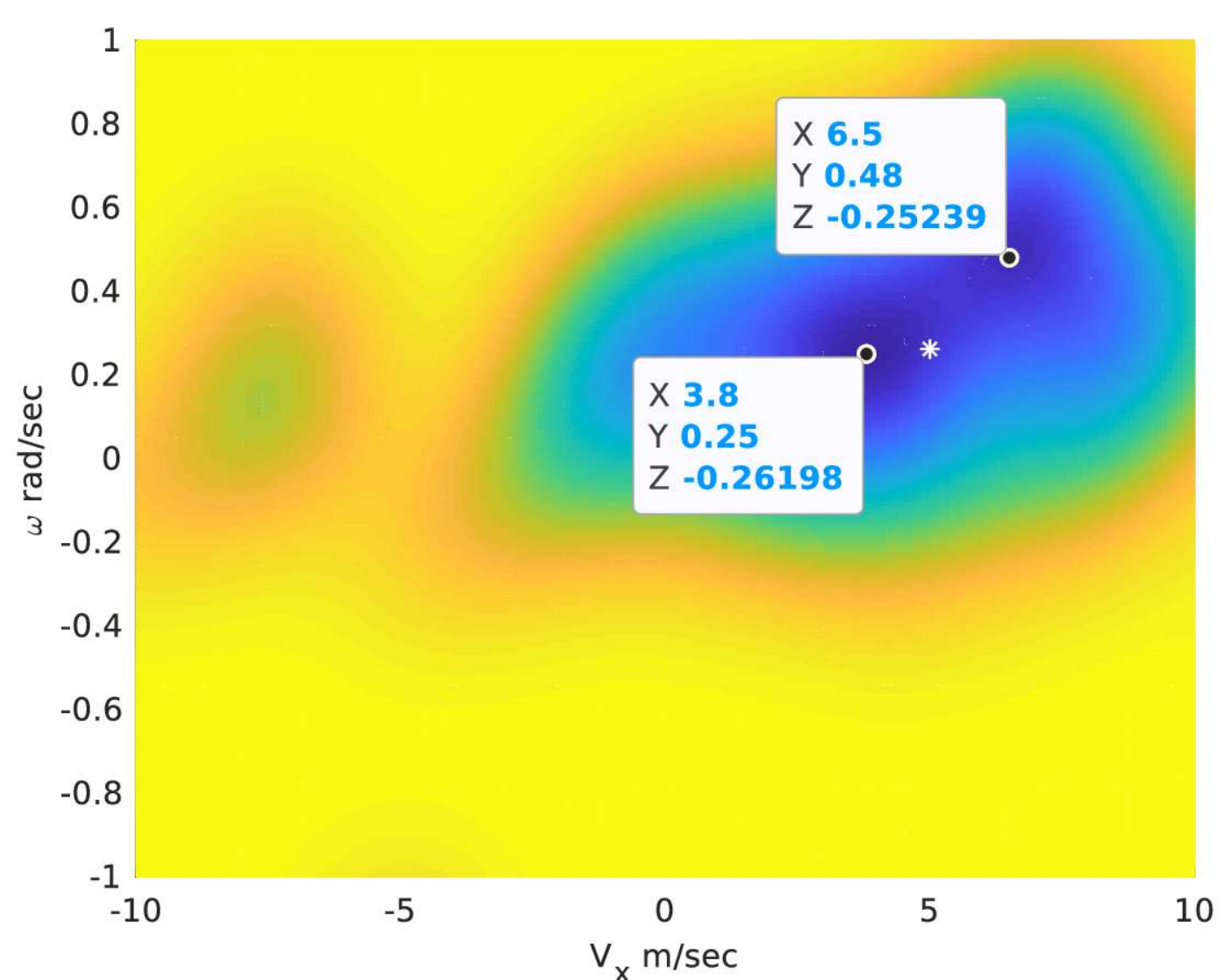}}
\subfigure[\textbf{The likelihood cost surface for the fully overallaped scenario without clustered points.}]{\label{fig:c}\includegraphics[width=70mm]{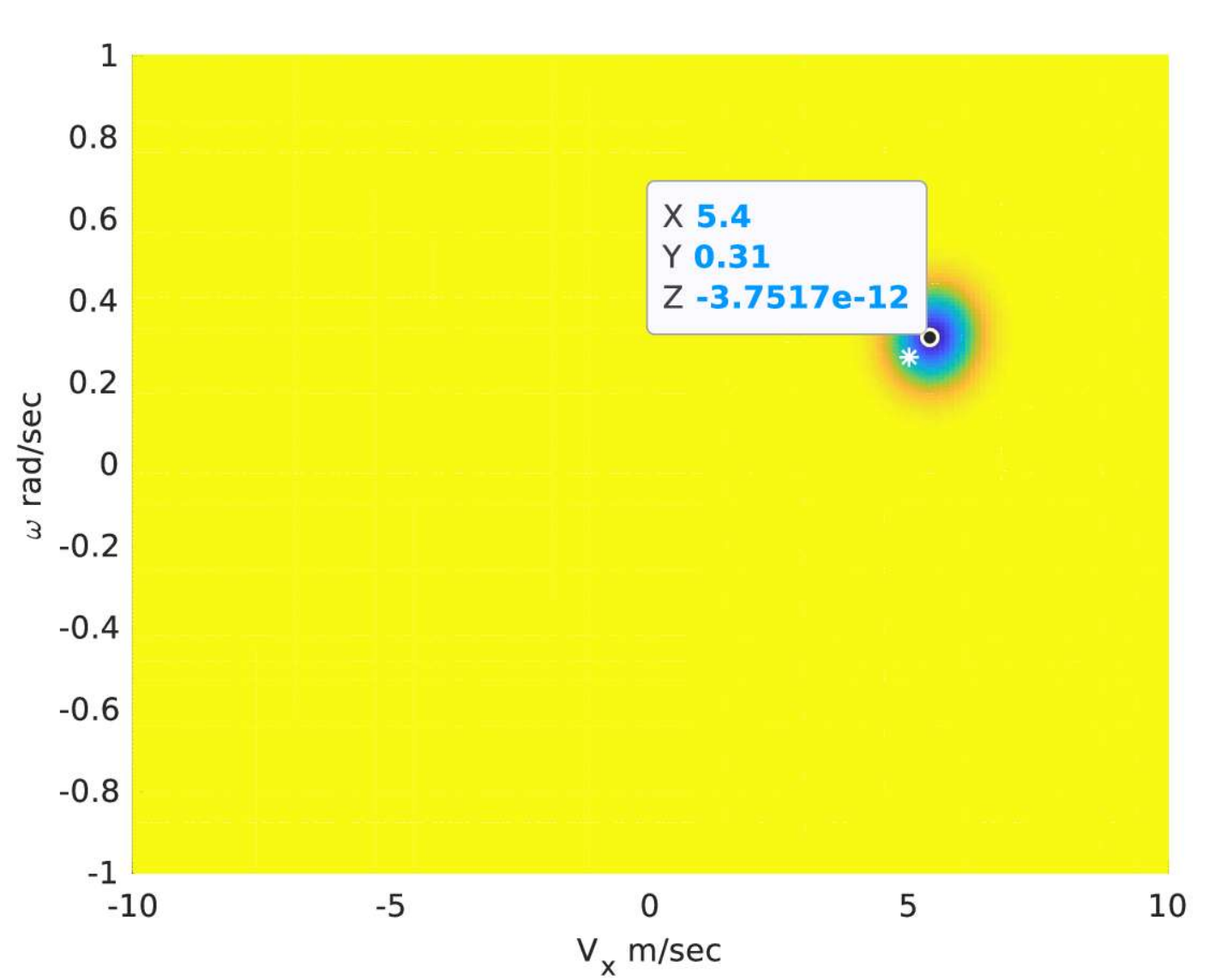}}
\subfigure[\textbf{The summing cost surface for the fully overlapped scenario without clustered points.}]{\label{fig:d}\includegraphics[width=70mm]{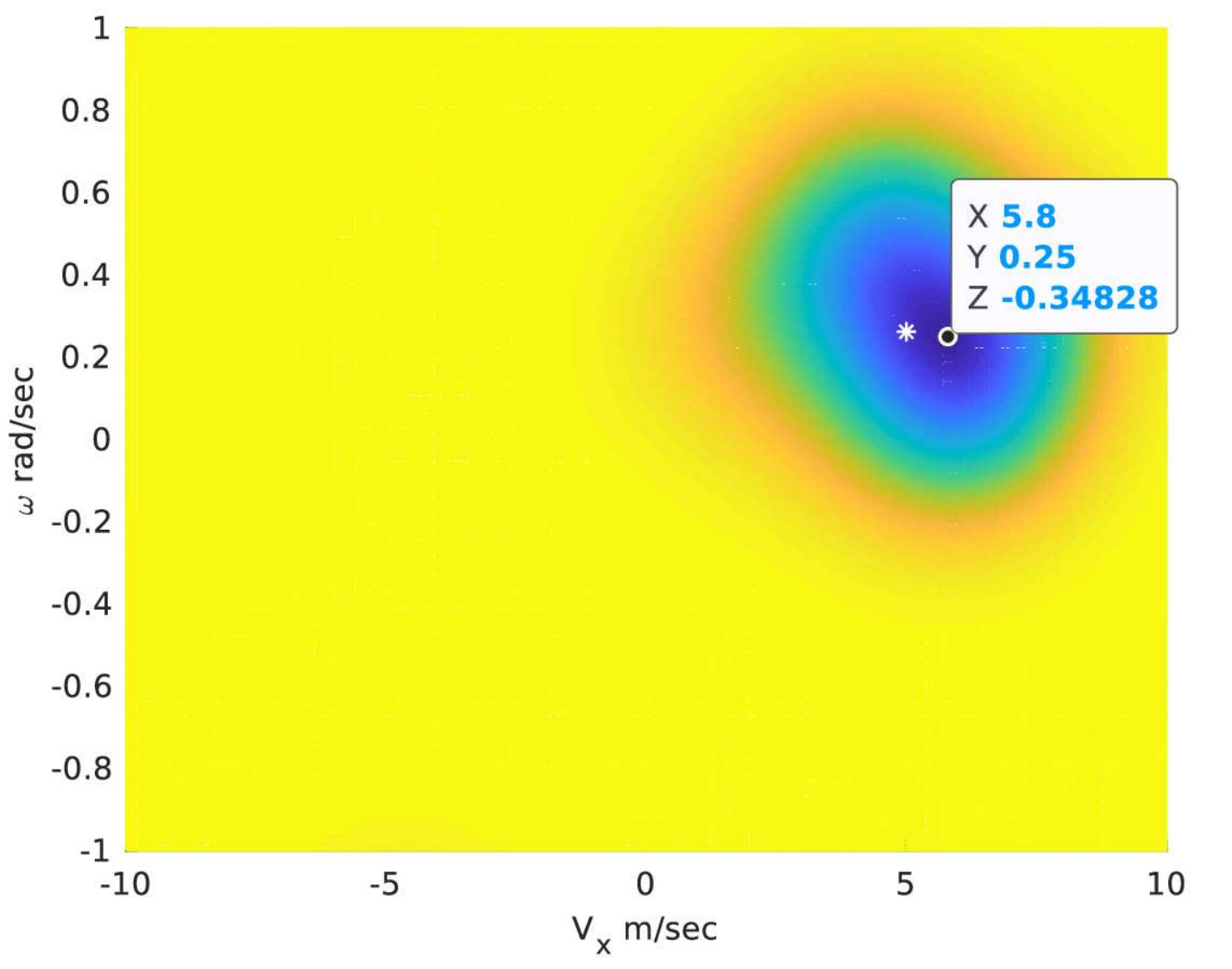}}
\caption{The cost surface for the clustered points, and comparing the cost surface to the fully overlapped scenario. The clustered points have a big impact on the summing approach, where even the optimizer starts from the ground truth as initial guess in the summing approach will converge in a wrong minimal(b).}
\label{fig:cost_3}
\end{figure}

\begin{figure}[htb!]
\centering
\subfigure[\textbf{The estimation error in the translation}]{\label{fig:c}\includegraphics[width=60mm,height=60mm]{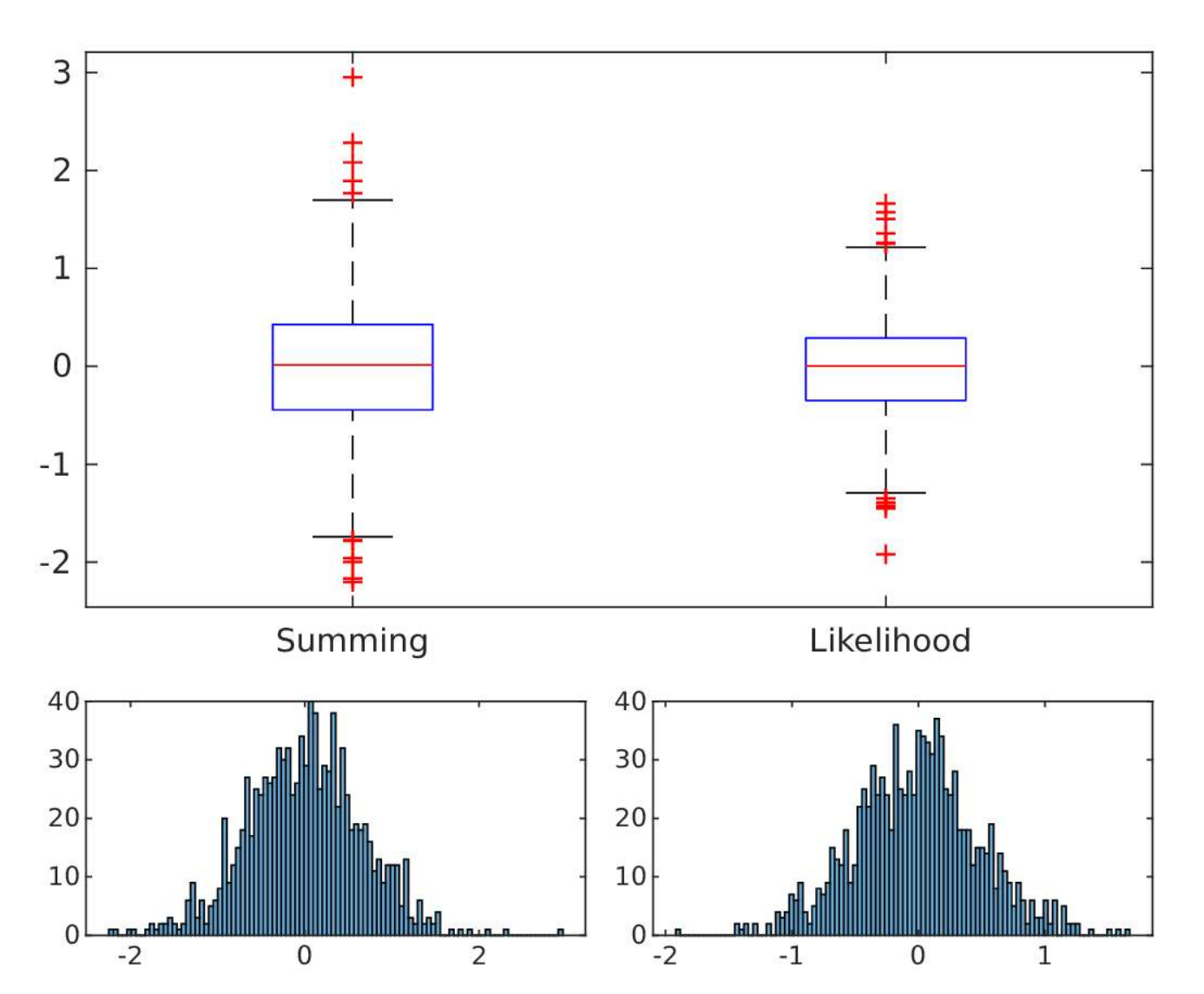}}
\subfigure[\textbf{ The estimation error in the rotation}]{\label{fig:a}\includegraphics[width=60mm,height=60mm]{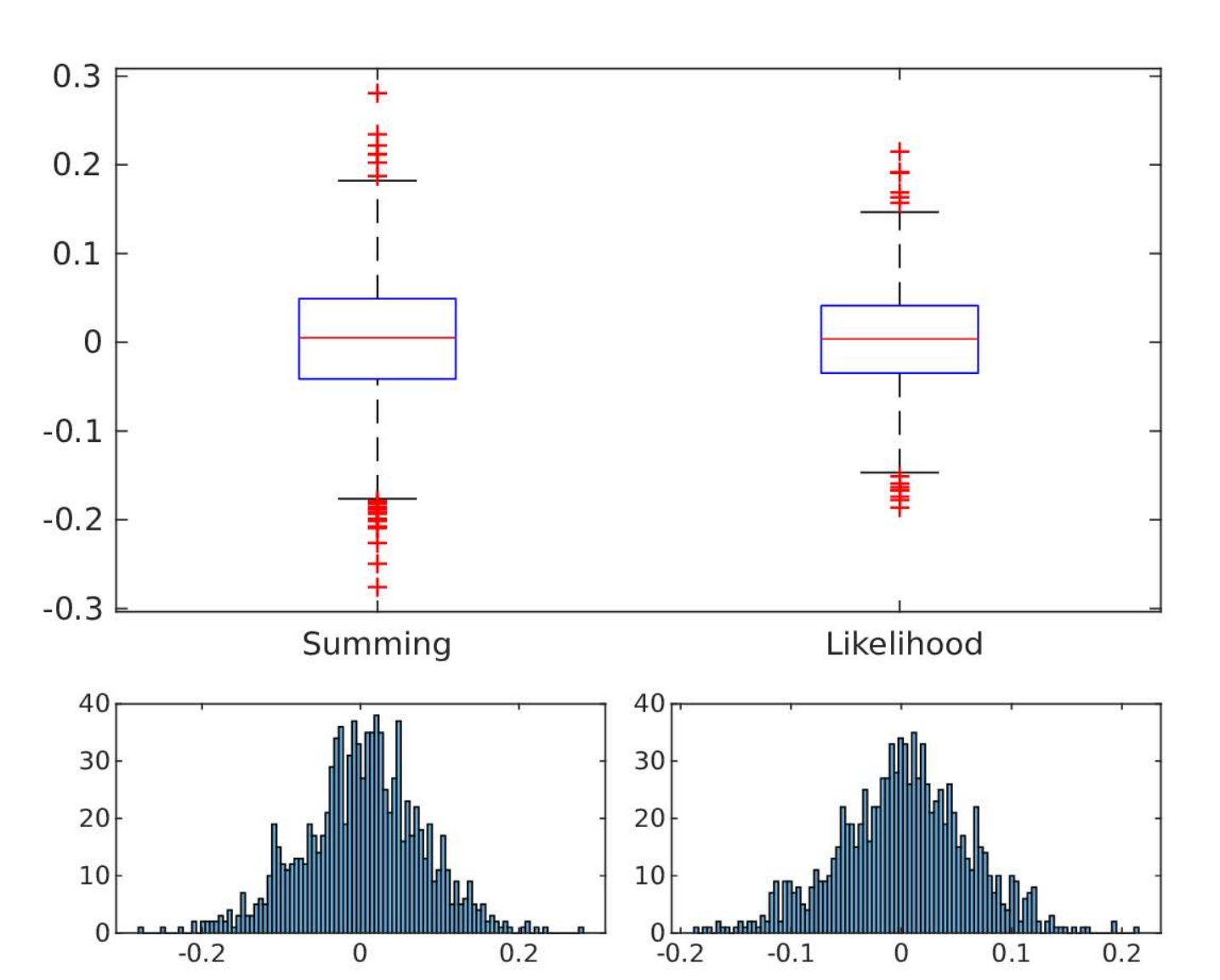}}
\caption{The estimation error for the summing and the likelihood, where the ground truth as the initial guess. The estimation error for the likelihood approach is lower than the summing approach.}
\label{fig:error_4}
\end{figure}

Firstly, the cost surface based on the summing approach for the clustered scenario is different from the cost surface for the fully overlapped scenario, as shown in Fig. \ref{fig:cost_3} (b) and (d). However, the likelihood approach behaves almost the same, as depicted in Fig. \ref{fig:cost_3} (a) and (c). Moreover, the global minimum in Fig .\ref{fig:cost_3} (b) is completely wrong, even if the optimization starts from the ground truth. Consequently, the estimation error for the clustered scenario based on the summing approach is higher than the fully overlapped scenario, as shown in Fig. \ref{fig:error_4}. Secondly, even this is a fully overlapped scenario, the estimation error based on the summing approach is increased compared to the likelihood approach, as shown in Fig. \ref{fig:error_4}.

\subsection*{Outlier With Clustering Scenario }

\begin{figure}[htb!]
\centering
\subfigure[\textbf{ The first scenario  }]{\label{fig:c}\includegraphics[width=70mm]{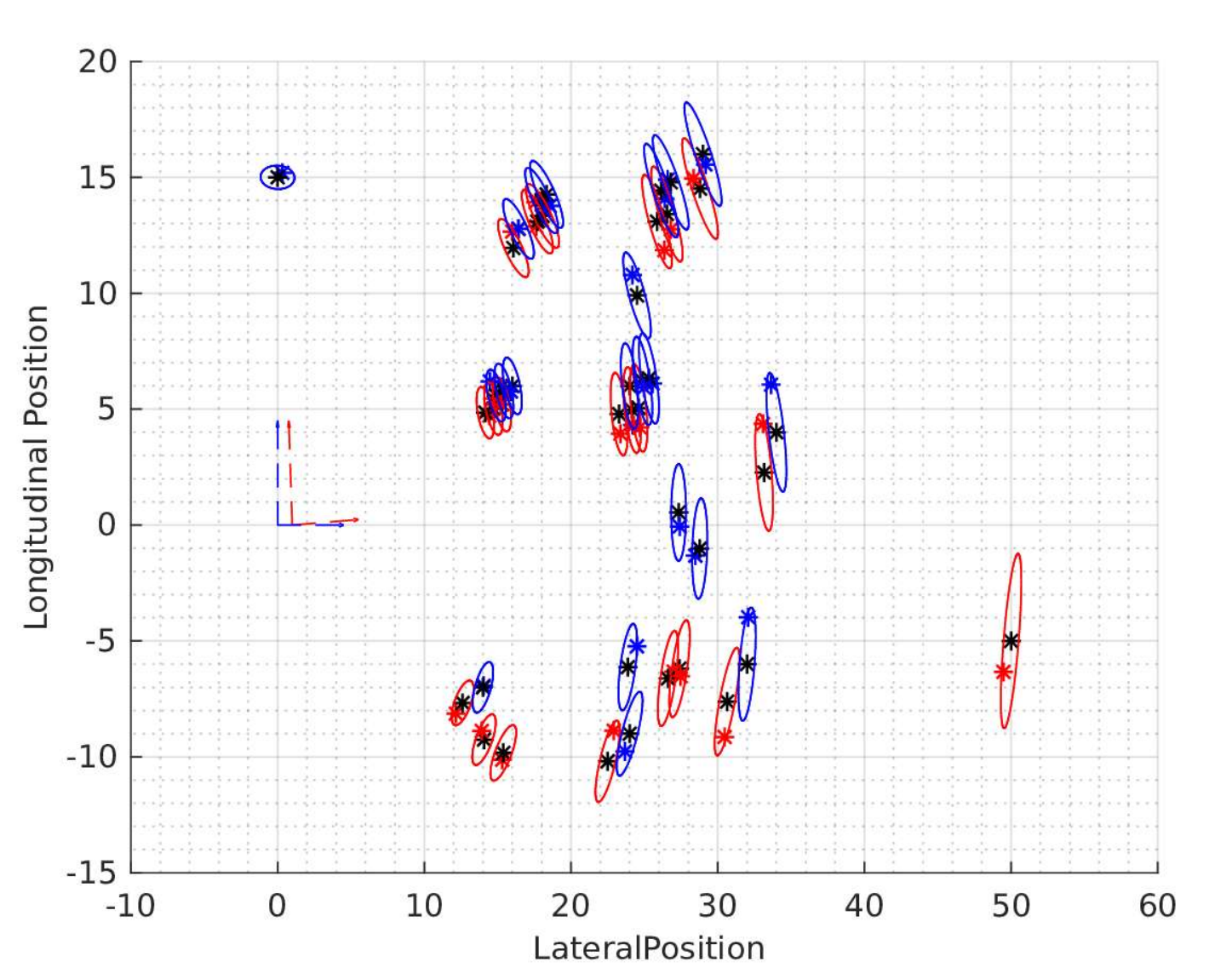}}
\subfigure[\textbf{ The second scenario }]{\label{fig:a}\includegraphics[width=75mm]{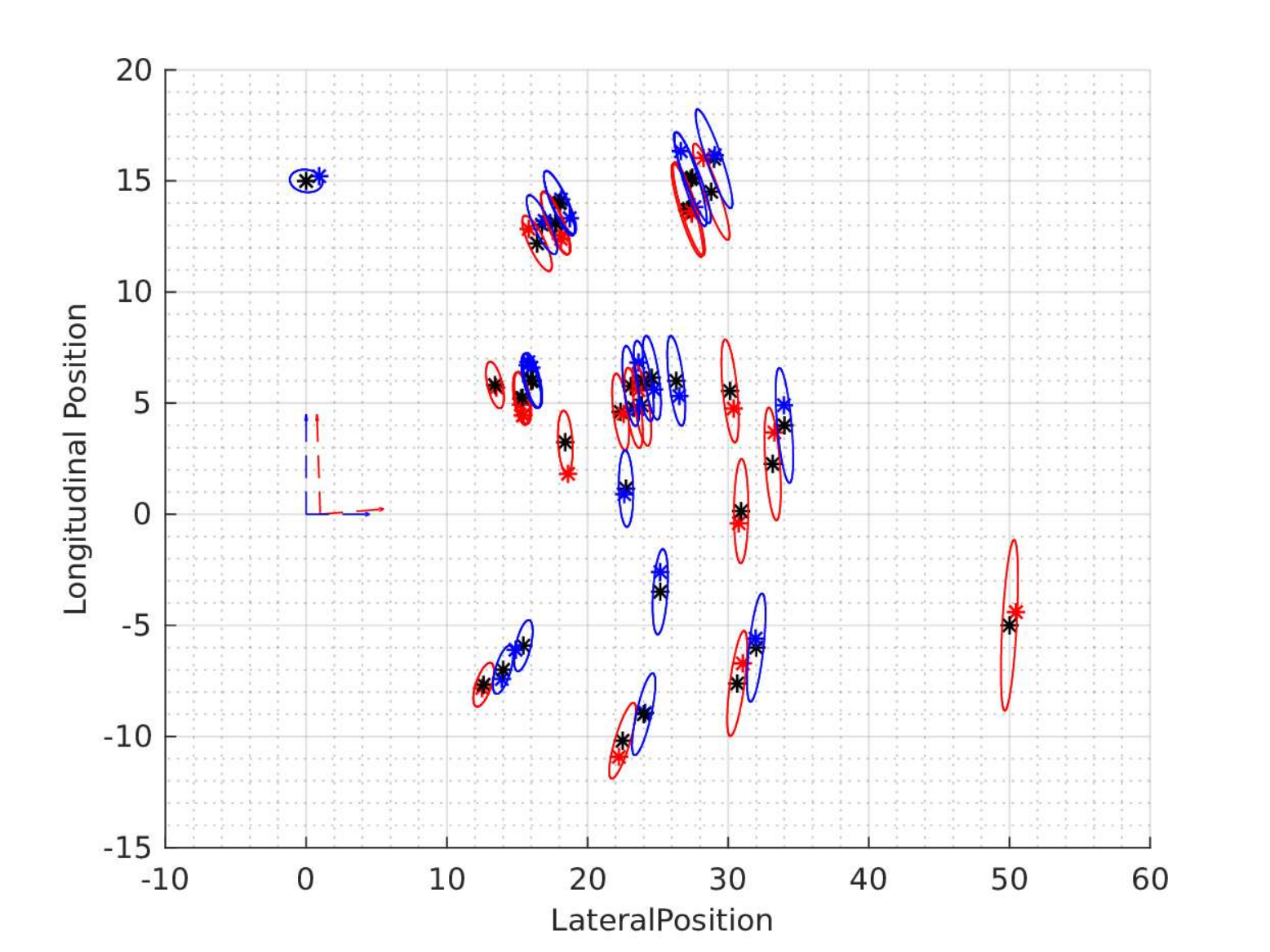}}
\subfigure[\textbf{For the first scenario, the summing result}]{\label{fig:a}\includegraphics[width=70mm]{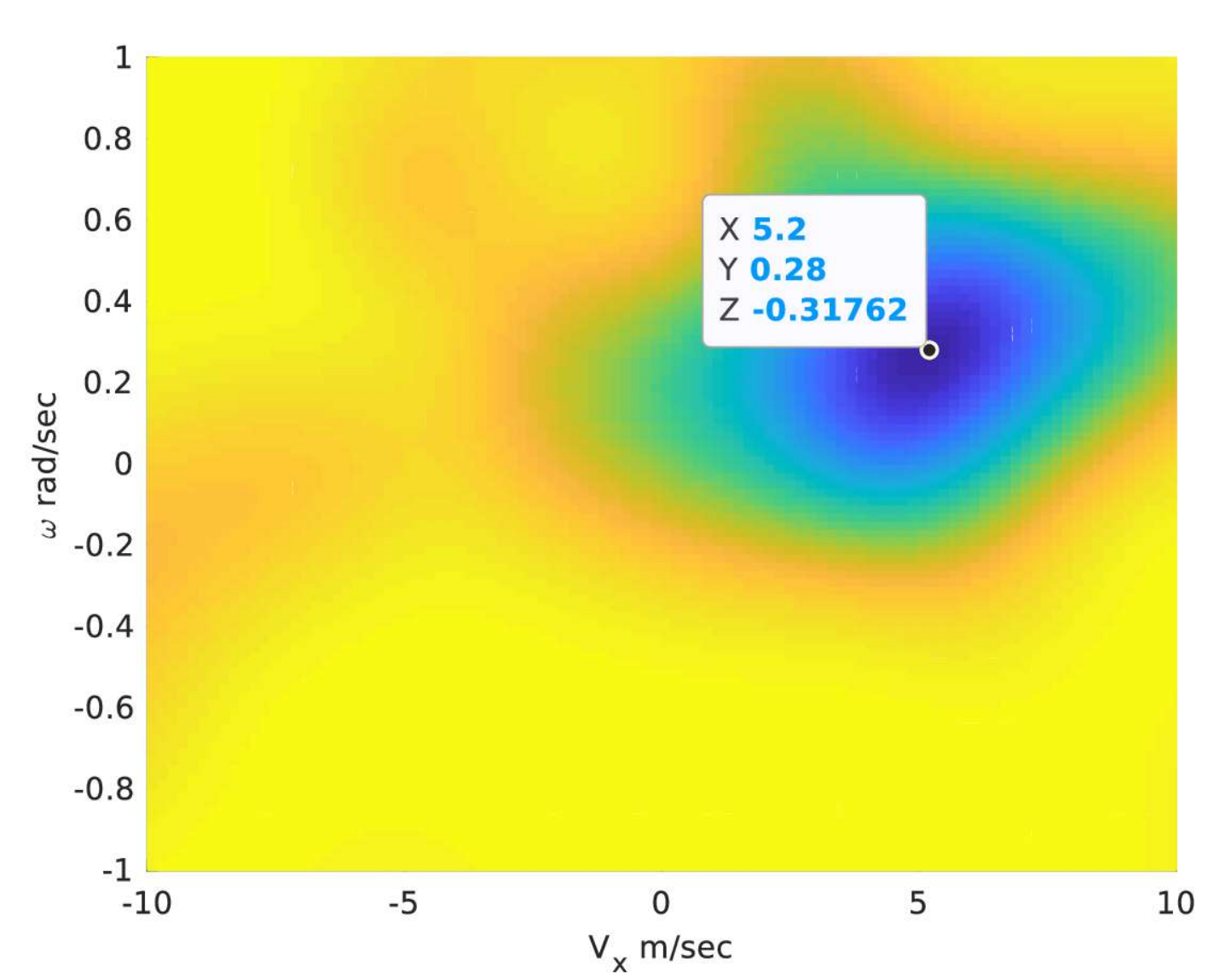}}
\subfigure[\textbf{For the second scenario, the summing result }]{\label{fig:a}\includegraphics[width=70mm]{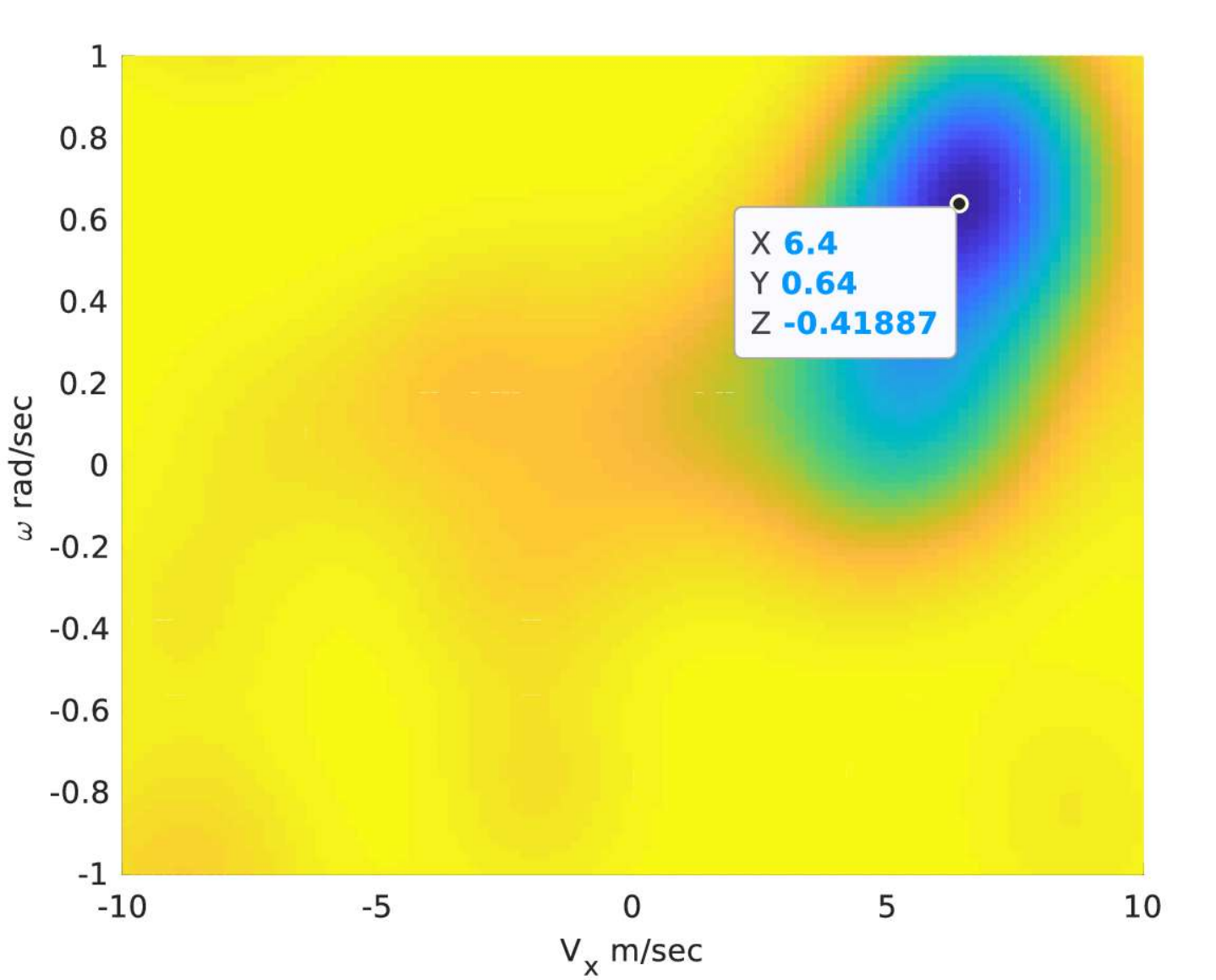}}
\subfigure[\textbf{For the first scenario, the likelihood result}]{\label{fig:a}\includegraphics[width=70mm]{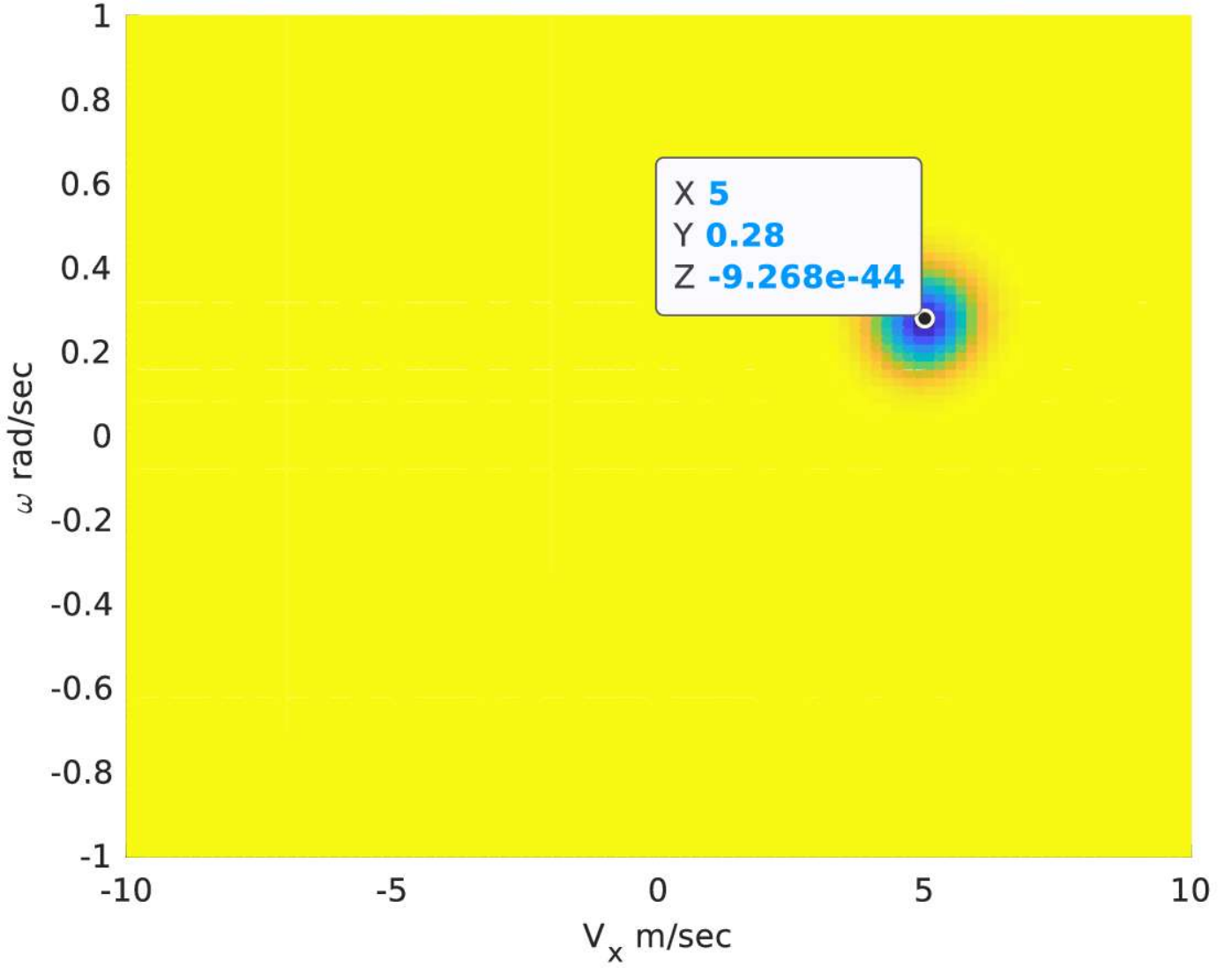}}
\subfigure[\textbf{For the second scenario, the likelihood result }]{\label{fig:a}\includegraphics[width=70mm]{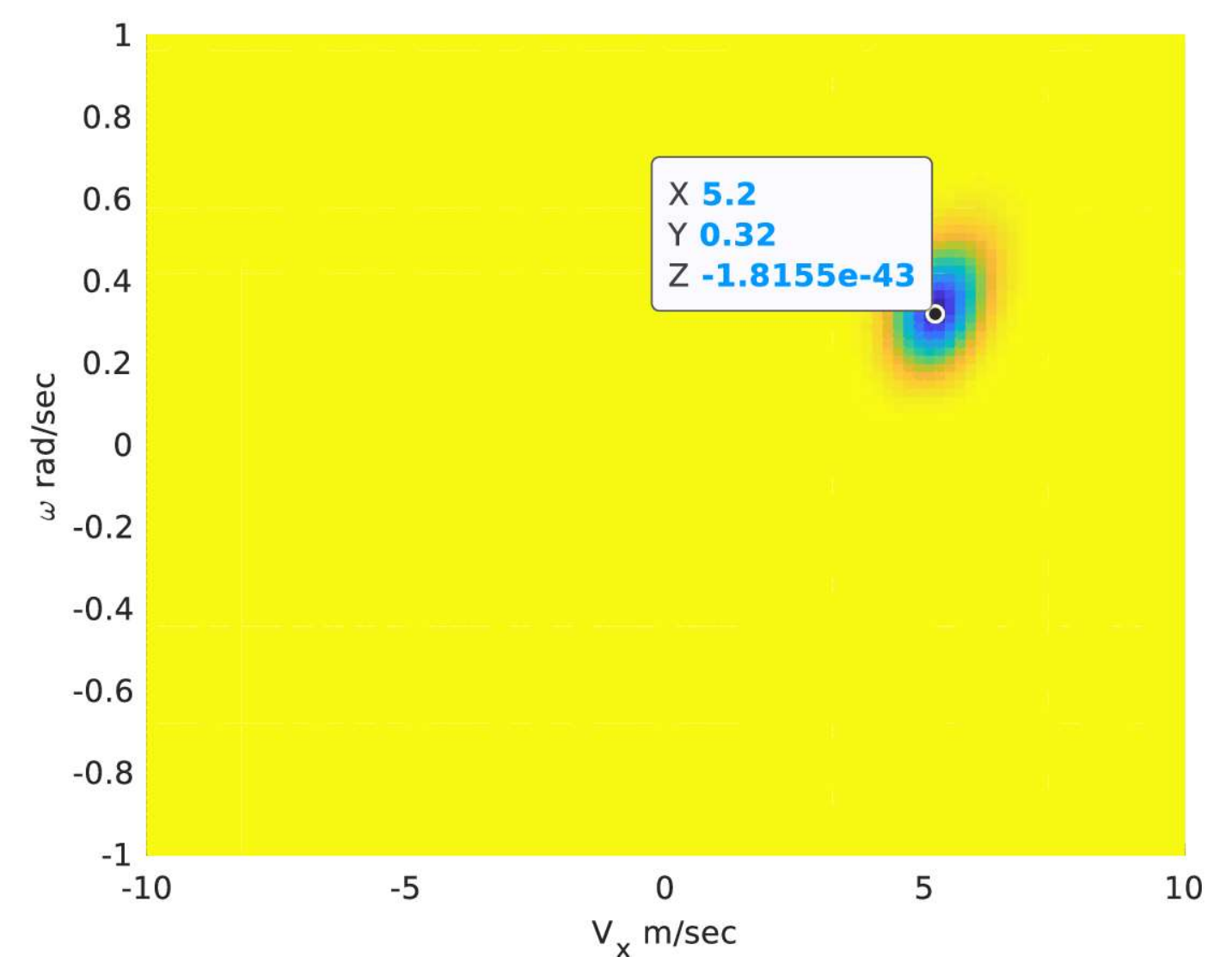}}
\caption{This scenario includes most of all test cases that have been discussed. Moreover, it is hard to clarify all aspects of the results. Therefore two scenarios are selected (a) and (b). ( c) shows the summing results for the first scenario and (e) shows the likelihood approach. (d) and (f) are the results for the second scenario the summing and the likelihood respectively.}
\label{fig:all_1}
\end{figure}

This scenario includes the 8 fully overlapped points, and the two types of the outlier, and the cluster points. The main idea is to stress both approaches with all test cases and evaluate the results. The initial guess is out of focus for this section, but the main focus here is the estimation error. It is hard to expect the results after including all evaluation aspects in one scenario. Therefore, this section introduces two synthetic scenarios, where both scenarios have the same condition, but with a random generation for the outlier and clustered points, as shown in Fig. \ref{fig:all_1} (a) and (b).

\begin{figure*}[htb!]
\centering
\includegraphics[width=0.8 \textwidth]{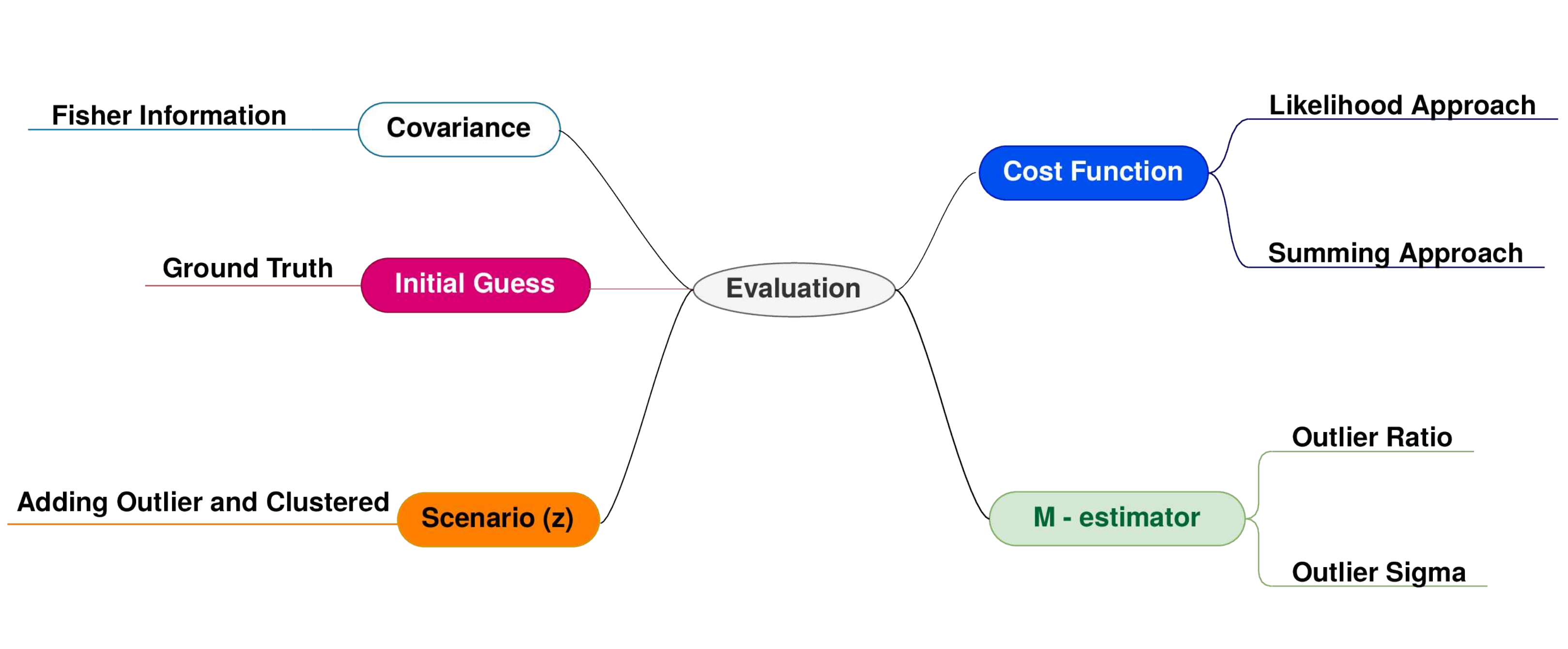}
\caption{ The evaluation map for the clustering points with the outlier scenario.}
\label{fig:eval_3}
\end{figure*}

\subsubsection*{Results and Discussion}

The cost surface for the likelihood for both scenarios has only one global minimum close to the ground truth $\boldsymbol{\theta_g}$, as shown in Fig. \ref{fig:all_1} (c) and (e). However, the summing approach behaves differently in both scenarios, and it is hard to clarify the main reason for those different behaviors. Therefore, the estimation error for both scenarios behaves differently. However, the likelihood almost behaves the same in both scenarios.

\section{Discussion}

The summing approach and the likelihood behave almost the same for the perfect scenarios, such as fully overlapped point sets. However, a good initial guess is quite essential for the summing approach, due to the local minimums. Besides, the optimal minimum in the summing approach usually has more curvature than the likelihood, so it gives more pessimistic results. In reality, the point set should have some outliers and some clustered points. Therefore, model the outlier in the cost function is crucial to suppress the outlier effect, following the M-estimator technique. The outlier component in the likelihood cost function changes the optimal minimum's curvature; thus, the results get more pessimistic. The clustered points could lead to a wrong global minimum, then a wrong estimation. In some cases, it is tough to explain what is the reason for the wrong cost surface, like, the fully overlapped clustered scenario. Lastly, \textbf{NCI} is not enough to evaluate the estimation, especially if the ground truth for covariance is unknown.



\printindex

\chapter{Ego Motion}

\begin{figure*}[h]
\centering
\includegraphics[width=0.8\textwidth]{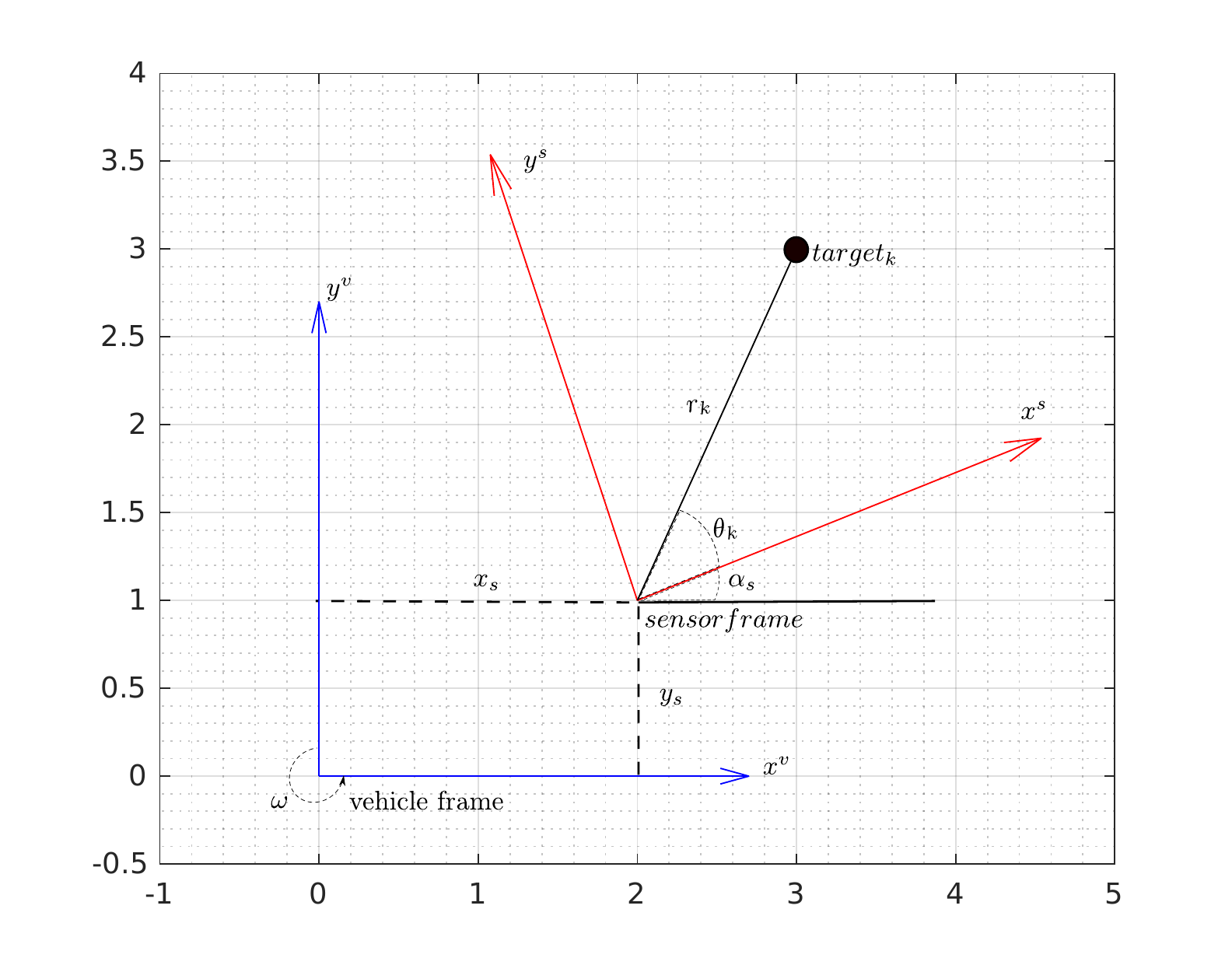}
\caption{The Vehicle and the sensor coordinate frames. Where, $[x_s,y_s,\alpha_s]$ are the offset parameters and $[r_k, \phi_k, v_k]$ are the measured parameters for $k-th$ target measured by the radar.}
\label{ego_1}
\end{figure*}

This work addresses the point set registration problem. It is the process of aligning two-point sets by finding spatial transformation. Automotive and mobile robot domains have the same problem as apart of the localization problem. Here we address the ego-motion estimation problem as apart of the localization problem using an automotive Doppler radar. Where, in each scan, the radar detects \textit{k} targets. Each target is represented as a vector  $\boldsymbol{s_k} =[r_k, \phi_k, v_k]$; where $r_k$ is the relative distance between the sensor and the target $k$, $\phi_k$ is the relative angle and $v_k $ is the Doppler velocity of the target. Moreover, the radar gives the standard deviation for each target $[\sigma_r, \sigma_\phi,\sigma_{v}]$. Ego-Motion estimation is an algorithm trying to optimize the motion vector \{$\boldsymbol{\theta}$\}. In a 2D scenario, the motion vector \{$\boldsymbol{\theta}$\} = $[v_x, v_y, \omega ]^T$ where $v_x$ is the velocity in $x$ direction, $v_y$ is the velocity in $y$ direction and $ \omega$ is the yaw-rate. Lastly, $[x_s,y_s,\alpha_s]$ represents the relative translation and rotation between the sensor coordinate system, and the vehicle coordinate system has to be considered. The complete configuration is shown in Fig. \ref{ego_1}. The main focus of this chapter is to get the best estimate for \{$\boldsymbol{\theta}$\} based on the proposed cost function, and the proposed method to estimate the covariance.\\

\begin{figure}[htb!]
\centering
\subfigure[\textbf{ Camera view}]{\label{fig:a}\includegraphics[width=100mm,height=70mm]{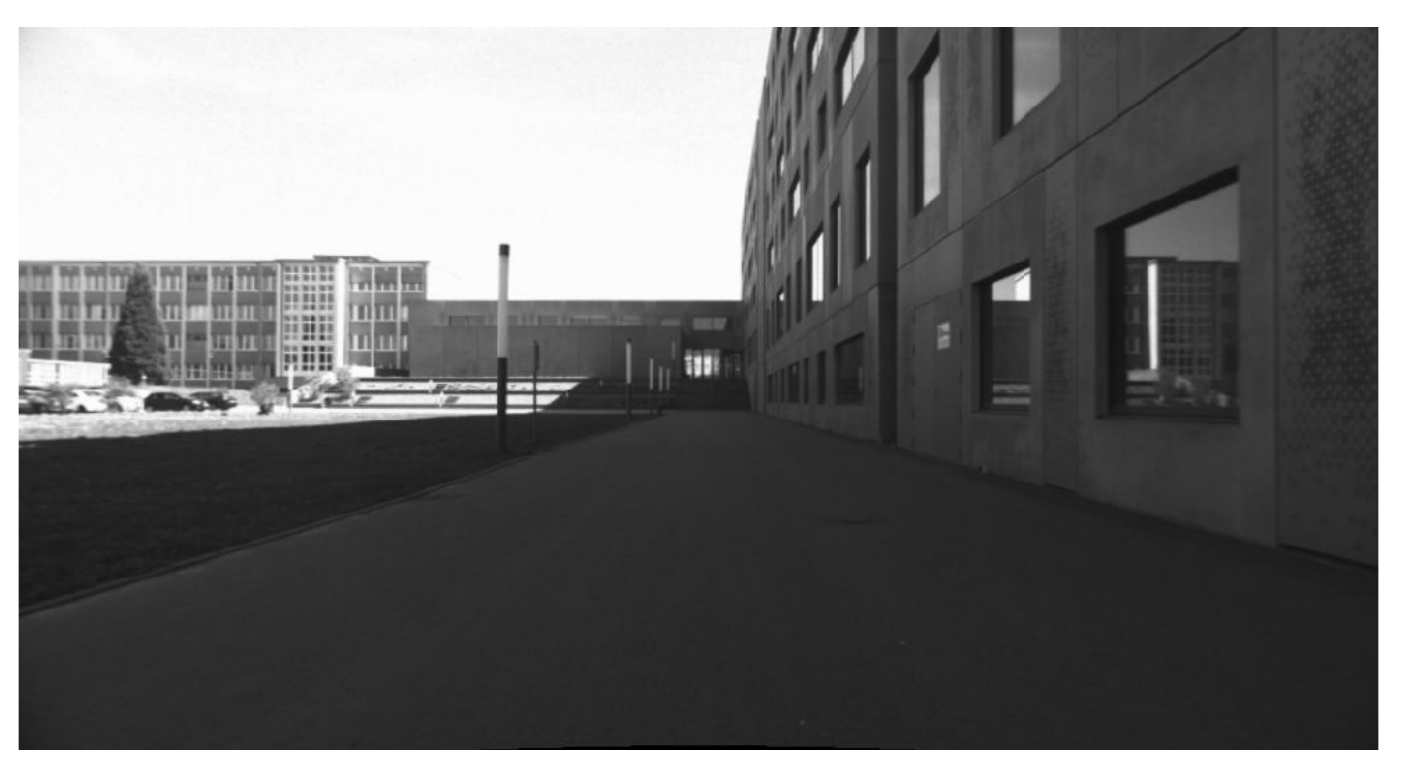}}
\subfigure[\textbf{ Radar Scan}]{\label{fig:b}\includegraphics[width=115mm]{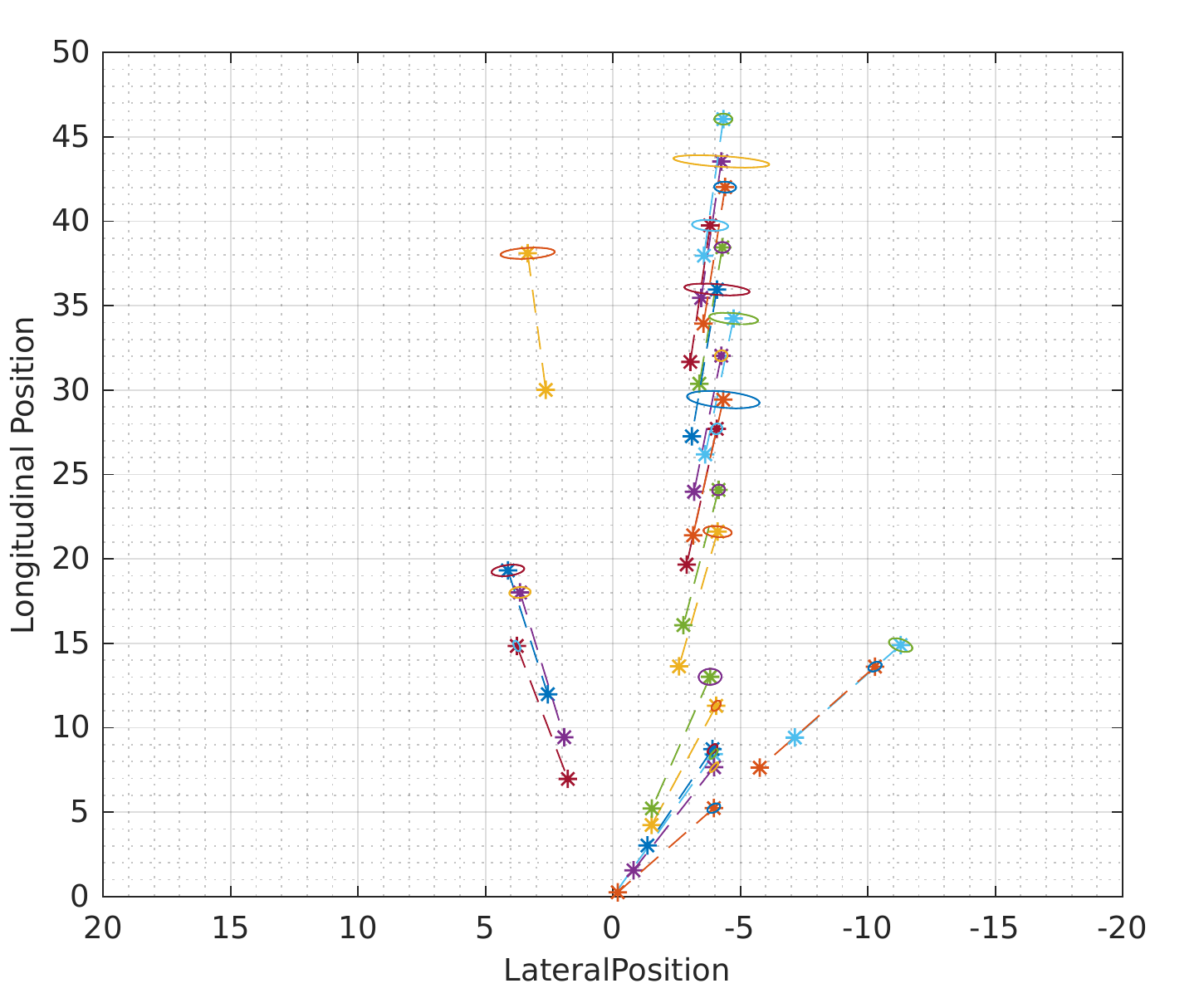}}
\caption{This plot shows one scan from the radar with the equivalent camera image, where the detected targets as represents by the stars, and the error ellipse represents the 3 Sigma bound of the spatial information. Moreover, the dashed line to each target represents the Doppler information.}
\label{fig:phobos_3}
\end{figure}

Fig. \ref{fig:phobos_3} visualize one scan from the radar with a similar camera image, where the Doppler radar gives the relative position for each target as well as the standard deviation. Moreover, the dashed line represents the Doppler velocity for each target.

\section{ Mathematical Formulation }

The radar gives the measurements $\boldsymbol{s_k}$ in polar space. Thus, the first step is to convert the measurements to the cartesian space:

\begin{equation}
\boldsymbol{m_k} = \left[\begin{array}{l}
{r_{k} \cos \left(\phi_{k}+\alpha_s\right)+x_{s}} \\
{r_{k} \sin \left(\phi_{k}+\alpha_s\right)+y_{s}}
\end{array}\right]
\label{eq:ego_01}
\end{equation}

Then, reference the current scan through \{$\boldsymbol{\theta}$\}:

\begin{equation}
\boldsymbol{\mathcal{T}}(\boldsymbol{m_k}, \theta) =  \left[\begin{array}{cc}
{\cos (\omega \tau)} & {-\sin (\omega \tau)} \\
{\sin (\omega \tau)} & {\cos (\omega \tau)}
\end{array}\right] \boldsymbol{m_k} +\left[\begin{array}{c}
{v_{x} \tau} \\
{v_{y} \tau}
\end{array}\right]
\label{eq:ego_02}
\end{equation}

Where $\boldsymbol{\theta} $ represents the transformation parameters, which need to be optimized.\\

Also, the sensor gives the standard deviation for each target, but in polar space. For the spatial component, the covariance can be written as:
\begin{equation}
\Sigma_{s_k} =  \left[\begin{array}{cc}
{\sigma_r^2} & 0 \\
0 & {\sigma_\phi^2}
\end{array}\right]
\label{eq:ego_03}
\end{equation}

Thus,  it needs to be also transformed into cartesian space. The transformation from the polar space to the cartesian space is nonlinear (\ref{eq:ego_01}). Therefore, the first-order Taylor expansion is used to propagate the uncertainty as follows:

\begin{equation}
\Sigma(m_{k}, \theta)  =J\left(m_{k}, \theta\right) \  \Sigma_{s_k}  \ J \left(m_{k}, \theta \right)^{\mathrm{T}}
\end{equation}

Where, $\boldsymbol{J}$ represents the jacobian matrix, which can be written as:

\begin{equation}
J\left(m_{k}, \theta \right)=\left[\frac{\partial \mathcal{T}\left(m_{k}, \theta \right)}{\partial \phi_{k}} \frac{\partial \mathcal{T}\left(m_{k}, \theta \right)}{\partial r_{k}} \frac{\partial \mathcal{T} \left(m_{k}, \theta \right)}{\partial v_{k}}\right]
\end{equation}

Therefore, for one target from the current scan, the inlier component in the cost function can be written as:

\begin{equation}
\boldsymbol{Inlier \ component}_k =  \sum_{i=1}^{|\mathcal{F}|} w_{i} \cdot p\left(0 | \mathcal{T}\left(m_{k}, \boldsymbol{\theta}\right)- m_{i} \ , \  \mathcal{T}\left(\Sigma_{k}, \boldsymbol{\theta}\right)+\Sigma_{i} \right)
\end{equation}

This work uses the corrupted gaussian method to model the outlier component in the cost function. Where, the mean for the outlier component is the mean of the previous scan targets, with a significant standard deviation, as follows:

\begin{equation}
\boldsymbol{Outlier \ component}_k = p\left(\mathcal{T}\left(m_{k}, \boldsymbol{\theta}\right) | m_{i}, \mathbf{\Sigma}_{\text {Outlier }}\right)
\end{equation}

Consequently, the total cost function can be written as follows:
\begin{equation}
\boldsymbol{f(\theta)}_{d2d} = \prod_{k \ \in  |\mathcal{M}|}\left( (1-\alpha) \ \boldsymbol{Inlier_{component}}_k \ + \ \alpha \ \boldsymbol{Outlier_ {component}}_k  \right)
\label{ego_cost_1}
\end{equation}

Then, the optimizate problem can be written as:

\begin{equation}
\boldsymbol{\hat{\theta}} = \arg min(f_{\mathrm{d2d}}(\boldsymbol{Initial \ guess}_ {\theta}))
\end{equation}

Equation (\ref{ego_cost_1}) uses only the spatial information to optimize the motion, but the Radar also gives precise velocity information for each target. In the same vein, Barjenbruch in his work~\cite{barjenbruch2015joint} provides a mathematical model to calculate the doppler information from the spatial information, as follows:
\begin{equation}
\begin{aligned}
& V (m_{k}, \theta) = -\left(v_{x}-\omega y_{\mathrm{S}}\right) \cos \left(\phi_{k}+\alpha_{\mathrm{S}}\right)\\
&-\left(v_{y}+\omega x_{\mathrm{S}}\right) \sin \left(\phi_{k}+\alpha_{\mathrm{S}}\right)
\end{aligned}
\end{equation}

And the uncertainty also:

\begin{equation}
\hat{\Sigma}\left(m_{k}, \theta \right) \ = \ \left(\frac{\partial V\left(m_{k}, \theta \right)}{\partial \phi_{k}}\right)^{2} \sigma_{\phi}^{2}
\end{equation}

Then, the {L2} metric is used to measure the distance between the calculated doppler information and the measured doppler:

\begin{equation}
\begin{aligned}
&t^{\mathrm{D}} = v_{k}^{\mathrm{D}}-V\left(m_{k}, \theta\right)\\
&\gamma^{\mathrm{D}}=\sigma_{v}^{2}+ \hat{\Sigma}\left(m_{k}, \theta \right)
\end{aligned}
\end{equation}

\begin{equation}
\boldsymbol{Doppler \ component}_k =  p \ \left(0 | t^{\mathrm{D}}, \gamma^{\mathrm{D}}\right)
\label{eq:doppler_1}
\end{equation}
Then, joint the spatial information (\ref{ego_cost_1}) and the Doppler information (\ref{eq:doppler_1}) as follows:
\begin{equation}
\boldsymbol{f(\theta)_{joint}} = \prod_{k \ \in  |\mathcal{M}|}\left( (1-\alpha) \ \boldsymbol{Inlier}_k \cdot \boldsymbol{Doppler}_k \ + \ \alpha \ \boldsymbol{Outlier}_k  \right)
\end{equation}

The doppler information suppresses the non-stationary targets. Because the calculated velocity for the non-stationary targets should not match the measured one. Therefore, the doppler component will minimize the impact of this target on the cost function. Besides, it decodes precise information about the velocity in the $x$ direction, which reduces the outlier effect on the cost function. Eventually, the optimization problem can be written as follows: \\

\begin{equation}
\hat{\boldsymbol{\theta}}=\arg \min _{\theta} \ - log(\boldsymbol{f(\theta)_{joint}})
\end{equation}

\begin{algorithm}
	\caption{Motion Estimation Algorithm}\label{alg:euclid}
	\begin{algorithmic}[1]
		\State K targets, current scan $\Rightarrow$ $m_k$ \Comment Required
		\State N targets, previous scan $\Rightarrow$ $f_n$
		\Comment Required
		\State $\Delta t$, time between two scans
		\Comment Calculated from the time stamp of the scan
		\State $\boldsymbol{\theta^{i-1}}  \gets \boldsymbol{\theta^{intial}}$ \Comment Good guess
		\Repeat
		\State $cost = f(m_k,w_n,\Delta t,\boldsymbol{\theta^{i-1}})$ \Comment objective function
		\State $ \boldsymbol{\theta^{i}}$ $\Leftarrow$  $\argminA(f(\theta),\boldsymbol{\theta^{i-1}})$  \Comment update $\boldsymbol{\theta}$
		\Until convergence

	\end{algorithmic}
\end{algorithm}

\vspace{5mm}

Fig. \ref{fig:algorithem_seq} depicts the sequence of the algorithm, where it consists of three main steps: Firstly, converting the polar scan into cartesian space. Secondly, constructing the spatial metric using the L2 metric and the likelihood approach. Thirdly, adding the Doppler information to the cost function. Then, it gives an initial guess for the motion parameters and iterates till conversion. Scaling the covariance will be discussed later in this chapter.\\

\begin{center}
\resizebox {\textwidth} {!} {
\begin{tikzpicture}[node distance=1.5cm,
every node/.style={fill=white, font=\sffamily}, align=center]
\node (c1)    [activityStarts]  at (0 , 0)   {Previous radar \\ scan $\{\mathcal{F} \}$};
\node (c2)	  [nonblock, right of=c1, xshift=3cm]   { };
\node (c3)	  [base, right of=c2, xshift=3cm] 	{Current radar \\ scan$\{\mathcal{M} \}$};

\node (c4)     [base, below of=c1]          {Polar to cartesian};
\node (c5)     [base, below of=c2]          {Initial guess \\ for \{$\boldsymbol{\theta}$\}};
\node (c6)     [base, below of=c3]          {Polar to cartesian};

\node (c7)     [nonblock, below of=c4]          { };
\node (c8)     [base, , below of=c5]  {Reference the current scan};
\node (c9)     [nonblock, below of=c6]          { };

\node (c10)     [nonblock, below of=c7]          { };
\node (c11)     [base, right of=c10, xshift=3cm]  {Spatial metric};
\node (c12)     [base, right of=c11, xshift=3cm]  {Calculate the \\ expected velocity };

\node (c13)     [nonblock, below of=c10]          { };
\node (c14)     [base, below of=c11]          { Calculate the cost };
\node (c15)     [base, below of=c12]          { Doppler metric };

\node (d12)     [base, below of=c13]          { Update Initial \\ guess \{$\boldsymbol{\theta}$\} };
\node (d13)     [nonblock, below of=c14]          { };
\node (d14)     [base, below of=c15]          { Scale $\boldsymbol{\Sigma}$};

\node (c16)     [nonblock, below of=d12]          { };
\node (c17)     [base, below of=d13]          { Do the cost optimization };
\node (c18)     [nonblock, below of=d14]          { };

\node (c19)     [nonblock, below of=c16]          { };
\node (c20)     [nonblock, below of=c17]          { \{ $\hat{\boldsymbol{\theta}}$ \} };
\node (c21)     [nonblock, below of=c18]          { };

\draw[->]      (c1) -- (c4);
\draw[->]      (c3) -- (c6);
\draw[->]      (c5) --  (c8);
\draw[->]      (c6) |- (c8);

\draw[->]      (c8)  -- (c11);
\draw[->]      (c4)  |-  (c11);
\draw[->]      (c12) -- (c15);
\draw[->]      (c15) -- (c14);
\draw[->]      (c11) -- (c14);

\draw[->]      (c14) -- (c17);
\draw[->]      (c17) -| (d12);

\draw[->]      (d12)  -- ++(-3,0) -- ++(0,4.5)--  (c8) ;

\draw[->]      (c3) -- ++(3,0) -- ++(0,0)  -- ++(0,0)  |-  (c12);
\draw[->]      (c3) -- ++(3,0) -- ++(0,0)  -- ++(0,0)  |-  (c15);

\draw[->]      (c17) -- (c20);
\draw[->]      (d14) -- ++(-2,0)-- (c17);
\draw[->]      (c17) -| (d14);
\end{tikzpicture}
}
 
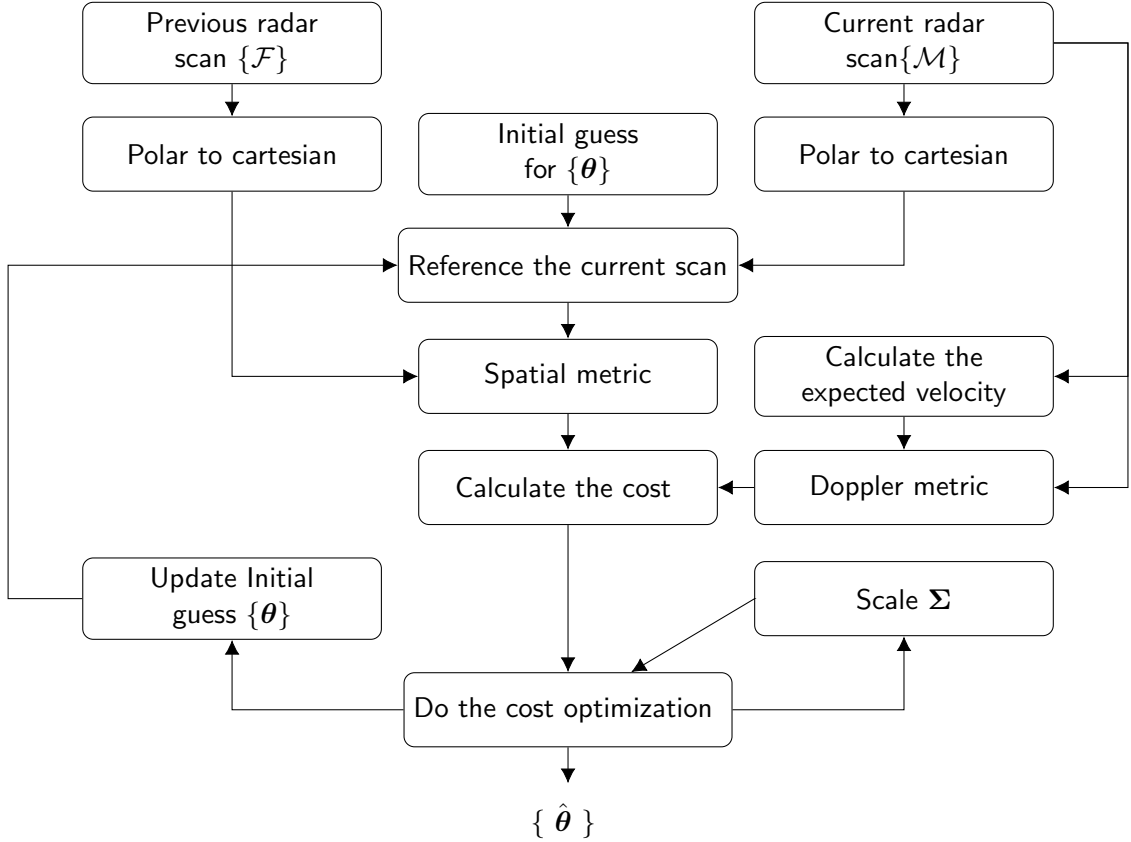
\captionof{figure}{Ego-Motion Estimation Frame Work \\}
 \label{fig:algorithem_seq}
\end{center}
\vspace{5mm}
\section{Platforms and environments}

\begin{figure*}[h]
\centering
\includegraphics[width=1\textwidth]{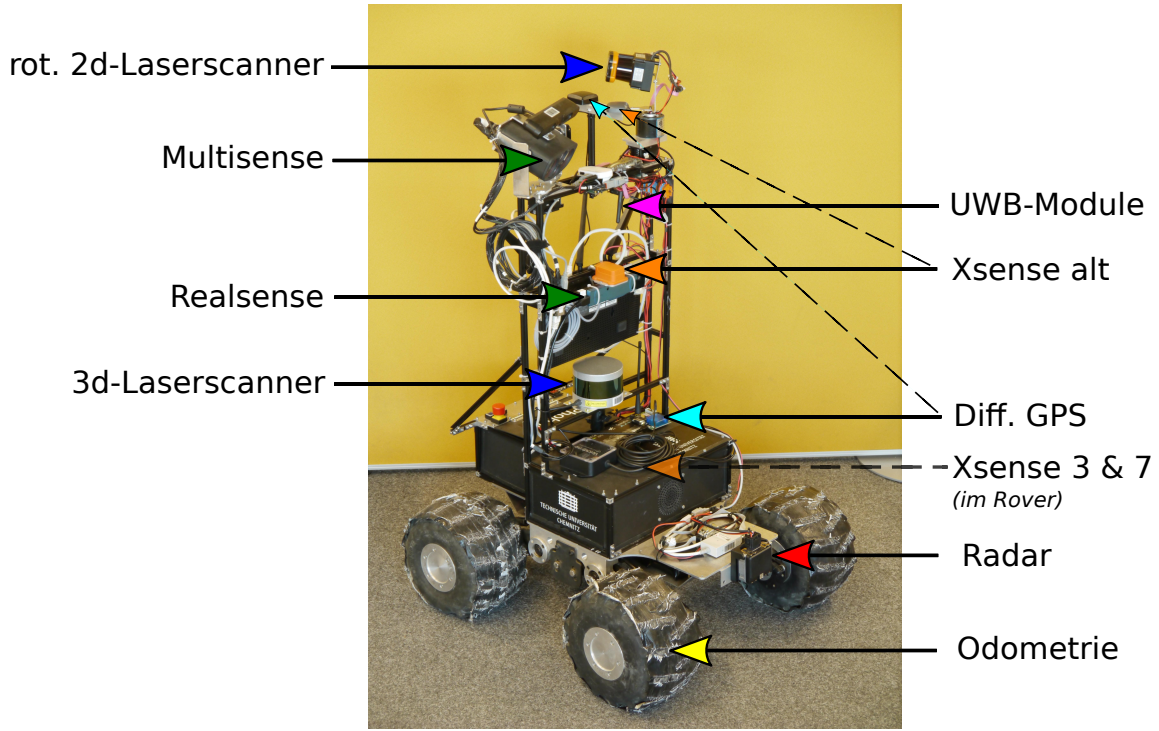}
\caption{The complete structure for the mobile robot Phobos, which is the platform have been used to collect the data.}
\label{fig:phobos_1}
\end{figure*}

Before getting involved in the evaluation part, it is time to exhibit the experimental platform, the used sensor, and the environment, where the data is collected. This work focuses on evaluating a robust cost function for point set registration problem, and the Ego-motion estimation is discussed as an applied application for the point set registration problem. Thus, an automotive Doppler radar is used to do the evaluation. The measurements are collected using an automotive 77 GHz radar with a bandwidth of 500 MHz and an update rate of 10 Hz. In each scan, the sensor can detect up to 48 features. For each feature: the relative distance, relative angle, and Doppler speed are measured. Moreover, the sensor provides the standard deviation for each measurement. The sensor is mounted in front of a mobile robot called Phobos, as depicted in Fig. \ref{fig:phobos_1}, which was used in 2015 in Space Bot Cup~\cite{lange2016two}.\\

\begin{figure*}[h]
\centering
\includegraphics[width=1\textwidth]{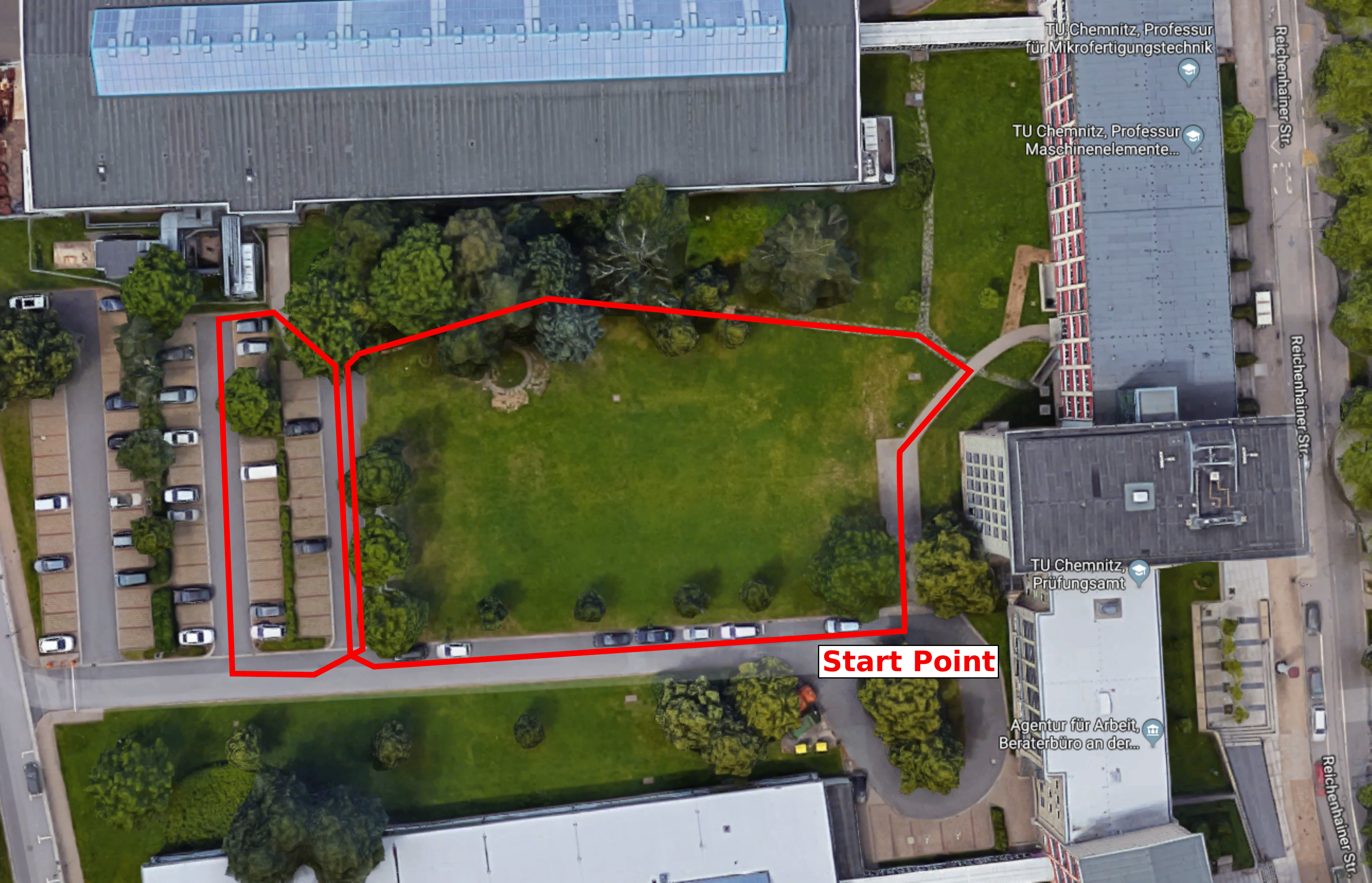}
\caption{A google map screenshot, where it shows the measurement environment and the robot trajectory showed in red.}
\label{fig:phobos_2}
\end{figure*}

One of the main focuses of the Automation Technology professorship in Tu Chemnitz is SLAM, and multi-sensor fusion~\cite{proaut}, and this chapter uses one of the data sets from this professorship. The measurements are taken behind the tower in TU Chemnitz, as shown in Fig. \ref{fig:phobos_2}.

\section{ Experimental Results }

The evaluation process in this chapter is different from the evaluation process in the point set registration chapter, where we do not have prior knowledge about the initial guess, how much overlap between each consecutive scans, and which scan has a clustered points. For a better initial guess, we used the constant-velocity model. Moreover,  we used 0.2 as the outlier ratio $(\alpha)$, and 10 meters as the standard deviation for the outlier component in each dimension $(x,y)$.\\

This chapter evaluates the estimation error and the estimated covariance, as well as the impact of the Doppler information on the estimation. Thus, the evaluation process runs two times: one time without incorporating the Doppler and the second time with including the Doppler information. The ground truth for this test scenario is optimized by fusing the motion from an odometry system and Radio based ranging system.\\

\begin{figure}[htb!]
\centering
\subfigure[\textbf{ The estimation error without Doppler}]{\label{fig:a}\includegraphics[width=70mm,height=70mm]{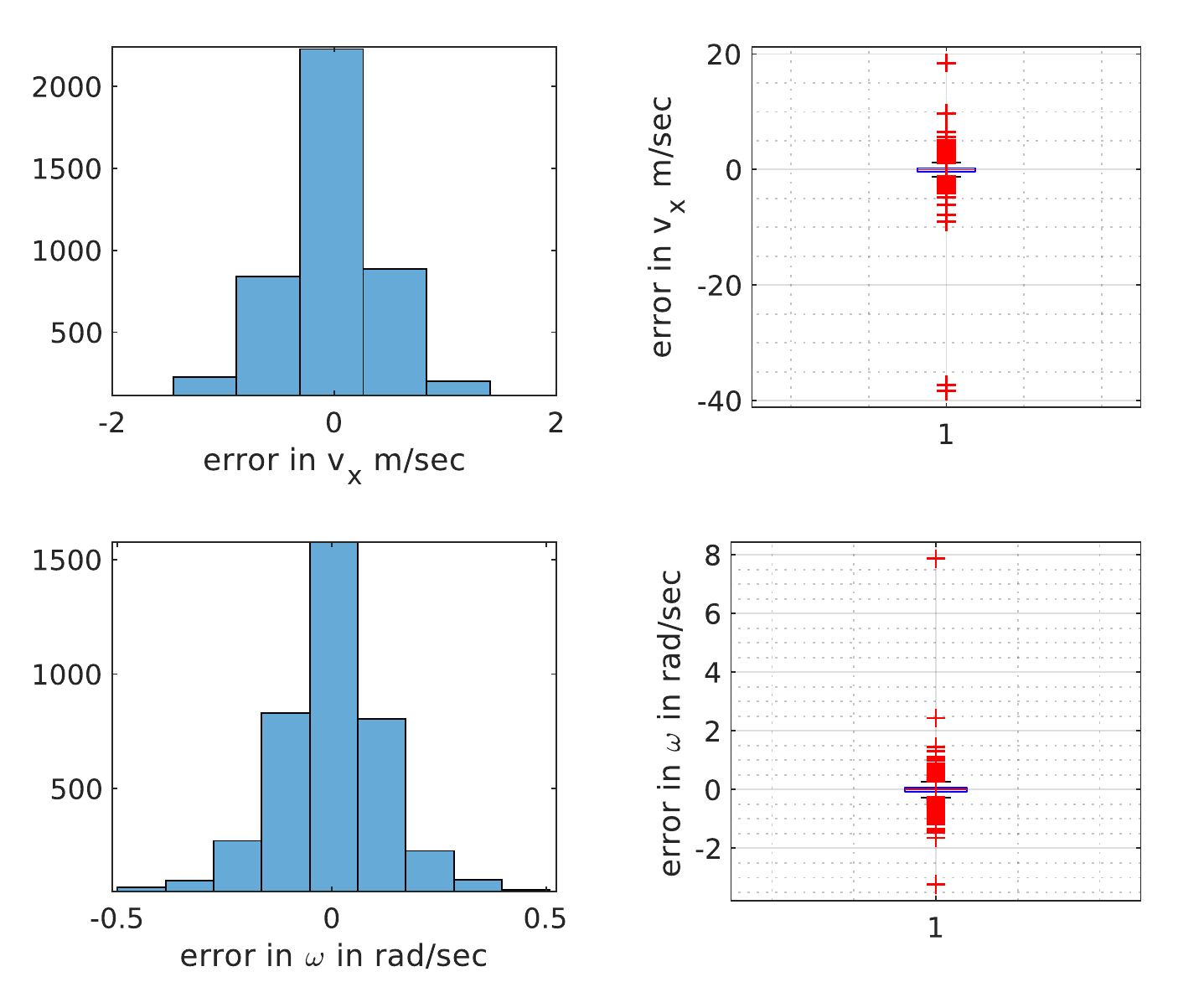}}
\subfigure[\textbf{ The estimation error with Doppler}]{\label{fig:b}\includegraphics[width=70mm,height=70mm]{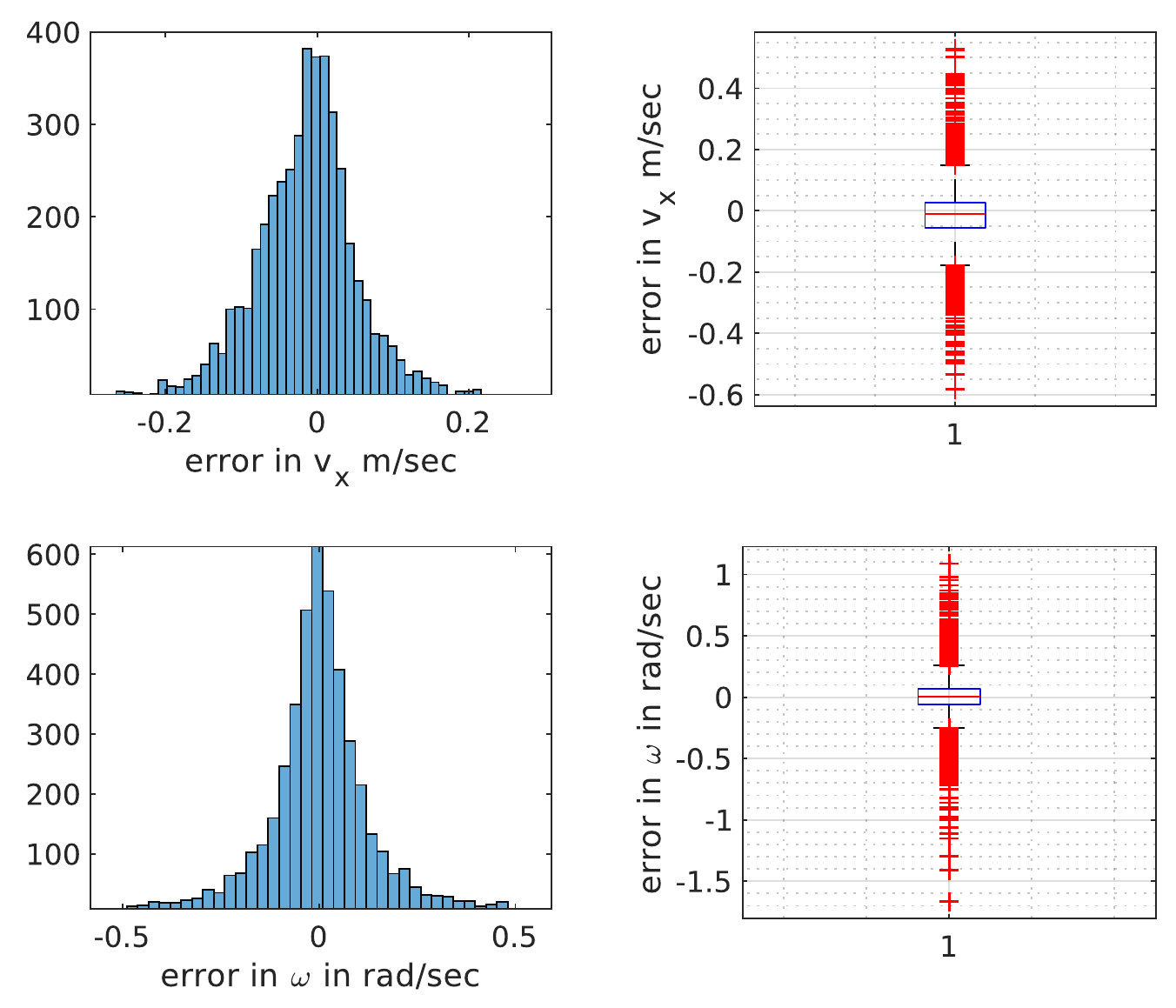}}
\subfigure[\textbf{ The estimation error with better intial guess}]{\label{fig:b}\includegraphics[width=70mm,height=70mm]{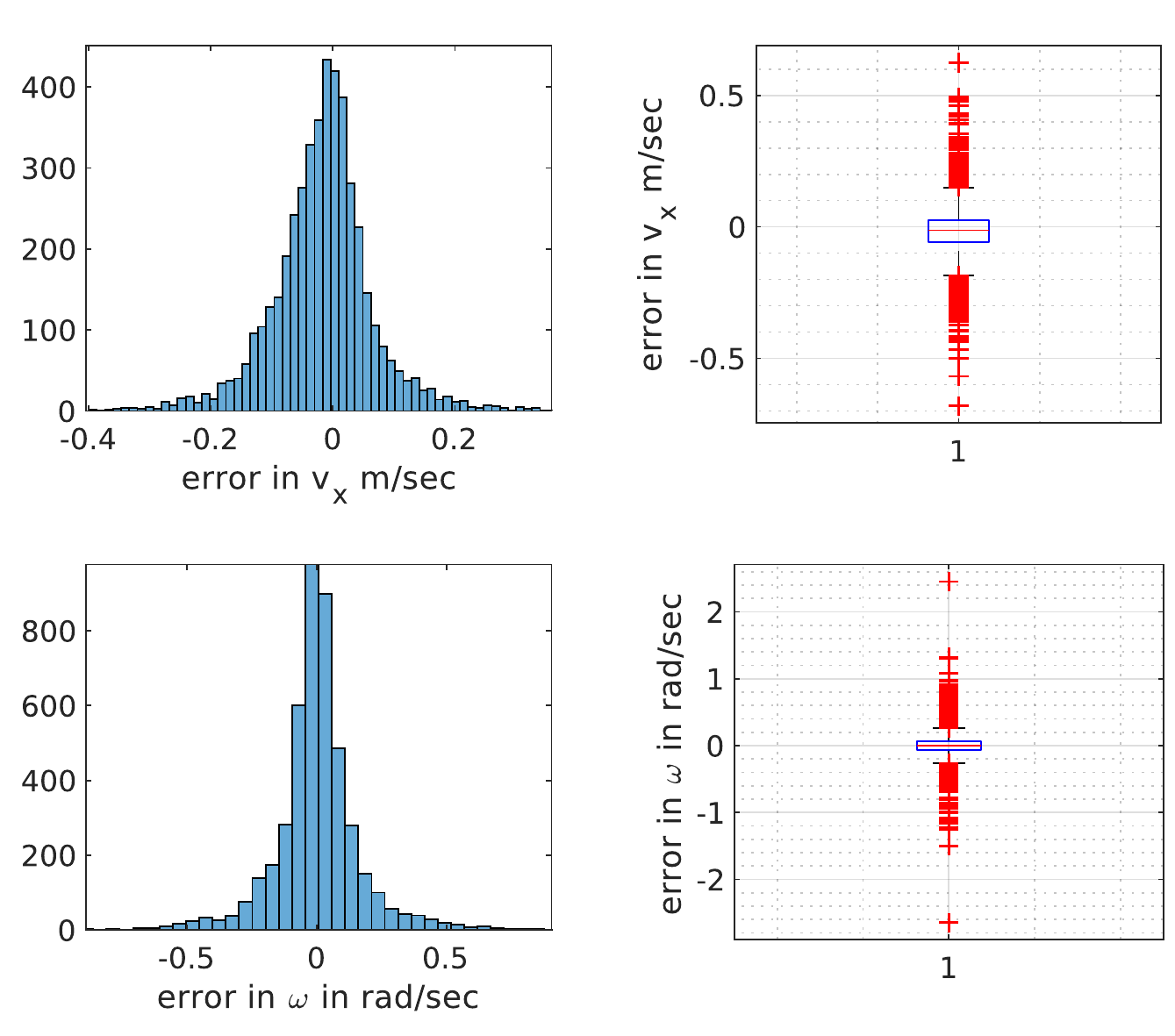}}
\caption{This figure shows the estimation error when the Doppler component is included and excluded from the cost function. Where it shows the Doppler component improves the estimation and minimizing the estimation errors.}
\label{fig:Exp_1}
\end{figure}

\begin{figure}[htb!]
\centering
\subfigure[\textbf{ The estimation without Doppler}]{\label{fig:a}\includegraphics[width=70mm,height=70mm]{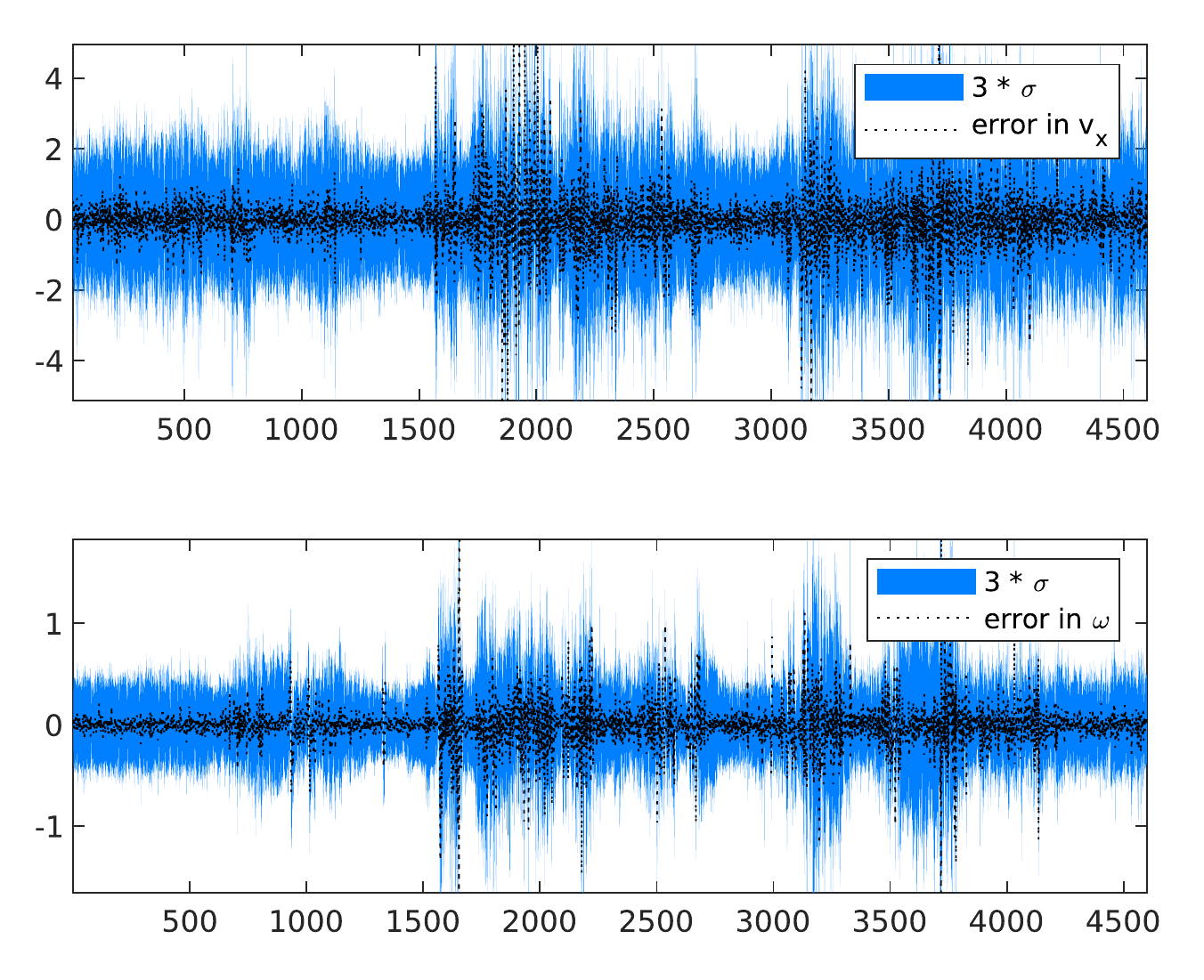}}
\subfigure[\textbf{ The estimation with Doppler}]{\label{fig:b}\includegraphics[width=70mm,height=70mm]{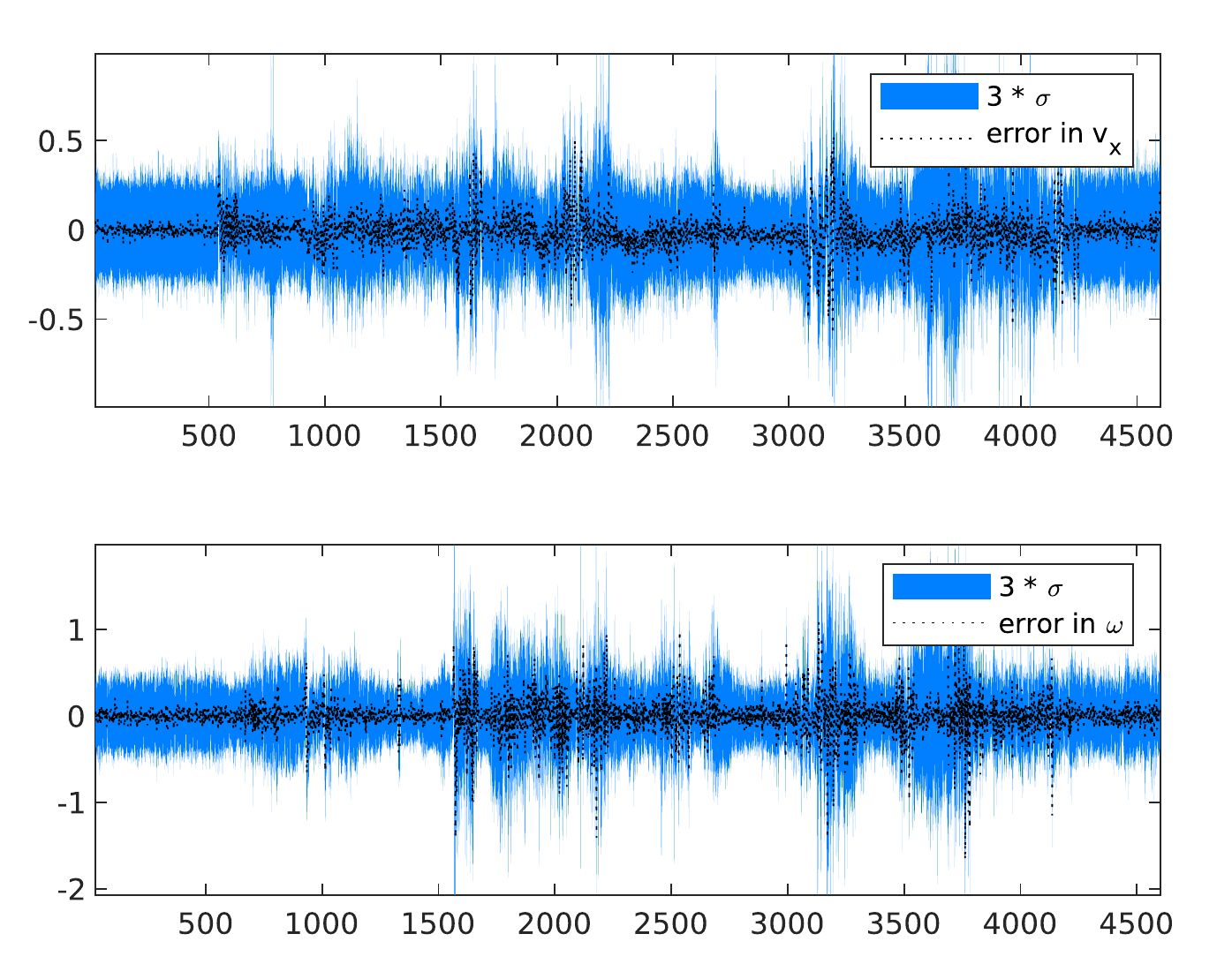}}
\caption{This figure exhibits the estimated error and the estimated covariance, where the blue area represents the $3 \sigma$ of the estimated covariance, and the blck dots are the estimation error. The Doppler information improves the estimated covariance comparing to the estimation without Doppler, where, the doppler is a joint component and the joint make the local minimum less curvature and that makes the estimation more certain. }
\label{fig:Exp_2}
\end{figure}

\begin{figure}[htb!]
\centering
\subfigure[\textbf{ Without Doppler}]{\label{fig:a}\includegraphics[width=70mm]{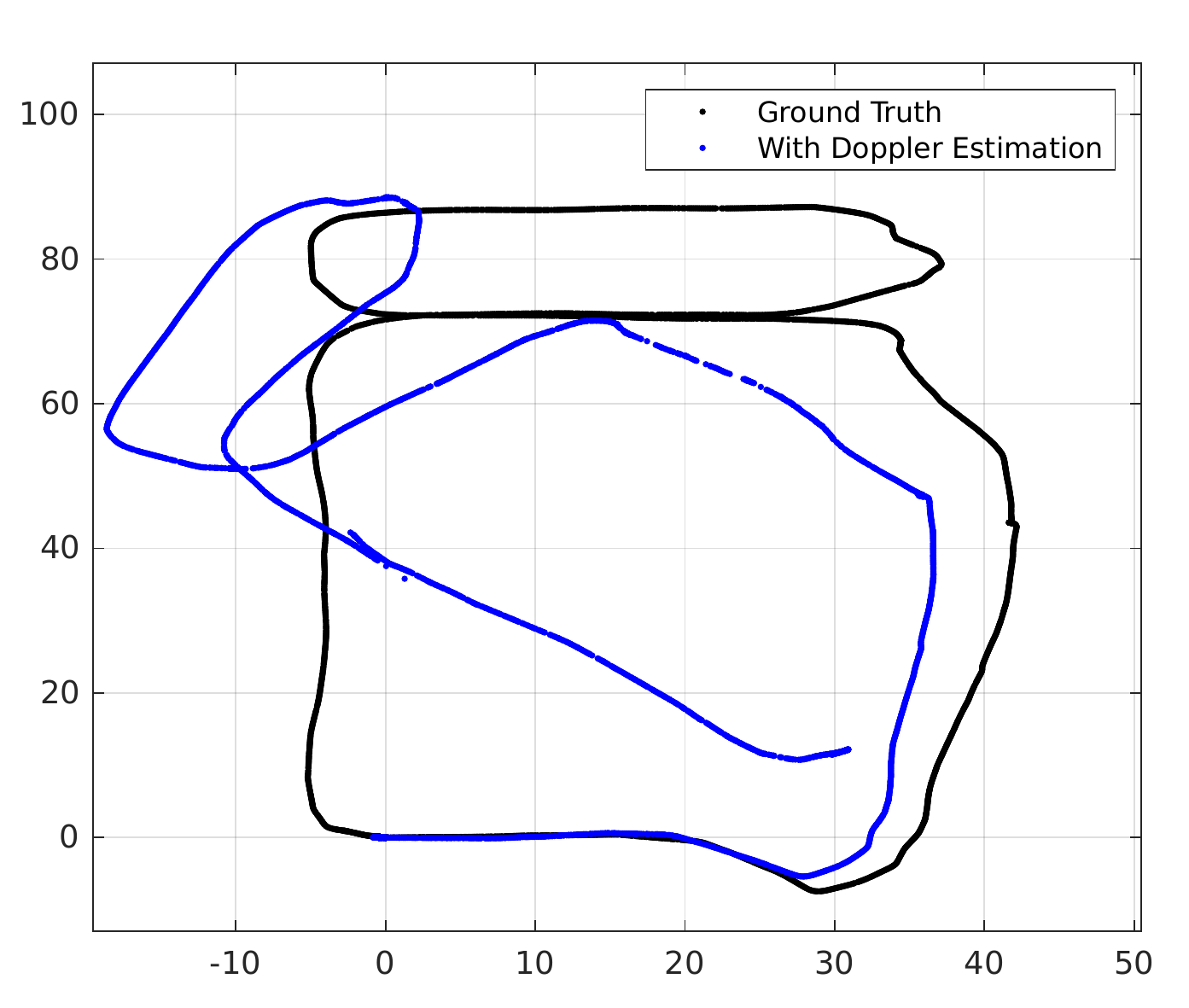}}
\subfigure[\textbf{ With Doppler}]{\label{fig:b}\includegraphics[width=70mm]{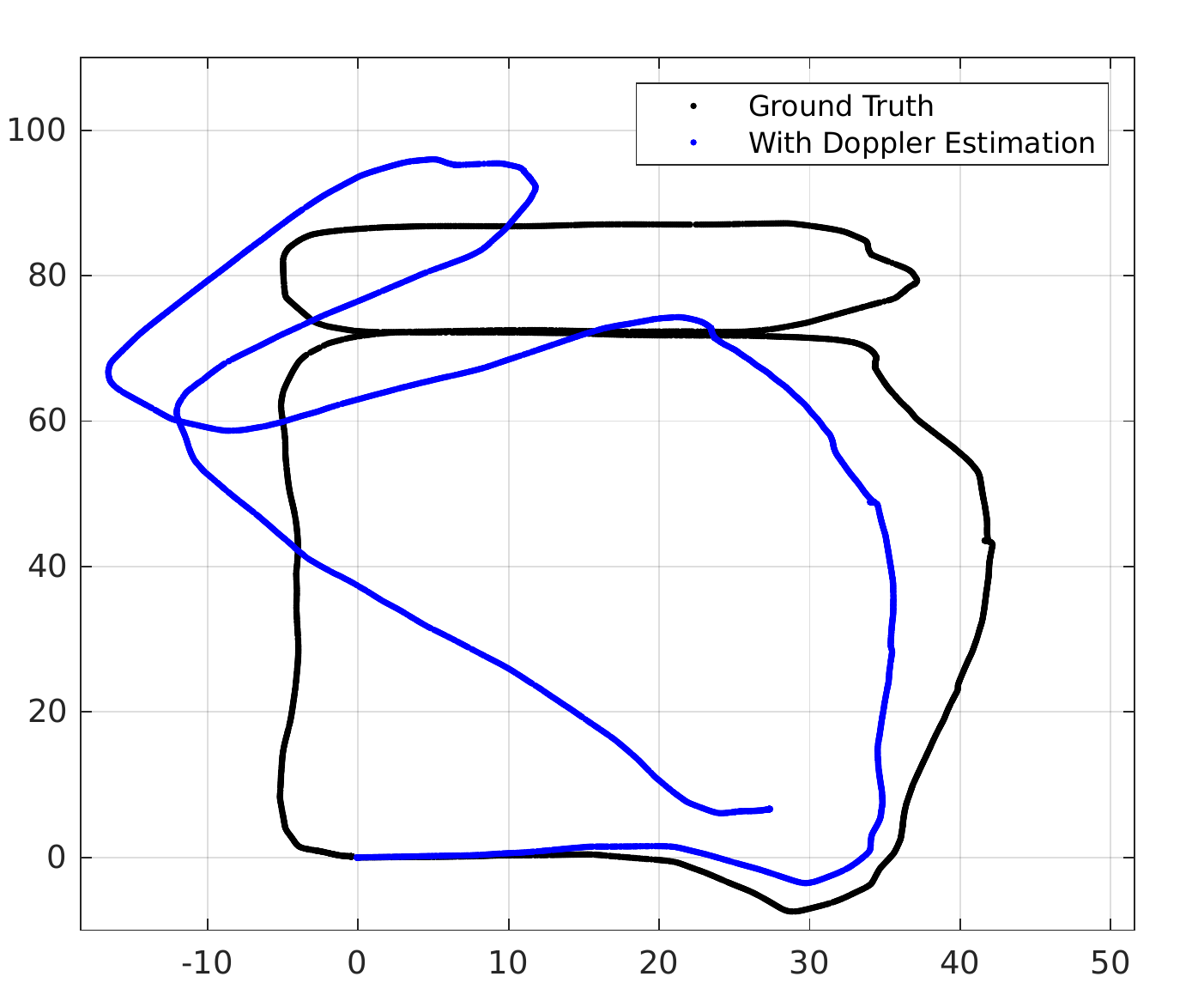}}
\caption{This figure shows the complete trajectory based on the estimated motion, where the Doppler information is included in the cost function in one time and excluded in another.}
\label{fig:Exp_3}
\end{figure}

Firstly, the estimation error without the Doppler information is much bigger than with including the Doppler information, as shown in Fig. \ref{fig:Exp_1}. Thus, the Doppler information reduces the estimation error. Moreover, it modifies the estimated uncertainty, where the estimation becomes certain. The main reason is the final cost function is joint, which reduces the curvature of the optimal minimum, and makes the estimation process more certain about the estimated motion, as shown in Fig. \ref{fig:Exp_2}. Fig. \ref{fig:Exp_3} shows the complete trajectory for the estimated motion in two cases: including and excluding the Doppler in the cost function. Both plots exhibit the accumulative error problem, which is a typical problem for relative motion estimation (local category).\\

Lastly, Jian in~\cite{jian2010robust} proposed to scale the standard deviation for the targets to get a better initial guess. Firstly, scale the standard deviation for the targets, then run the algorithm with a random initial guess, secondly,  scale down the standard deviation to the original values, and run the algorithm again with the optimized value as a better initial guess. Fig. \ref{fig:Exp_1}(c) shows the estimation error after following the extending approach to get a better initial guess, but there is no big difference compared to the normal constant velocity model. One reason for the results is the likelihood approach has only one optimal minimum, and following the constant velocity model for the initial is enough, but this for that specific data set. Nevertheless, extending the covariance could have more impact on other scenarios.

\printindex

\addchap*{Conclusion and Future Work}

This work discussed the point set registration problem to build a robust cost function, where constructing the cost function is a challenging task. ICP is a famous algorithm in point set registration, where it uses explicit correspondence to build the cost function. However, this work avoids explicit correspondence to construct the cost function following a probabilistic framework. The cost function collects the distance metric for individual targets. For each target, the distribution-to-distribution metric shows better behavior than the point-to-distribution because it includes the covariance of both point sets in the target's cost function. One possibility to collect the overall cost surface is, the summing approach, in most cases, it can give an optimal minimum close to the ground truth. However, it is sensitive to the initial guess due to the local minimum problem, which leads to a false convergence. Besides, in the clustering points scenario, it fails to build the optimal minimum close to the ground truth.\\

A second possibility is the likelihood approach, where it gives one optimal minimum, but it sufferers from the outlier effect. However, modeling the outlier component in the cost function overcomes the outlier effect.  Therefore, this work proposed to use the distribution-to-distribution metric to build the cost function for each target, then the likelihood approach to collect all of them, and the corrupted Gaussian model to the outlier component. Moreover, this work proposed to use the Fisher information method to estimate the estimation error as long as the cost function incorporates the variances. \\

Previous works proposed to use the error propagation model to estimate the covariance matrix. Therefore, one possible future work is to estimate the covariance based on the error propagation model and compare the results with the Fisher information method. A second possible future work to do is more exhaustive evaluation, to understand when the algorithm gives a significant estimation error.\\

For the automotive Doppler radar, the target material influences the radar detection. Thus, a third possible future work is to record data set recorded in an environment that is more suitable for the radar specification and examins more and more the radar data, especially the standard deviation for each target.

\bibliographystyle{unsrt}
\bibliography{bibliography}
\end{document}